\documentclass{article}

\usepackage{style/arxiv}

\usepackage{amssymb}
\usepackage{array}
\usepackage{booktabs}       
\usepackage[utf8]{inputenc} 
\usepackage[T1]{fontenc}    
\usepackage[english]{babel}
\usepackage{csquotes}
\usepackage[inline]{enumitem}
\usepackage{float}
\usepackage{graphicx}
\usepackage{hyperref}       
\usepackage{multirow}
\usepackage[round]{natbib}
\usepackage{url}            
\usepackage{siunitx}
\usepackage{subcaption}
\usepackage{textcomp}       
\usepackage{tikz}
\usepackage{xcolor}
\usepackage{soul}
\usepackage{xspace, setspace} 

\newcommand{\eg}{e.g.\xspace}
\newcommand{\ie}{i.e.\xspace}
\newcommand{\bspline}{B-spline\xspace}
\newcommand{\vs}{\textit{vs.}\xspace}
\newcommand{\tsne}{t-SNE\xspace}
\newcommand{\percent}{\%\xspace}
\newcommand{\fold}{x}

\makeatletter 
\newcount\SOUL@minus
\makeatother  

\title{Generative Sampling in Bundle Tractography using Autoencoders (GESTA)}

\author{
Jon Haitz Legarreta \\
Sherbrooke Connectivity Imaging Laboratory (SCIL) \\
Videos \& Images Theory and Analytics Laboratory (VITAL) \\
Department of Computer Science, Universit\'{e} de Sherbrooke, Canada \\
\And
Laurent Petit \\
Universit\'{e} Bordeaux, CNRS, CEA, IMN, GIN, UMR 5293, F-33000 Bordeaux, France \\
\And
Pierre-Marc Jodoin* \\
Videos \& Images Theory and Analytics Laboratory (VITAL) \\
Department of Computer Science, Universit\'{e} de Sherbrooke, Canada \\
Imeka Solutions inc., 195 rue Belv\'{e}d\`{e}re Nord, Sherbrooke (Qu\'{e}bec) J1H 4A7, Canada \\
\And
Maxime Descoteaux* \\
Sherbrooke Connectivity Imaging Laboratory (SCIL) \\
Department of Computer Science, Universit\'{e} de Sherbrooke, Canada \\
Imeka Solutions inc., 195 rue Belv\'{e}d\`{e}re Nord, Sherbrooke (Qu\'{e}bec) J1H 4A7, Canada \\
\\
*{Co-senior author. These authors contributed equally.}
}

\begin{document}

\maketitle

\vspace*{-0.1cm}

\begin{abstract}
Current tractography methods use the local orientation information to propagate streamlines from seed locations. Many such seeds provide streamlines that stop prematurely or fail to map the true white matter pathways because some bundles are ``harder-to-track'' than others. This results in tractography reconstructions with poor white and gray matter spatial coverage. In this work, we propose a generative, autoencoder-based method, named GESTA (\textit{Generative Sampling in Bundle Tractography using Autoencoders}), that produces streamlines achieving better spatial coverage. Compared to other deep learning methods, our autoencoder-based framework uses a single model to generate streamlines in a bundle-wise fashion, and does not require to propagate local orientations. GESTA produces new and complete streamlines for any given white matter bundle, including hard-to-track bundles. Applied on top of a given tractogram, GESTA is shown to be effective in improving the white matter volume coverage in poorly populated bundles, both on synthetic and human brain \textit{in vivo} data. Our streamline evaluation framework ensures that the streamlines produced by GESTA are anatomically plausible and fit well to the local diffusion signal. The streamline evaluation criteria assess anatomy (white matter coverage), local orientation alignment (direction), and geometry features of streamlines, and optionally, gray matter connectivity. The GESTA framework offers considerable gains in bundle overlap using a reduced set of seeding streamlines with a \num{1.5}{\fold} improvement for the ``Fiber Cup'', and \num{6}{\fold} for the ISMRM 2015 Tractography Challenge datasets. Similarly, it provides a \num{4}{\fold} white matter volume increase on the BIL\&GIN callosal homotopic dataset, and successfully populates bundles on the multi-subject, multi-site, whole-brain \textit{in vivo} TractoInferno dataset. GESTA is thus a novel deep generative bundle tractography method that can be used to improve the tractography reconstruction of the white matter.
\end{abstract}


\section{Introduction}
\label{sec:introduction}

White matter (WM) brain tractography has become an essential tool to study structural connectivity and track-specific properties in a wide range of applications. Tractography is the computational process of integrating local fiber orientation reconstructions of the white matter into long-range pathways reaching the gray matter (GM). Most commonly, such mapping is done through streamline propagation methods: given a map of local orientations, such as a voxel-wise fiber Orientation Distribution Function (fODF) estimated from diffusion Magnetic Resonance Imaging (dMRI) data, a map of initial locations (seeds), and possibly some anatomical constraints, continuous fiber trajectories are reconstructed using a numerical integration method \citep{Jeurissen:NMRBiomed:2019, ODonnell:NMRBiomed:2019}. The result of this process is a ``tractogram'', composed of a set of three-dimensional curves, called ``streamlines'', estimating the white matter fiber pathways.

Conventional streamline propagation methods can either be deterministic or probabilistic, depending on whether they assume a unique fiber orientation in each voxel, or else, whether a distribution of possible trajectories are allowed at each location \citep{Descoteaux:TMI:2009}. A number of other methods, including global optimization approaches \citep{Reisert:Neuroimage:2011}, particle filtering methods \citep{Girard:Neuroimage:2014}, or surface-enhanced \citep{StOnge:Neuroimage:2018, StOnge:BrainConnectivity:2021}, have emerged to try to adequately reconstruct the white matter pathways. A summary of these methods can be found in \citet{Jeurissen:NMRBiomed:2019} and \citet{ODonnell:NMRBiomed:2019}. More recently, deep learning-based tractography methods have been proposed as an alternative to conventional methods \citep{Poulin:MRI:2019}.

However, modern tractography pipelines still miss to extract streamlines for some fiber pathways \citep{Maffei:Neuroimage:2022, Maier-Hein:NatureComm:2017}. Essentially, models propagating local orientations produce tractography results that are inherently limited in their accuracy \citep{Schilling:MRI:2019}. Several works \citet{Rheault:JNeuralEng:2020, Schilling:Neuroimage:2019, Schilling:HBM:2021, Zhang:Neuroimage:2022} have studied the challenges faced by tractography methods, such as propagating streamlines through hard-to-track regions, which results in streamline groups (bundles) that show a poor volume occupancy and are relatively under-represented, among others \citep{Jeurissen:NMRBiomed:2019}. Incorporating bundle-specific priors (anatomical or orientational, amongst others) to tractography was proposed in \citep{Chamberland:HBM:2017, Rheault:Neuroimage:2019} as a way to increase the likelihood of reconstructing streamlines in hard-to-track regions.

Following our previous work using autoencoders in tractography \citep{Legarreta:MIA:2021}, we show that an autoencoder can be used to generate new and complete streamlines to better fill-up the white matter, especially the hard-to-track regions and bundles. Our contributions are twofold: \begin{enumerate*}[label=(\roman*)]\item a generative, autoencoder-based method for extracting new streamline candidates from the latent space in a bundle-wise fashion; and \item an evaluation framework to asses the anatomical plausibility and dMRI signal fit of the generated streamline data.\end{enumerate*} We show that our approach can be successfully applied to both synthetic and human brain \textit{in vivo} data with extensive experiments. Initiated on the results of an existing tractogram, our generative tractography method yields anatomically plausible streamlines that can be used to improve the spatial coverage of hard-to-track bundles or hard-to-extract clusters of streamlines. The proposed anatomical plausibility framework ensures that the provided streamlines comply with the required geometrical, brain tissue occupancy, diffusion signal fit, and connectivity features. To the best of our knowledge, GESTA (\textit{Generative Sampling in Bundle Tractography using Autoencoders}) is the first deep generative tractography method.

\subsection{Related work}
\label{subsec:related_work}


A number of works have proposed ways to improve the results of tractography methods to provide a better white matter spatial coverage and/or cortical coverage, either using bundle-specific or whole-brain approaches.

\citet{Chamberland:FrontiersNeuroinf:2014} proposed to iteratively improve the output of a tracking method by interactively changing the tractography parameters. By selectively seeding regions of interest, such as those with a poor spatial coverage, and adjusting the streamline propagation parameters, authors showed that their method produces tractograms with an improved white matter occupancy. The same authors later introduced a method \citep{Chamberland:HBM:2017} to improve tracking of the optic radiation by modifying the streamline propagation equation according to orientational priors within some given regions of interest. Bundle-specific tractography \citep{Rheault:biorxiv:2018, Rheault:Neuroimage:2019} requires a set of bundle templates to scale the fiber orientation distributions accordingly, and employs a multi-parametric approach to extract streamlines in a bundle-wise fashion. In \citet{Poulin:ISMRM:2018} authors used a different deep recurrent neural network to reconstruct streamlines for each of the considered bundles.

Whole-brain tractography strategies have been introduced to offer generalized solutions to enhance the reconstruction of hard-to-track bundles. In \citet{Battocchio:MICCAI-CDMRI:2020, Battocchio:ISMRM:2021}, the authors presented a method to improve the WM spatial coverage of a tractogram by re-parameterizing the existing set of streamlines using \bspline functions and computing new streamline trajectories using Markov chain Monte Carlo (MCMC) methods. \citet{StOnge:BrainConnectivity:2021} proposed to seed the WM/GM interface in an adaptive manner based on GM local features. They also proposed to optionally dynamically (iteratively) add seeds in regions presenting a low streamline endpoint density to provide a new tractogram that features an improved spatial and cortical surface coverage. Although this strategy excels in improving the surface coverage, it relies on the surface data availability, and is still hindered by the limitations in the underlying local tracking procedure, which might still track preferably along given orientations in the deep white matter.

Numerous methods using neural networks, based on optimizing the propagation direction predictions according to a loss function, have been proposed for whole-brain tractography in recent years \citep{Poulin:MRI:2019}. Similarly, deep reinforcement learning has been applied to whole-brain tractography \citep{Theberge:MIA:2021} as another choice to avoid detrimental local detours in streamline propagation. In this case, choosing the most appropriate streamline propagation step is learned according to the values of a reward function.

In the broader context of diffusion MRI, deep generative models have emerged as a method capable of successfully performing image synthesis and super-resolution. Adversarial methods have been proposed to synthesize diffusion data (or its derivatives) using structural data (\eg \citet{Anctil-Robitaille:MICCAI-CDMRI:2020}), or to provide high-resolution diffusion MRI from low-resolution images (\eg \citet{Luo:MRI:2022}). These methods employ an \textit{adversarial} discriminator to iteratively allow the network to improve the generated data so as to increase its anatomical reliability. Instead of using a discriminator network, autoencoder-based generative methods with explicit anatomical constraints have also been used successfully in cardiac image segmentation tasks (\eg \citet{Painchaud:TMI:2020}). However, to the best of our knowledge, deep generative models have not been employed to generate a tractography product yet.

In this work, we propose a generative, autoencoder-based tractography approach that, sourcing from a set of streamlines, is able to reconstruct new, anatomically plausible streamlines globally. Our method can be readily applied to both synthetic and human brain \textit{in vivo} tractography data. Compared to other solutions, our method \begin{enumerate*}[label=(\roman*)]\item uses a single model to yield new and complete streamlines; \item it does not involve iterative optimizations of the seeding strategy or streamline trajectories; and \item it does not require locally propagating an orientational field.\end{enumerate*} Our procedure works by generating new streamline candidates for a bundle of choice in the representation space of an autoencoder, and evaluating the anatomical plausibility of the streamlines to accept or to discard them. Our anatomical plausibility evaluation framework includes structural and connectivity constraints, streamline geometry properties, and fixel-based features \citep{Raffelt:Neuroimage:2015}.

\section{Material and methods}
\label{sec:materials_methods}

We leverage the FINTA autoencoder architecture presented in \citet{Legarreta:MIA:2021} to introduce a generative tractography method, GESTA (\textit{Generative Sampling in Bundle Tractography using Autoencoders}). GESTA uses the same convolutional deep neural autoencoder network model to allow extracting new, complete streamlines for tractography. To this end, the autoencoder does not need to be retrained to accomplish the generative task; the learned latent space is re-used as a ``streamline yard'' to produce new streamlines through a bundle-wise sampling process in GESTA. The newly generated latent vectors are then decoded into streamline space, and once their anatomical plausibility established, a new, anatomically plausible (generative) tractogram is provided.

The autoencoder is trained to minimize the mean squared-error loss between the input streamlines and their reconstructions at the output. It is assumed that the training process allows the autoencoder's latent space to hold a (statistically) meaningful representation of the input data points \citep{Bengio:TPAMI:2013}. GESTA takes advantage of such representational space to estimate new points within the data distribution.

GESTA uses a sampling strategy and a streamline acceptance/rejection procedure. A sampling strategy of choice probes new streamline candidates in the latent space using a set of available streamlines as reference, hereinafter named \textit{seed streamlines}. Due to the relatively high dimensionality of the latent space and the uncertainty of the sampling procedure, anatomically implausible streamlines can be generated when probing the latent space. Thus, the newly produced latent samples are decoded into the streamline space, and their features are evaluated according to a number of anatomical, geometrical, diffusion signal fit, and connectivity criteria to determine their plausibility. Given the streamlines of a tractogram as the input, the procedure works as follows:
\begin{enumerate}
\item Train once an autoencoder using raw, unlabeled streamlines.
\item Select a subset of seed streamlines of a poorly covered bundle that will serve as the reference for the sampling procedure.
\item Project the set of seed streamlines to the latent space of the trained autoencoder.
\item Use a sampling procedure to yield new samples in the latent space.
\item Decode the new streamline samples.
\item Compute the streamlines' features, and keep the anatomically plausible streamlines.
\end{enumerate}

Note that the autoencoder needs to be trained only once and remains unaltered to perform all considered tasks (including those performed in \citet{Legarreta:MIA:2021}). Input streamlines are resampled to contain \num{256} equidistant vertices, and the latent space dimensionality and network parameter values are fixed to those used in FINTA \citep{Legarreta:MIA:2021}. 

\begin{figure*}[!htbp]
\centering
\includegraphics[scale=0.985, trim=0.06in 0.2in 0.06in 0.2in, clip=true, width=0.985\linewidth, keepaspectratio=true]{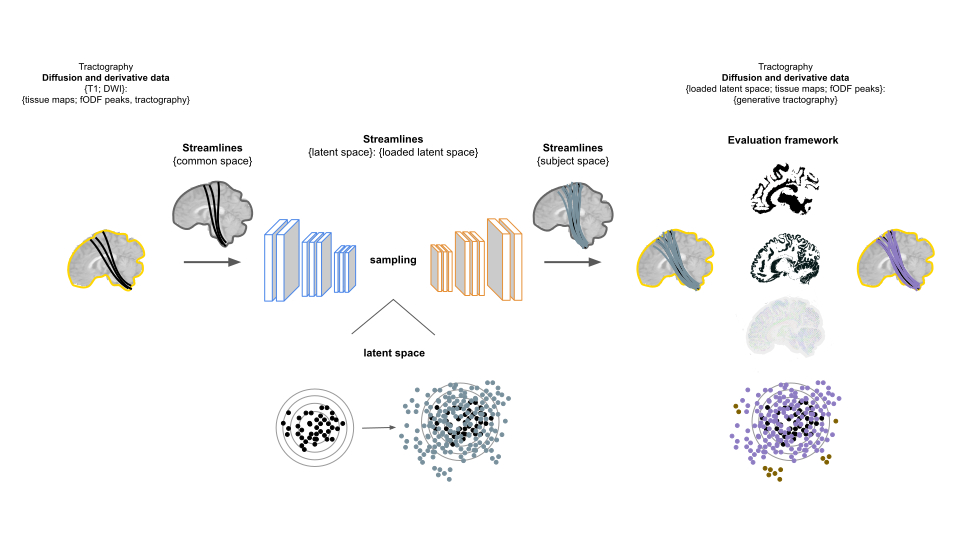} \caption{\label{fig:generative_tractography_pipeline}Illustration of the GESTA (\textit{Generative Sampling in Bundle Tractography using Autoencoders}) generative tractography pipeline using autoencoders. The streamlines of a given subject are put into a common space, such as the MNI-space for example. Given a set of seed streamlines of some bundle (black color streamlines and dots) at the input, the learned latent space of a trained autoencoder holds a meaningful representation of the input streamlines. Such data points are assumed to embody some unknown distribution. A sampling method applied to the subject- and bundle-wise estimated distribution is used to generate new streamline candidates (gray color streamlines and dots). The latent samples, decoded and brought back to the subject space, are accepted (purple streamlines and dots) or rejected (dark gold dots) depending on some plausibility criteria involving anatomical and diffusion-related constraints. Note that the autoencoder training procedure is not shown in the figure.}
\end{figure*}

Figure \ref{fig:generative_tractography_pipeline} depicts the GESTA pipeline applied to a subject at test time. At first, streamlines are brought to a common, standard space (the same space used to train the autoencoder). The learned latent space of the already trained autoencoder is used to generate new streamlines for each fascicle of interest (\eg to fill the spatial coverage of bundles) using a set of seed streamlines. These seed streamlines are assumed to embody a representative set of plausible streamlines of a bundle. Typically, the bundle would include a group of streamlines defined according to some anatomical organizational consistency criterion (e.g. the anatomical regions they traverse or connect \citep{Catani:Neuroimage:2002, Catani:Cortex:2008}), and whose WM volume might not be covered appropriately by a given tractography method. The plausibility of the generated streamlines is evaluated using different criteria, including WM occupancy (anatomy), local orientation alignment (direction), geometry, and GM occupancy (connectivity) features.

GESTA can accept an additional number of seeds, sourced from an atlas, if the bundle at issue contains an excessively low number of seeds or if such seeds are not representative enough for the assumed pathway configurations. We call the mode when no auxiliary seeds are used the \textit{unassisted seeding mode} \vs the \textit{assisted seeding mode}, where a helper atlas is used to provide additional seeds for the latent distribution estimation step (see section \ref{sec:experimental}).

\subsection{Latent space sampling}
\label{subsec:meth:latent_space_sampling}

GESTA uses the rejection sampling method to generate new observations (\ie latent vectors) from the unknown distribution of the (encoded) seed streamlines of a bundle. Given a hard-to-sample distribution $p(z)$, rejection sampling uses a simpler proposal distribution $q(z)$, which is easier to sample from, in order to generate values from the original distribution. It assumes that a constant $k$ exists such that $k q(z) \geq p(z), \, \forall z$. First, a $z_{0}$ value is obtained from the distribution $q(z)$. A number $u_{0}$ is then generated from the uniform distribution over $\left[ 0, k q(z_{0}) \right]$. If $u_{0} > p(z_{0})$, the sample is rejected; otherwise, $u_{0}$ is accepted. In our case, we do not know the $p(z)$ distribution of our seed streamlines in the latent space, and thus, we estimate it using a Parzen estimator. In our experiments, $q(z)$ is set to be a multivariate Gaussian distribution with mean and variance values estimated from the seed data.

\begin{figure*}[!htbp]
\centering
\includegraphics[scale=0.95, trim=0.075in 0.2in 0.075in 0.2in, clip=true, width=0.75\linewidth]{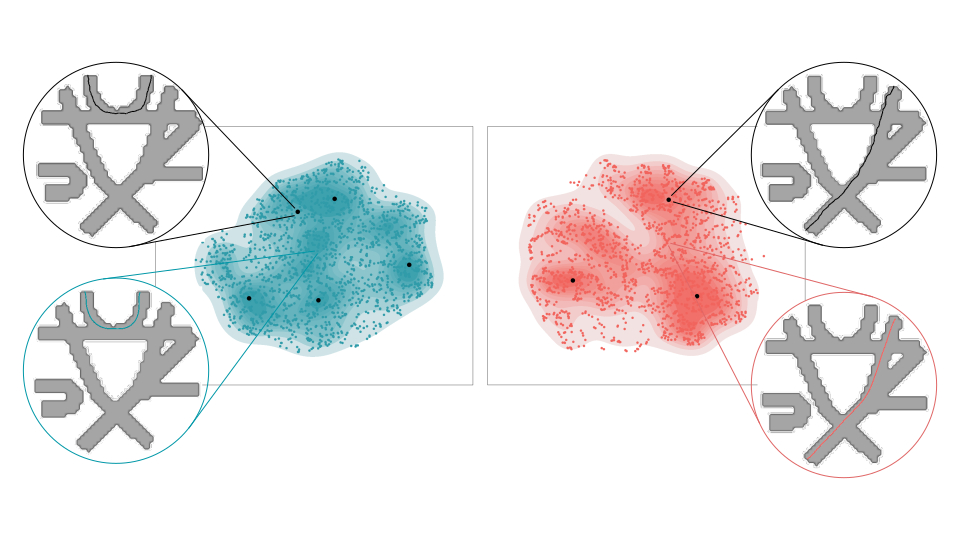}
\caption{\label{fig:fibercup_generative_dimensionality_reduction}\tsne projection of the autoencoder-based generative streamlines corresponding to the ``Fiber Cup'' dataset bundles 1 (turquoise) and 7 (red). \num{3}{\percent} of the test set streamlines' were used to seed each bundle, and the data distribution was estimated using a bandwidth of \num{1.0} (see section \ref{subsec:sampling_params}). \num{2000} samples were generated for each bundle. The generated streamlines are shown as colored dots; the background color represents their density distribution; black dots represent the seed streamlines.}
\end{figure*}

Figure \ref{fig:fibercup_generative_dimensionality_reduction} shows the streamlines sampled using the autoencoder-based generative tractography method as their \tsne projection \citep{vanderMaaten:JMLR:2008} in the latent space for the ``Fiber Cup'' dataset (see section \ref{subsubsec:fibercup}) bundles 1 and 7 (following the numbering in \citet{Cote:MIA:2013}). Using a few seed streamlines for each bundle (shown in black), the latent space sampling retrieves streamlines that are issued by the same distribution. Once their fit to anatomical and diffusion features has been determined, these generative samples are considered as reliable streamlines for all purposes and intents.

\subsection{Streamline evaluation framework}
\label{subsec:streamline_evaluation_framework}

The latent space sampling procedure yields a number of latent vectors $N$. These latent vectors constitute streamline candidates whose anatomical plausibility needs to be established to yield the final generative tractogram. The streamline evaluation framework accepts or discards the latent sampled streamlines depending on a number of features and some given thresholds. The following anatomical reliability criteria are established for streamline acceptance:

\begin{itemize}
\item \textit{ADG} (\textit{Anatomy, Direction, Geometry}): constrains the streamlines' geometry-, fixel-, and WM occupancy-related features. The geometry attributes are determined in terms of the streamlines' length and winding. The streamline-to-fixel compliance is measured in terms of the streamline's local orientation to fODF peak alignment cone. The local orientation is computed as the orientation vector between two consecutive streamline vertices, and the corresponding fODF peak values are interpolated at each of the vertices. Additionally, streamlines are required to be located within the WM tissue. Alternatively, streamlines' WM occupancy compliance can be softened to be set as the ratio of the streamline vertices that lie in the WM tissue over the total number of streamline vertices.
\item \textit{ADGC} (\textit{Anatomy, Direction, Geometry, Connectivity}): in addition to the above criteria, evaluates the degree at which streamlines reach the GM (or the voxels that simulate the expected terminations in the case of synthetic datasets).
\end{itemize}

\begin{figure*}[!htbp]
\centering
\begin{tabular}{cc}
\includegraphics[scale=0.95, trim=0.75in 0.15in 0.75in 0.15in, clip=true, width=0.4\linewidth, keepaspectratio=true]
{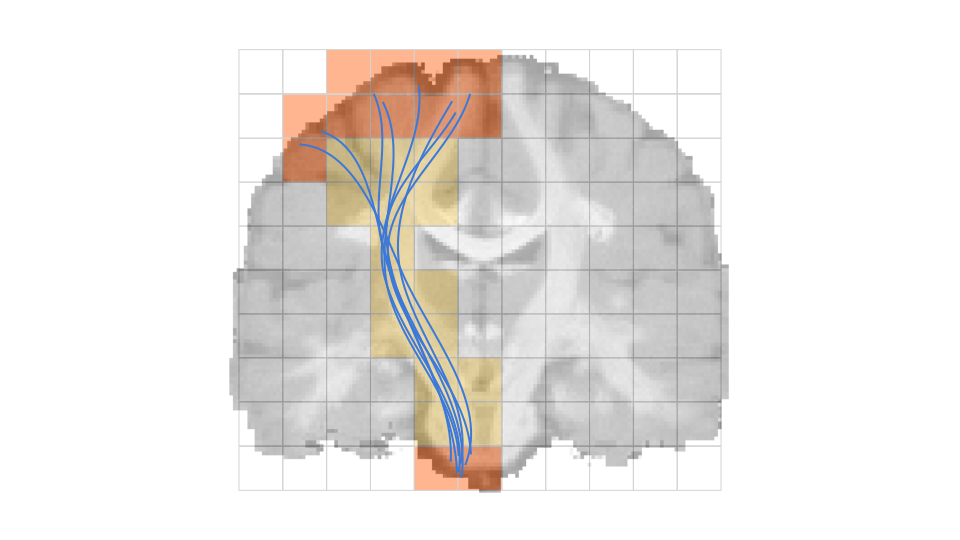} &
\includegraphics[scale=0.95, width=0.335\linewidth, keepaspectratio=true]
{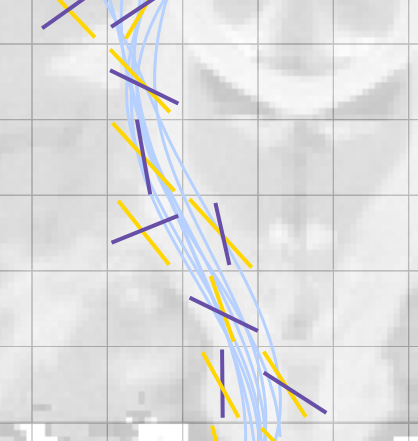} \\
\textbf{(a)} & \textbf{(b)} \\
\end{tabular}
\caption{\label{fig:streamline_evaluation_criteria}Streamline evaluation criteria. (a) Anatomy (WM occupancy) and connectivity (GM occupancy); (b) Direction (local orientation alignment). Yellow color depicts voxels belonging to the WM (anatomy), and orange designates voxels belonging to GM tissue (connectivity). The local orientation is depicted as yellow and purple sticks corresponding to different fODF peaks. Only \num{2} fODF peaks are shown in each voxel for illustrative purposes. The streamline geometry criteria (namely, the length and the winding) are not shown. The \textit{ADG} criterion evaluates the streamlines' WM occupancy (anatomy), the local orientation alignment (direction), and geometry features; the \textit{ADGC} adds to these the GM occupancy (connectivity) criterion. Note that the grid resolution has been oversized for illustrative purposes, and its resolution is not representative of the employed datasets'.}
\end{figure*}

For either of the two criteria, a streamline is only considered plausible when its attributes are within a set of thresholds for all considered features. Figure \ref{fig:streamline_evaluation_criteria} depicts the features involved in the streamline evaluation framework.

\subsection{Data}
\label{subsec:data}

GESTA is tested across a variety of datasets and target bundles. Our use cases include bundles from \begin{enumerate*}[label=(\roman*)]\item synthetic tractography data (``Fiber Cup'' dataset); \item clinical-style realistic data (ISMRM 2015 Tractography Challenge dataset); \item multi-subject, partial tractography \textit{in vivo} human data (BIL\&GIN callosal homotopic data); and \item multi-subject, multi-site, whole-brain \textit{in vivo} human tractography data (TractoInferno dataset \citep{Poulin:NatureSciData:2022}).\end{enumerate*} Particularly, the BIL\&GIN callosal homotopic data is chosen to showcase \begin{enumerate*}[label=(\roman*)]\item how the generative tracking framework can sample the white matter pathways using a different bundling organization without modification; and \item how the generative framework can improve the spatial coverage of bundles that might be affected after a tractography processing step.\end{enumerate*}

\subsubsection{``Fiber Cup''}
\label{subsubsec:fibercup}

A synthetic ``Fiber Cup'' dataset generated using Fiberfox \citep{Neher:MRM:2014} is used to mimic the original ``Fiber Cup'' phantom \citep{Cote:MIA:2013, Fillard:Neuroimage:2011}. This ``Fiber Cup'' phantom contains \num{7} bundles, and the ground truth consists of a total of \num{7833} streamlines. The raw diffusion data were generated using \num{30} gradient directions, a diffusion gradient strength of \SI{1000}{\second\per\milli\metre\squared}, and a \SI{3}{\milli\metre} isotropic spatial resolution for a $64 \times 64 \times 3$ volume.

\subsubsection{ISMRM 2015 Tractography Challenge}
\label{subsubsec:ismrm2015}

The ISMRM 2015 Tractography Challenge dataset of \citet{Maier-Hein:Zenodo:2015} is used as a human brain model quantitative testbed. The dataset consists of a clinical-style, realistic single subject tractogram (containing approximately \num{200000} ground truth streamlines) with \num{25} ground truth fiber bundles, raw diffusion-weighted data and a structural T1-weighted MRI volume generated using Fiberfox \citep{Neher:MRM:2014}. The raw dMRI data were generated at a \SI{2}{\milli\metre} isotropic spatial resolution, with \num{32} gradient directions, and a b-value of \SI{1000}{\second\per\milli\metre\squared}. According to \citet{Maier-Hein:NatureComm:2017}, the dataset contains \num{18} bundles that are considered hard-to-track or very hard-to-track.

\subsubsection{BIL\&GIN callosal homotopic data}
\label{subsubsec:bil_gin_homotopic_cc}

\num{39} subjects of the BIL\&GIN (\textit{Brain Imaging of Lateralization} by the \textit{Groupe d'Imagerie Neurofonctionnelle}) human brain dataset \citep{Chenot:BrainStructFunc:2019, Mazoyer:Neuroimage:2016} are used to demonstrate the autoencoder-based generative tractography framework on clinical-style, single-site, \textit{in vivo} data. Acquisitions were done with a Philips Achieva \SI{3}{\tesla} MR scanner using \num{21} non-colinear diffusion gradient directions and a diffusion gradient encoding strength of \SI{1000}{\second\per\milli\metre\squared} with four averages and an isotropic spatial resolution of \SI{2}{\milli\metre}. In this work, only the corpus callosum is considered, and within the bundle, homotopic streamlines lying within any of \num{26} gyral-based segment pairs are considered as plausible streamlines. Readers are referred to \citet{Legarreta:MIA:2021} for the details about the gyral-based callosal segments used for the extraction of the streamlines. The number of segments that have a mean streamline count larger than \num{5}{\percent} of the mean homotopic streamline count across subjects is \num{3}. Hence, most segments can be considered as poorly populated and very hard-to-track.

\subsubsection{TractoInferno}
\label{subsubsec:tractoinferno}

A subset of the TractoInferno dataset \citep{Poulin:NatureSciData:2022} is used for whole-brain multi-subject, multi-site experiments. The dataset contains diffusion MRI and derivative data corresponding to \num{284} subjects drawn from six (\num{6}) different datasets. The provided \textit{gold standard} tractography data consists of an ensemble tractogram built using local deterministic and probabilistic, Particle Filtering \citep{Girard:Neuroimage:2014}, and Surface-Enhanced Tractography \citep{StOnge:Neuroimage:2018} methods. In this work, a subset of \num{107} randomly selected subjects is used to train the model, each of the tractograms being limited to \num{200000} randomly selected streamlines, evenly split across each tracking method. The model is evaluated on the same \num{28} subject test split used by the original authors. A subset of seven (\num{7}) bundles (left/right AF; CC\_Fr\_1; left/right OR\_ML; left/right PYT) are chosen to demonstrate the generative ability of GESTA.

\section{Experiments}
\label{sec:experimental}

Conventional streamline propagation methods were used on all datasets to obtain the tractography data for this work. Local probabilistic tracking was used for the ``Fiber Cup'' and ISMRM 2015 Tractography Challenge datasets. The BIL\&GIN callosal homotopic tractograms were obtained using the probabilistic setting of the Particle Filtering Tracking (PFT) \citep{Girard:Neuroimage:2014} method. Further details relevant to the tracking parameters can be found in our previous work \citet{Legarreta:MIA:2021}. The experiments on the TractoInferno dataset comprise drawing seeds from local deterministic and probabilistic tractography data as provided with the dataset. They both rank lower among the conventional tractography methods (see \citep{Poulin:NatureSciData:2022}), and thus, are appropriate to demonstrate how GESTA can improve the WM coverage.

The BIL\&GIN callosal homotopic {and the TractoInferno }tractograms were registered \citep{Avants:Neuroimage:2011} to the MNI template common space using the MNI152 2009 standard-space T1-weighted average structural \SI{1}{\milli\metre} isotropic resolution template image \citep{Fonov:Neuroimage:2011}.

The autoencoder trained on each of the mentioned data is maintained unmodified (and was trained using the same parameters for the TractoInferno dataset) with respect to \citet{Legarreta:MIA:2021} for the upcoming experiments.

Appropriate bundles were extracted from each dataset's test set of anatomically plausible streamlines to be seeded separately in GESTA. For the ``Fiber Cup'' dataset, the bundle assignments were obtained using the ground truth bundle endpoints. The bundles for the ISMRM 2015 Tractography Challenge dataset were obtained using the Tractometer tool \citep{Cote:MIA:2013, Maier-Hein:NatureComm:2017}. The BIL\&GIN callosal homotopic data were segmented from the FINTA-determined plausible streamlines according to twenty-six (\num{26}) gyral homotopic regions of interest of the JHU template \citep{Zhang:Neuroimage:2010}. Finally, the bundles for the TractoInferno dataset were extracted using RecoBundlesX \citep{Garyfallidis:Neuroimage:2018, Rheault:PhD:2020}, as proposed by the dataset authors. Section \ref{subsec:acronyms} summarizes the bundle and segment naming used in this work.

To demonstrate GESTA's ability to improve the white matter spatial coverage of a tractogram, we pull a subset of the streamlines in the mentioned datasets' test sets at different rates, and measure the performance on the generative tractogram. Starting from a \textit{complete} (\num{100}{\percent} set) of plausible streamlines for each bundle, different subsets are generated ($P${\percent} sets) by bundle-wise subsampling to serve as seeds for the ``Fiber Cup'', ISMRM 2015 Tractography Challenge, and BIL\&GIN datasets, and a fixed number $S$ of seed streamlines is used for the TractoInferno dataset. The generative tractography framework is then run to extract new, complete, plausible streamlines from such seeds without requiring any iterative propagation procedure to induce an improvement on the measured feature.

Given a trained autoencoder, and the aforementioned complete streamline data, the experiments were completed as follows:
\begin{enumerate}
\item For each bundle, a set of streamlines is selected randomly to serve as the seeds in the latent space. The ratio of the selected streamlines is varied in the set $P = \;$\{\numlist[list-separator={,},list-final-separator={,}]{3;5;10;100}\}{\percent} for the ``Fiber Cup'' and ISMRM 2015 Tractography Challenge datasets to test the generative tractography procedure under different hard-to-track conditions across all bundles. Due to the low streamline availability, all available test set streamlines ($P = 100${\percent}) are used as seeds in the BIL\&GIN callosal homotopic dataset experiment. For TractoInferno, the number of seed streamlines is fixed to $S = 1000$ for each bundle (see section \ref{subsubsec:tractoinferno_considerations} for details).
\item New streamline candidates are generated from the latent space using the rejection sampling algorithm.
\item The generated latent samples are decoded and are accepted/rejected using an anatomical plausibility evaluation framework.
\end{enumerate}

The numerical measures used to grade the white matter spatial coverage improvement offered by GESTA are described in section \ref{subsec:wm_coverage_quantification}.

\subsection{Sampling parameters}
\label{subsec:sampling_params}

The Parzen estimator used in the rejection sampling method was allowed a degree of freedom by using different bandwidth factor values to vary the span of the explored latent space to some degree. The values used ranged from \numrange[]{1.0}{10.0}: as a general rule, smaller values result in a weaker sampling performance (both in terms of the necessary time to accept a sample and the wealth of streamlines generated), and larger values result in a large proportion of streamlines that are not fitting the seed streamlines, and hence yield excessively low plausibility ratios. Larger values do not provide significantly better bundle overlap scores either, with only a slight improvement on the ``Fiber Cup'' dataset. Thus, the parameter is fixed to a value of \num{1.0} in all experiments.

GESTA is run in its \textit{unassisted seeding mode} for the ``Fiber Cup'', ISMRM 2015 Tractography Challenge, and BIL\&GIN callosal homotopic datasets, and the \textit{assisted seeding mode} is used for the TractoInferno dataset. A total number of $N =$ \num{2000} streamlines are sampled for each the ``Fiber Cup'' bundles; $N =$ \num{15000} streamlines are generated for each bundle of the ISMRM 2015 Tractography Challenge dataset; $N =$ \num{6000} for the BIL\&GIN callosal homotopic dataset; and the value is fixed to $N =$ \num{40000} for the TractoInferno dataset, prior to the plausibility assessment. The values were set so that a presumably \textit{sufficient} number of streamlines could be generated as a trade-off between the streamline count on the available reference data, the sampling time, and the potentially implausible streamlines generated. They were fixed to the same values across bundles on each dataset for the sake of simplicity.

\subsection{Streamline plausibility thresholds}
\label{subsec:streamline_plausibility_thresholds}

To consider as plausible any streamline generated from the latent space, a set of thresholds are established for the \textit{ADG(C)} (Anatomy, Direction, Geometry, Connectivity) criteria introduced in section \ref{subsec:streamline_evaluation_framework}. The thresholds are set as sensible values in tractography practice. For the sake of simplicity, the constraints that required numerical thresholding values are set to the same values across all bundles for the same dataset. The \textit{ADG} constraints are considered as a fundamental set of features, and are also required to fully accept a streamline under the \textit{ADGC} criterion.

Due to the nature of the ``Fiber Cup'' dataset (where the construct simulating the white matter volume is perfectly delineated, and thus, the structural image can be considered free from any partial volume effects), the WM occupancy is set as a hard, binary requirement. The criterion is relaxed to a soft requirement for the rest of the datasets, where \num{95}{\percent} of the streamline vertices are required to be in the WM tissue. See section \ref{subsec:generative_evaluation_framework} for a more comprehensive discussion. Table \ref{tab:plausibility_criteria_values} summarizes the values used for each dataset.

\begin{table*}[!htbp]
\caption{\label{tab:plausibility_criteria_values}Plausibility criteria values. LOA: local orientation angle; WM: white matter occupancy; T/F: binary requirement (true/false). Note that \textit{ADGC} includes the \textit{ADG} criteria. ``c/h'' denotes ``callosal homotopic'' for the BIL\&GIN data.}
\centering
\begin{tabular}{cc|cccc}
\hline
& & \textbf{``Fiber Cup''} & \textbf{ISMRM 2015} & \textbf{BIL\&GIN c/h} & \textbf{TractoInferno}\\
\cmidrule{2-6}
\multirow{4}{*}{\textit{ADG}} & Length (\si{\milli\metre}) & \numrange[range-phrase=--]{20}{220} & \numrange[range-phrase=--]{20}{220} & \numrange[range-phrase=--]{20}{220} & \numrange[range-phrase=--]{20}{220}\\
& Winding (\si{\degree}) & \num{< 330} & \num{< 330} & \num{< 360} & \num{< 360} \\
& LOA-to-fODF peak (\si{\degree}) & \num{< 30} & \num{< 30} & \num{< 40} & \num{< 40} \\
& WM & T/F & - & - & -\\
& WM ratio ({\percent}) & - & \num{> 95} & \num{> 95} & \num{> 95} \\
\midrule
\textit{ADGC} & GM & T/F & T/F & T/F & T/F\\
\end{tabular}
\end{table*}

The streamline evaluation framework using the WM occupancy binary constraint is designated as \textit{ADG\textsubscript{B}} (and \textit{ADGC\textsubscript{B}}); \textit{ADG\textsubscript{R}} (and \textit{ADGC\textsubscript{R}}) is used when using the WM occupancy ratio hereafter. Readers are referred to section \ref{subsec:evaluation_framework_parameterization} for further details concerning the evaluation framework parameterization.

\subsection{White matter coverage improvement quantification}
\label{subsec:wm_coverage_quantification}

The performance of the GESTA method is graded according to the following measures:
\begin{itemize}
\item \textbf{Bundle volume overlap} (OL). Measures the ratio of voxels occupied within the volume of a ground truth bundle traversed by at least one plausible streamline associated with the bundle. The overlap allows to quantitatively measure the white matter spatial coverage improvement induced by the generative tractography framework.
\item \textbf{Bundle volume overreach} (OR). Measures the ratio of voxels outside the ground truth volume of a bundle that are traversed by at least one (valid) streamline associated with the bundle over the total number of voxels within the ground truth bundle.
\item \textbf{Dice's coefficient}. Computes the harmonic mean of the voxel precision (the fraction of ground truth voxels over the total number of voxels traversed by at least one plausible streamline) and sensitivity (which is equivalent to the overlap in our context).
\item \textbf{Bundle volume}. The volume (in \si{\milli\metre\cubed}) occupied by the streamlines.
\item \textbf{Bundle detection score}. Fraction of bundles detected over the total number of ground truth bundles.
\end{itemize}

For each $P = \;$\{\numlist[list-separator={,},list-final-separator={,}]{3;5;10;100}\}{\percent} seed streamline ratio value used in the ``Fiber Cup'' and ISMRM 2015 Tractography Challenge data, the bundle volume overlap and overreach of the reconstructed tractogram is compared to the generative tractogram for the corresponding ratio. For the ``Fiber Cup'' dataset, the ensemble tracking method results reported in \citet{Joanisse:ISMRM:2021} are used as a baseline (see section \ref{subsec:ensemble_tracking_baselines} for further details). For the ISMRM 2015 Tractography Challenge dataset, the average score across all Challenge submissions, the submissions using ensemble tracking methods (submission identifiers \num{16} through \num{17}), and the best overlap-scoring submission (local probabilistic tracking, submission identifier 12\_2) values are reported as baselines. The data was obtained from the Challenge organizers (see section \ref{subsec:ensemble_tracking_baselines} for details).

For the BIL\&GIN callosal homotopic dataset, the volume occupied by the streamlines (both the seed and the latent-generated ones) is reported.

GESTA is compared to the baseline tractography methods (local deterministic, local probabilistic, PFT, and SET) used by the authors of the TractoInferno dataset. The overlap, overreach, Dice's coefficient and bundle detection score values are reported.

\subsection{Corpus callosum considerations}
\label{subsec:cc_considerations}

The dissected corpus callosum in the ISMRM 2015 Tractography Challenge dataset is further split into \num{6} regions (CC\_Fr\_1; CC\_Fr\_2; CC\_Oc; CC\_Pa; CC\_Pr\_Po; CC\_Te) according to the atlas used by the RecobundlesX bundling method \citep{Rheault:Zenodo:2021} to test GESTA. Only the streamlines recognized as belonging to these regions are kept. In practice, this approach allows the sampling method to effectively generate observations in the corpus callosum system.

Due to both a poor quality under visual inspection and an insufficient amount of streamlines available for the sampling process (see section \ref{subsec:limitations}), the CC\_Te streamlines were discarded. Thus, no generative streamline is extracted for the temporal section of the corpus callosum. Finally, note that the BIL\&GIN callosal homotopic dataset is split into different callosal segments by definition, and uses another delineation atlas, as mentioned in section \ref{subsubsec:bil_gin_homotopic_cc}.

\subsection{TractoInferno considerations}
\label{subsubsec:tractoinferno_considerations}

The local deterministic and probabilistic tractography splits of the test subjects are used to demonstrate the GESTA method. The choice is motivated by \begin{enumerate*}[label=(\roman*)]\item a lower average overlap compared to the PFT and SET tractograms; \item the ability to improve the white matter coverage on the poorly populated deterministic tractography bundles; and \item a richer intra-bundle streamline configuration of probabilistic tractography.\end{enumerate*}

The bundles to sample are identified using the RecoBundlesX method on each local deterministic and probabilistic tractograms. Bundles are not uniformly extracted across subjects, and thus, we use the RecoBundlesX provided atlas as an unbiased, dataset-independent resource to provide at least \num{50}\percent of the seeds in every case for the probabilistic tractography seeds, and \num{70}\percent for the deterministic tracking. A minimum \num{50}\percent value is chosen as a trade-off to potentially allow generating streamlines with a richer intra-bundle configuration. A larger value is used for deterministic seeding in order to allow sampling each bundle within a reasonable time (< \SI{30}{\minute}).

A fixed number of seed streamlines is used for the rejection sampling method; the value is set to $S = 1000$ for all bundles in this work as a trade-off between providing a ``sufficiently'' representative population for seeding appropriately and the sampling time (see section \ref{subsec:sampling_measures}).

The bundles to sample comprise examples of association (AF, OR\_ML), commissural (CC\_Fr\_1), and projection (PYT) pathways. Additionally, for all considered bundles (at least on one of the hemispheres), the overlap for both the deterministic and probabilistic tractography is one (\num{1}) standard deviation below the average in at least \num{10}\percent of subjects.

\section{Results}
\label{sec:results}

First, we show how GESTA can generate streamlines on the ``Fiber Cup'' and ISMRM 2015 Tractography Challenge datasets close to completing their ground truth coverage using only a small proportion of seed streamlines. GESTA's performance using the full set of available seed streamlines (still relatively low for many segments) is analyzed thereafter for the BIL\&GIN callosal homotopic dataset. The TractoInferno results are finally compared to the conventional tractography baselines provided by the dataset authors.

\subsection{``Fiber Cup''}
\label{subsec:fibercup_results}

Table \ref{tab:fibercup_generative_tractography_measures} shows the improvement in the white matter spatial coverage (measured in terms of the bundle volume overlap, OL), and the incurred excess in terms of the overreach (OR) using GESTA for the ``Fiber Cup'' dataset. The seed ratio represents the ratio of streamlines used as seeds in the latent space over the total number of available streamlines for a given bundle in the test set. Results are averaged across bundles. Results show significant gains in bundle overlap while keeping the overreach measures at low values. The proposed generative tractography method is able to raise the overlap measure using a very limited set of seed streamlines: for the $P = 3${\percent} case, it is increased by \num{1.5} times. As more seed streamlines become available, the overlap provided by the latent-generated streamlines increases. The bundle overlap measures are maintained even when the plausibility requirements become more demanding using the \textit{ADGC\textsubscript{B}} criterion. This reveals that prematurely terminated streamlines might be avoided by GESTA (see section \ref{subsec:generative_global_features} for further evidence). The baseline ensemble tractography method outperforms GESTA, at the cost of producing ``no connection'' (prematurely terminated) streamlines (reported at 60\percent by the authors).

\begin{table*}[!htbp]
\caption{\label{tab:fibercup_generative_tractography_measures}``Fiber Cup'' dataset overlap and overreach. (a) GESTA: reconstructed seed streamlines' and generative streamlines' measures. $N =$ \num{2000} streamlines are generated for each bundle with a bandwidth factor of value \num{1.0}, and the plausibility is evaluated using the \textit{ADG\textsubscript{B}} and \textit{ADGC\textsubscript{B}} criteria; (b) Ensemble tracking \citep{Joanisse:ISMRM:2021}. Mean and standard deviation values across bundles. N/R: Not reported.}
\begin{subtable}{\linewidth}
\centering
\caption{GESTA}
\begin{tabular}{ccccc|cc}
\toprule
& & & \multicolumn{2}{c}{\textit{ADG\textsubscript{B}}} & \multicolumn{2}{c}{\textit{ADGC\textsubscript{B}}} \\
\cmidrule(lr){4-5}\cmidrule(lr){6-7}
\textbf{Seed ratio} ($P${\percent}) & \textbf{OL} & \textbf{OR} & \textbf{OL} ($\uparrow$) & \textbf{OR} ($\downarrow$) & \textbf{OL} ($\uparrow$) & \textbf{OR} ($\downarrow$) \\
\midrule
3 & 0.57 (0.11) & 0.01 (0.01) & 0.83 (0.16) & 0.01 (0.02) & 0.83 (0.16) & 0.01 (0.02) \\
5 & 0.68 (0.1) & 0.01 (0.01) & 0.9 (0.08) & 0.02 (0.01) & 0.9 (0.08) & 0.01 (0.01) \\
10 & 0.76 (0.09) & 0.03 (0.02) & 0.91 (0.05) & 0.02 (0.02) & 0.91 (0.05) & 0.02 (0.01) \\
100 & 0.90 (0.07) & 0.09 (0.04) & 0.94 (0.05) & 0.02 (0.01) & 0.94 (0.05) & 0.02 (0.01) \\
\bottomrule
\end{tabular}
\end{subtable}
\newline
\vspace*{0.15in}
\newline
\begin{subtable}{\linewidth}
\centering
\caption{``Fiber Cup'' dataset ensemble tracking.}
\begin{tabular}{ccc}
\toprule
\textbf{Method} & \textbf{OL} ($\uparrow$) & \textbf{OR} ($\downarrow$) \\
\midrule
Ensemble & 0.98 (N/R) & N/R \\
\bottomrule
\end{tabular}
\end{subtable}
\end{table*}

\begin{figure*}[!htbp]
\centering
\setlength{\tabcolsep}{0pt}
\begin{tabular}{cccc}
\includegraphics[scale=0.95, trim=2.5in 2.25in 2.5in 3in, clip=true, width=0.2\linewidth, keepaspectratio=true]{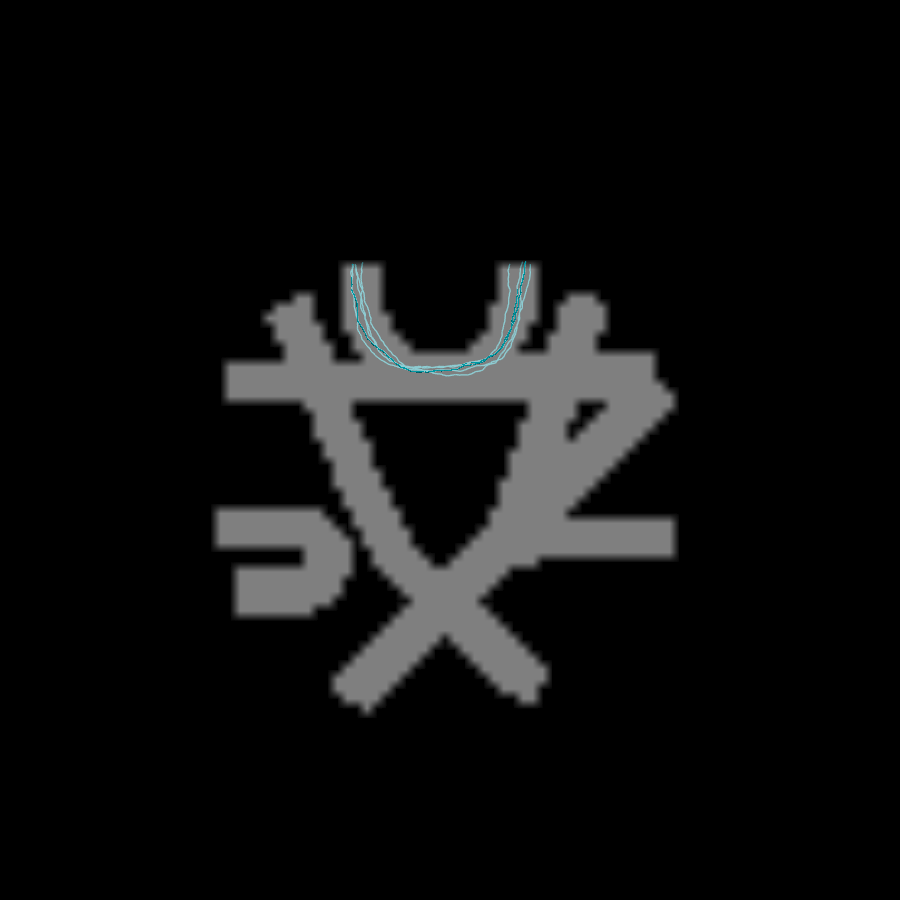} &
\includegraphics[scale=0.95, trim=2.5in 2.25in 2.5in 3in, clip=true, width=0.2\linewidth, keepaspectratio=true]{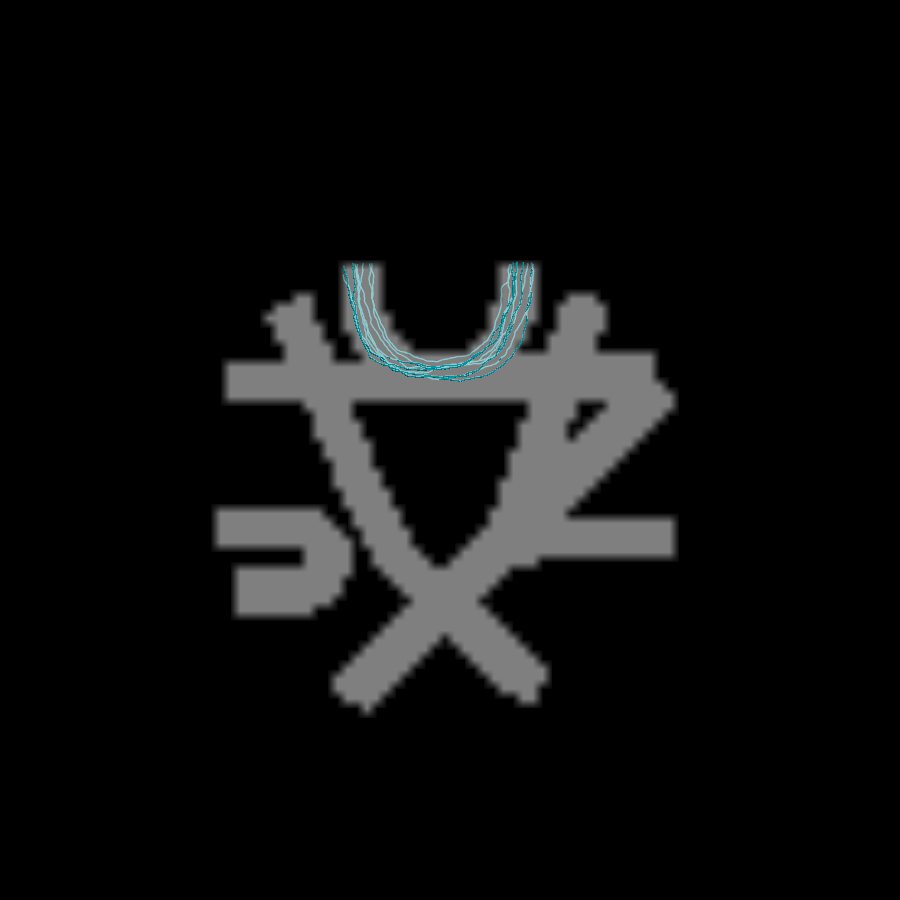} &
\includegraphics[scale=0.95, trim=2.5in 2.25in 2.5in 3in, clip=true, width=0.2\linewidth, keepaspectratio=true]{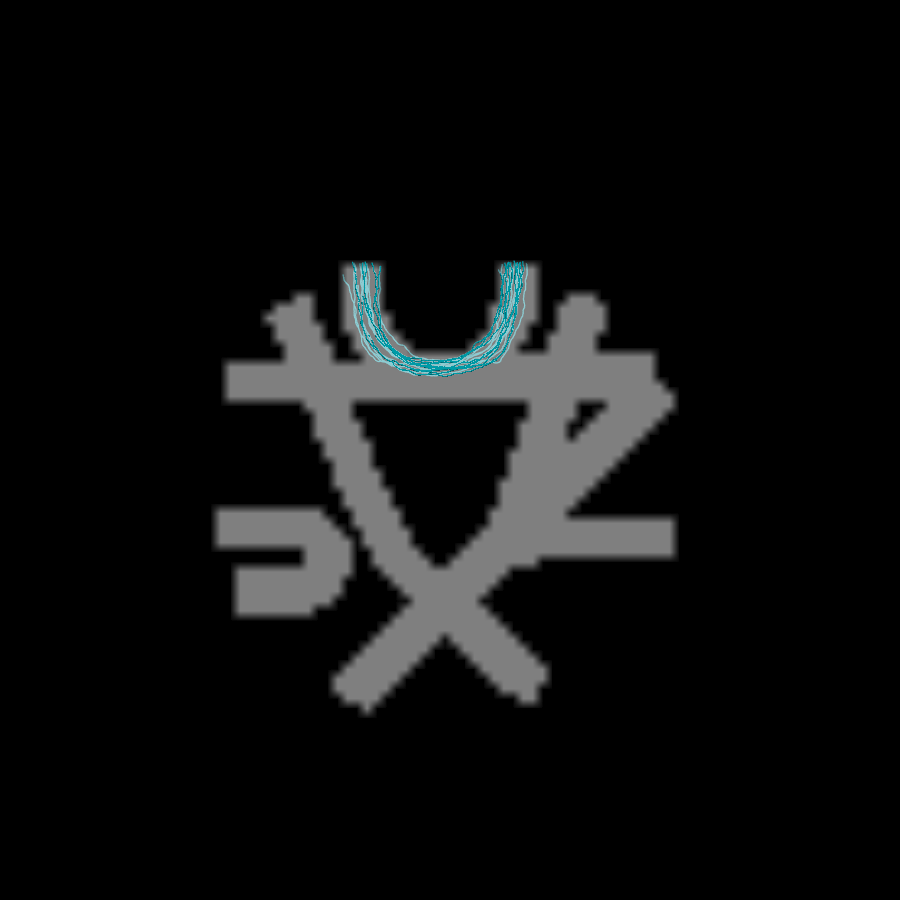} \\
\includegraphics[scale=0.95, trim=2.5in 2.25in 2.5in 3in, clip=true, width=0.2\linewidth, keepaspectratio=true]{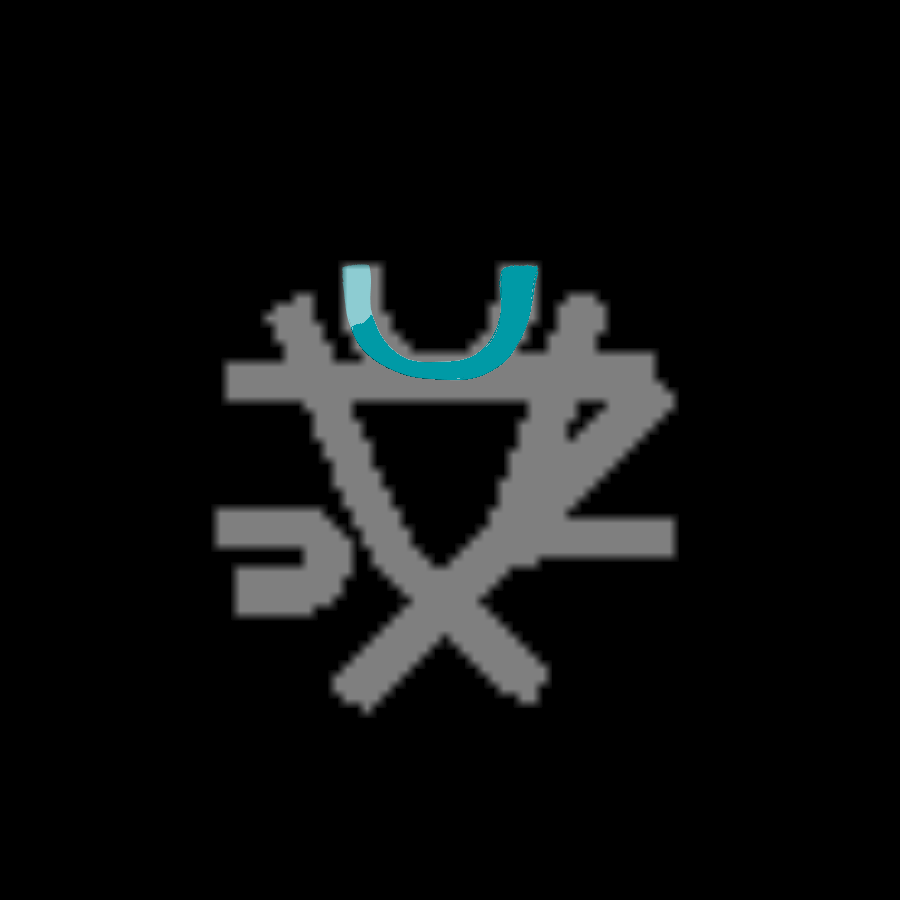} &
\includegraphics[scale=0.95, trim=2.5in 2.25in 2.5in 3in, clip=true, width=0.2\linewidth, keepaspectratio=true]{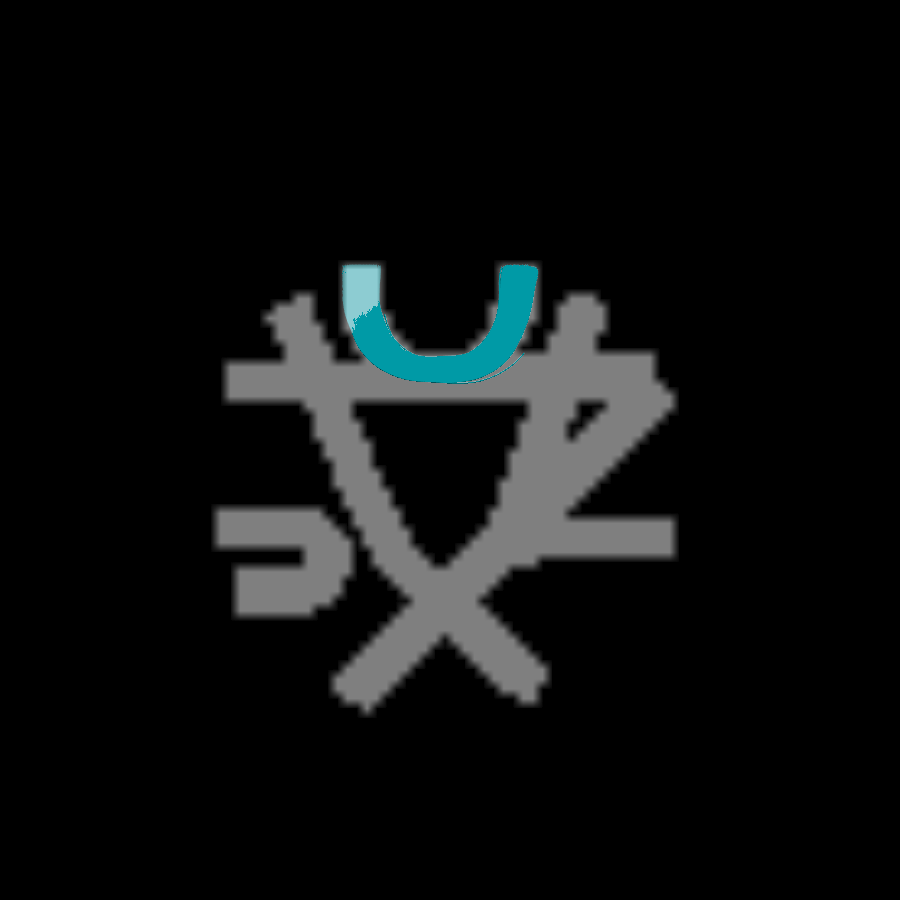} &
\includegraphics[scale=0.95, trim=2.5in 2.25in 2.5in 3in, clip=true, width=0.2\linewidth, keepaspectratio=true]{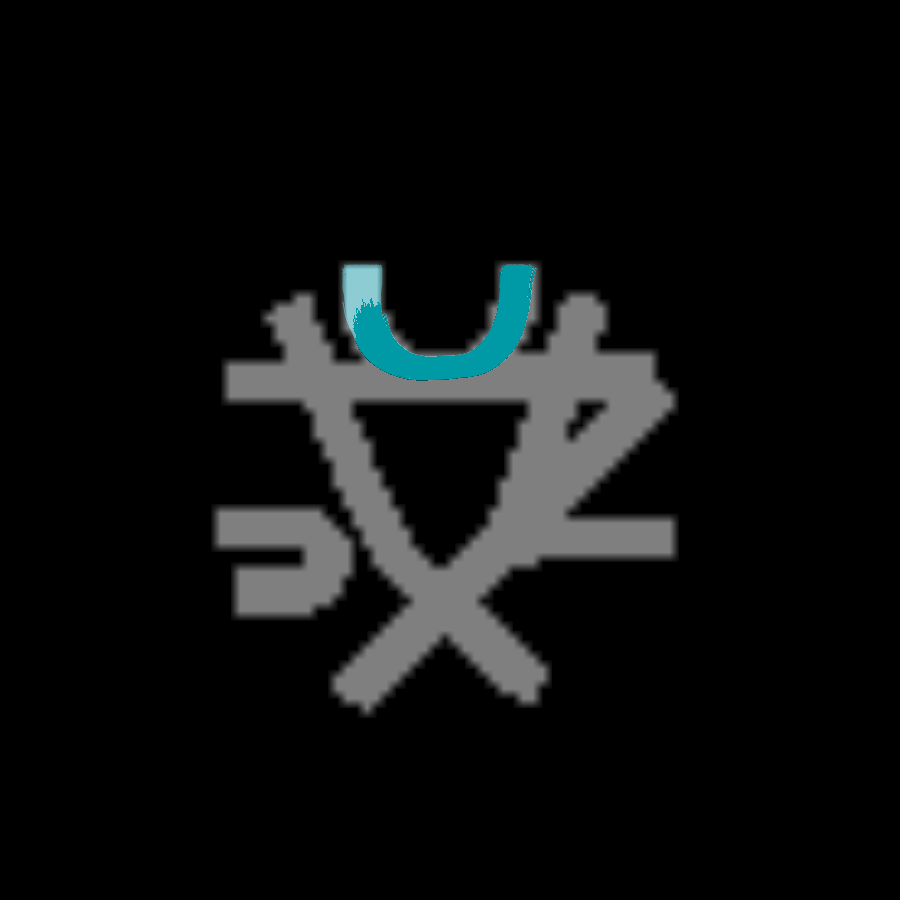} \\
\includegraphics[scale=0.95, trim=2.5in 2.25in 2.5in 3in, clip=true, width=0.2\linewidth, keepaspectratio=true]{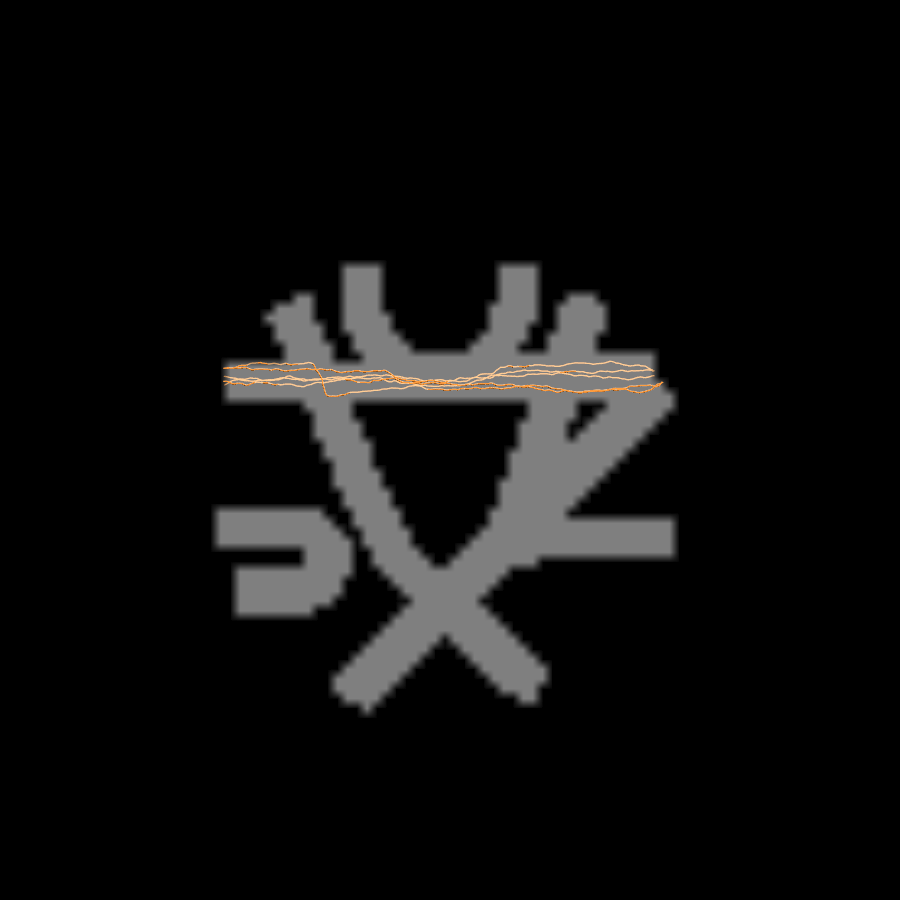} &
\includegraphics[scale=0.95, trim=2.5in 2.25in 2.5in 3in, clip=true, width=0.2\linewidth, keepaspectratio=true]{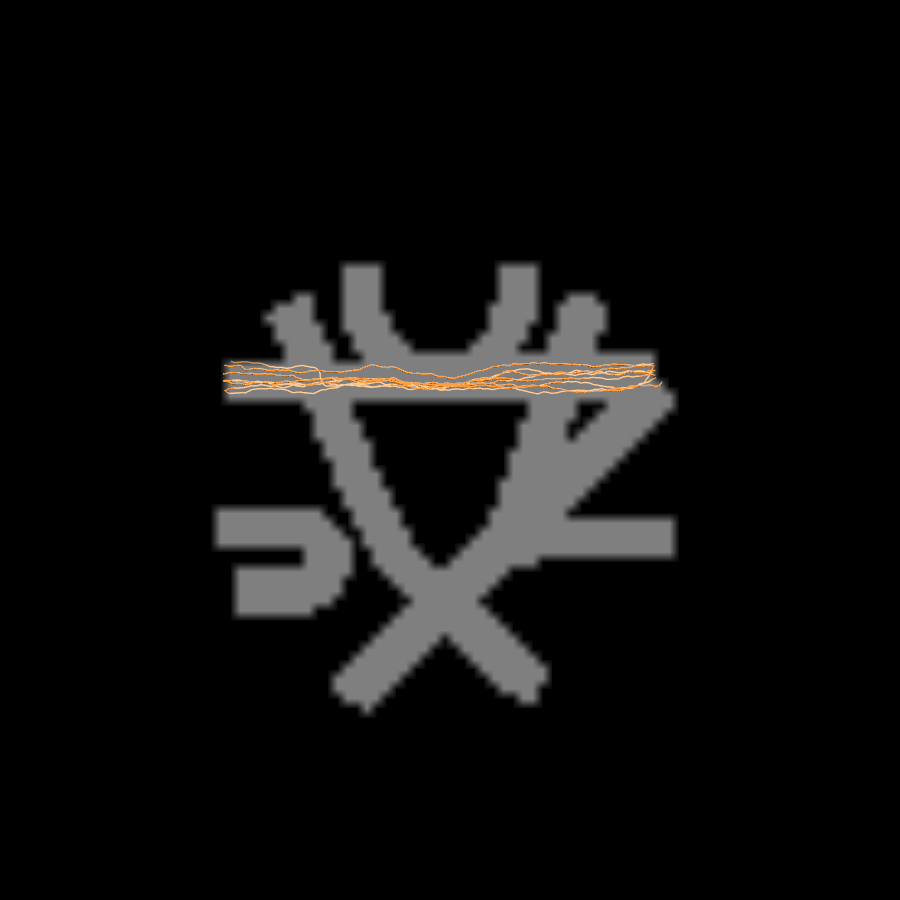} &
\includegraphics[scale=0.95, trim=2.5in 2.25in 2.5in 3in, clip=true, width=0.2\linewidth, keepaspectratio=true]{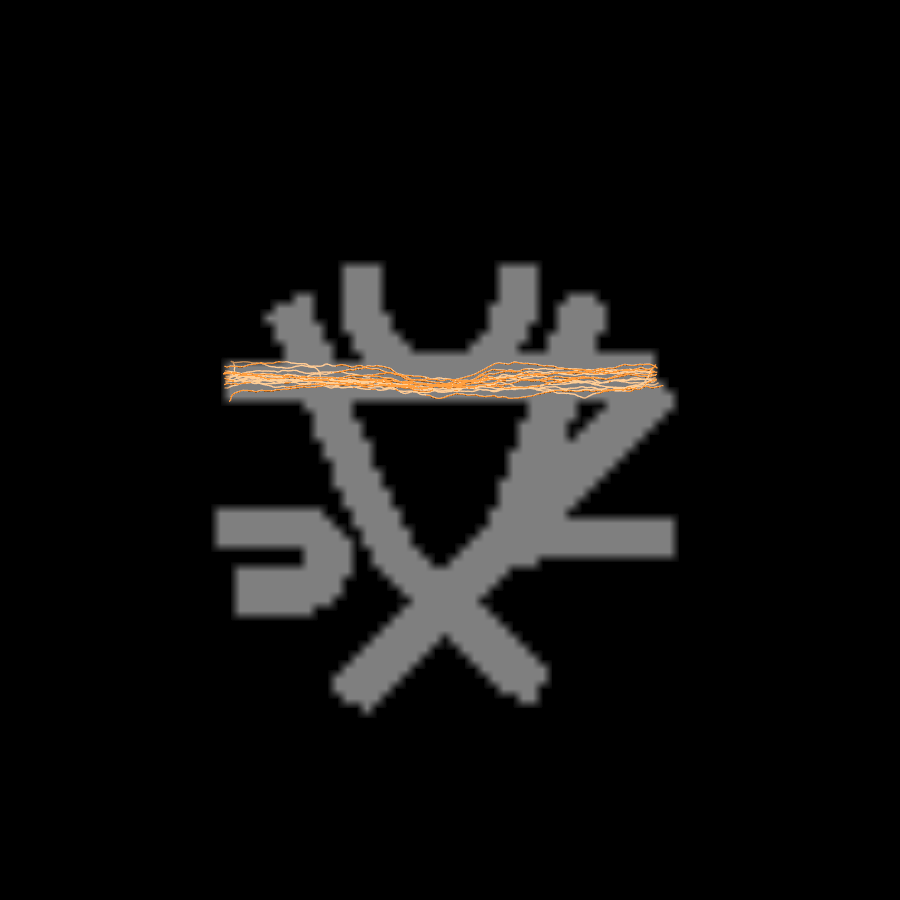} \\
\includegraphics[scale=0.95, trim=2.5in 2.25in 2.5in 3in, clip=true, width=0.2\linewidth, keepaspectratio=true]{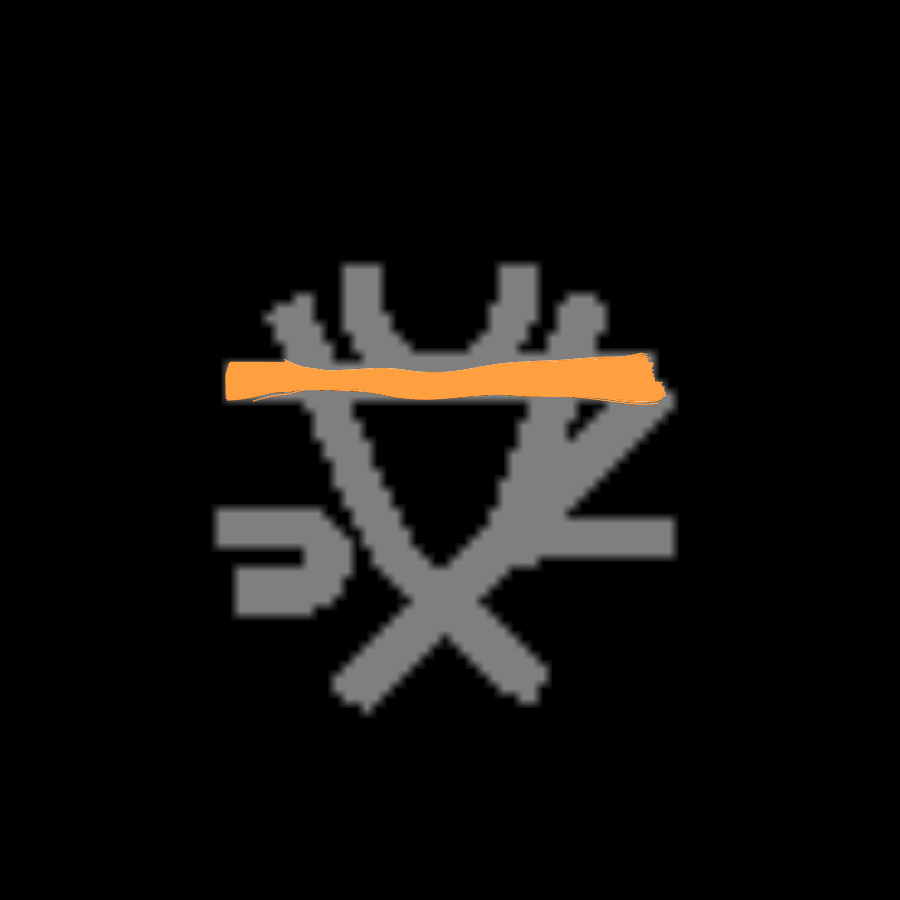} &
\includegraphics[scale=0.95, trim=2.5in 2.25in 2.5in 3in, clip=true, width=0.2\linewidth, keepaspectratio=true]{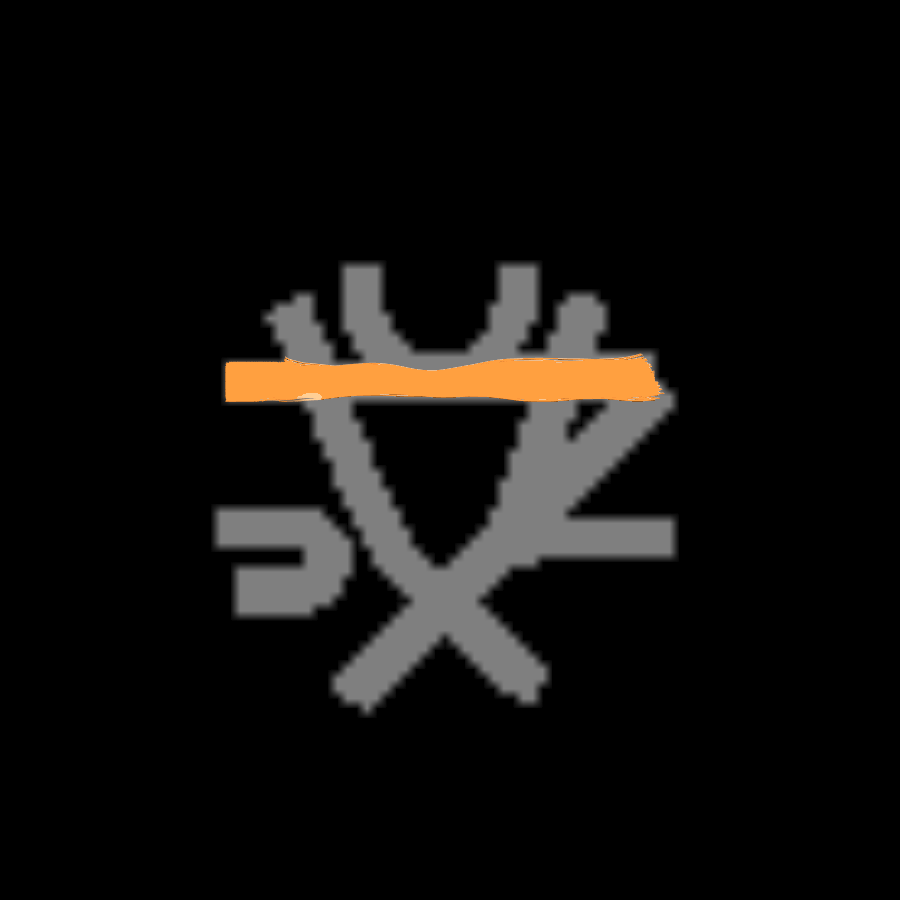} &
\includegraphics[scale=0.95, trim=2.5in 2.25in 2.5in 3in, clip=true, width=0.2\linewidth, keepaspectratio=true]{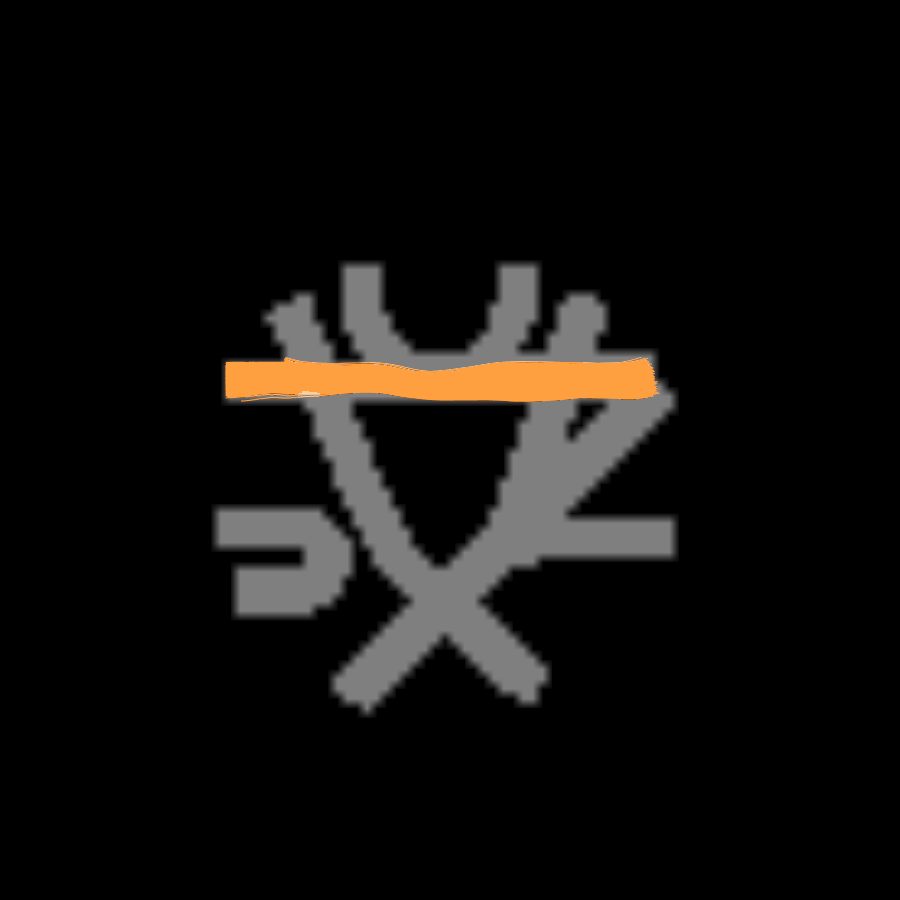} \\
\includegraphics[scale=0.95, trim=2.5in 2.25in 2.5in 3in, clip=true, width=0.2\linewidth, keepaspectratio=true]{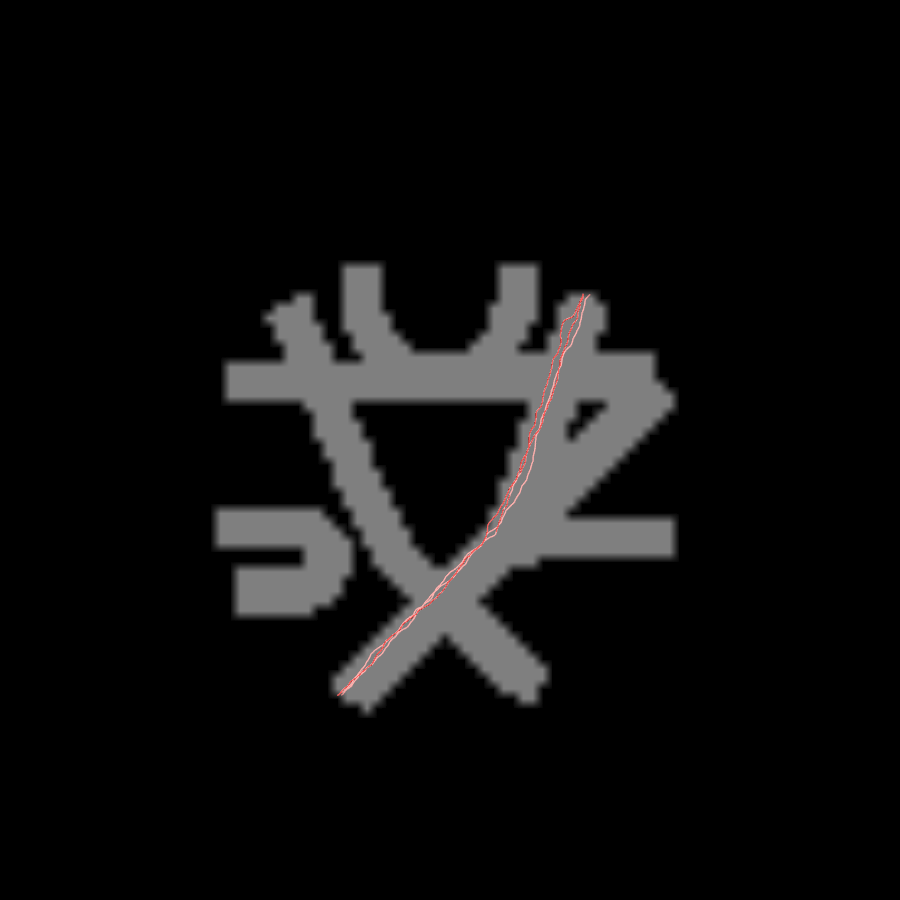} &
\includegraphics[scale=0.95, trim=2.5in 2.25in 2.5in 3in, clip=true, width=0.2\linewidth, keepaspectratio=true]{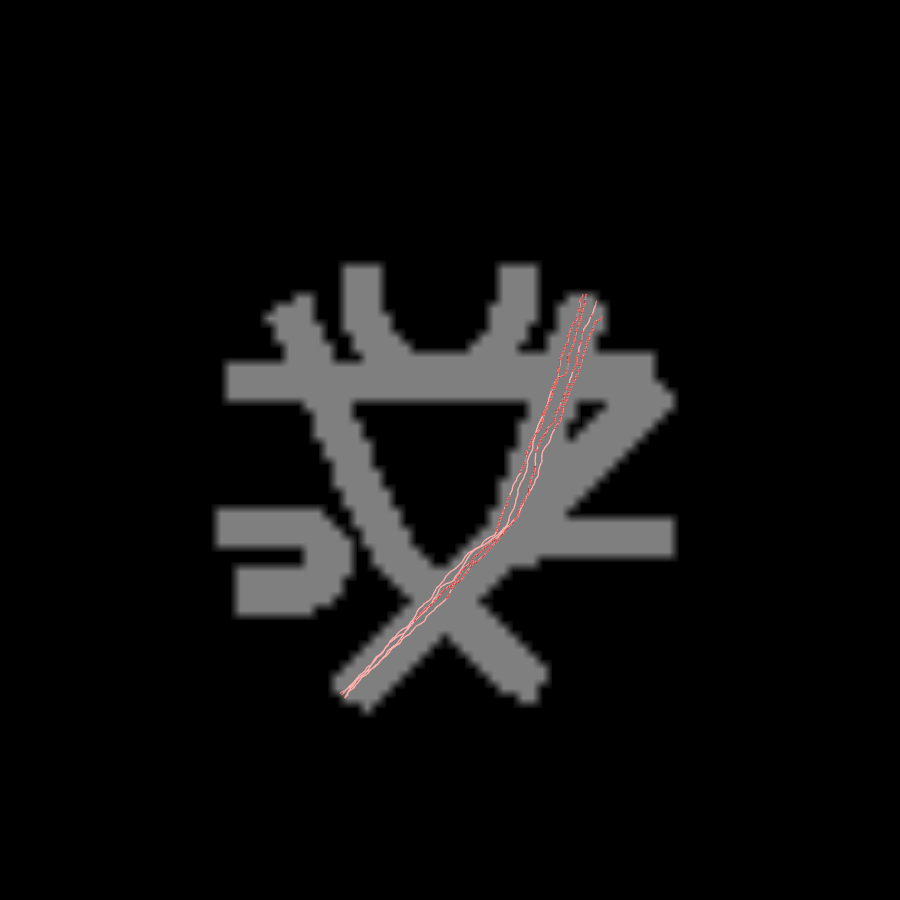} &
\includegraphics[scale=0.95, trim=2.5in 2.25in 2.5in 3in, clip=true, width=0.2\linewidth, keepaspectratio=true]{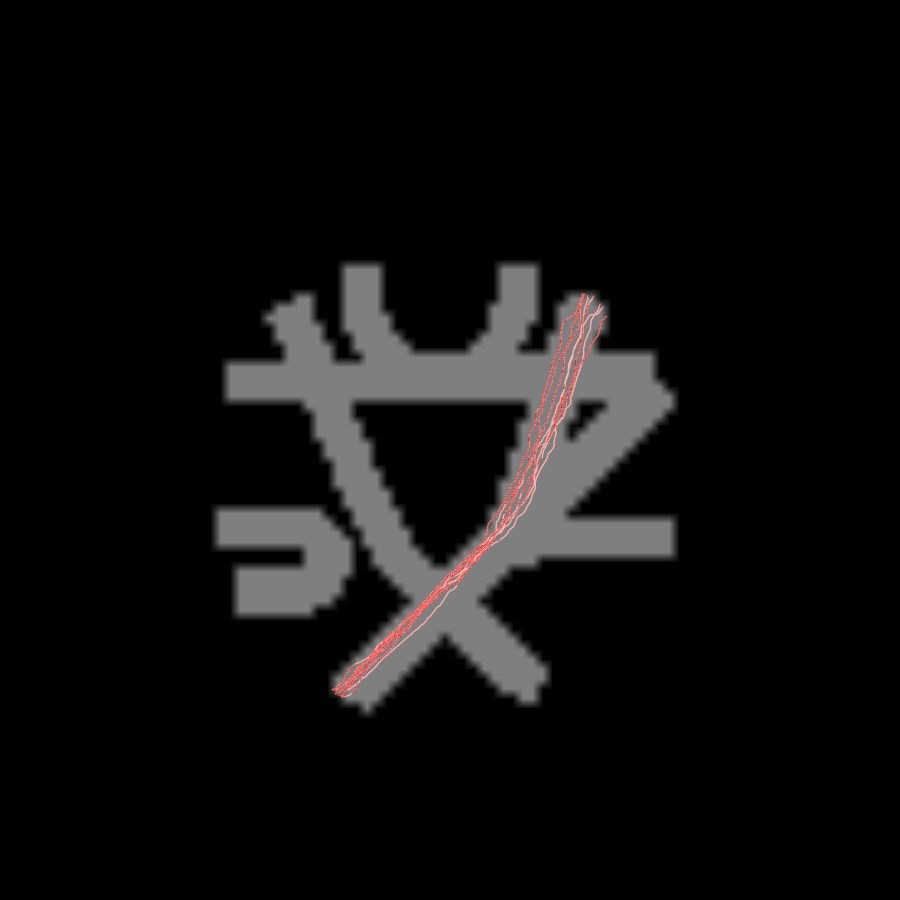} \\
\includegraphics[scale=0.95, trim=2.5in 2.25in 2.5in 3in, clip=true, width=0.2\linewidth, keepaspectratio=true]{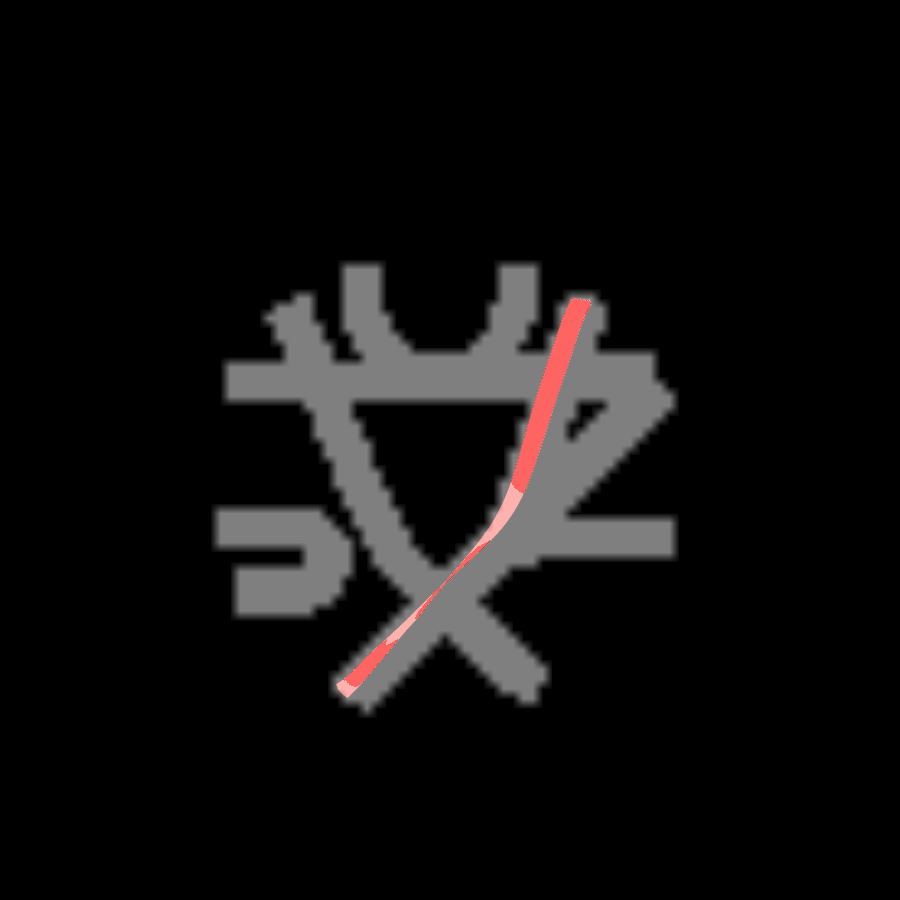} &
\includegraphics[scale=0.95, trim=2.5in 2.25in 2.5in 3in, clip=true, width=0.2\linewidth, keepaspectratio=true]{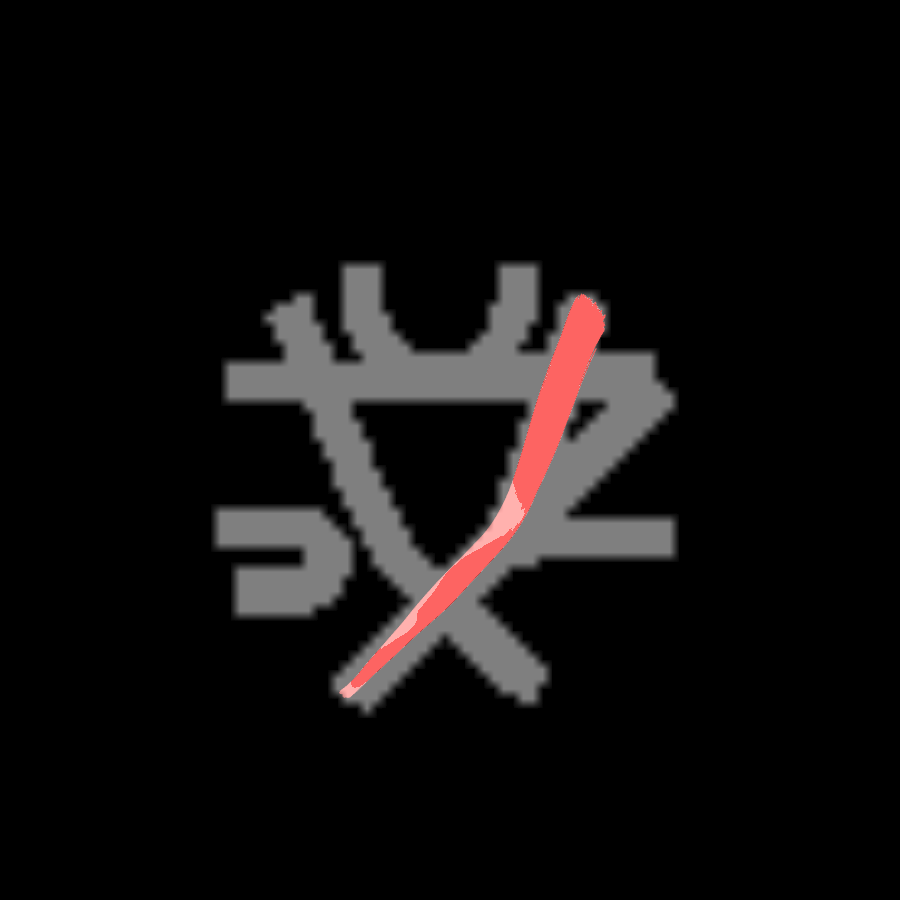} &
\includegraphics[scale=0.95, trim=2.5in 2.25in 2.5in 3in, clip=true, width=0.2\linewidth, keepaspectratio=true]{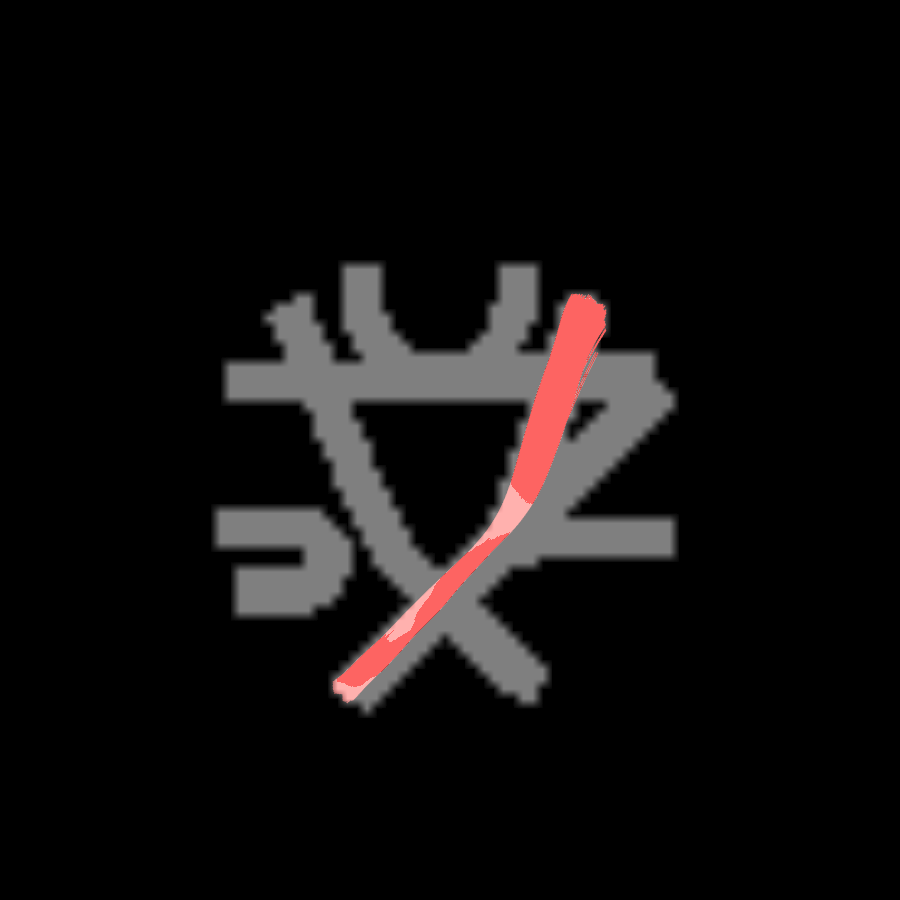} \\
\textbf{(a)} & \textbf{(b)} & \textbf{(c)} \\
\end{tabular}
\caption{\label{fig:fibercup_generative_streamlines}Latent space-generated streamlines for bundles 1 (turquoise), 3 (orange), and 7 (red) corresponding to ``Fiber Cup'' dataset. For each bundle: top: seed streamlines; bottom: latent-generated plausible streamlines. All axial superior views. The seed streamlines represent $P = 3${\percent} (a), $P = 5${\percent} (b), and $P = 10${\percent} (c) of the test set streamlines for each bundle, and streamlines' plausibility is assessed following the \textit{ADG\textsubscript{B}} criterion. Seed streamlines are depicted with a slightly larger diameter compared to the generative streamlines.}
\end{figure*}

Plausible streamlines generated for the ``Fiber Cup'' dataset bundles 1, 3 and 7 are shown in figure \ref{fig:fibercup_generative_streamlines} together with their corresponding seed streamlines. Due to the fact that seed streamlines are chosen randomly, the picked seeds might show an unfavorable spatial distribution for the generative process. This is especially noticeable when the ratio of streamlines used is low (\eg $P = 3${\percent} in bundle 7). The generative framework then suffers from a limited ability to explore the latent space (for a given Parzen estimator bandwidth factor value), and the generated streamlines improve the bundle overlap to a lesser extent.

\clearpage
\newpage

\subsection{ISMRM 2015 Tractography Challenge}
\label{subsec:ismrm2015_results}

Table \ref{tab:ismrm2015_generative_tractography_measures} shows the bundle overlap and overreach results (averaged across bundles) on the ISMRM 2015 Tractography Challenge dataset using GESTA. As it is the case with the ``Fiber Cup'' dataset, the generative tractography method offers a considerable improvement on the bundle spatial coverage: at the $P = 3${\percent} lower-end seed ratio, the overlap is raised to \num{0.74} from \num{0.13} (a \num{5.7}{\fold} factor) for the \textit{ADG\textsubscript{R}} criterion. The ISMRM 2015 Tractography Challenge dataset being a more complex one, as the plausibility requirements become more demanding using the \textit{ADGC\textsubscript{R}} criterion, less streamlines qualify as plausibles, and thus a decrease in the bundle overlap is observed with respect to the \textit{ADG\textsubscript{R}} criterion. Despite the streamline connectivity constraint limiting the gain, a value of \num{0.74} is achieved for the same $P = 3${\percent} seed ratio. Results show that GESTA's performance is competitive against the baselines reported in the Challenge submission. The larger overreach values recorded for GESTA result from the need to relax the tissue occupancy criteria (see section \ref{subsec:limitations} for a discussion).

\begin{table*}[!htbp]
\caption{\label{tab:ismrm2015_generative_tractography_measures}ISMRM 2015 Tractography Challenge dataset reconstructed seed streamlines' and generative streamlines' overlap and overreach. $N =$ \num{15000} streamlines are generated for each bundle with a bandwidth factor of value \num{1.0}, and the plausibility is evaluated using the \textit{ADG\textsubscript{R}} and \textit{ADGC\textsubscript{R}} criteria. Mean and standard deviation values across bundles.}
\begin{subtable}{\linewidth}
\centering
\caption{GESTA}
\begin{tabular}{ccccc|cc}
\toprule
& & & \multicolumn{2}{c}{\textit{ADG\textsubscript{R}}} & \multicolumn{2}{c}{\textit{ADGC\textsubscript{R}}} \\
\cmidrule(lr){4-5}\cmidrule(lr){6-7}
\textbf{Seed ratio} ($P${\percent}) & \textbf{OL} & \textbf{OR} & \textbf{OL} ($\uparrow$) & \textbf{OR} ($\downarrow$) & \textbf{OL} ($\uparrow$) & \textbf{OR} ($\downarrow$) \\
\midrule
3 & 0.13 (0.05) & 0.02 (0.01) & 0.74 (0.15) & 0.79 (0.32) & 0.65 (0.27) & 0.64 (0.35) \\
5 & 0.19 (0.07) & 0.03 (0.02) & 0.78 (0.12) & 0.91 (0.34) & 0.69 (0.23) & 0.7 (0.36) \\
10 & 0.29 (0.1) & 0.05 (0.03) & 0.81 (0.09) & 0.95 (0.32) & 0.7 (0.24) & 0.71 (0.34) \\
100 & 0.65 (0.12) & 0.03 (0.1) & 0.84 (0.09) & 1.04 (0.36) & 0.74 (0.24) & 0.78 (0.37) \\
\bottomrule
\end{tabular}
\end{subtable}
\newline
\vspace*{0.15in}
\newline
\begin{subtable}{\linewidth}
\centering
\caption{ISMRM 2015 Tractography Challenge \citep{Maier-Hein:NatureComm:2017} submission data: average values across all methods; ensemble tracking methods (submission identifiers \num{16} through \num{17}); and best overlap method performance.}
\begin{tabular}{ccc}
\toprule
\textbf{Method} & \textbf{OL} ($\uparrow$) & \textbf{OR} ($\downarrow$) \\
\midrule
Mean submissions & 0.36 (0.16) & 0.29 (0.26) \\
\midrule
Ensemble (subm 16\_0) & 0.37 (0.22) & 0.25 (0.2) \\
Ensemble (subm 16\_1) & 0.22 (0.17) & 0.11 (0.11) \\
Ensemble (subm 16\_2) & 0.27 (0.17) & 0.15 (0.12) \\
Ensemble (subm 16\_3) & 0.25 (0.17) & 0.13 (0.11) \\
Ensemble (subm 16\_4) & 0.34 (0.22) & 0.22 (0.18) \\
\midrule
Ensemble (subm 17\_0) & 0.32 (0.2) & 0.22 (0.17) \\
Ensemble (subm 17\_1) & 0.2 (0.15) & 0.12 (0.11) \\
Ensemble (subm 17\_2) & 0.23 (0.17) & 0.12 (0.11) \\
Ensemble (subm 17\_3) & 0.22 (0.16) & 0.13 (0.1) \\
Ensemble (subm 17\_4) & 0.19 (0.15) & 0.12 (0.1) \\
\midrule
Probabilistic; best OL (subm 12\_2) & 0.77 (0.24) & 0.89 (0.44) \\
\bottomrule
\end{tabular}
\end{subtable}
\end{table*}

\begin{figure*}[!htp]
\centering
\includegraphics[scale=0.95, trim=0.0in 0.0in 0.0in 0.0in, clip=true, width=0.5\linewidth, keepaspectratio=true]{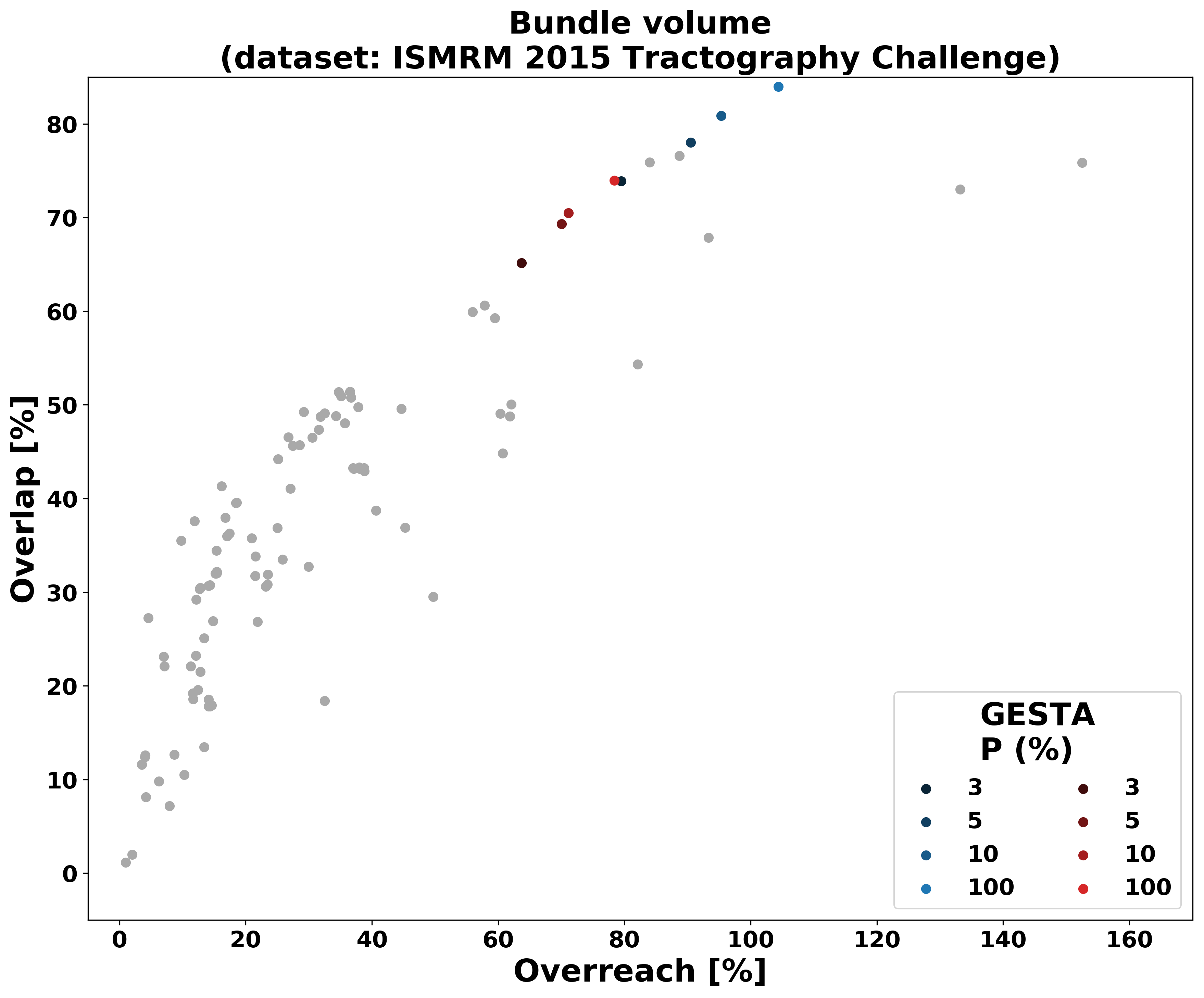} \\
\caption{\label{fig:ismrm2015_challenge_ol_vs_or}Average bundle volume overlap and overreach measures of GESTA compared to all submissions to the ISMRM 2015 Tractography Challenge. Gray color: Challenge submissions; blue shades: generative streamlines assessed with the \textit{ADG\textsubscript{R}} criterion; red shades: generative streamlines assessed with \textit{ADGC\textsubscript{R}} criterion. Note that the values are given as a percentage, as originally reported in \citet{Maier-Hein:NatureComm:2017}.}
\end{figure*}

Figure \ref{fig:ismrm2015_challenge_ol_vs_or} shows the bundle volume performance of GESTA compared to all submissions to the ISMRM 2015 Tractography Challenge in terms of the average bundle overlap and overreach values. Results reveal that GESTA is competitive against the compared methods, even when the more strict connectivity requirement is imposed. However, it also shows an increased overreach with respect to most methods.

\begin{figure*}[!htbp]
\centering
\setlength{\tabcolsep}{0pt}
\begin{tabular}{ccc}
\includegraphics[scale=0.95, trim=2.25in 2.25in 2.25in 3.25in, clip=true, width=0.22\linewidth, keepaspectratio=true]{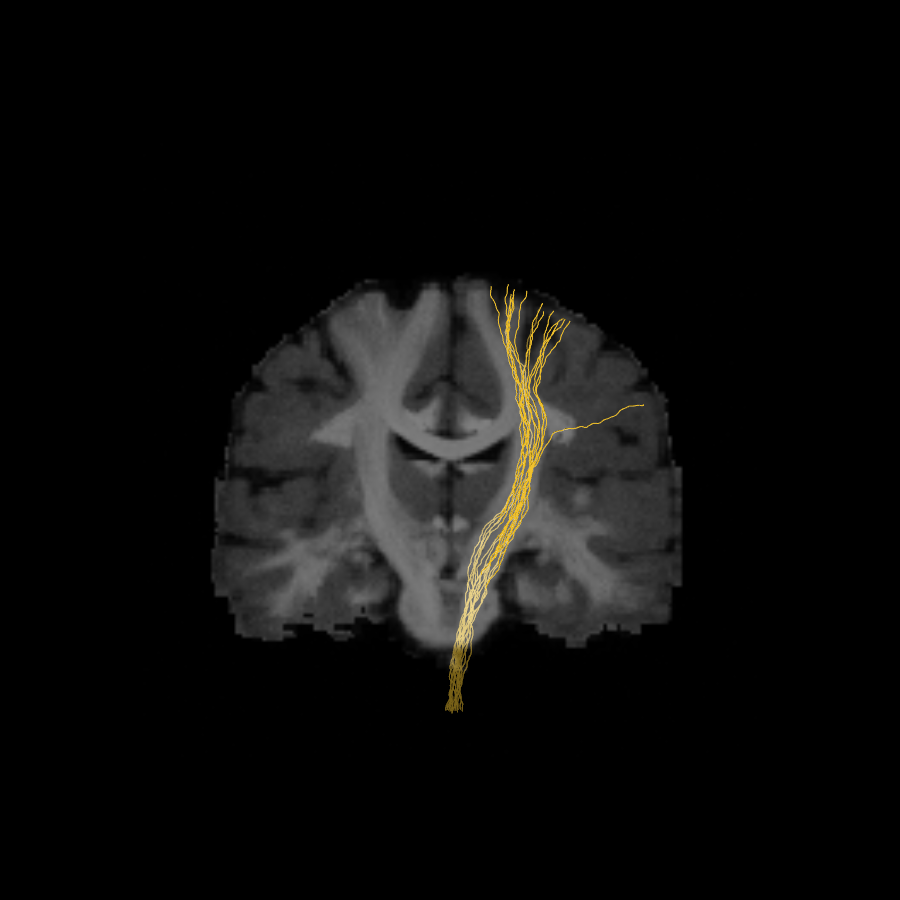} &
\includegraphics[scale=0.95, trim=2.25in 2.25in 2.25in 3.25in, clip=true, width=0.22\linewidth, keepaspectratio=true]{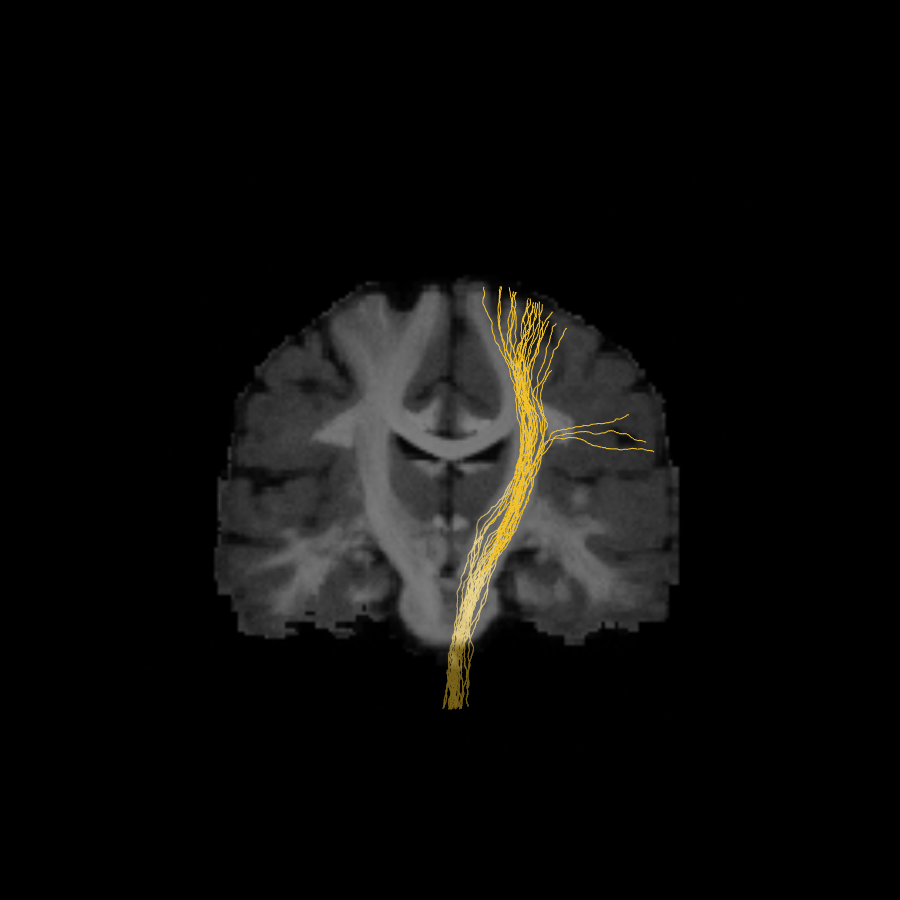} &
\includegraphics[scale=0.95, trim=2.25in 2.25in 2.25in 3.25in, clip=true, width=0.22\linewidth, keepaspectratio=true]{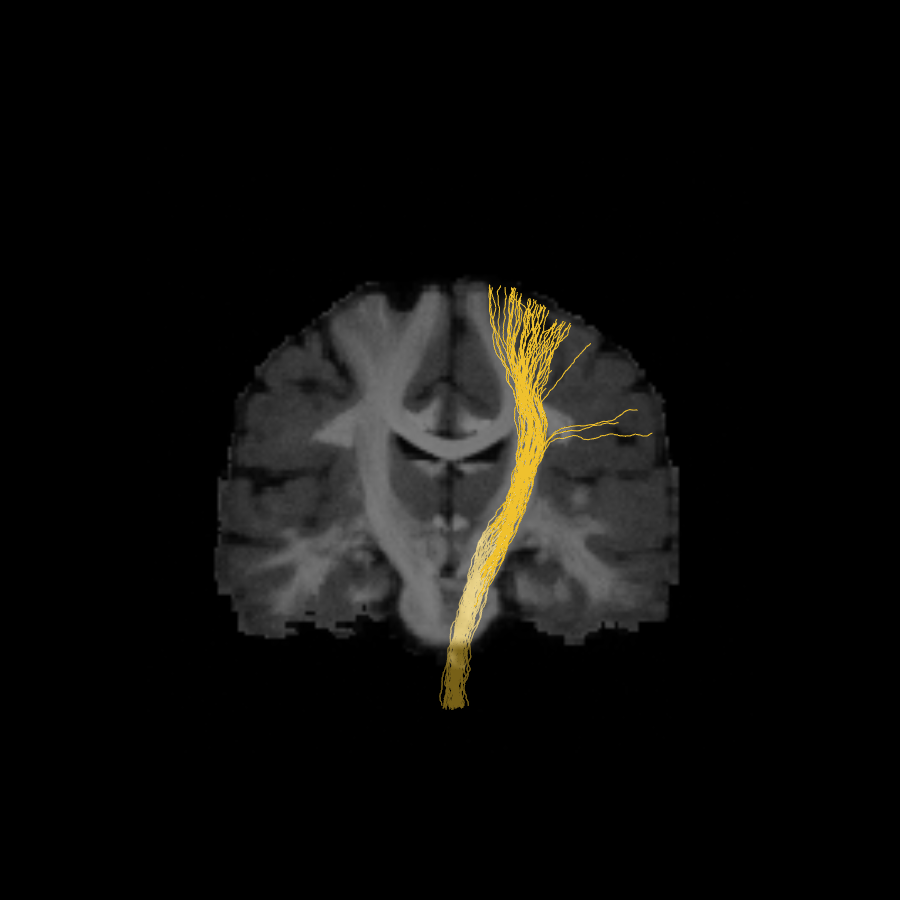} \\
\includegraphics[scale=0.95, trim=2.25in 2.25in 2.25in 3.25in, clip=true, width=0.22\linewidth, keepaspectratio=true]{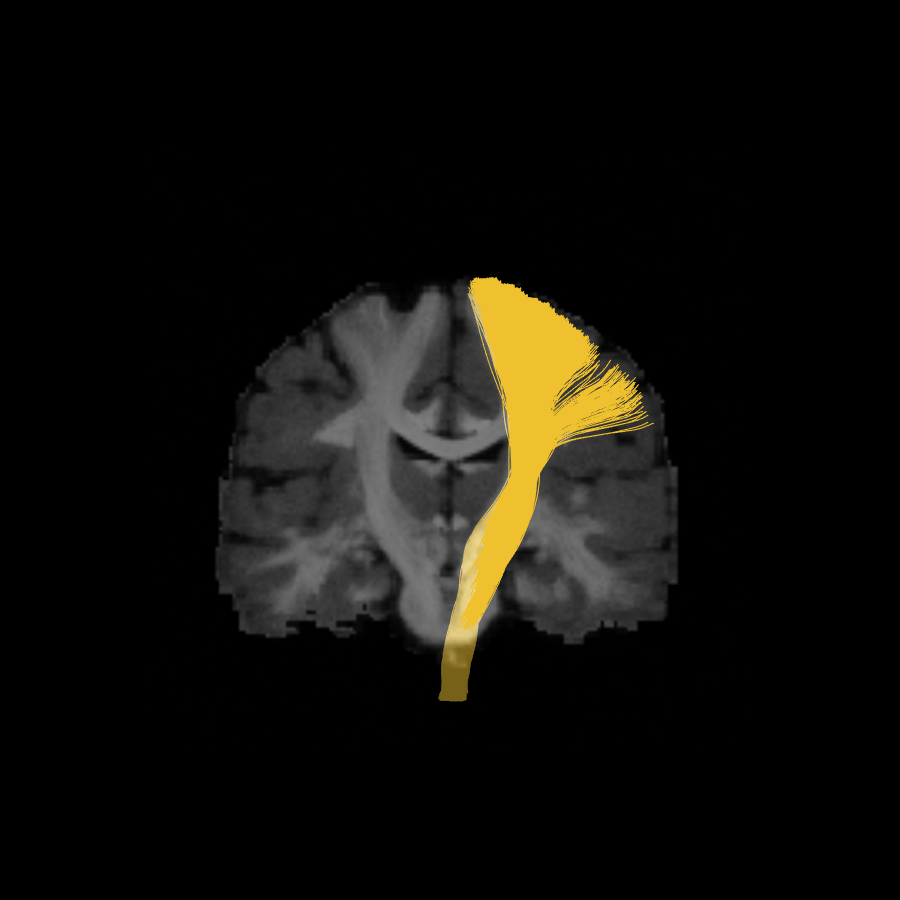} &
\includegraphics[scale=0.95, trim=2.25in 2.25in 2.25in 3.25in, clip=true, width=0.22\linewidth, keepaspectratio=true]{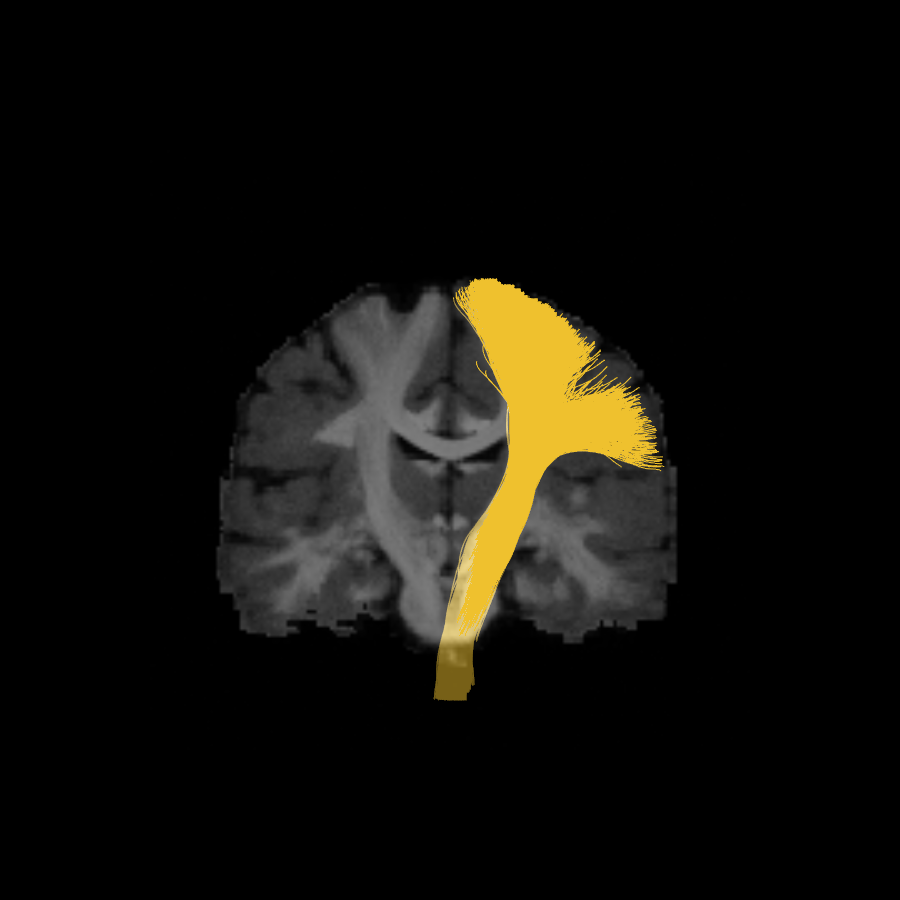} &
\includegraphics[scale=0.95, trim=2.25in 2.25in 2.25in 3.25in, clip=true, width=0.22\linewidth, keepaspectratio=true]{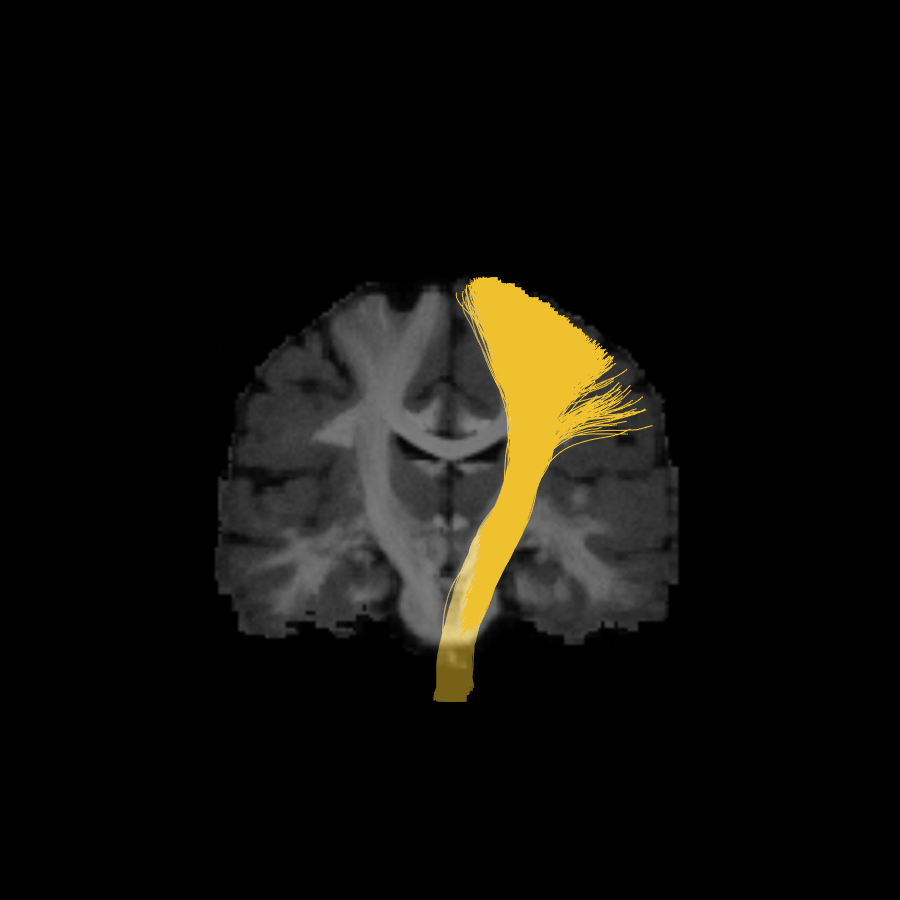} \\
\includegraphics[scale=0.95, trim=2.25in 2.25in 2.25in 3.25in, clip=true, width=0.22\linewidth, keepaspectratio=true]{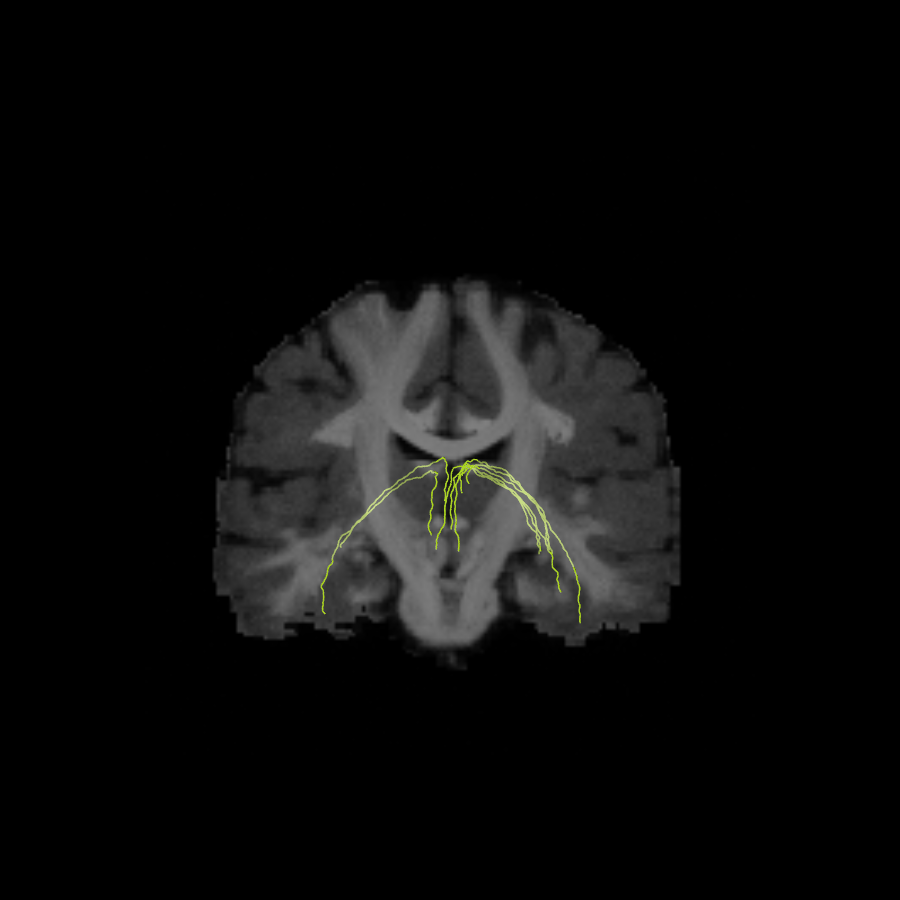} &
\includegraphics[scale=0.95, trim=2.25in 2.25in 2.25in 3.25in, clip=true, width=0.22\linewidth, keepaspectratio=true]{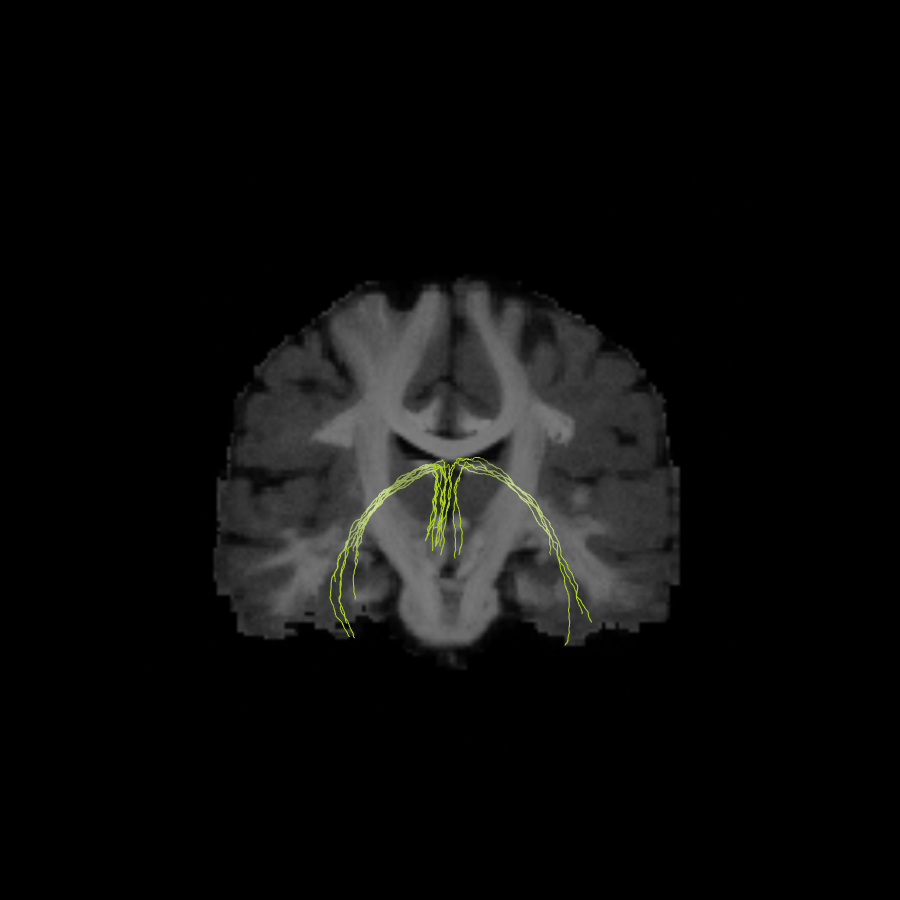} &
\includegraphics[scale=0.95, trim=2.25in 2.25in 2.25in 3.25in, clip=true, width=0.22\linewidth, keepaspectratio=true]{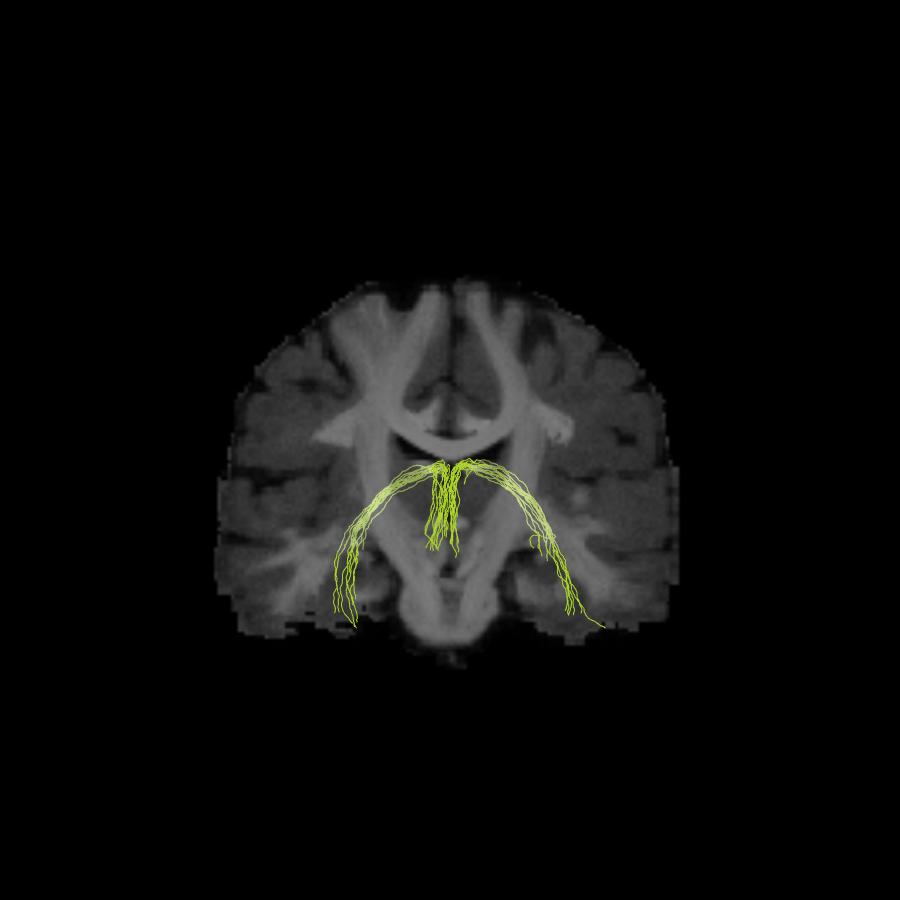} \\
\includegraphics[scale=0.95, trim=2.25in 2.25in 2.25in 3.25in, clip=true, width=0.22\linewidth, keepaspectratio=true]{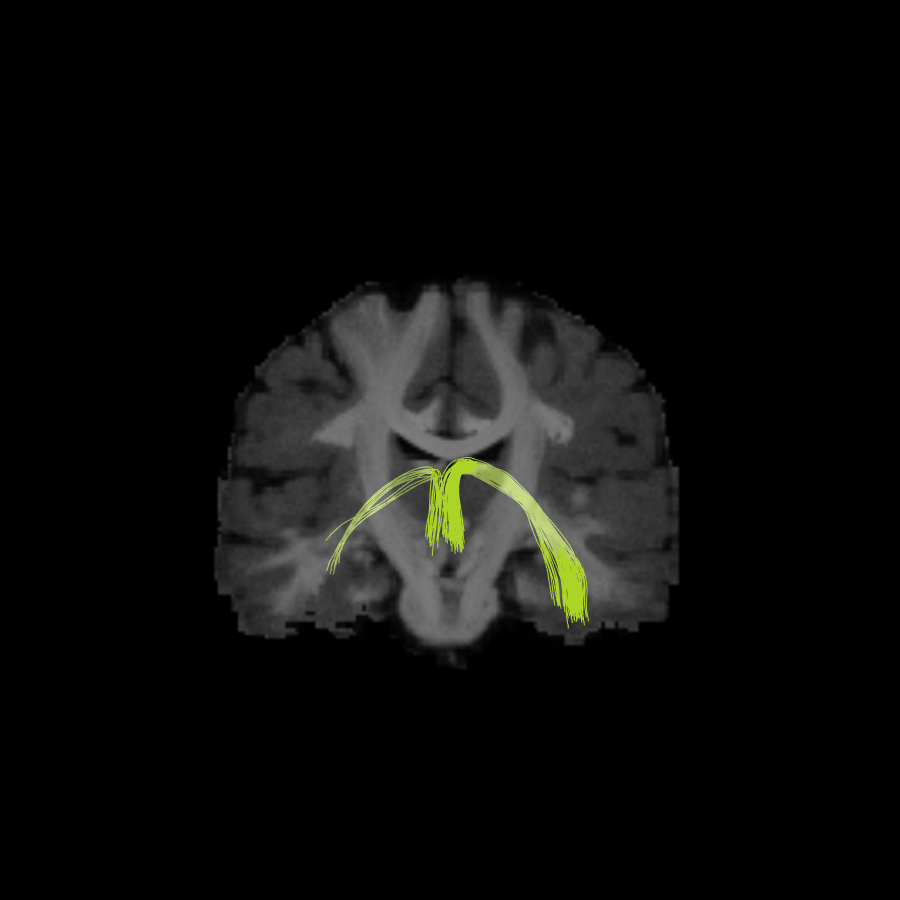} &
\includegraphics[scale=0.95, trim=2.25in 2.25in 2.25in 3.25in, clip=true, width=0.22\linewidth, keepaspectratio=true]{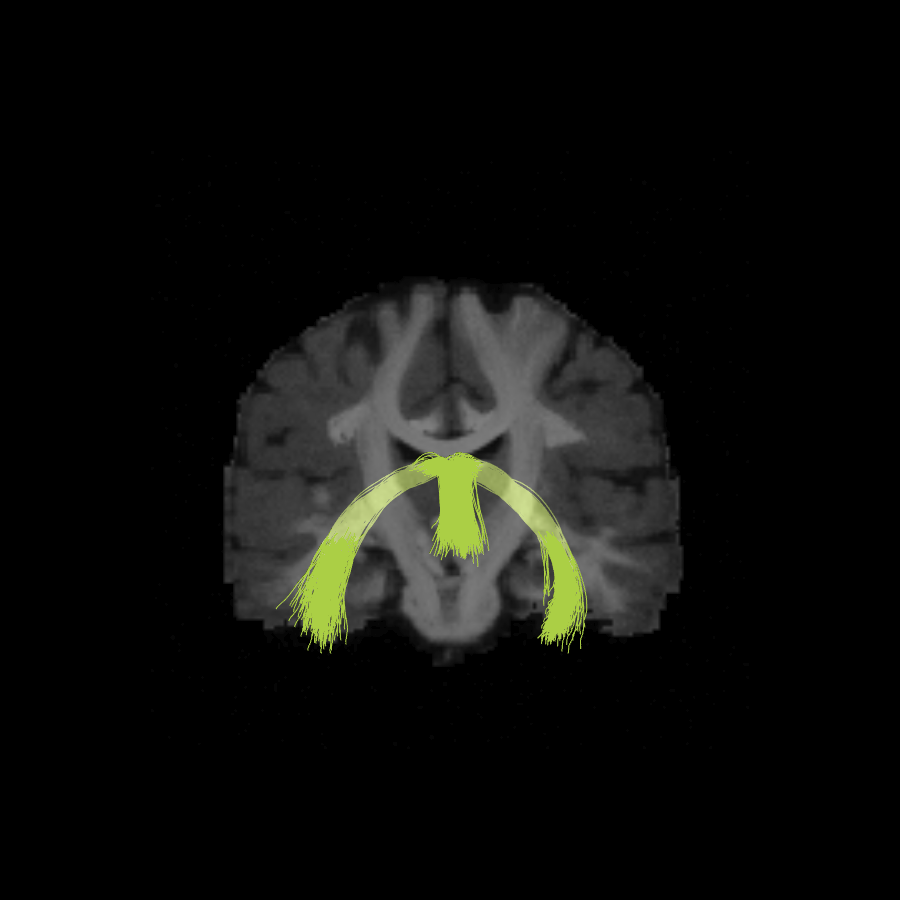} &
\includegraphics[scale=0.95, trim=2.25in 2.25in 2.25in 3.25in, clip=true, width=0.22\linewidth, keepaspectratio=true]{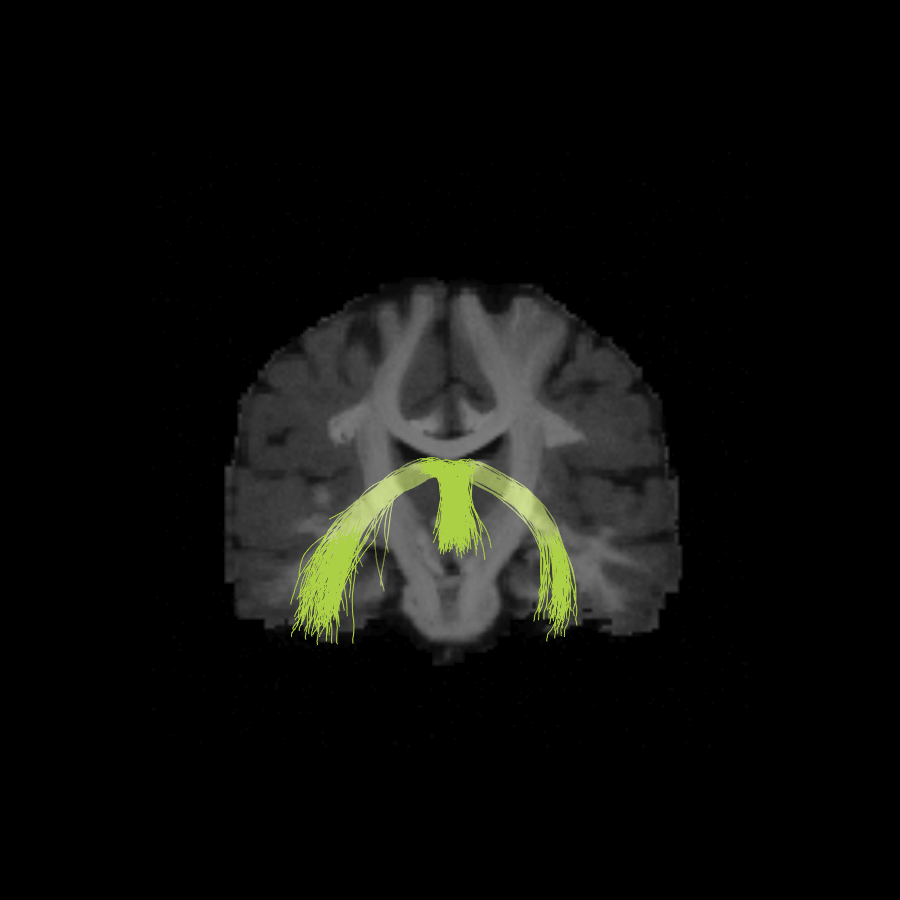} \\
\includegraphics[scale=0.95, trim=2.25in 2.25in 2.25in 3.25in, clip=true, width=0.22\linewidth, keepaspectratio=true]{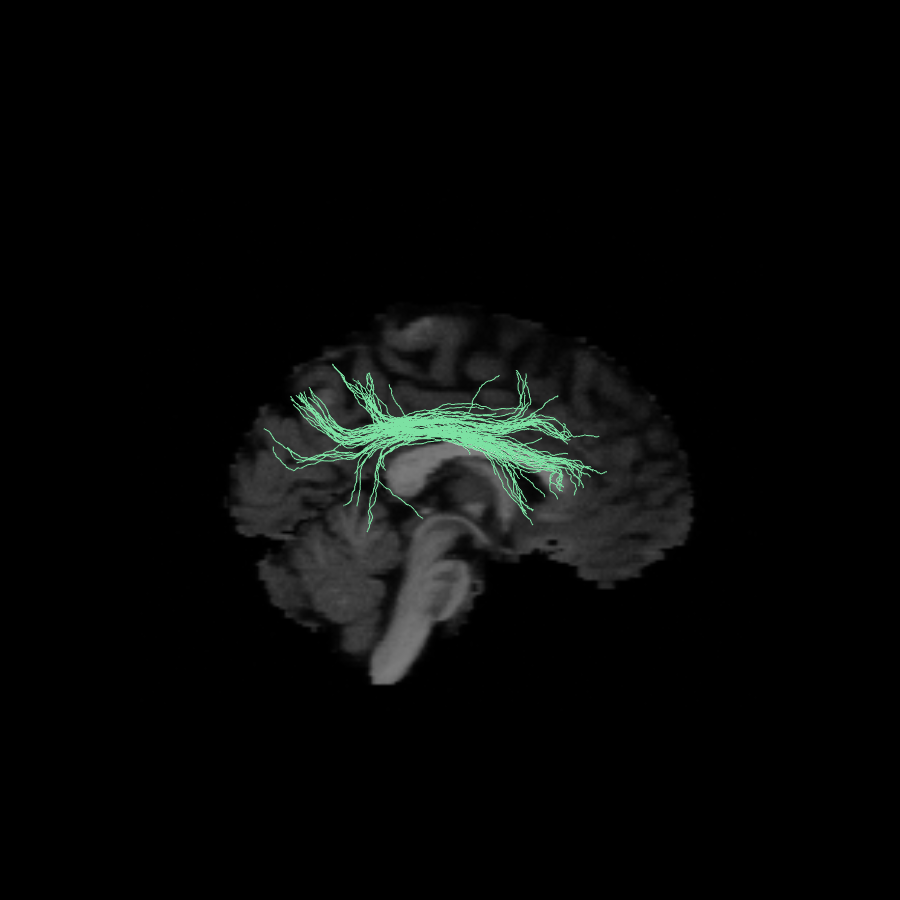} &
\includegraphics[scale=0.95, trim=2.25in 2.25in 2.25in 3.25in, clip=true, width=0.22\linewidth, keepaspectratio=true]{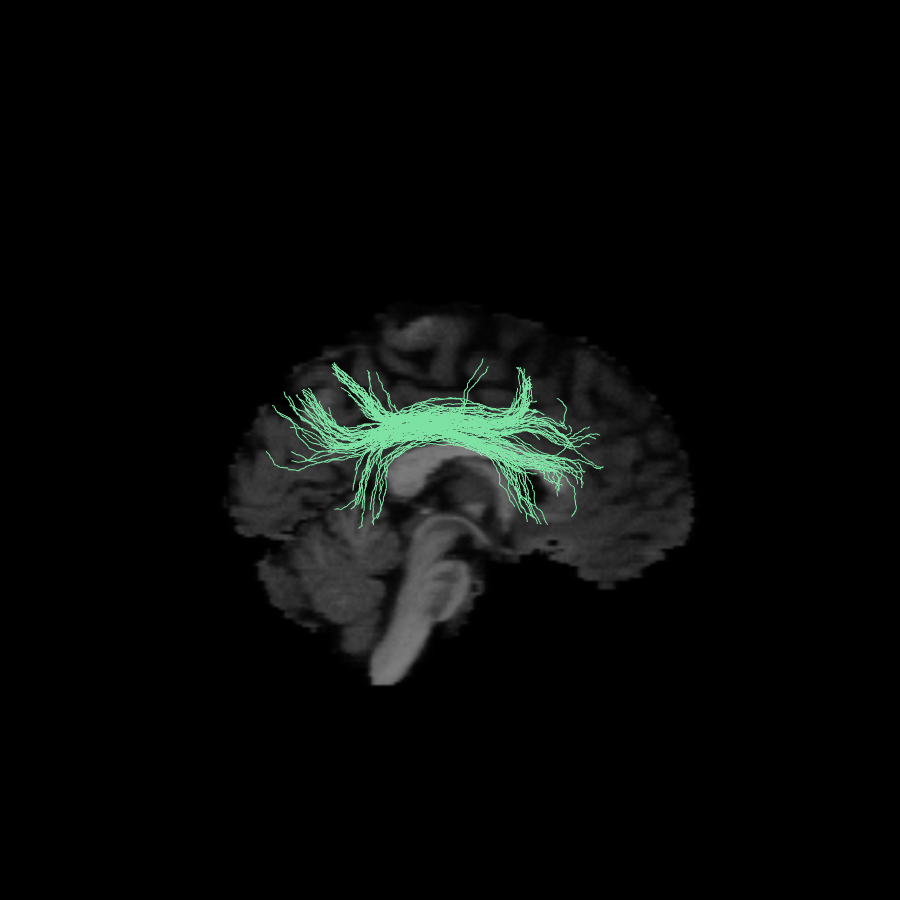} &
\includegraphics[scale=0.95, trim=2.25in 2.25in 2.25in 3.25in, clip=true, width=0.22\linewidth, keepaspectratio=true]{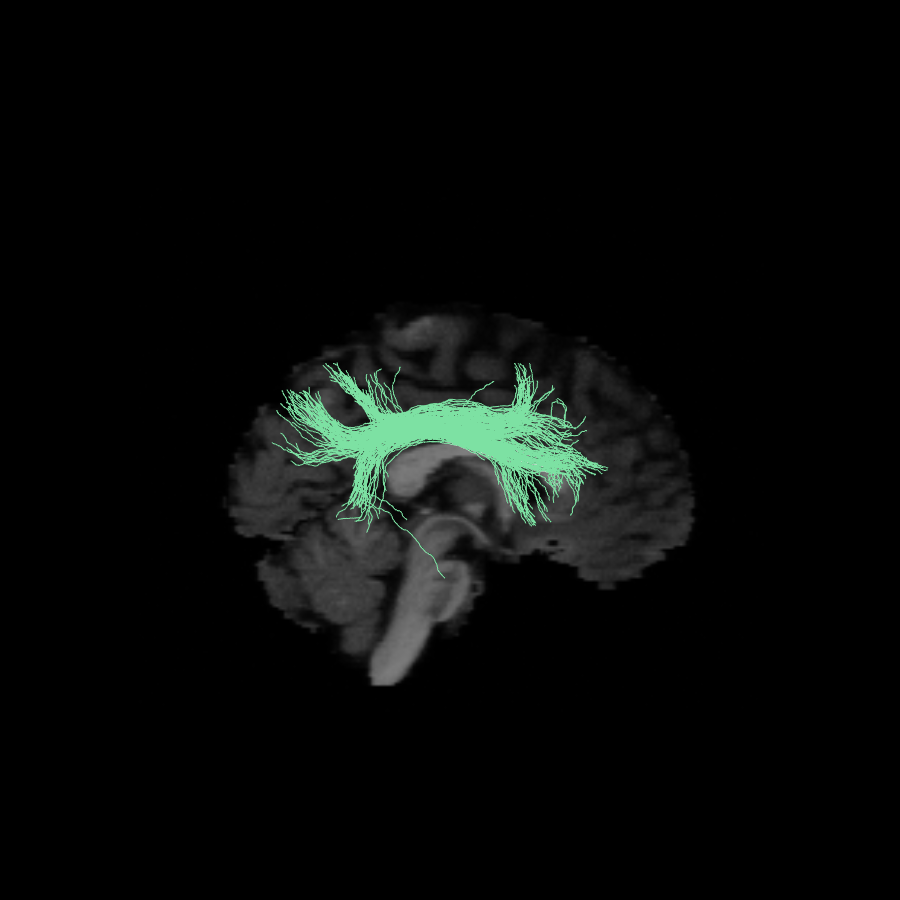} \\
\includegraphics[scale=0.95, trim=2.25in 2.25in 2.25in 3.25in, clip=true, width=0.22\linewidth, keepaspectratio=true]{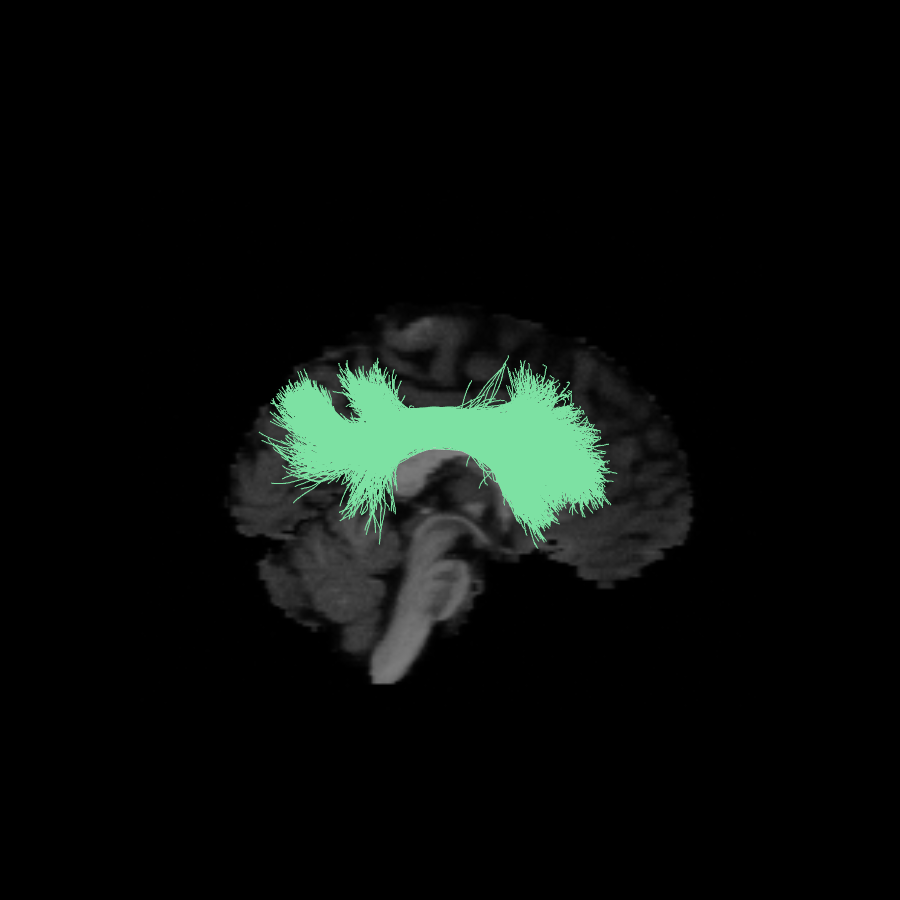} &
\includegraphics[scale=0.95, trim=2.25in 2.25in 2.25in 3.25in, clip=true, width=0.22\linewidth, keepaspectratio=true]{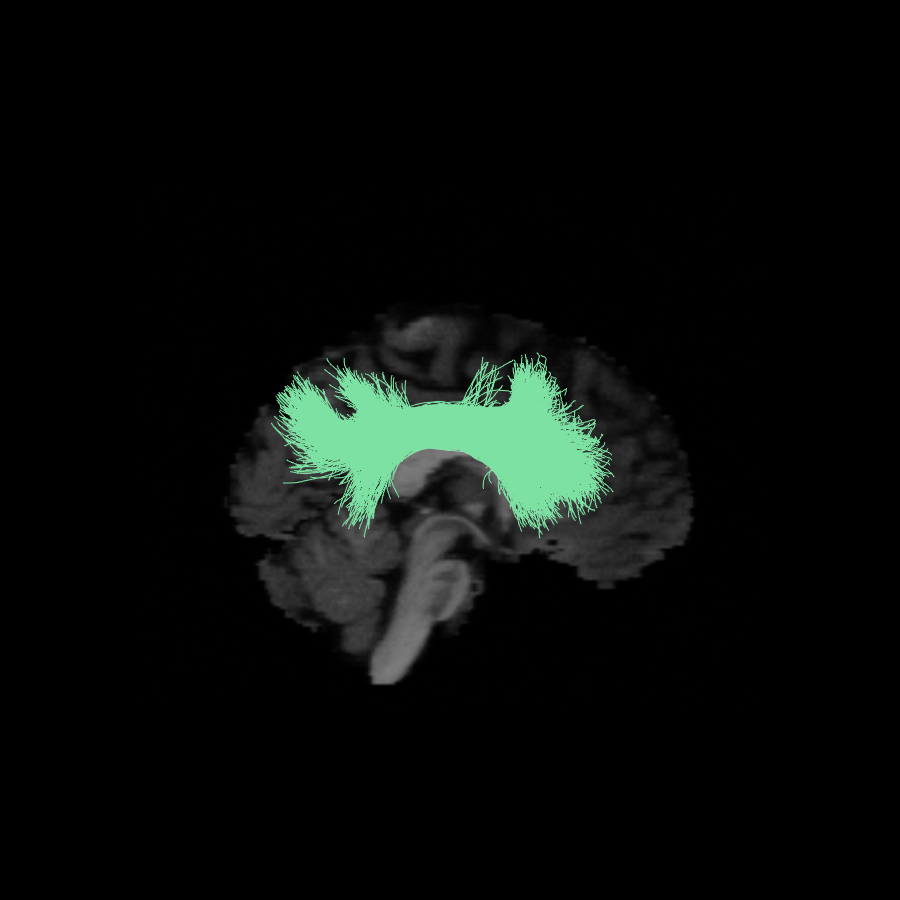} &
\includegraphics[scale=0.95, trim=2.25in 2.25in 2.25in 3.25in, clip=true, width=0.22\linewidth, keepaspectratio=true]{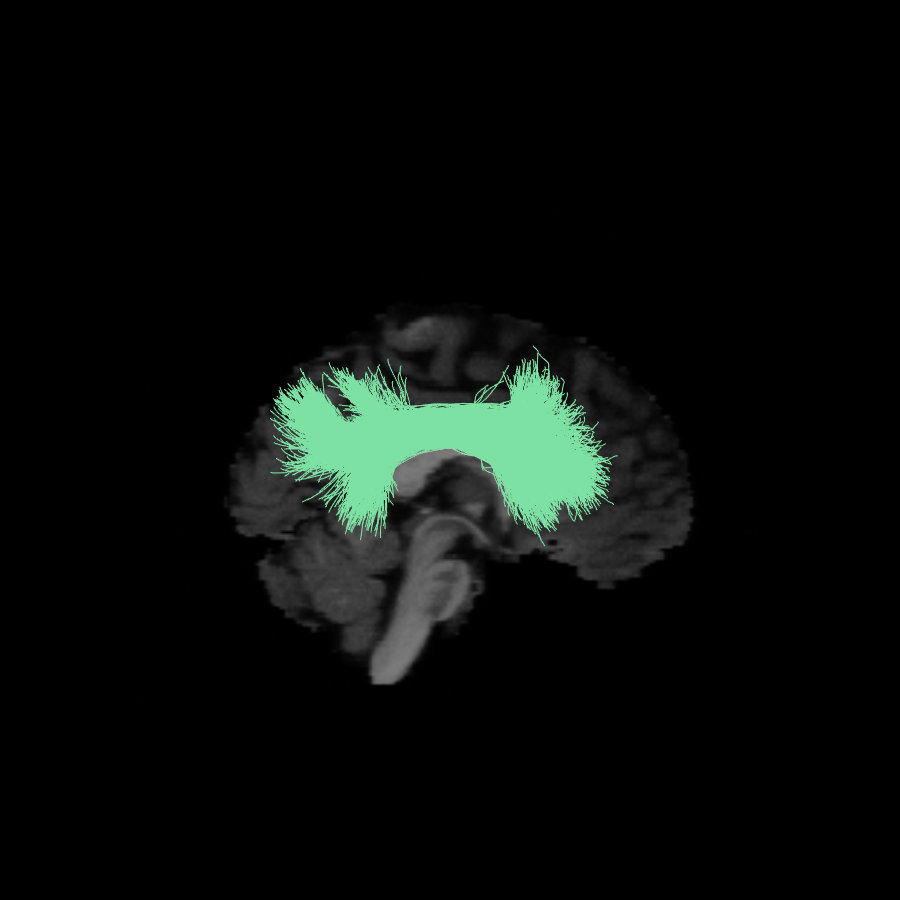} \\
\textbf{(a)} & \textbf{(b)} & \textbf{(c)} \\
\end{tabular}
\caption{\label{fig:ismrm2015_generative_streamlines}Generative tractography applied to the ISMRM 2015 Tractography Challenge dataset left CST (yellow orange color), fornix (lime), and right SLF (sea green) bundles. For each bundle: top: seed streamlines; bottom: latent-generated plausible streamlines. Coronal anterior (left CST and fornix) and sagittal right (right SLF) views. The seed streamlines represent $P = 3${\percent} (a), $P = 5${\percent} (b), and $P = 10${\percent} (c) of the test set streamlines for each bundle, and streamlines are evaluated according to the \textit{ADGC\textsubscript{R}} criterion.}
\end{figure*}

Figure \ref{fig:ismrm2015_generative_streamlines} shows the generative tractography process applied to the left CST, fornix and right SLF bundles of the ISMRM 2015 Tractography Challenge dataset for different seed streamline ratio values (from left to right, $P = $\{\numlist[list-separator={,},list-final-separator={,}]{3;5;10}\}{\percent}). According to the ISMRM 2015 Tractography Challenge results \citep{Maier-Hein:NatureComm:2017}, the right SLF bundle reconstruction was one whose difficulty was medium, whereas the left CST and fornix were considered to be hard. The recovered spatial coverage has considerably improved already using as little as $P = 3${\percent} of the available streamlines in each bundle's test set. As more seed streamlines are available, a better coverage can be expected. However, as the seeds are added randomly in our experiments, the underlying data distribution might be shifted, and the sampling method might preferentially draw samples around some given seed streamlines.

\begin{figure*}[!htp]
\centering
\setlength{\tabcolsep}{0pt}
\begin{tabular}{ccccc}
\includegraphics[scale=0.95, trim=4.5in 2.0in 0.5in 3.25in, clip=true, width=0.225\linewidth, keepaspectratio=true]{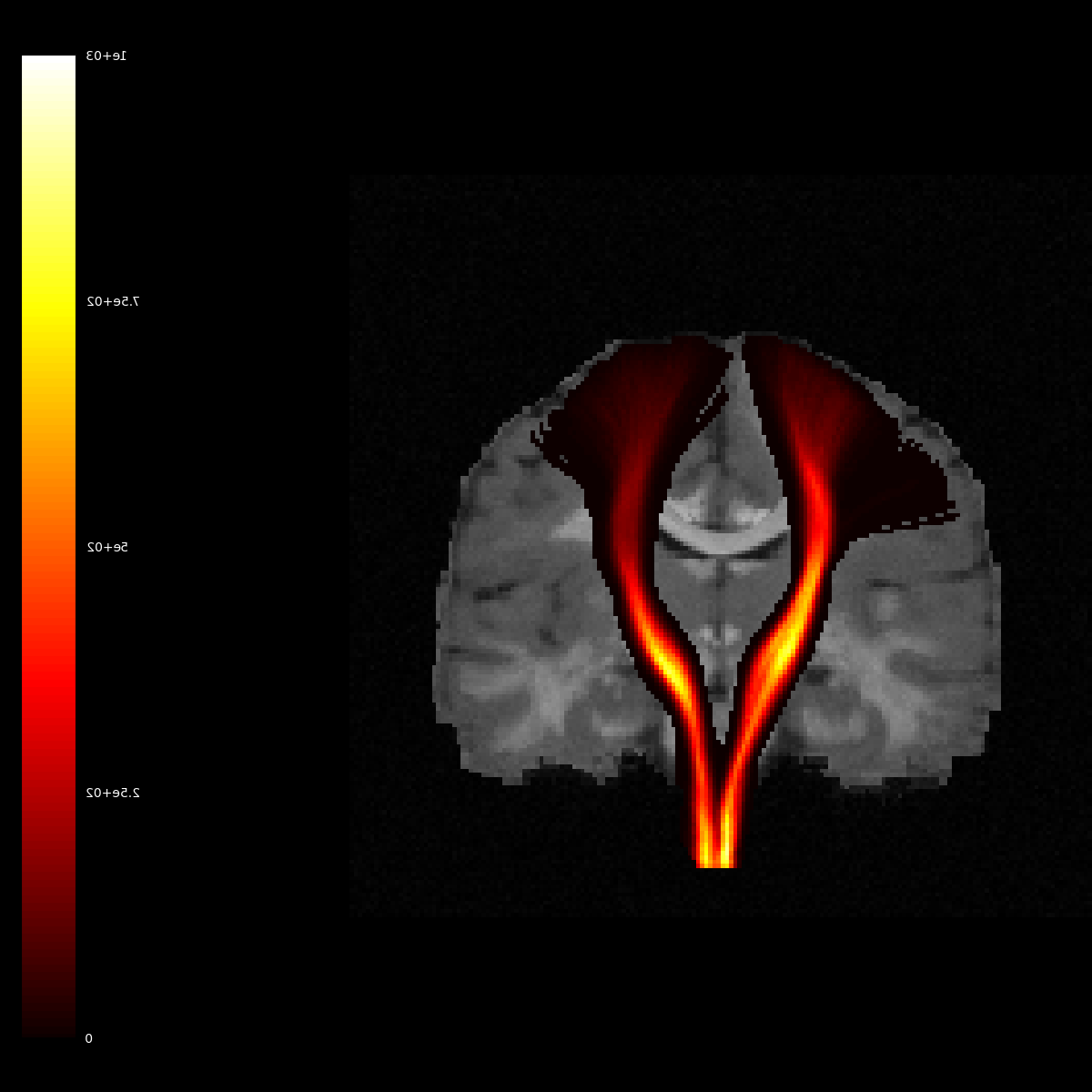} &
\includegraphics[scale=0.95, trim=4.5in 2.0in 0.5in 3.25in, clip=true, width=0.225\linewidth, keepaspectratio=true]{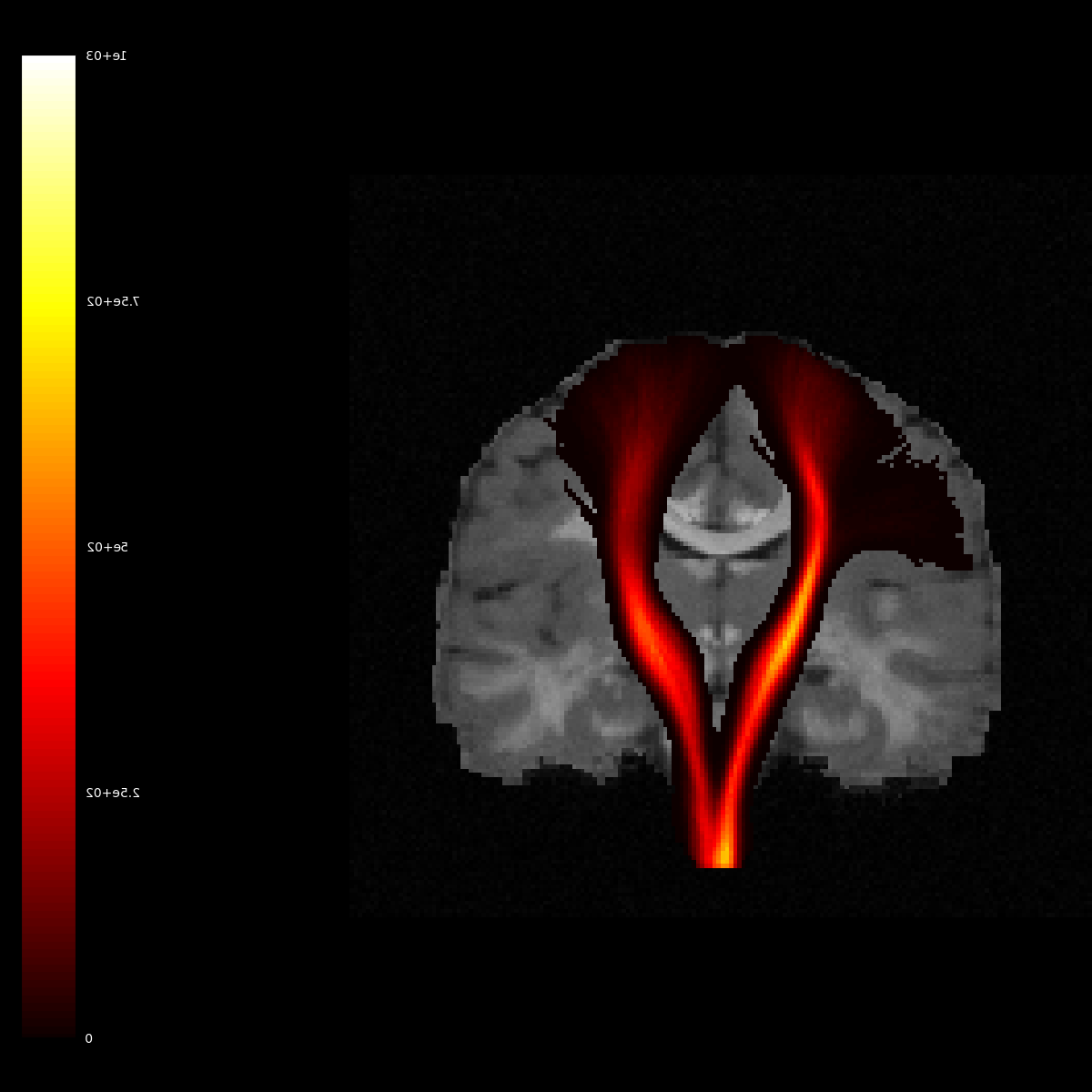} &
\includegraphics[scale=0.95, trim=4.5in 2.0in 0.5in 3.25in, clip=true, width=0.225\linewidth, keepaspectratio=true]{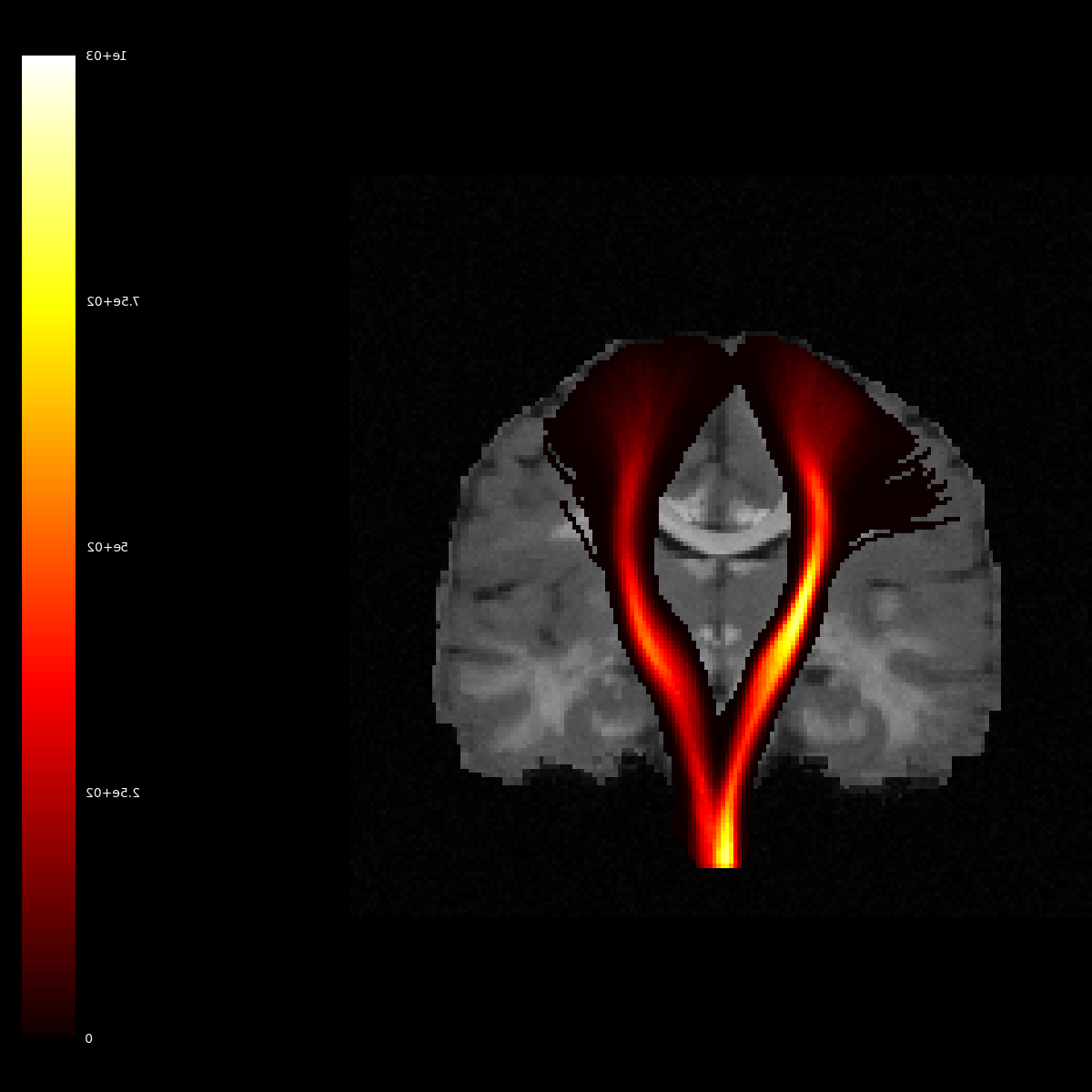} &
\includegraphics[scale=0.95, trim=4.5in 2.0in 0.5in 3.25in, clip=true, width=0.225\linewidth, keepaspectratio=true]{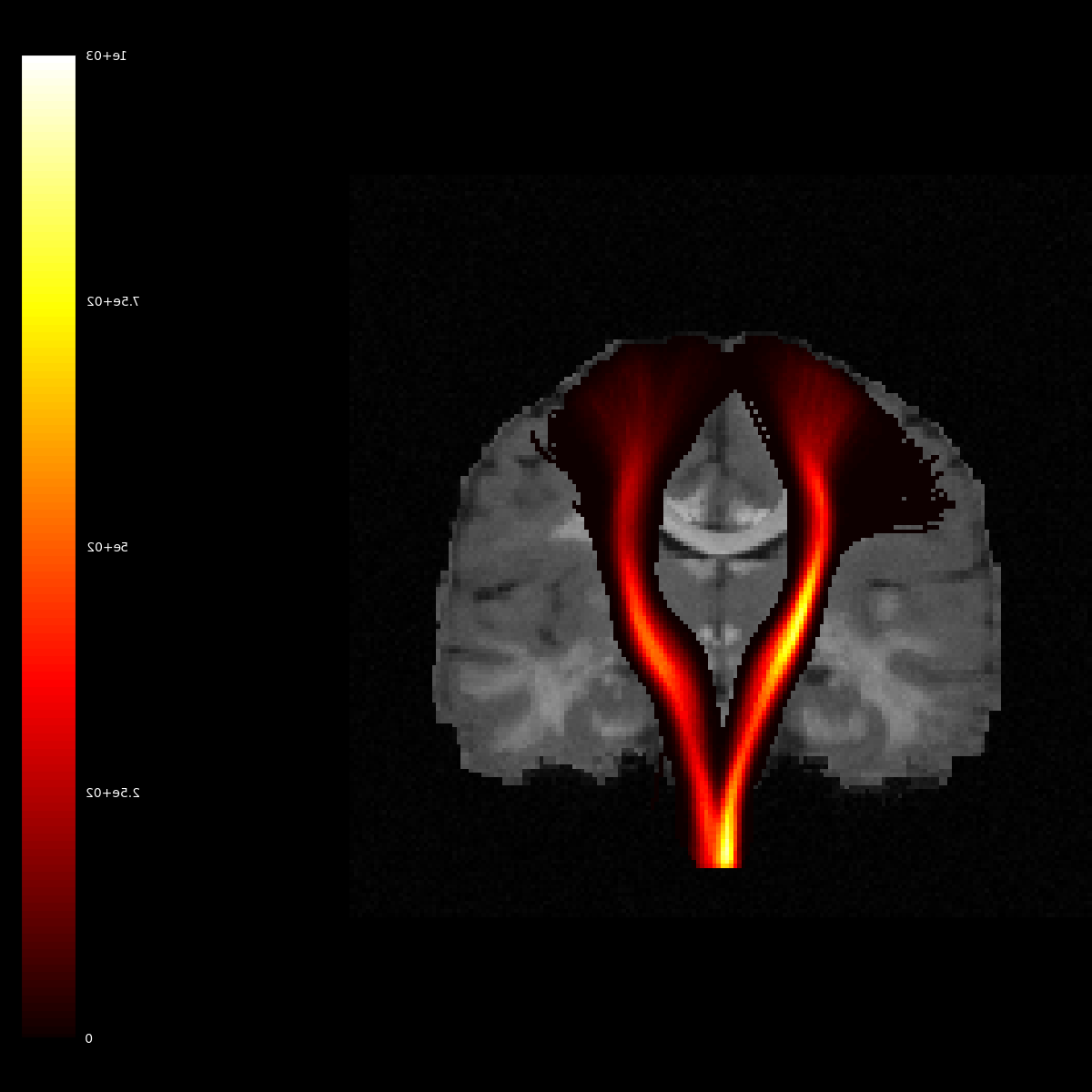} &
\includegraphics[scale=0.95, trim=9.25in 0.25in 0.0in 0.25in, clip=true, width=0.05875\linewidth, keepaspectratio=true]{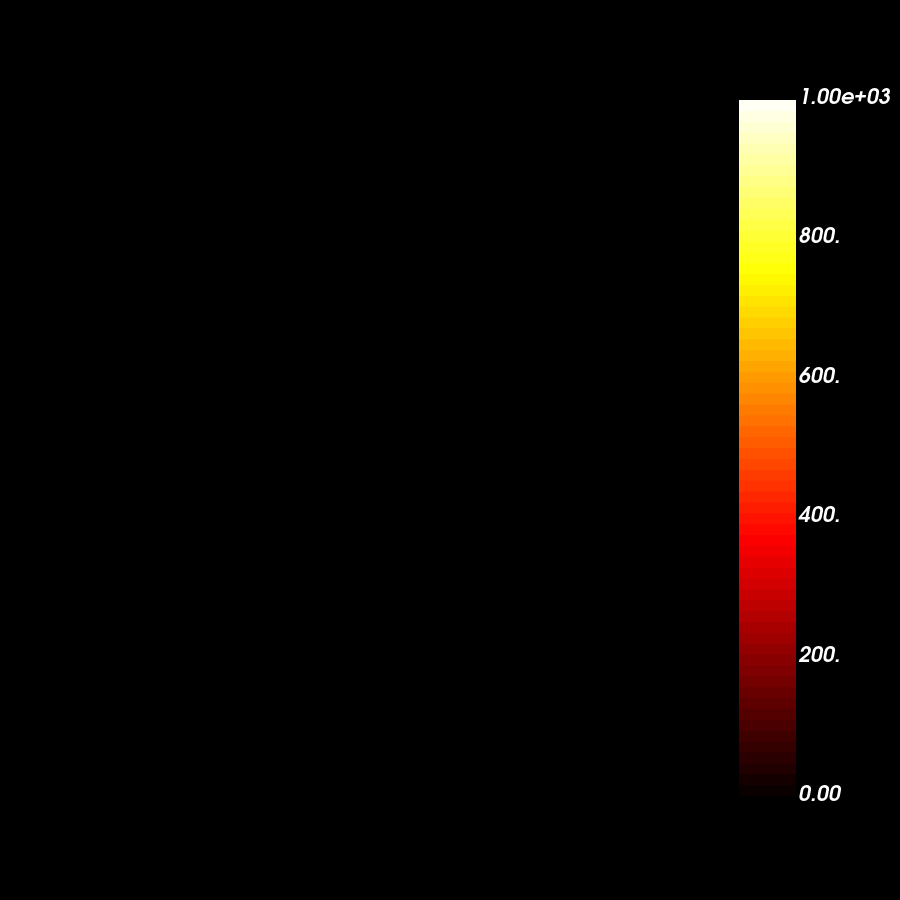} \\
\textbf{(a)} & \textbf{(b)} & \textbf{(c)} & \textbf{(d)} \\
\end{tabular}
\caption{\label{fig:ismrm2015_generative_track_density}Track density on the ISMRM 2015 Tractography Challenge dataset left/right CST bundle generative streamlines. All coronal anterior views. The seed streamlines represent $P = 3${\percent} (a), $P = 5${\percent} (b), $P = 10${\percent} (c); and $P = 100${\percent} (d) of the test set streamlines; and streamlines are evaluated according to the \textit{ADGC\textsubscript{R}} criterion.}
\end{figure*}

Figure \ref{fig:ismrm2015_generative_track_density} shows the track density images on the ISMRM 2015 Tractography Challenge dataset left/right CST bundles generative streamlines for the $P = \;$\{\numlist[list-separator={,},list-final-separator={,}]{3;5;10;100}\}{\percent} seed streamline ratio values, and evaluated according to the \textit{ADGC\textsubscript{R}} criterion. As the seed count increases, results show (especially across the $P = \;$\{\numlist[list-pair-separator={,}]{3;5}\}{\percent} and $P = \;$\{\numlist[list-pair-separator={,}]{10;100}\}{\percent} pairs), that the sampling procedure can generate streamlines over a larger area, \ie the streamline density can be balanced across the bundle.

\begin{figure*}[!htp]
\centering
\setlength{\tabcolsep}{0pt}
\begin{tabular}{ccc}
\includegraphics[scale=0.95, trim=2.25in 2.75in 2.25in 2.75in, clip=true, width=0.325\linewidth, keepaspectratio=true]{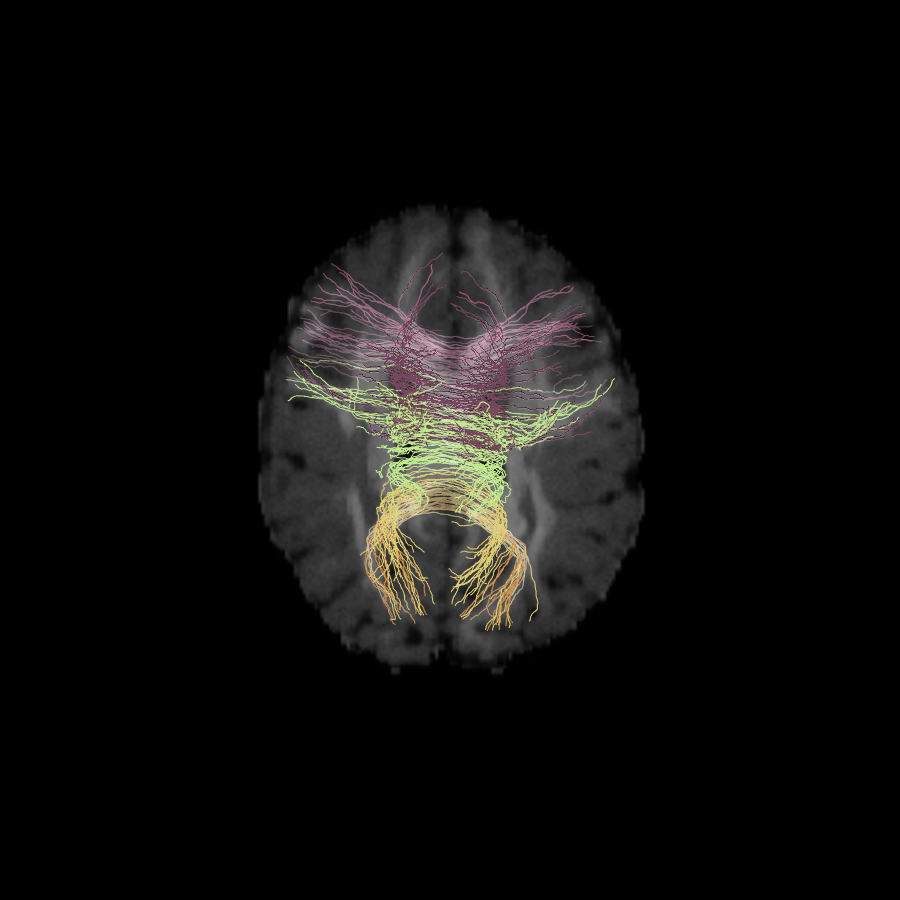} &
\includegraphics[scale=0.95, trim=2.25in 2.25in 2.25in 3.25in, clip=true, width=0.325\linewidth, keepaspectratio=true]{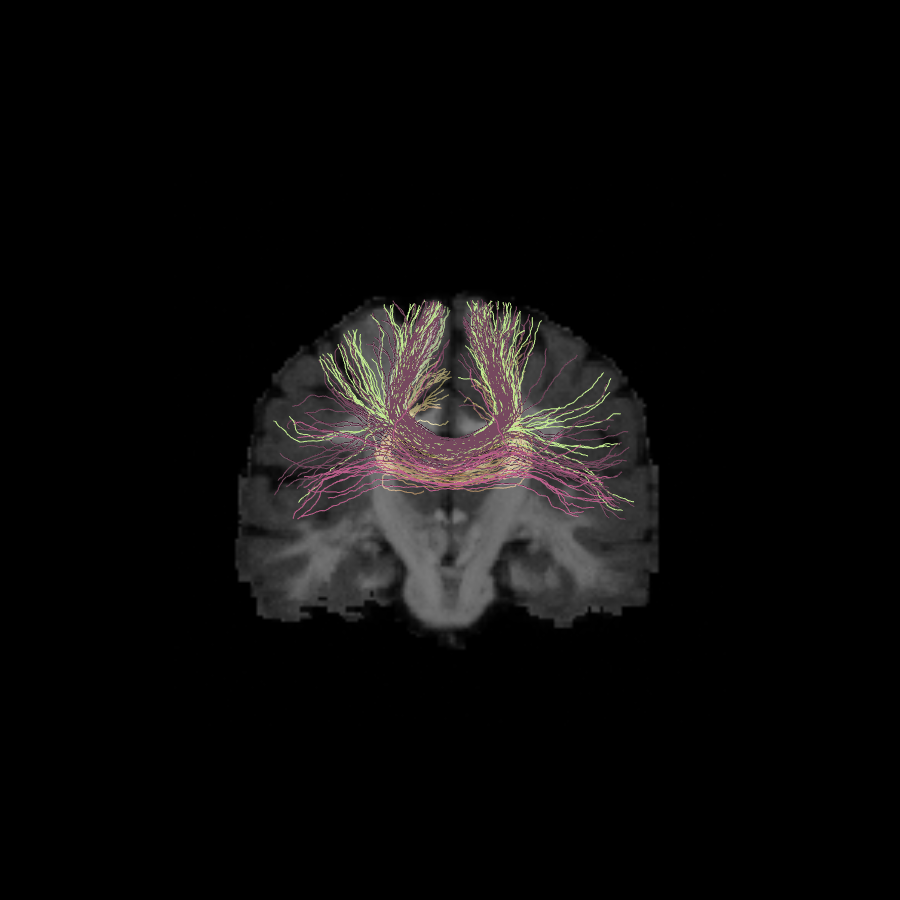} &
\includegraphics[scale=0.95, trim=2.25in 2.25in 2.25in 3.25in, clip=true, width=0.325\linewidth, keepaspectratio=true]{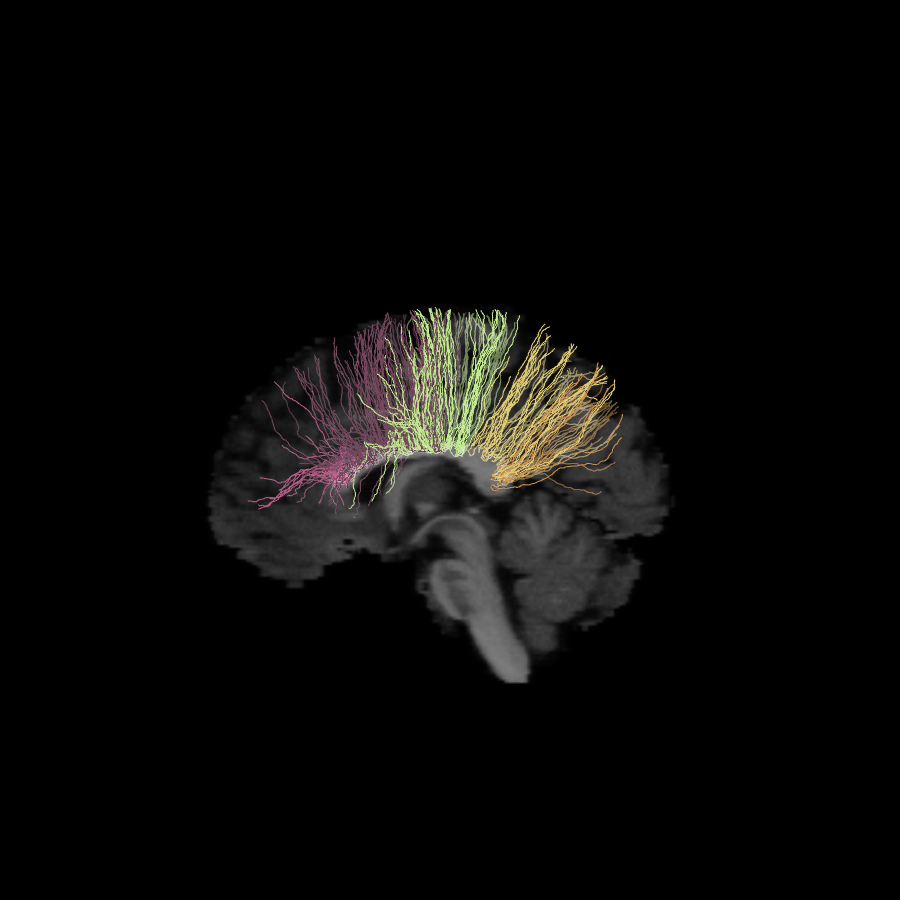} \\
\includegraphics[scale=0.95, trim=2.25in 2.75in 2.25in 2.75in, clip=true, width=0.325\linewidth, keepaspectratio=true]{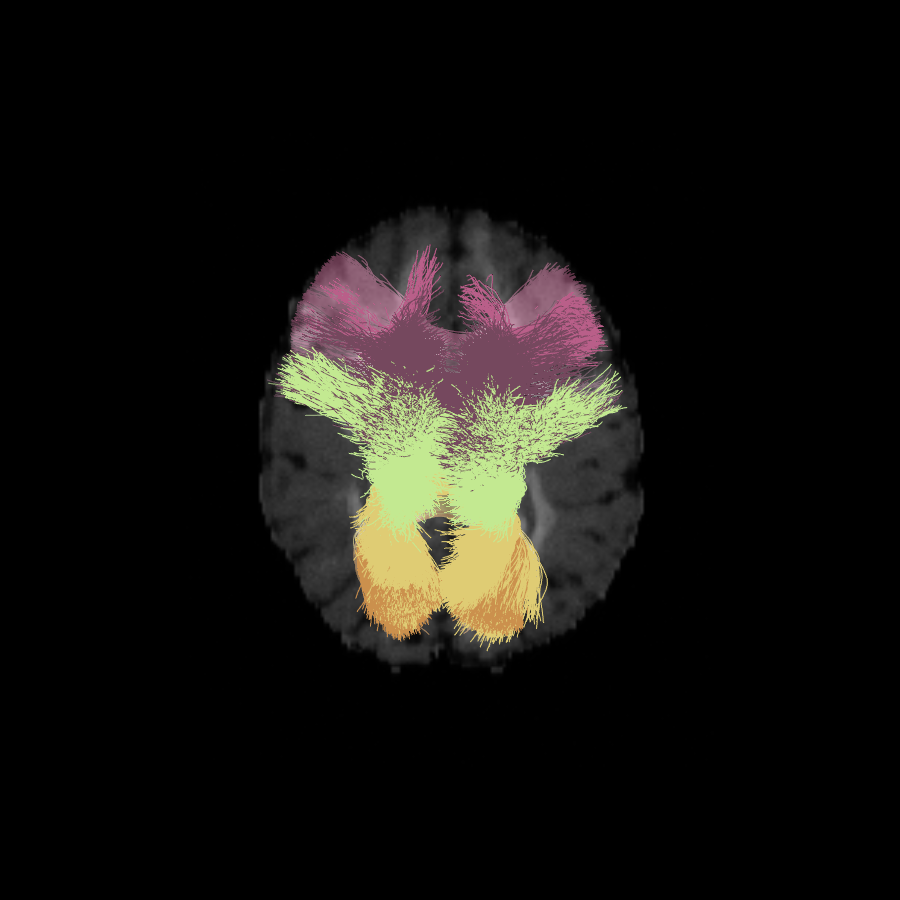} &
\includegraphics[scale=0.95, trim=2.25in 2.25in 2.25in 3.25in, clip=true, width=0.325\linewidth, keepaspectratio=true]{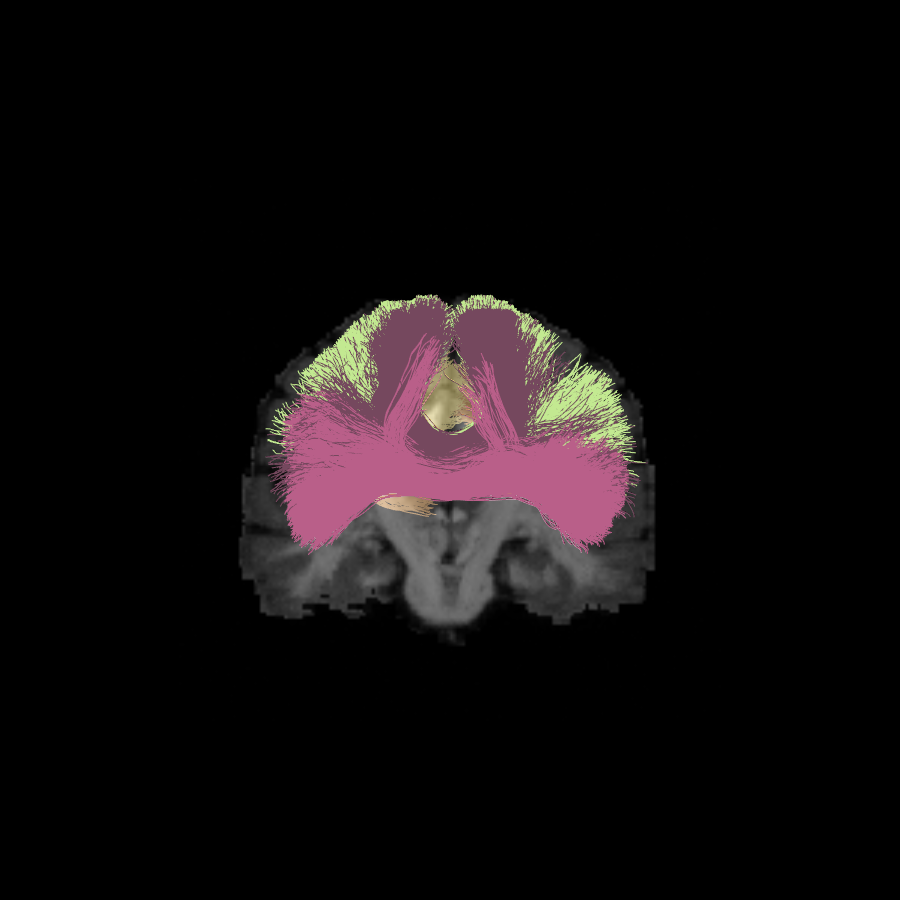} &
\includegraphics[scale=0.95, trim=2.25in 2.25in 2.25in 3.25in, clip=true, width=0.325\linewidth, keepaspectratio=true]{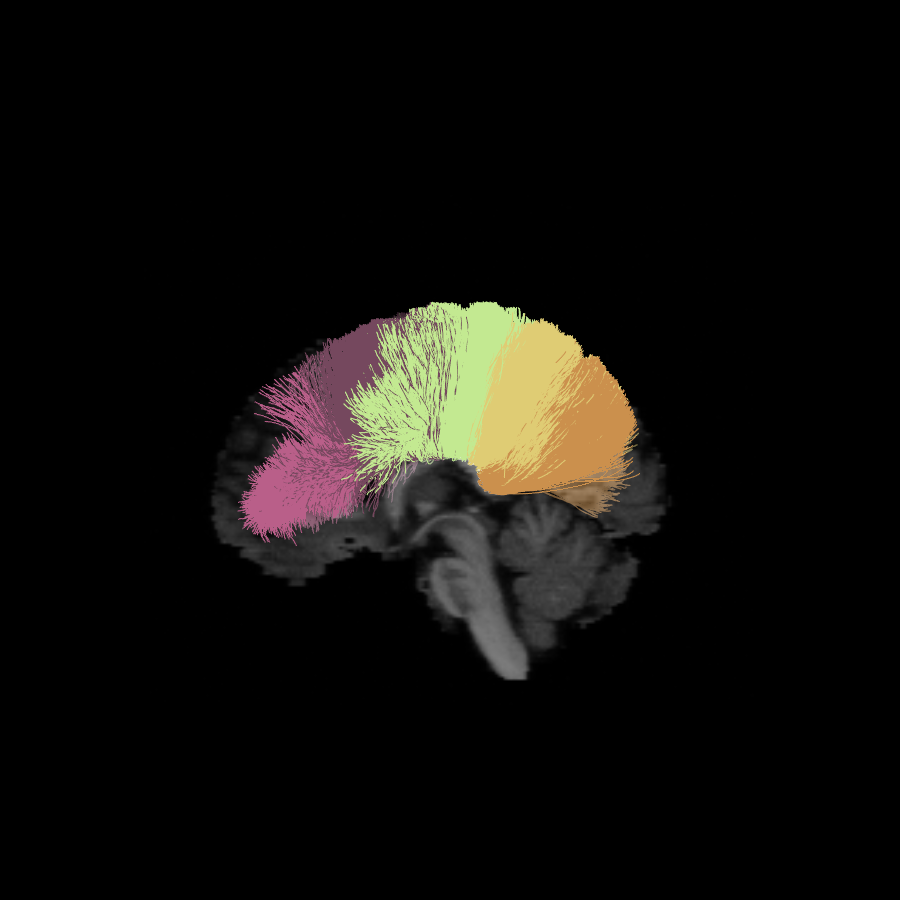} \\
\textbf{(a)} & \textbf{(b)} & \textbf{(c)} \\
\end{tabular}
\caption{\label{fig:ismrm2015_generative_bundle_cc}Generative tractography applied to the ISMRM 2015 Tractography Challenge dataset corpus callosum. Top row: seed streamlines; bottom row: latent-generated plausible streamlines. (a) Axial superior; (b) coronal anterior; (c) sagittal left views. The seed streamlines represent $P = 3${\percent} of the corpus callosum test set streamlines, and streamlines are evaluated using the \textit{ADGC\textsubscript{R}} criterion. Atlas region to color code: CC\_Fr\_1: pink; CC\_Fr\_2: purple; CC\_Pr\_Po: yellow green; CC\_Pa: sand; CC\_Oc: orange.}
\end{figure*}

Figure \ref{fig:ismrm2015_generative_bundle_cc} shows the result of generating streamlines in the latent space for the ISMRM 2015 Tractography Challenge dataset corpus callosum. Using $P = 3${\percent} of the available seeds for each segment, GESTA is able to successfully recover the potential white matter pathways of interest.

\clearpage
\newpage

\subsection{BIL\&GIN callosal homotopic}
\label{subsec:bil_gin_homotopic_cc_results}

Table \ref{tab:bil_gin_cc_homotopic_generative_tractography_measures} shows the volume of the seed streamlines and the generative streamlines for the BIL\&GIN callosal homotopic dataset. GESTA produces new streamlines that successfully increase the volume coverage using a limited set of streamlines (\num{3.9} times on average across all segments for the \textit{ADG\textsubscript{R}} criterion; \num{3.7} times for the \textit{ADGC\textsubscript{R}} criterion).

\begin{table*}[!htbp]
\caption{\label{tab:bil_gin_cc_homotopic_generative_tractography_measures}BIL\&GIN callosal homotopic dataset seed streamlines' and generative streamlines' volume measures (mean and standard deviation). Measures are averaged over all available subjects and segments. All segments having at least one streamline are included in the computation of the seed streamline volume measure; however, as mentioned in section \ref{subsec:limitations} segments that have only one streamline cannot be effectively sampled, and hence, are not considered in the generative framework. All measures are in \si{\milli\metre\cubed}.}
\centering
\begin{tabular}{ccc|c}
\toprule
& & \textit{ADG\textsubscript{R}} & \textit{ADGC\textsubscript{R}} \\
\cmidrule(lr){3-3}\cmidrule(lr){4-4}
\textbf{Seed ratio} ($P${\percent}) & \textbf{Volume} & \textbf{Volume} & \textbf{Volume} \\
\midrule
100 & 9349 (1973) & 36736 (4143) & 34843 (3991) \\
\bottomrule
\end{tabular}
\end{table*}

\begin{figure*}[!htbp]
\centering
\setlength{\tabcolsep}{0pt}
\begin{tabular}{cccc}
\includegraphics[scale=0.95, trim=2.25in 2.25in 2.25in 2.25in, clip=true, width=0.245\linewidth, keepaspectratio=true]{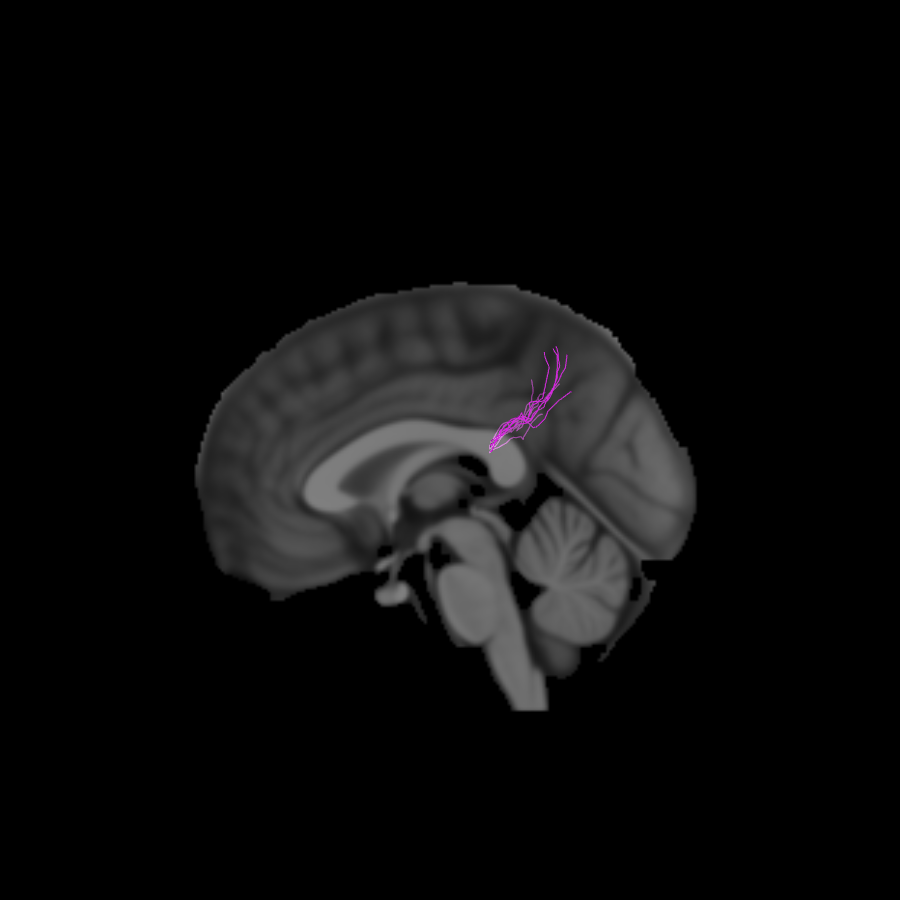} &
\includegraphics[scale=0.95, trim=2.25in 2.25in 2.25in 2.25in, clip=true, width=0.245\linewidth, keepaspectratio=true]{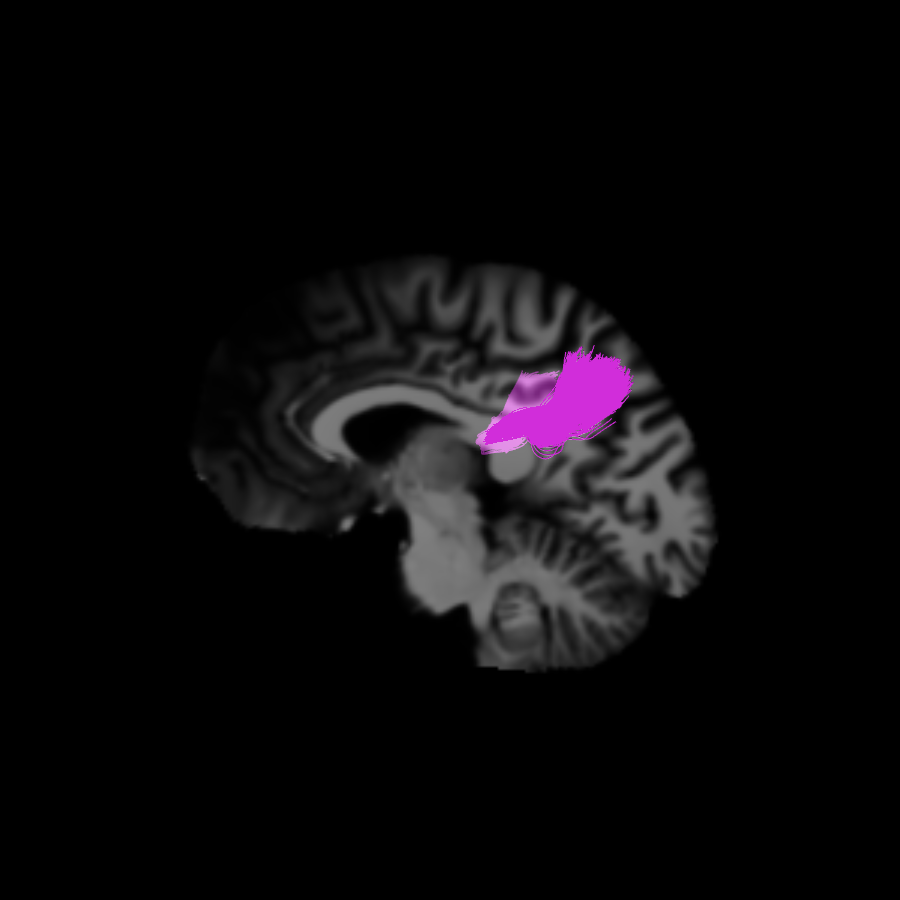} &
\hspace{0.01in}
\includegraphics[scale=0.95, trim=2.25in 2.25in 2.25in 2.25in, clip=true, width=0.245\linewidth, keepaspectratio=true]{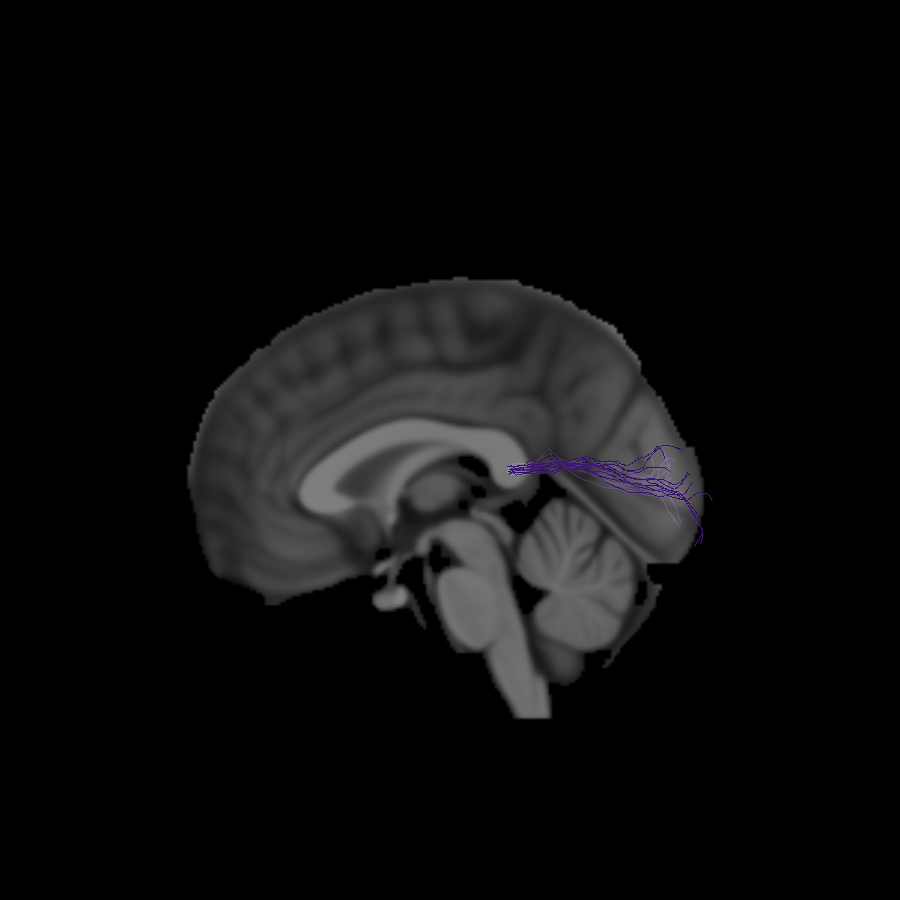} &
\includegraphics[scale=0.95, trim=2.25in 2.25in 2.25in 2.25in, clip=true, width=0.245\linewidth, keepaspectratio=true]{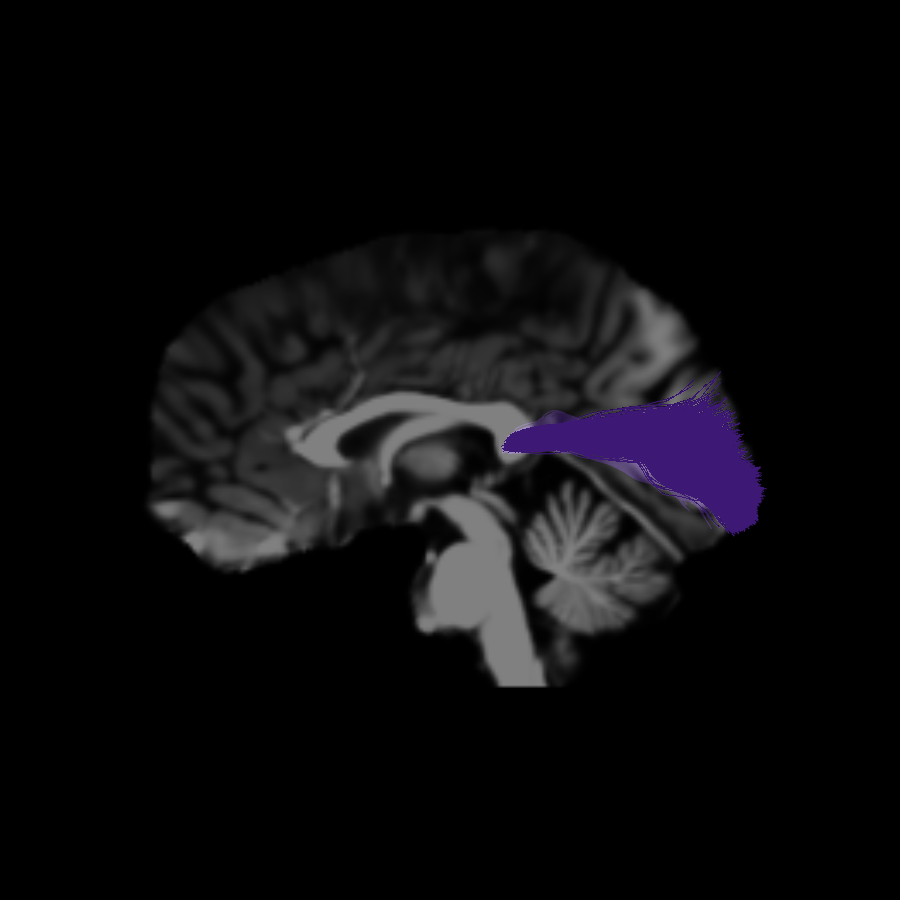} \\
\multicolumn{2}{c}{\textbf{(a)}} & \multicolumn{2}{c}{\textbf{(b)}} \\
\includegraphics[scale=0.95, trim=2.25in 2.25in 2.25in 2.25in, clip=true, width=0.245\linewidth, keepaspectratio=true]{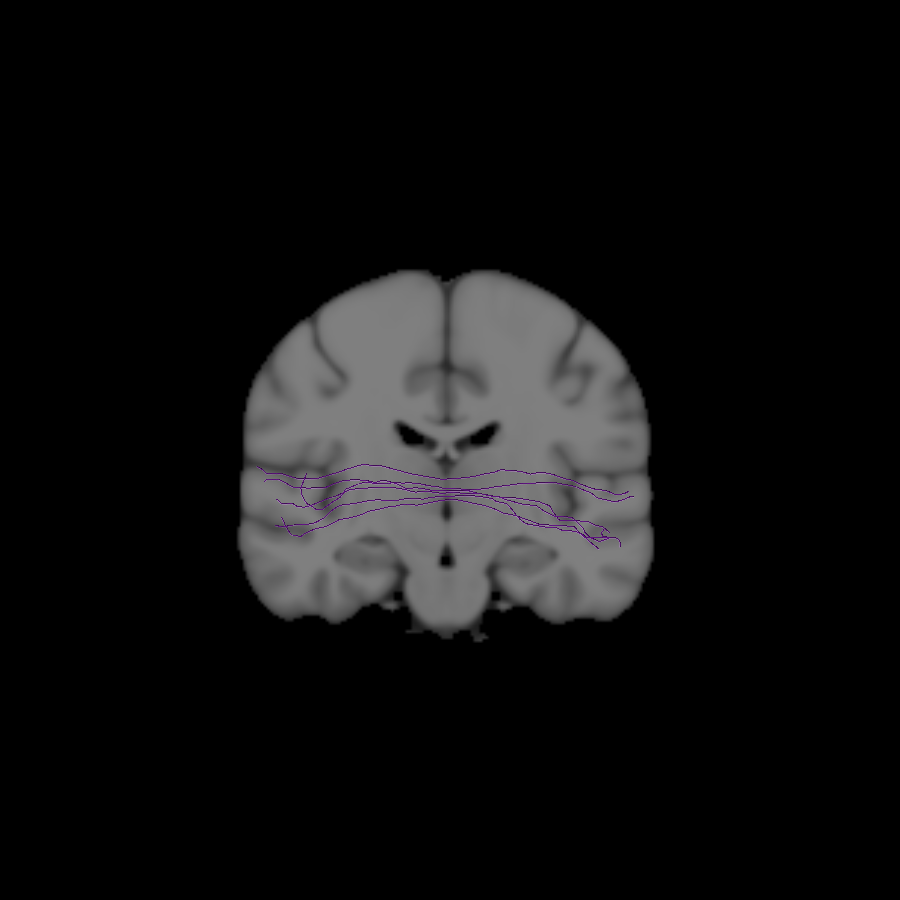} &
\includegraphics[scale=0.95, trim=2.25in 2.25in 2.25in 2.25in, clip=true, width=0.245\linewidth, keepaspectratio=true]{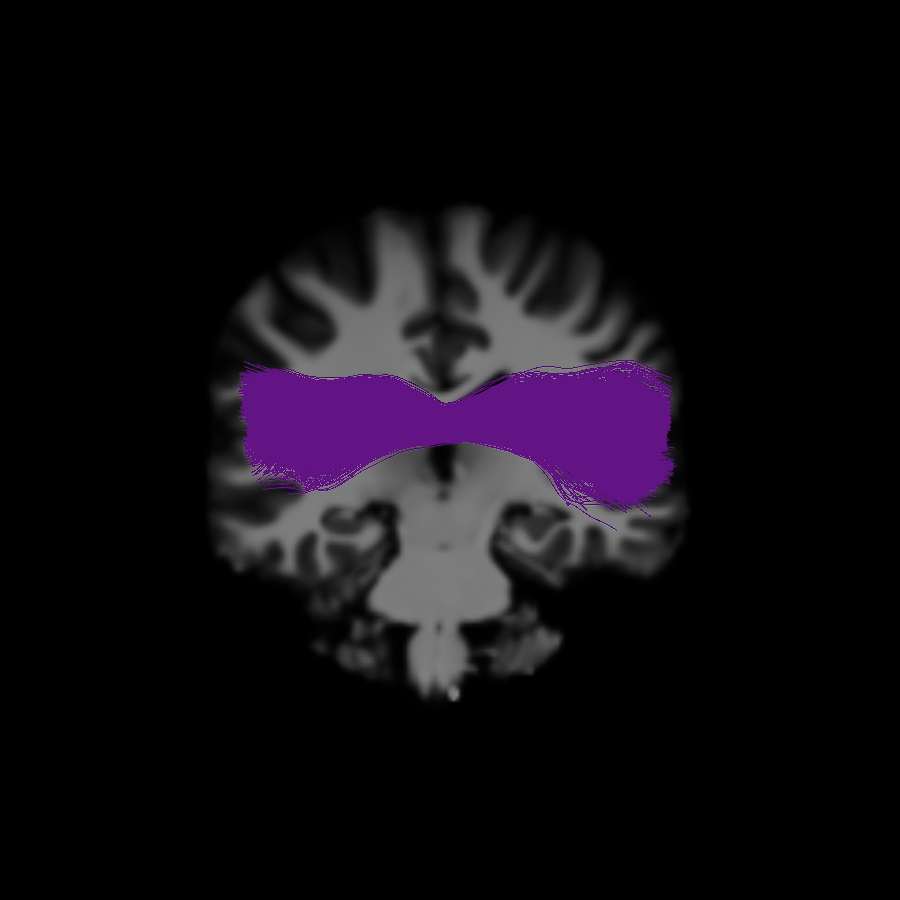} &
\hspace{0.01in}
\includegraphics[scale=0.95, trim=2.25in 2.25in 2.25in 2.25in, clip=true, width=0.245\linewidth, keepaspectratio=true]{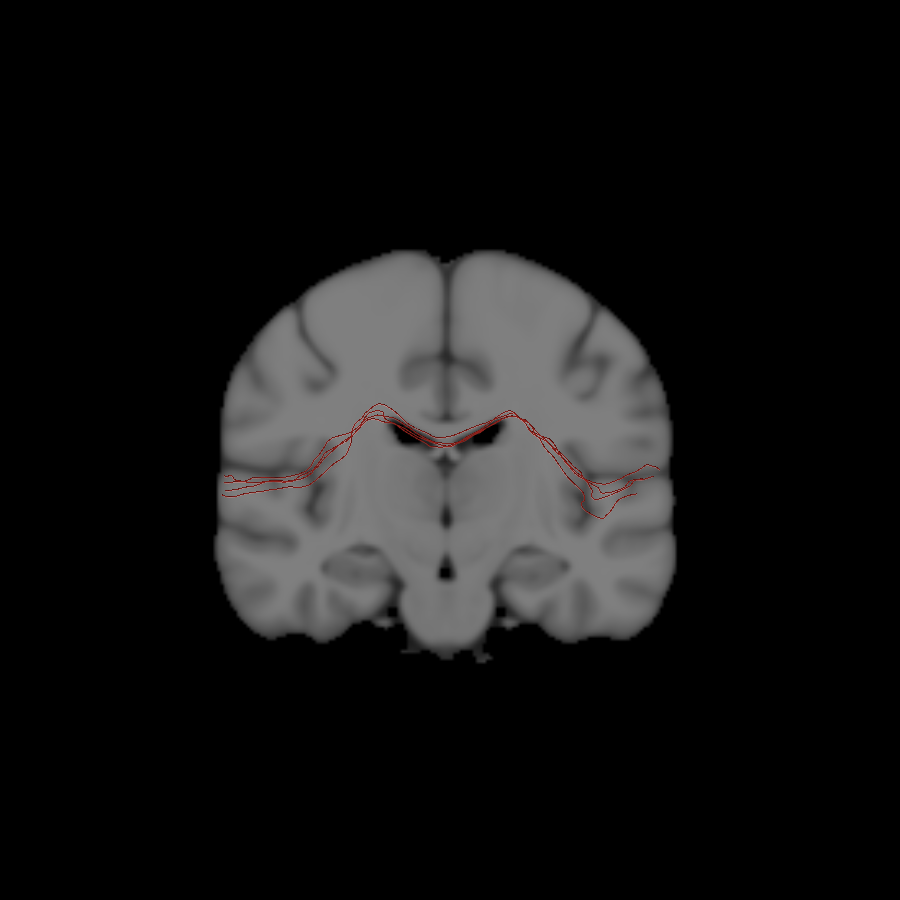} &
\includegraphics[scale=0.95, trim=2.25in 2.25in 2.25in 2.25in, clip=true, width=0.245\linewidth, keepaspectratio=true]{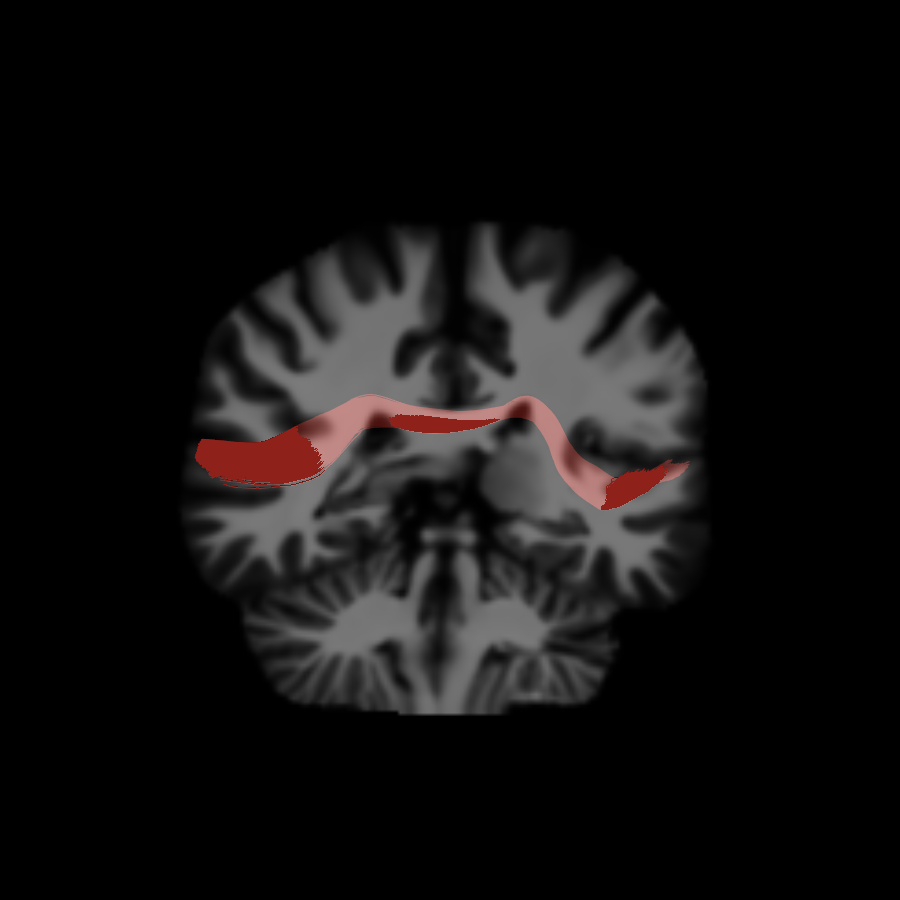} \\
\multicolumn{2}{c}{\textbf{(c)}} & \multicolumn{2}{c}{\textbf{(d)}} \\
\includegraphics[scale=0.95, trim=2.25in 2.25in 2.25in 2.25in, clip=true, width=0.245\linewidth, keepaspectratio=true]{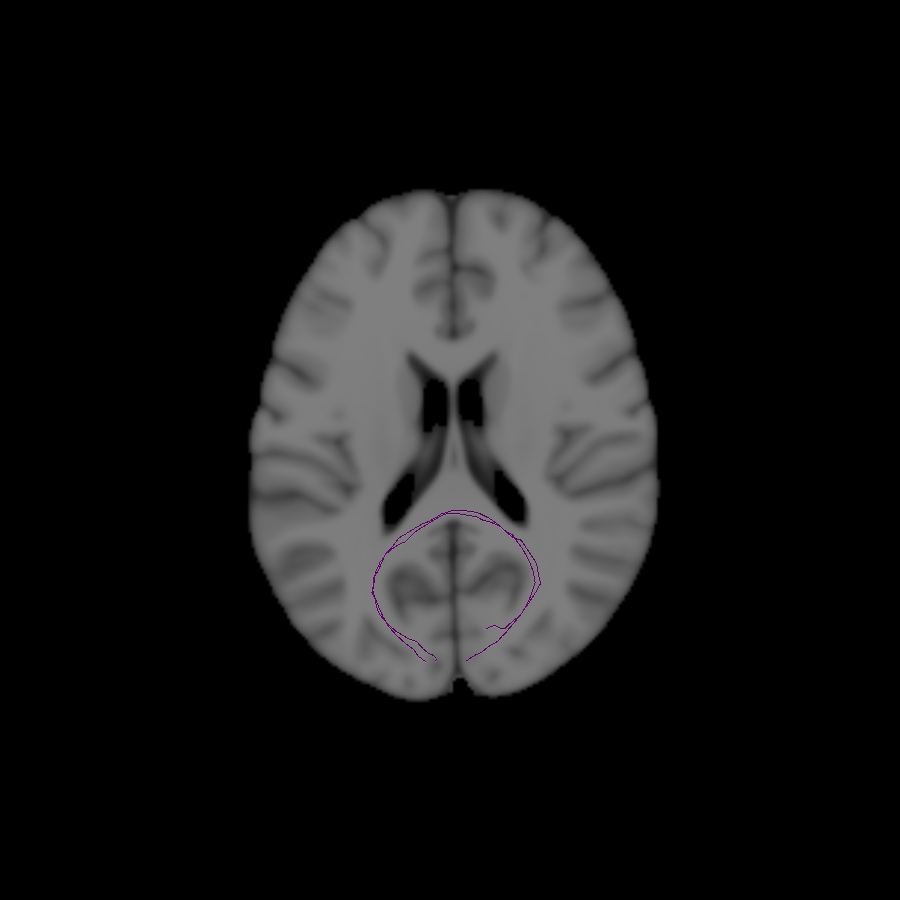} &
\includegraphics[scale=0.95, trim=2.25in 2.25in 2.25in 2.25in, clip=true, width=0.245\linewidth, keepaspectratio=true]{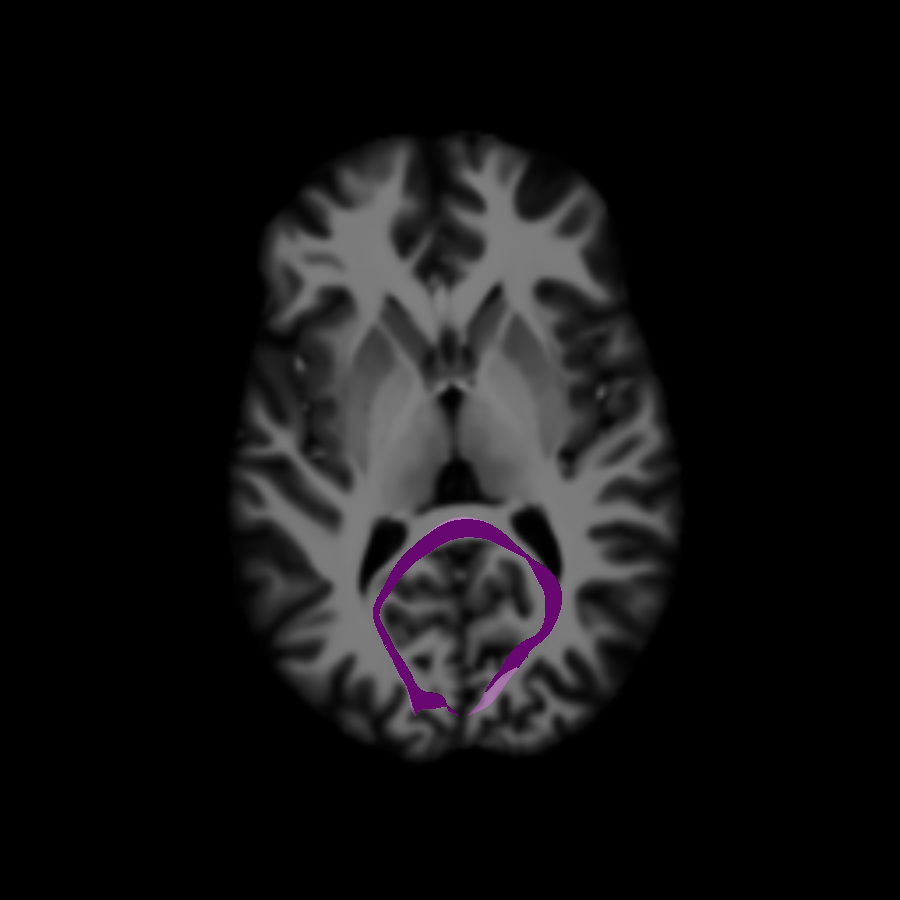} &
\hspace{0.01in}
\includegraphics[scale=0.95, trim=2.25in 2.25in 2.25in 2.25in, clip=true, width=0.245\linewidth, keepaspectratio=true]{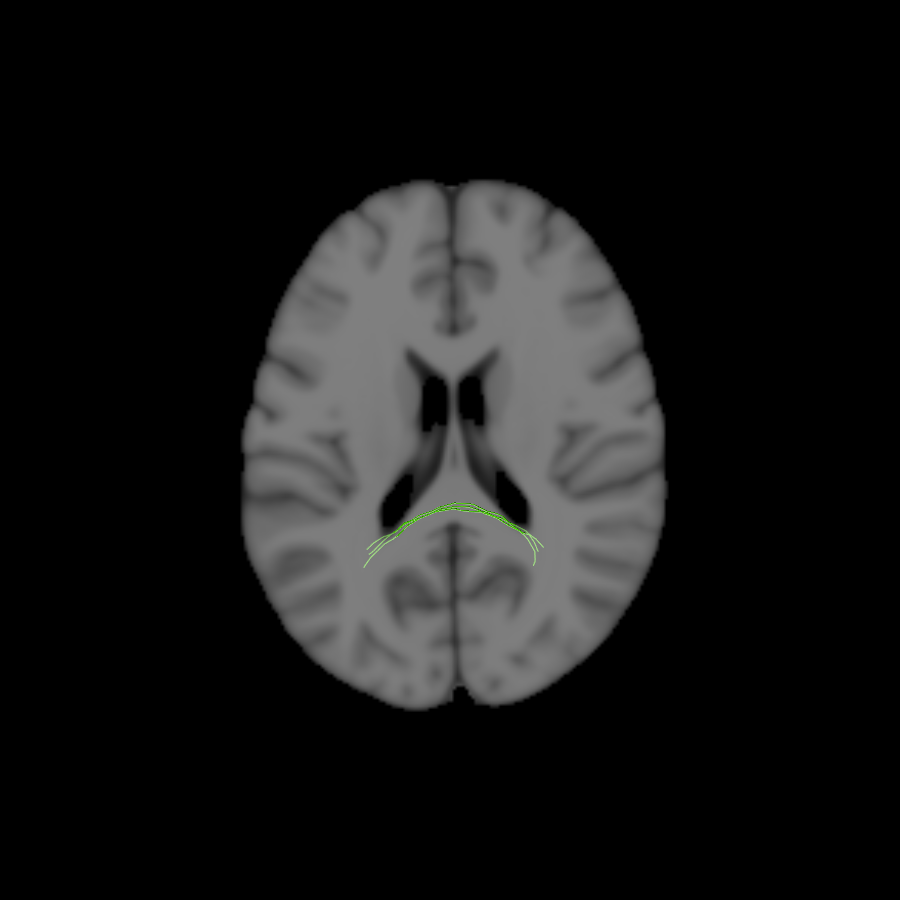} &
\includegraphics[scale=0.95, trim=2.25in 2.25in 2.25in 2.25in, clip=true, width=0.245\linewidth, keepaspectratio=true]{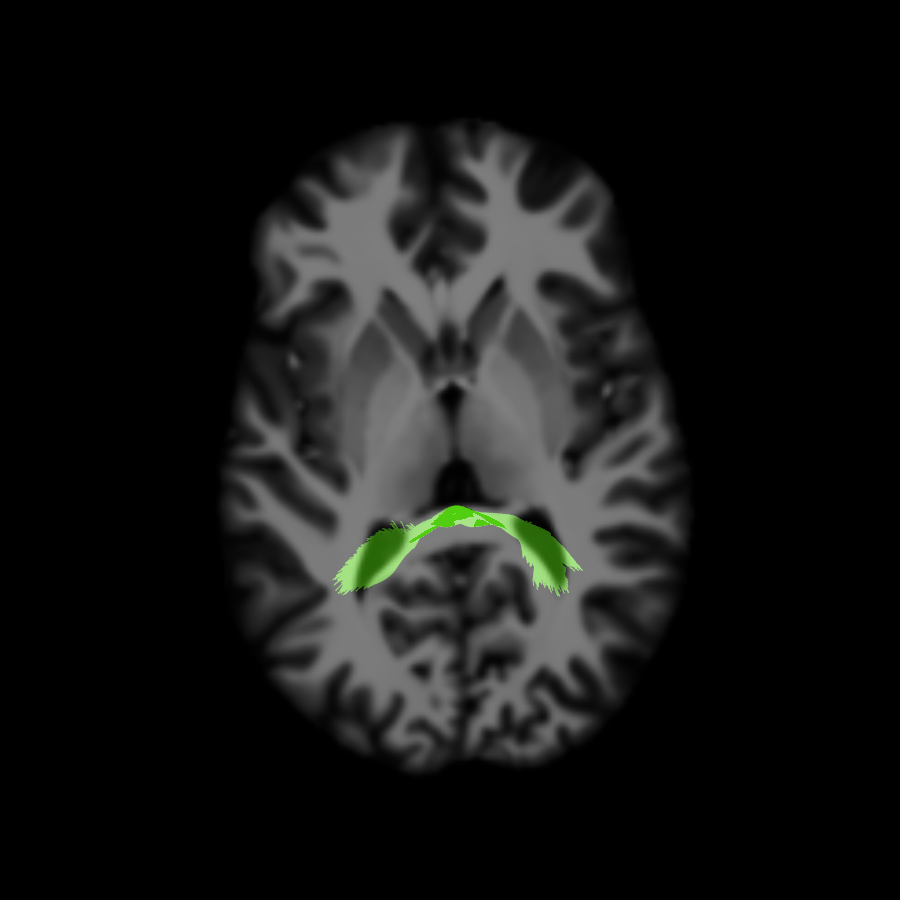} \\
\multicolumn{2}{c}{\textbf{(e)}} & \multicolumn{2}{c}{\textbf{(f)}} \\
\end{tabular}
\caption{\label{fig:bil_gin_cc_homotopic_generative_bundles_selected}GESTA applied to the BIL\&GIN callosal homotopic dataset. For each segment: left: seed streamlines; right: GESTA-generated plausible streamlines. Segments: (a) AG; (b) MOG; (c) IFG; (d) STG; (e) Cu; (f) LG. All available streamlines filtered with FINTA are used for seeding in the latent space, and streamlines are evaluated using the \textit{ADG\textsubscript{R}} criterion. Data corresponds to different test subjects (note that not all subjects contain streamlines in all segments). Views have been chosen to best visualize the bundles.}
\end{figure*}

Figure \ref{fig:bil_gin_cc_homotopic_generative_bundles_selected} shows the result of the generative tractography performed on \num{6} of the gyral-based segments, namely the AG, Cu, IFG, LG, MOG, and the STG, of the BIL\&GIN callosal homotopic dataset. The FINTA autoencoder-based tractography filtering method presented in \citet{Legarreta:MIA:2021} showed a poor performance on these segments. The figure shows that the GESTA generative tractography method can be used to recover the lost white matter spatial coverage.

\begin{figure*}[!htbp]
\centering
\setlength{\tabcolsep}{0pt}
\begin{tabular}{ccc}
\includegraphics[scale=0.95, trim=0.5in 0.6in 0.5in 0.6in, clip=true, width=0.3\linewidth, keepaspectratio=true]{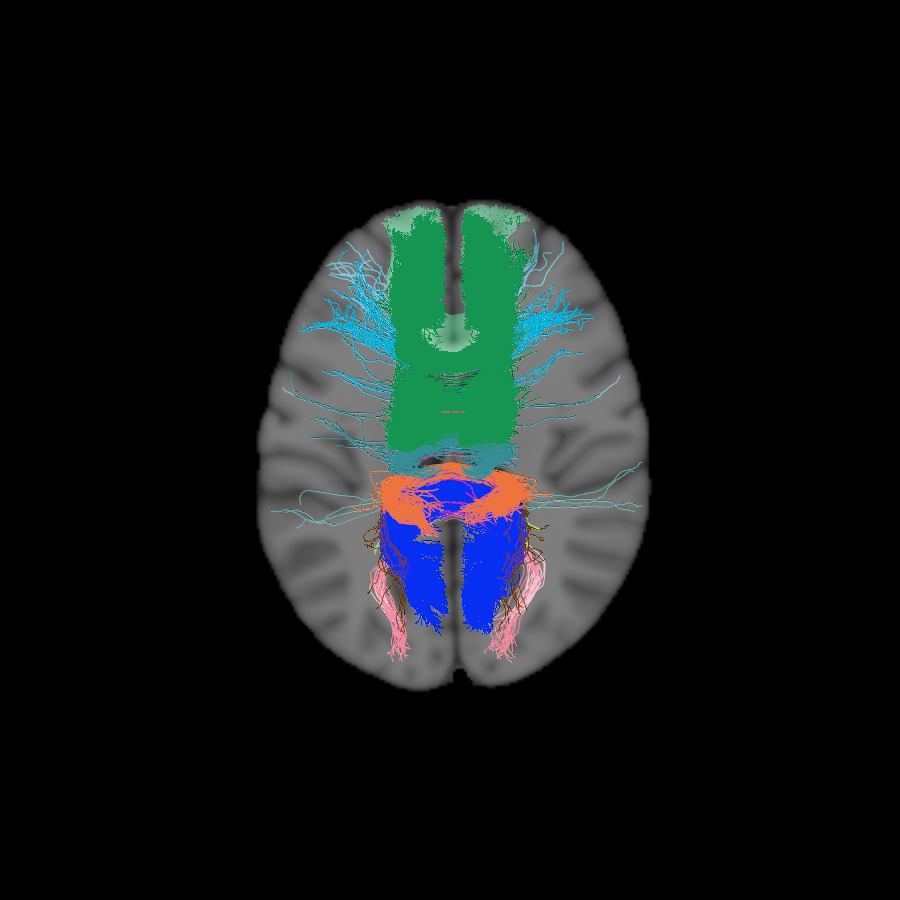} &
\includegraphics[scale=0.95, trim=0.5in 0.6in 0.5in 0.60in, clip=true, width=0.3\linewidth, keepaspectratio=true]{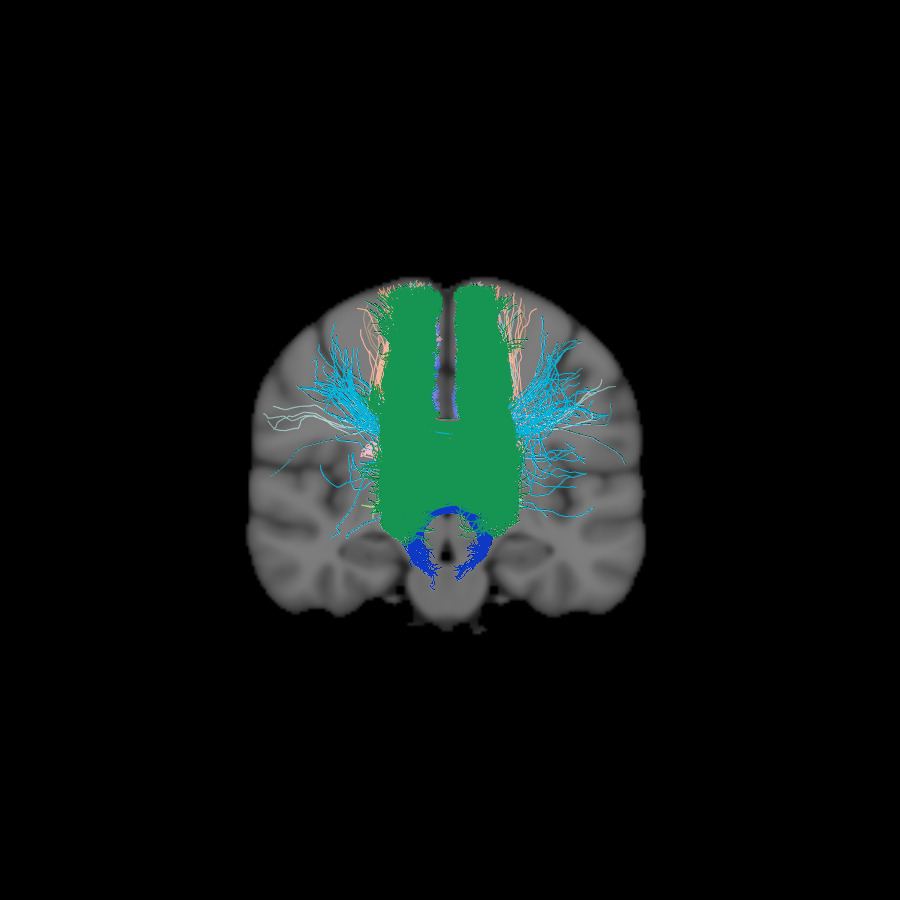} &
\includegraphics[scale=0.95, trim=0.5in 0.5in 0.5in 0.70in, clip=true, width=0.3\linewidth, keepaspectratio=true]{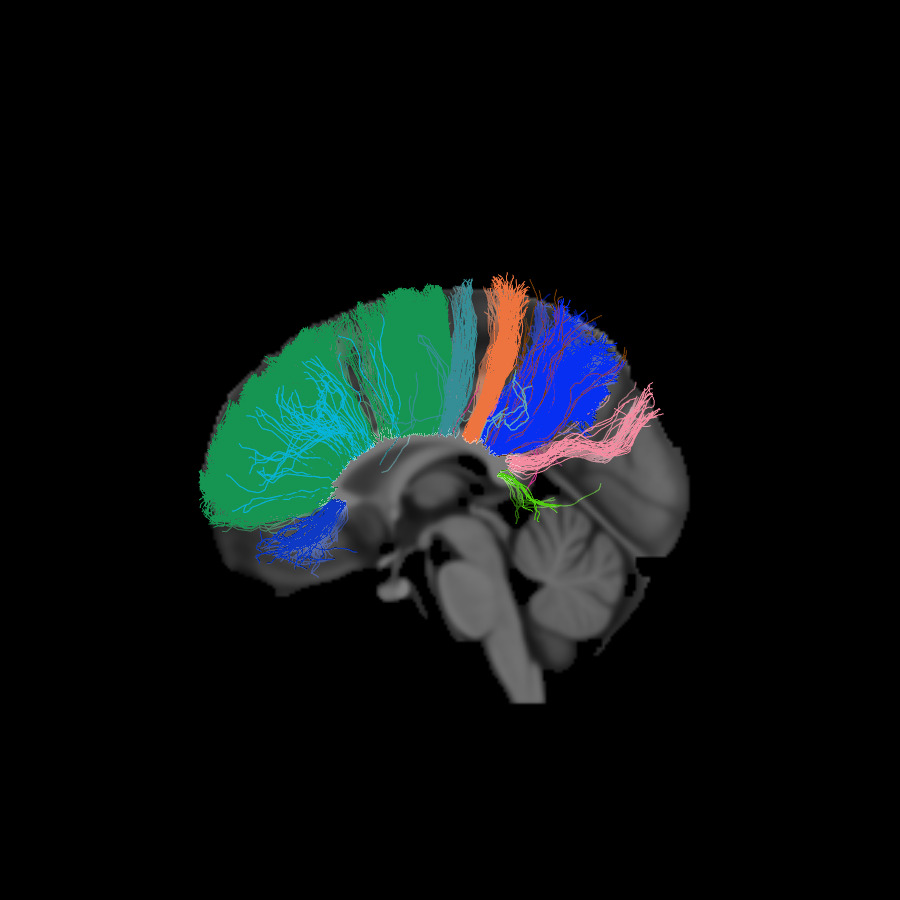} \\
\includegraphics[scale=0.95, trim=0.5in 0.55in 0.5in 0.55in, clip=true, width=0.3\linewidth, keepaspectratio=true]{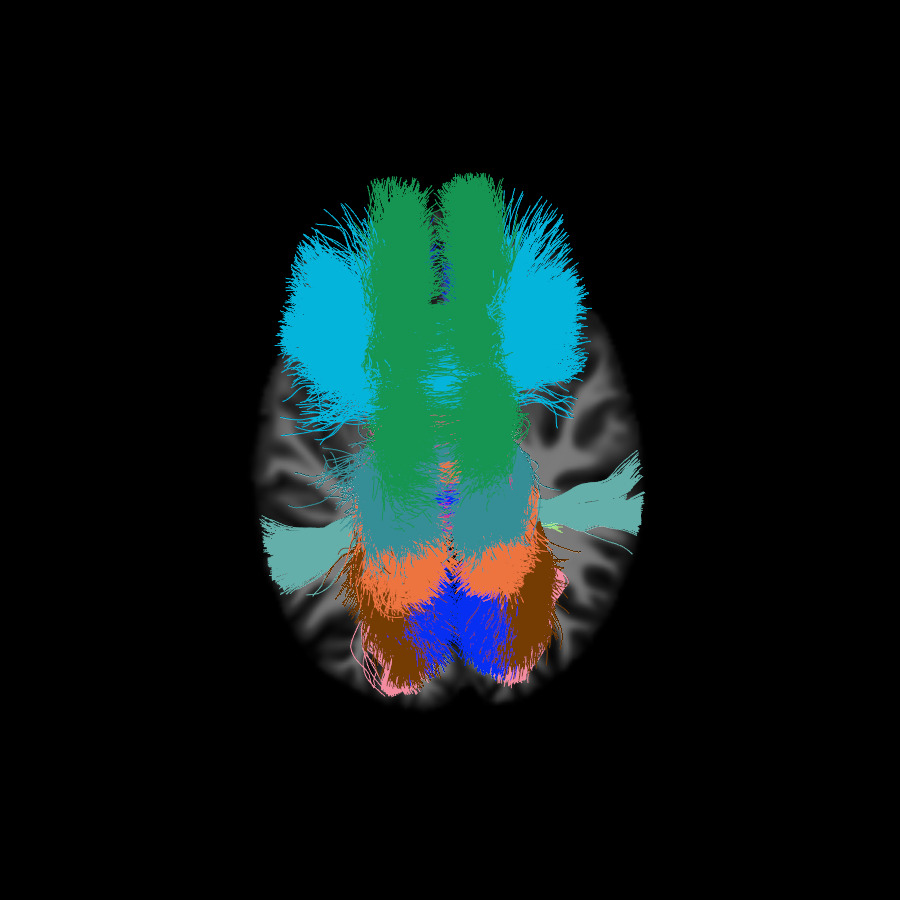} &
\includegraphics[scale=0.95, trim=0.5in 0.55in 0.5in 0.55in, clip=true, width=0.3\linewidth, keepaspectratio=true]{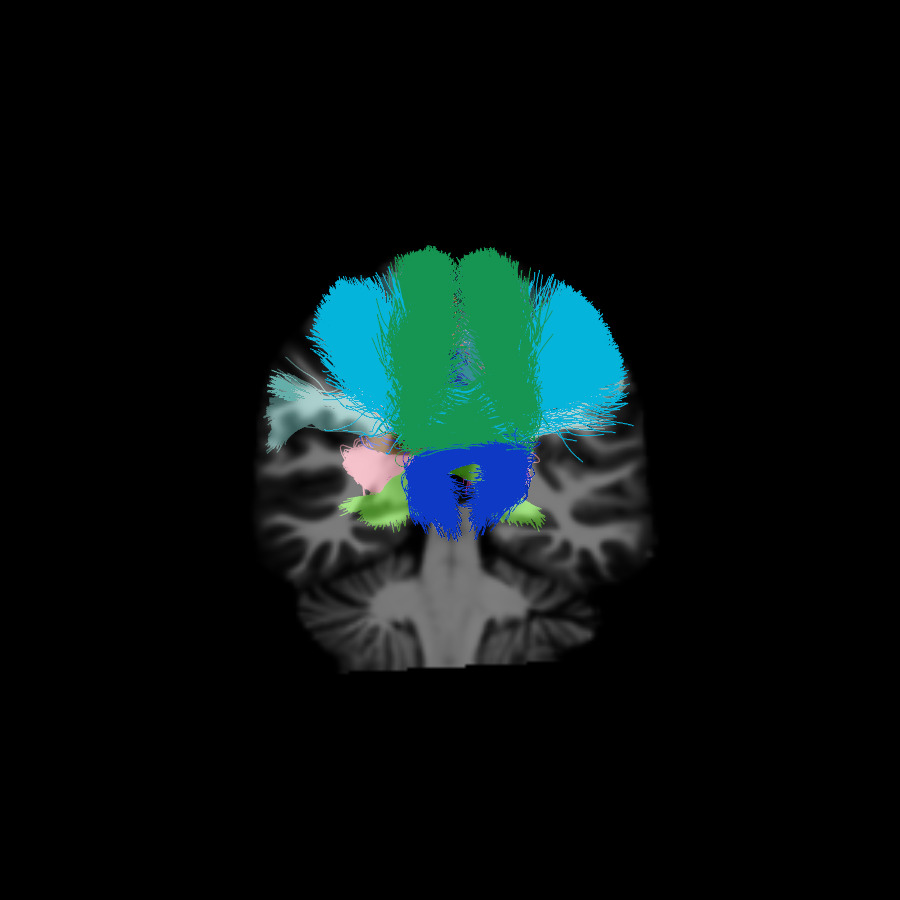} &
\includegraphics[scale=0.95, trim=0.5in 0.55in 0.5in 0.55in, clip=true, width=0.3\linewidth, keepaspectratio=true]{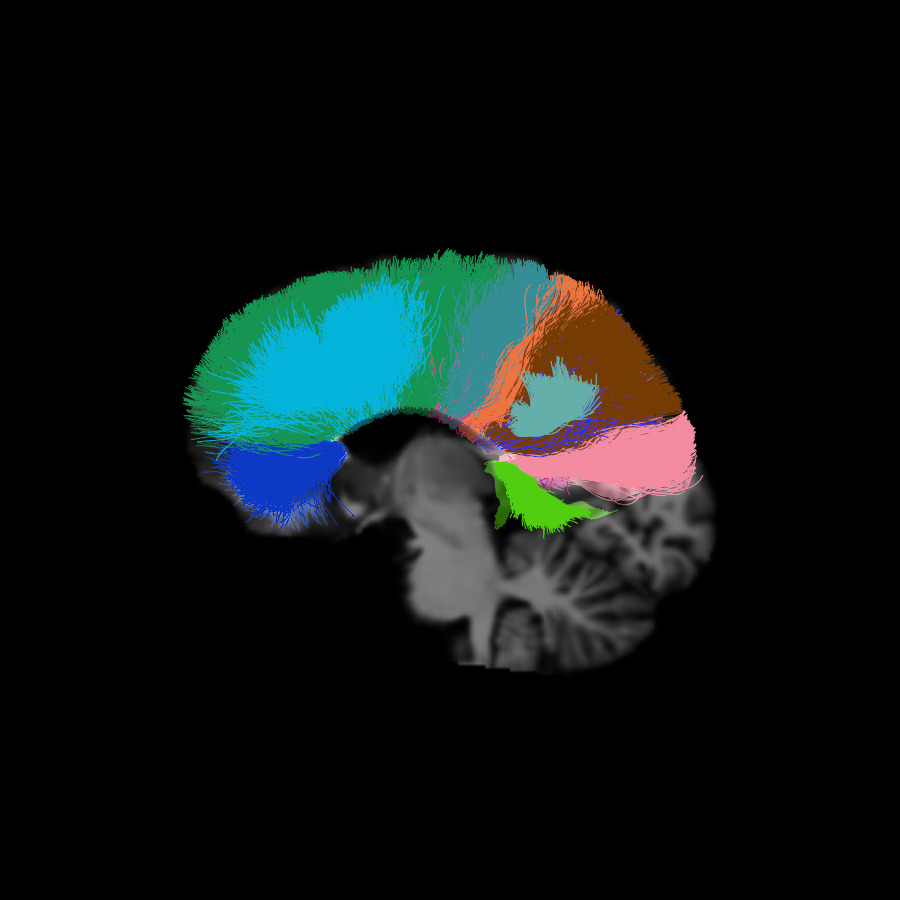} \\
\textbf{(a)} & \textbf{(b)} & \textbf{(c)} \\
\end{tabular}
\caption{\label{fig:bil_gin_cc_homotopic_generative_bundles_subj}Seed streamlines and latent-generated plausible streamlines for a given BIL\&GIN callosal homotopic dataset test subject. All available streamlines filtered with FINTA are used for seeding in the latent space, and streamlines are evaluated using the \textit{ADG\textsubscript{R}} criterion. (a) Axial superior; (b) coronal anterior; (c) sagittal left views. Segment color code (approximate): Cing: pink (not visible); LG: lime; MFG: cyan; PoCG: orange; PrCG: turquoise; PrCu: blue; RG: cobalt blue; SFG: emerald; SMG: aero blue; SOG: cotton candy; SPG: brown.}
\end{figure*}

Figure \ref{fig:bil_gin_cc_homotopic_generative_bundles_subj} shows the seed streamlines and the latent-generated streamlines for a given BIL\&GIN callosal homotopic dataset test subject. As it can be seen, GESTA improves the white matter spatial occupancy across all segments. Particularly, in the segments where the available seed streamline count is extremely low, such as the LG, SMG, or the SPG, there is a notable increase in the number of streamlines that are incorporated by the generative framework.

\begin{figure*}[!htbp]
\centering
\begin{tikzpicture}[      
every node/.style={anchor=south west,inner sep=0pt},x=0.05in,y=0.05in]
\node (fig1) at (0,0)
{\includegraphics[scale=0.95, trim=0.6in 1.45in 0.6in 0.35in, clip=true, width=0.65\linewidth, keepaspectratio=true]{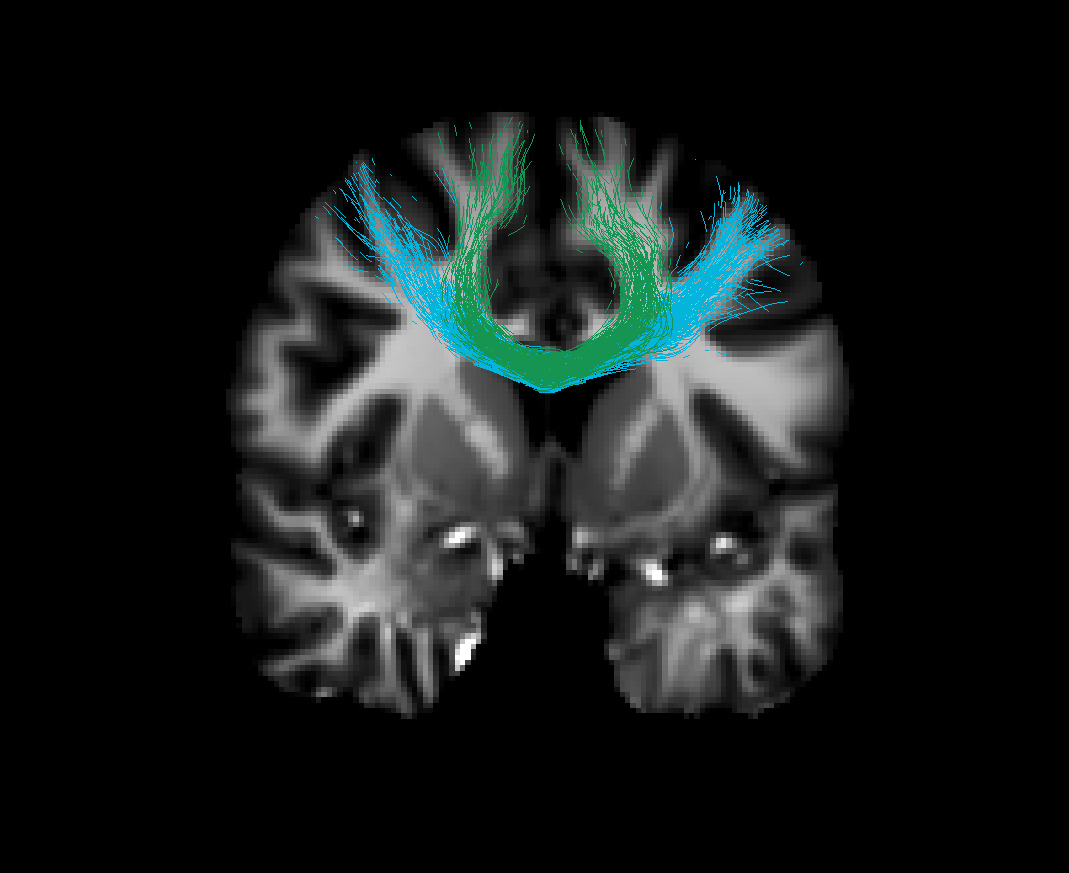}};
\node (fig2) at (2.0,2.0)
{\includegraphics[scale=0.4, trim=0.5in 0.45in 0.5in 0.2in, clip=true]{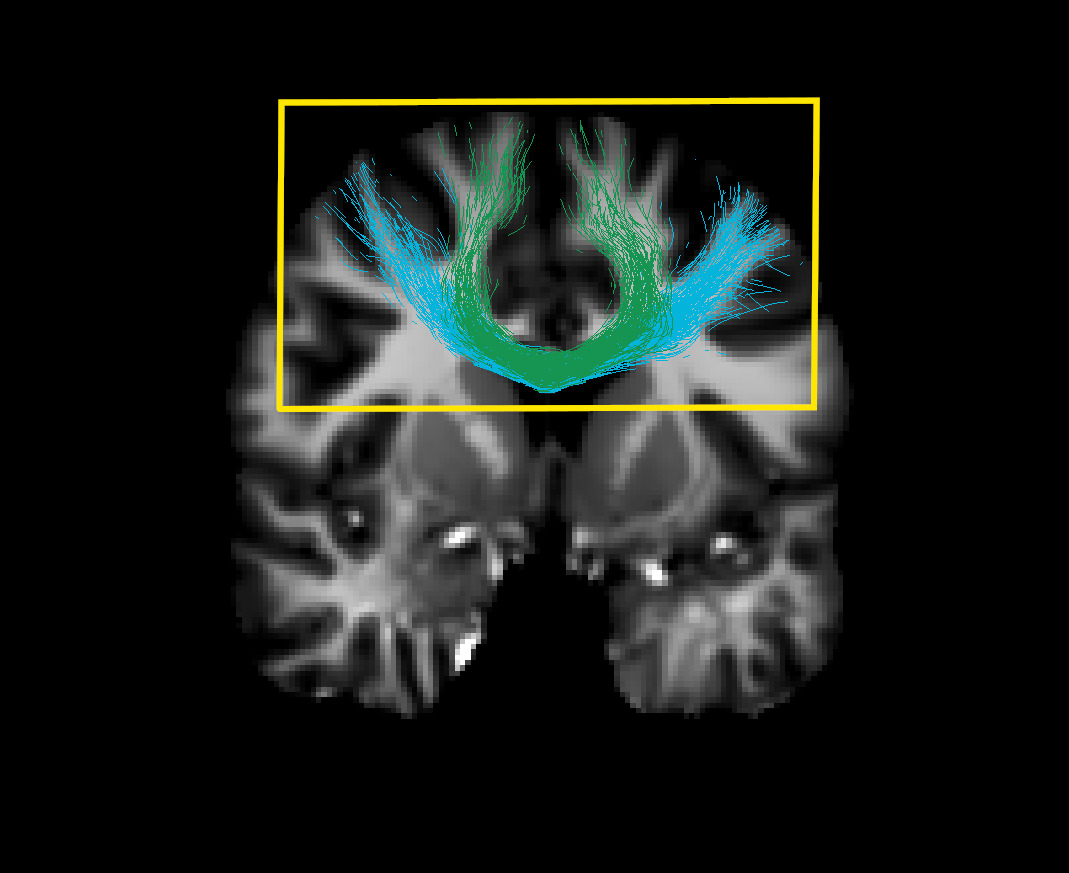}};  
\end{tikzpicture}
\caption{\label{fig:bil_gin_cc_homotopic_generative_bundles_subj_sfg_smg_selection}Generative streamlines for a given BIL\&GIN callosal homotopic dataset subject. Streamlines shown correspond to SFG (emerald) and MFG (cyan) segments. Coronal anterior view.}
\end{figure*}

The proposed generative tractography framework reliably reconstructs streamlines in regions where multiple fibers exist. The method does not involve propagating streamlines in a local orientation field, and thus, the reconstruction process is not influenced by potentially wrong local decisions. The close-up coronal view in figure \ref{fig:bil_gin_cc_homotopic_generative_bundles_subj_sfg_smg_selection} demonstrates that GESTA successfully creates multiple streamline populations in a sector of the corpus callosum where both the SFG and MFG homotopic segments contain fibers.

\clearpage
\newpage

\subsection{TractoInferno}
\label{subsubsec:tractoinferno_results}

Table \ref{tab:tractoinferno_generative_tractography_measures} shows the overlap, overreach, Dice's coefficient, and bundle detection scores of GESTA and the baseline tractography methods used by the authors of the dataset. Results show that using a fraction of either the deterministic or probabilistic seeds for the generative process, does not alter results significantly for any of the streamline evaluation criteria. GESTA improves the overlap and Dice's coefficient scores of its seeding tractography counterparts, at the cost of an increased overreach. As indicated by the bundle detection score, the generative tractography procedure is able to successfully generate streamlines across all bundles.

\begin{table*}[!hbp]
\caption{\label{tab:tractoinferno_generative_tractography_measures}Tractography evaluation measures averaged across the sampled bundles and subjects. The latent space seeding modes (deterministic and probabilistic tractography seeds) used for GESTA are denoted as ``Det'' and ``Prob'', respectively. The seed count applies to each bundle independently.}
\centering
\begin{tabular}{ccccccc}
\toprule
\textbf{Method} & \textbf{Seed count ($S$)} & \textbf{Criterion} & \textbf{OL} ($\uparrow$) & \textbf{OR} ($\downarrow$) & \textbf{Dice} ($\uparrow$) & \textbf{Bundle detection} ($\uparrow$) \\
\midrule
\multirow{2}{*}{GESTA-Det} & \multirow{2}{*}{1000} & \textit{ADG\textsubscript{R}} & 0.76 (0.1) & 1.33 (1.33) & 0.55 (0.13) & 1.0 \\
& & \textit{ADGC\textsubscript{R}} & 0.72 (0.11) & 1.15 (1.17) & 0.55 (0.12) & 1.0 \\
\multirow{2}{*}{GESTA-Prob} & \multirow{2}{*}{1000} & \textit{ADG\textsubscript{R}} & 0.78 (0.09) & 1.38 (1.29) & 0.55 (0.13) & 1.0 \\
& & \textit{ADGC\textsubscript{R}} & 0.75 (0.09) & 1.19 (1.14) & 0.55 (0.13) & 1.0 \\
\midrule
Deterministic & - & - & 0.28 (0.11) & 0.03 (0.07) & 0.41 (0.14) & 0.96 (0.07) \\
Probabilistic & - & - & 0.41 (0.18) & 0.06 (0.09) & 0.54 (0.2) & 0.95 (0.06) \\
PFT & - & - & 0.73 (0.22) & 0.31 (0.44) & 0.71 (0.19) & 0.99 (0.02) \\
SET & - & - & 0.55 (0.24) & 0.16 (0.22) & 0.62 (0.19) & 0.99 (0.02) \\
\bottomrule
\end{tabular}
\end{table*}

\begin{figure*}[!htbp]
\centering
\setlength{\tabcolsep}{0pt}
\begin{tabular}{ccc}
\includegraphics[scale=0.95, trim=0.2in 0.2in 0.2in 0.2in, clip=true, width=0.5\linewidth, keepaspectratio=true]{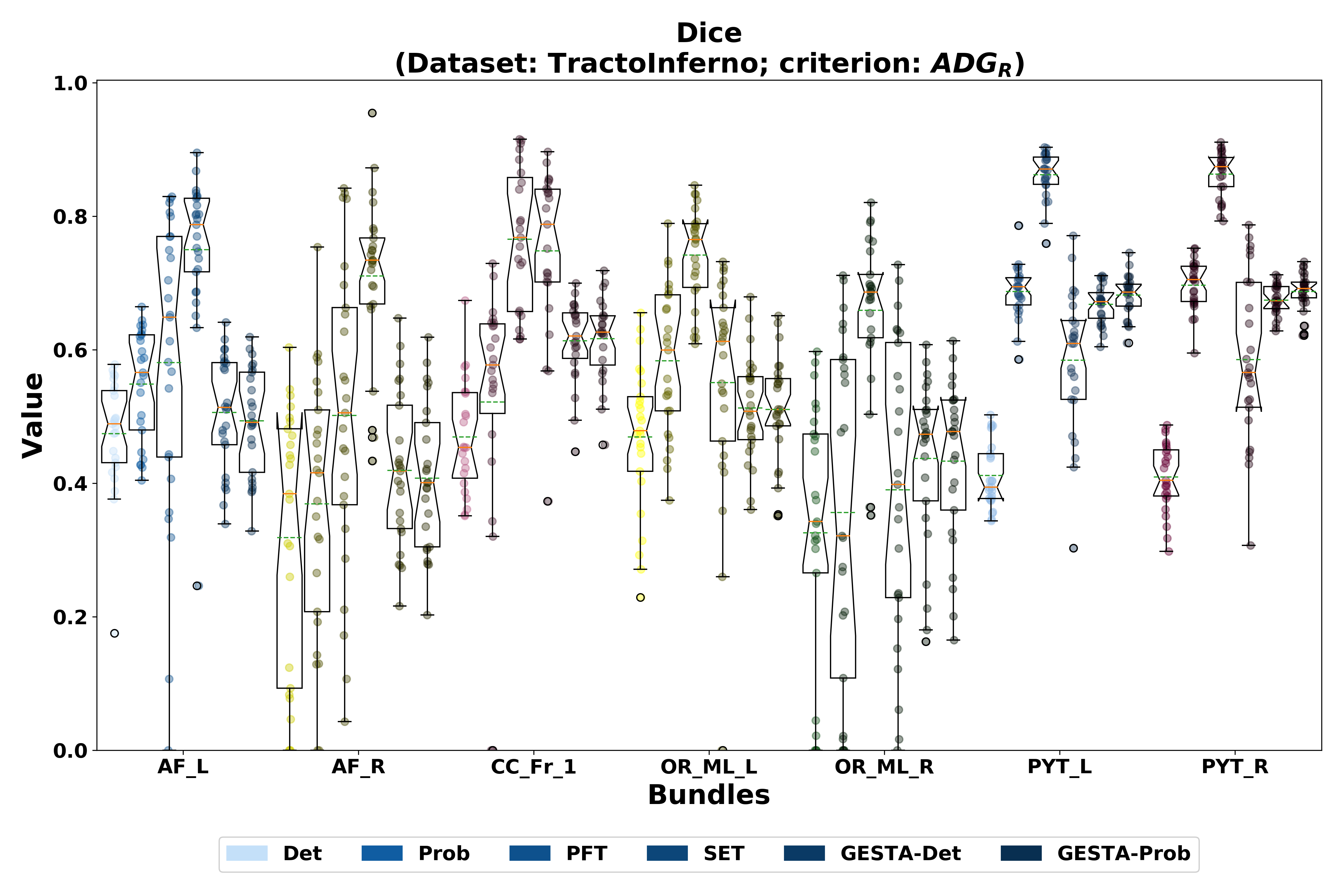} &
\includegraphics[scale=0.95, trim=0.2in 0.2in 0.2in 0.2in, clip=true, width=0.5\linewidth, keepaspectratio=true]{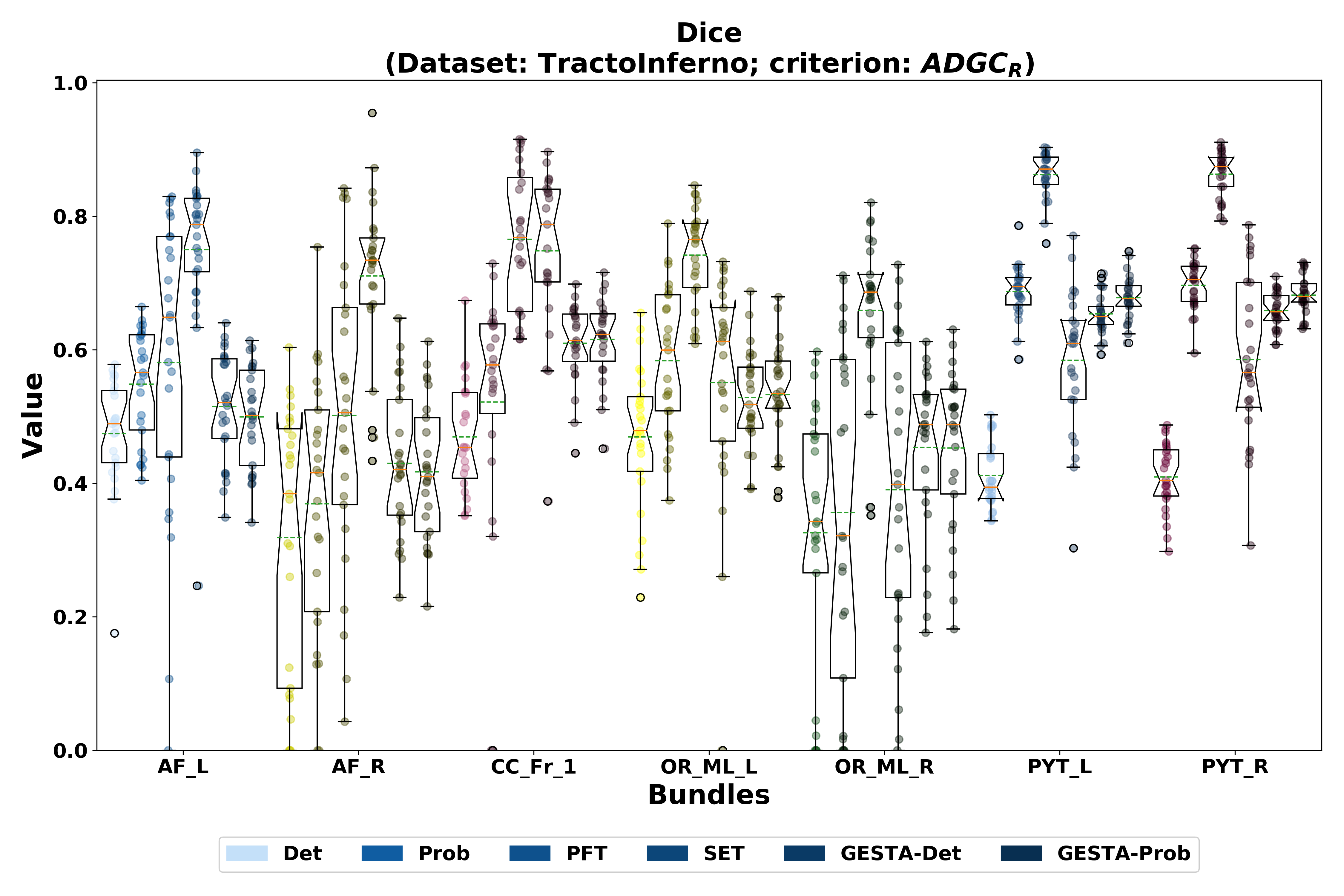} \\
\textbf{(a)} & \textbf{(b)} \\
\end{tabular}
\caption{\label{fig:tractoinferno_dice}TractoInferno bundle-wise Dice's coefficient across subjects for different tracking methods. (a) \textit{ADG}\textsubscript{R}; (b) \textit{ADGC}\textsubscript{R} streamline evaluation criteria. Note that the criterion only applies to the generative streamlines. The color shade identifies each method within each bundle, with the same shade pattern being repeated across bundles and sorted according to the legend.}
\end{figure*}

Figure \ref{fig:tractoinferno_dice} shows the bundle-wise Dice's coefficient across subjects for both latent space seeding modes (deterministic and probabilistic tractography seeds), and the \textit{ADG}\textsubscript{R} and \textit{ADGC}\textsubscript{R} criteria applied to the generative streamlines. Results show that GESTA is able to improve the scores of its seed tractography counterparts. It is systematically better than the deterministic tractography, and matches the probabilistic tractography for AF\_R, CC\_Fr\_1, OR\_ML\_R, and PYT\_R (\num{4} bundles out of \num{7}), with close scores for the rest of the bundles. It is also able to outperform SET on some bundles (OR\_ML\_R, PYT\_L, PYT\_R). Additionally, it has a reduced variability across bundles compared to the baseline methods.

Figure \ref{fig:tractoinferno_subj} shows the baseline tractography and the latent-generated plausible streamlines (using either deterministic or probabilistic tractography seeds, and evaluated with the \textit{ADGC}\textsubscript{R} criterion) for two randomly chosen TractoInferno test subjects. For either seeding method, GESTA is able to extract streamlines for bundles that other baseline methods fail to reconstruct (e.g. subject \textit{s1}'s deterministic tractography AF\_R, or subject \textit{s2}'s probabilistic tractography AF\_R and CC\_Fr\_1 bundles) or reconstruct poorly in terms of their spatial extent (e.g. subject \textit{s1}'s PFT baseline AF\_L and AF\_R, and subject \textit{s2}'s OL\_ML\_R bundle across all baselines). The generative streamlines successfully reconstruct the Meyer's loop in the OR\_ML bundle, and provide a reasonable coverage for the most lateral PYT bundle streamlines.

\begin{figure*}[!htbp]
\begin{subtable}{\linewidth}
\caption{TractoInferno baselines \vs generative tractography. Subject \textit{s1}.}
\centering
\setlength{\tabcolsep}{0pt}
\begin{tabular}{ccccc}
\includegraphics[scale=0.95, trim=1.5in 2.25in 1.5in 2.6in, clip=true, width=0.205\linewidth, keepaspectratio=true]{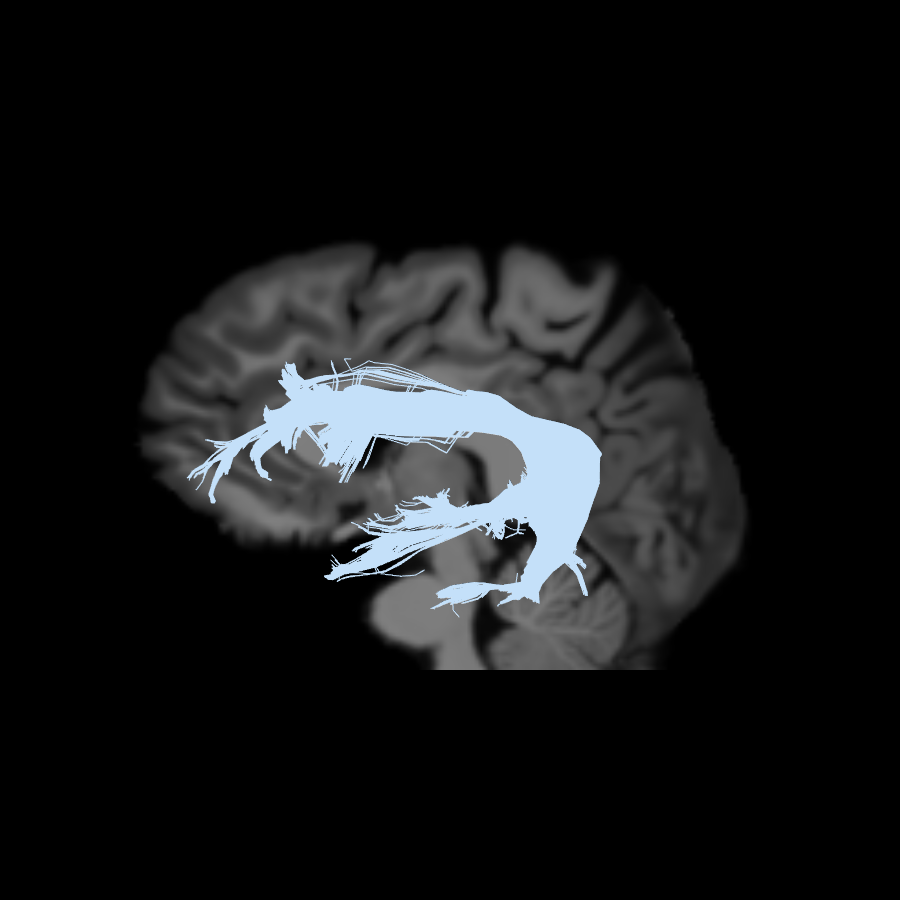} &
\includegraphics[scale=0.95, trim=1.2in 2.0in 1.15in 2.325in, clip=true, width=0.205\linewidth, keepaspectratio=true]{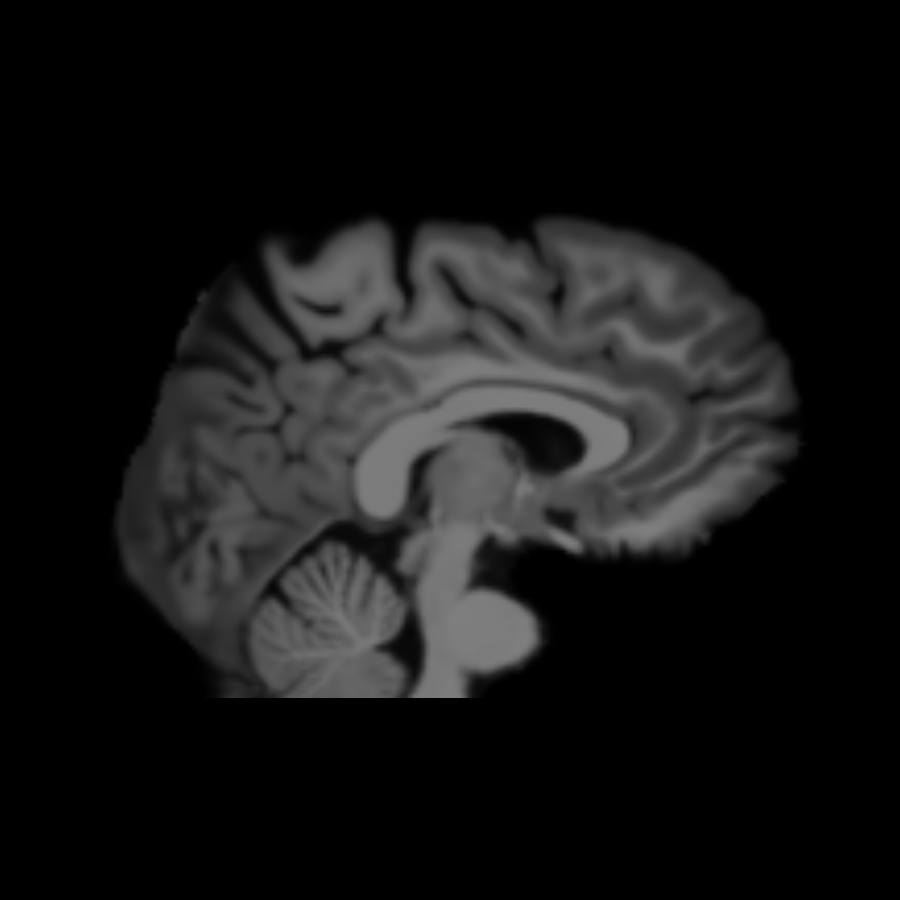} &
\includegraphics[scale=0.95, trim=1.5in 2.25in 1.5in 2.6in, clip=true, width=0.205\linewidth, keepaspectratio=true]{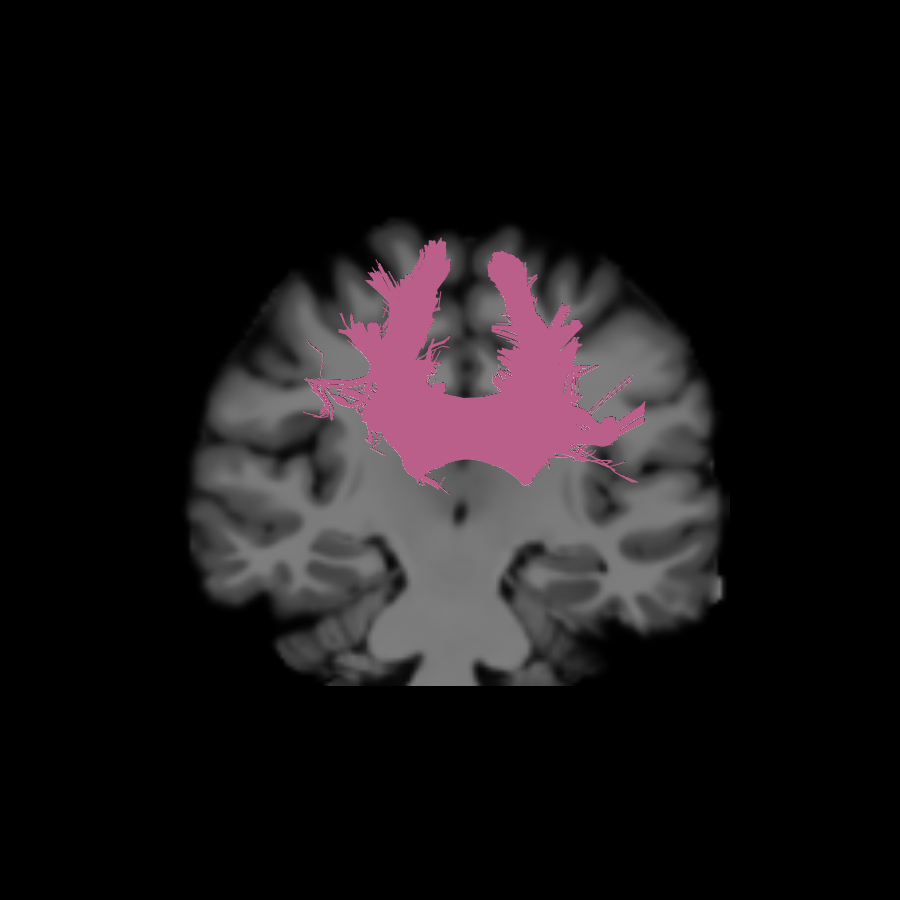} &
\includegraphics[scale=0.95, trim=2.25in 1.95in 2.25in 1.745in, clip=true, width=0.15\linewidth, keepaspectratio=true]{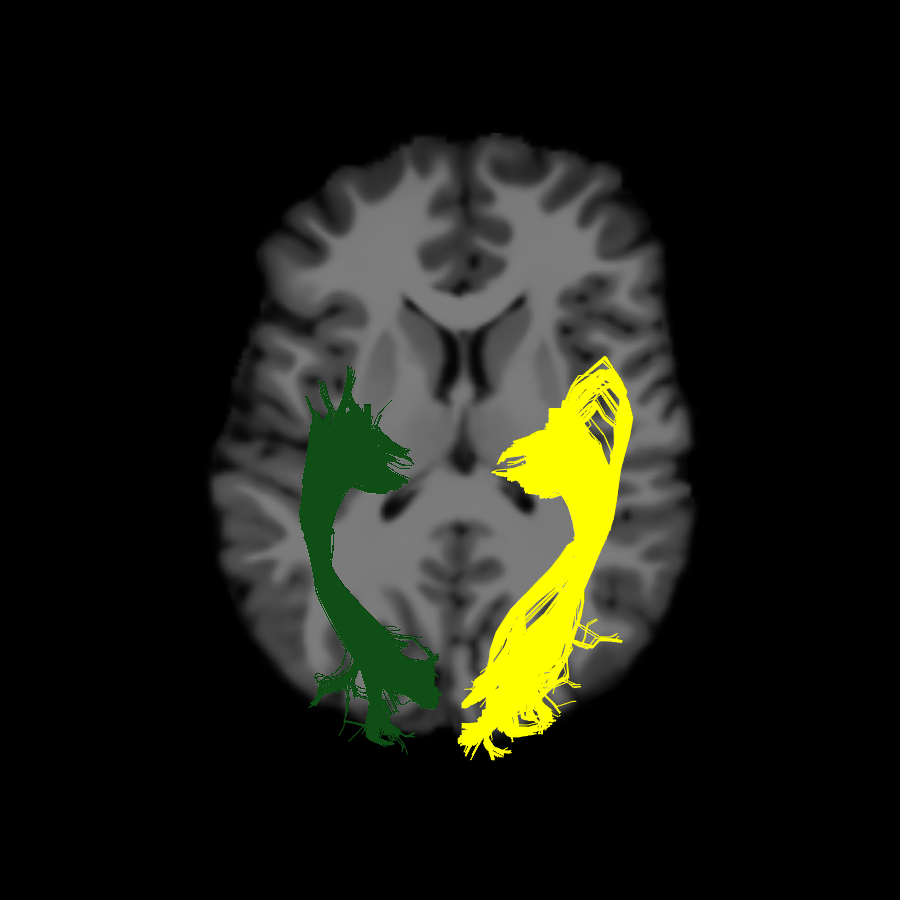} &
\includegraphics[scale=0.95, trim=1.5in 2.25in 1.5in 2.6in, clip=true, width=0.205\linewidth, keepaspectratio=true]{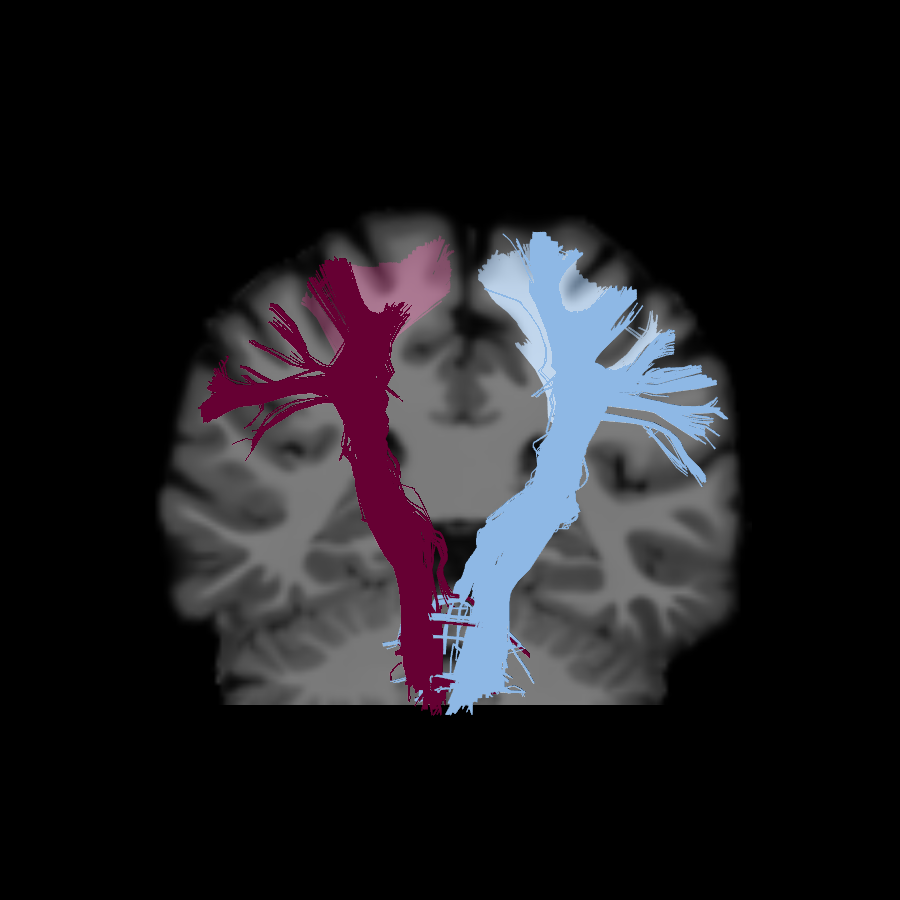} \\
\includegraphics[scale=0.95, trim=1.5in 2.25in 1.5in 2.6in, clip=true, width=0.205\linewidth, keepaspectratio=true]{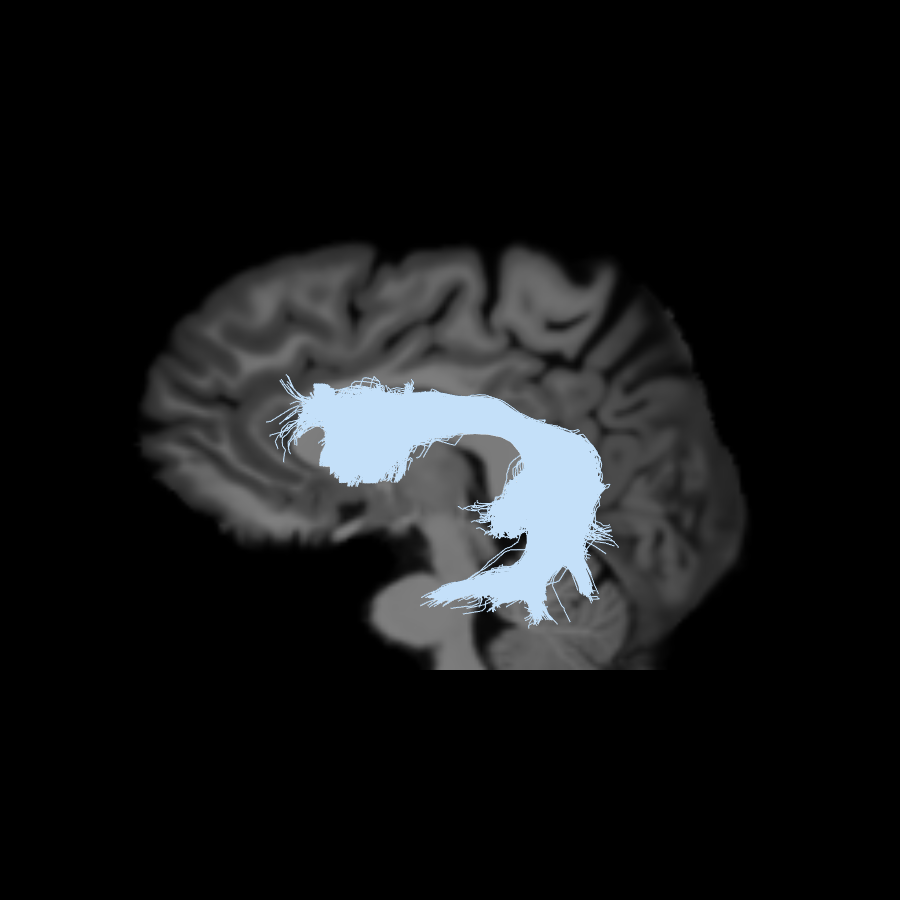} &
\includegraphics[scale=0.95, trim=1.5in 2.25in 1.5in 2.6in, clip=true, width=0.205\linewidth, keepaspectratio=true]{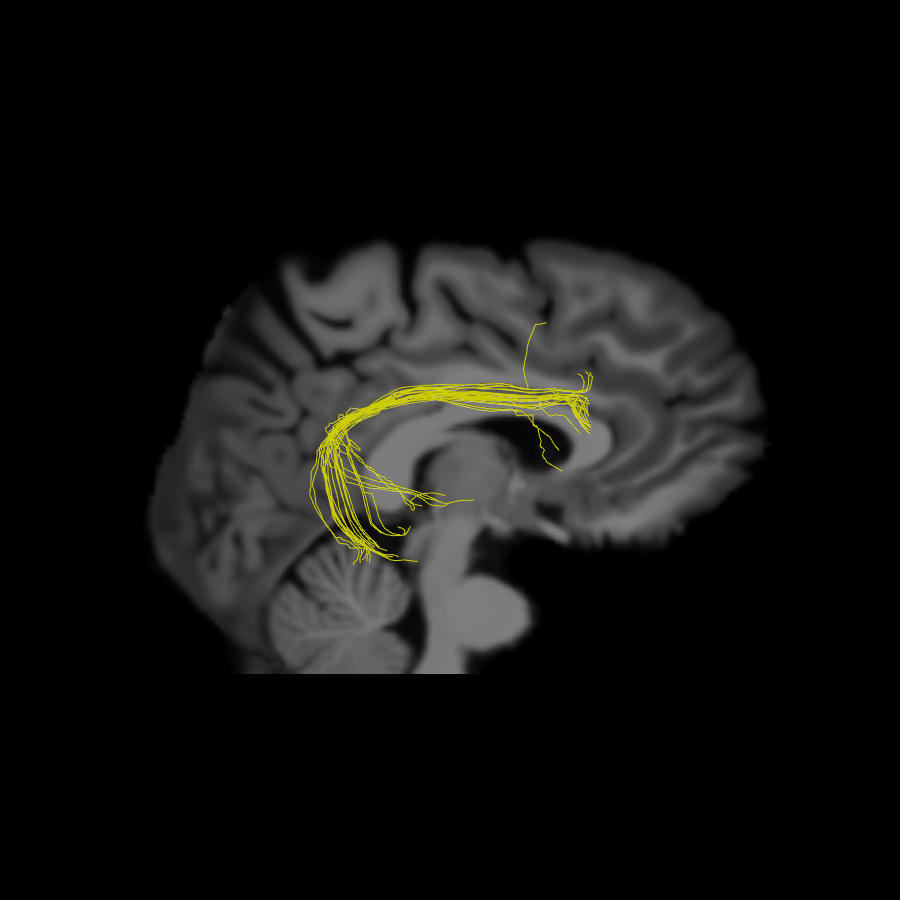} &
\includegraphics[scale=0.95, trim=1.5in 2.25in 1.5in 2.6in, clip=true, width=0.205\linewidth, keepaspectratio=true]{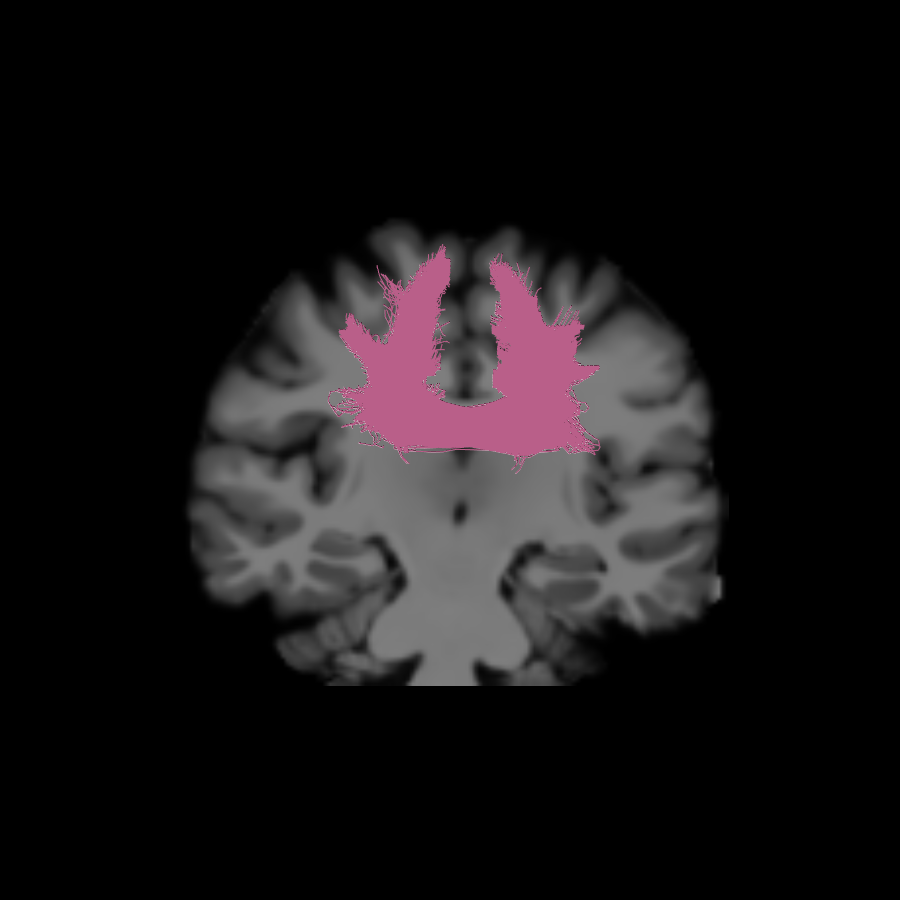} &
\includegraphics[scale=0.95, trim=2.25in 1.95in 2.25in 1.745in, clip=true, width=0.15\linewidth, keepaspectratio=true]{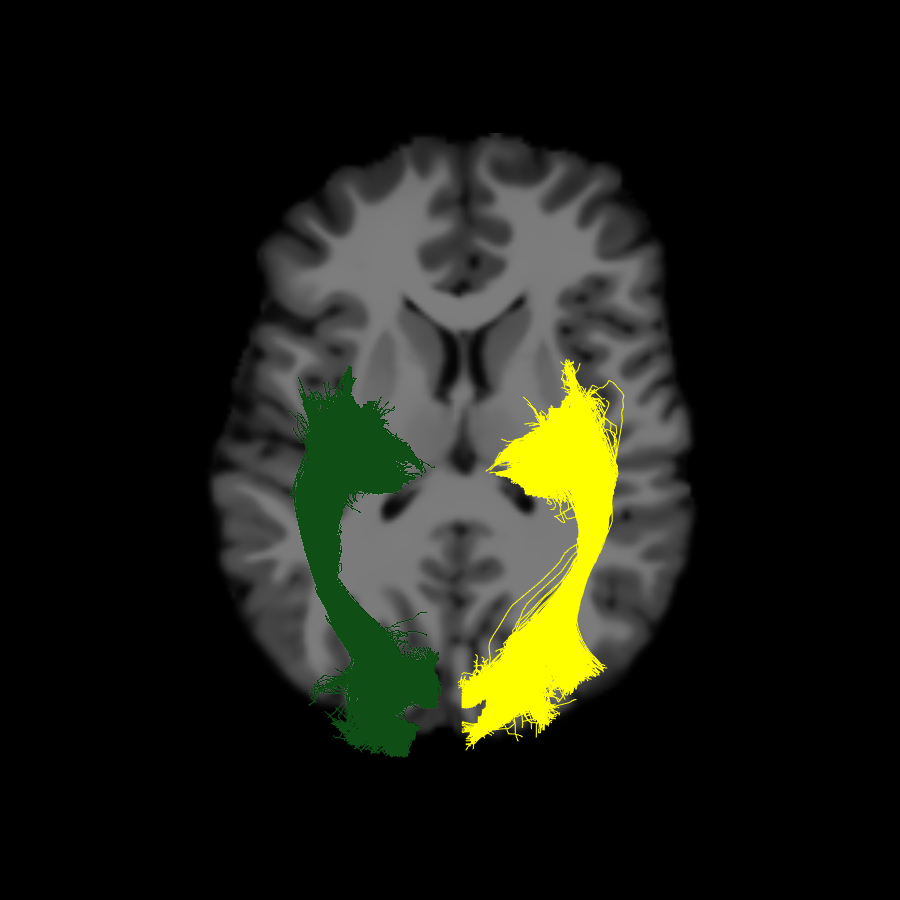} &
\includegraphics[scale=0.95, trim=1.5in 2.25in 1.5in 2.6in, clip=true, width=0.205\linewidth, keepaspectratio=true]{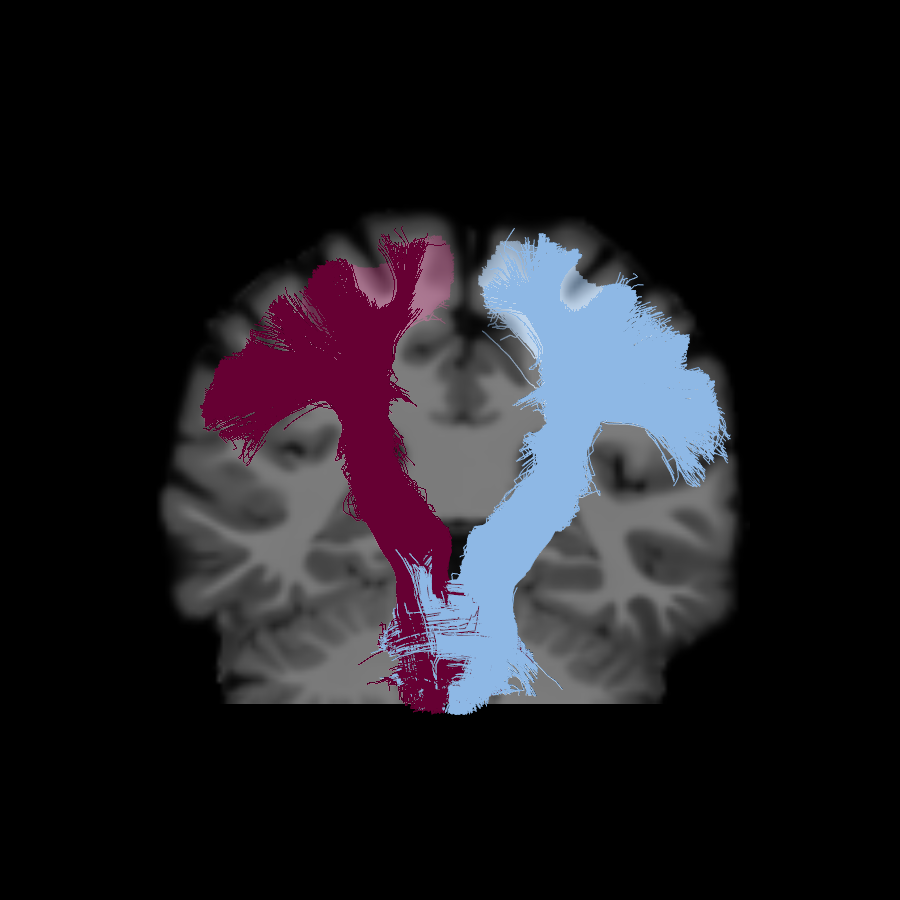} \\
\includegraphics[scale=0.95, trim=1.5in 2.25in 1.5in 2.6in, clip=true, width=0.205\linewidth, keepaspectratio=true]{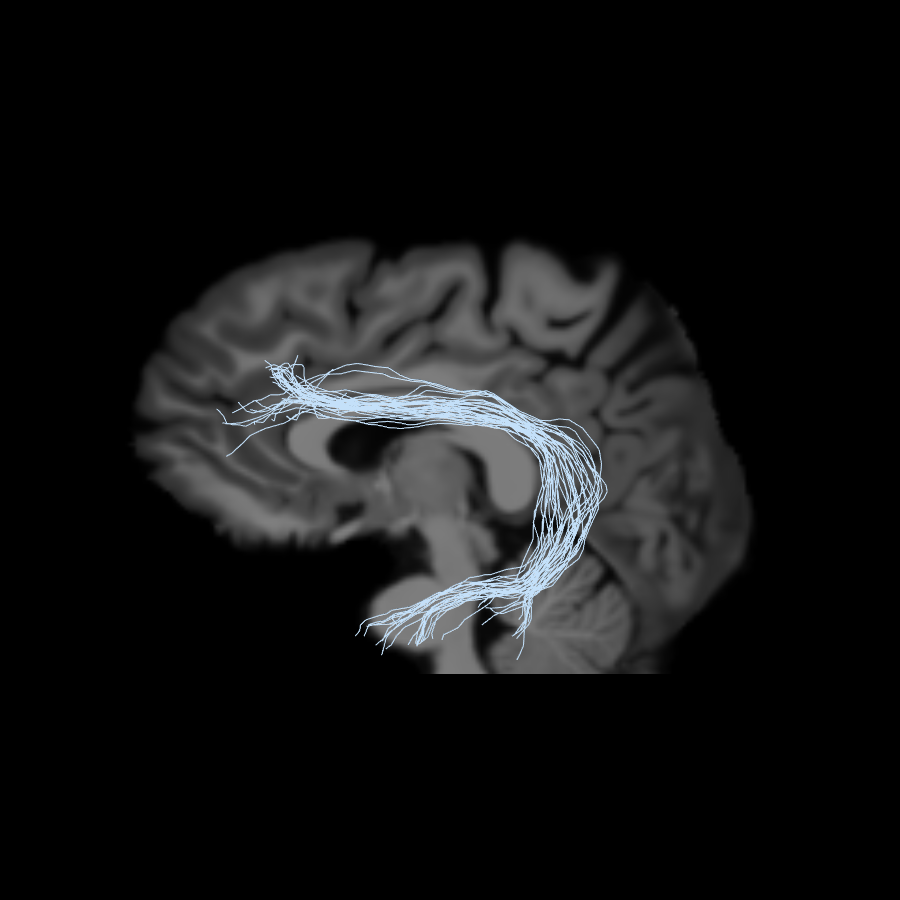} &
\includegraphics[scale=0.95, trim=1.5in 2.25in 1.5in 2.6in, clip=true, width=0.205\linewidth, keepaspectratio=true]{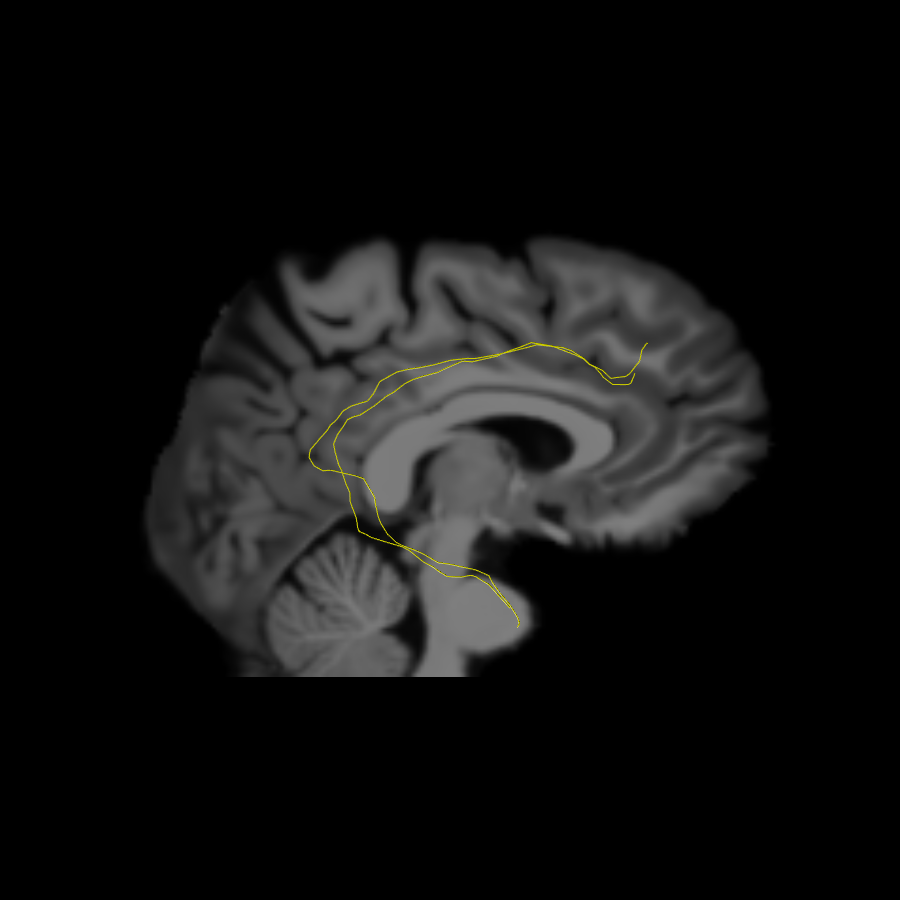} &
\includegraphics[scale=0.95, trim=1.5in 2.25in 1.5in 2.6in, clip=true, width=0.205\linewidth, keepaspectratio=true]{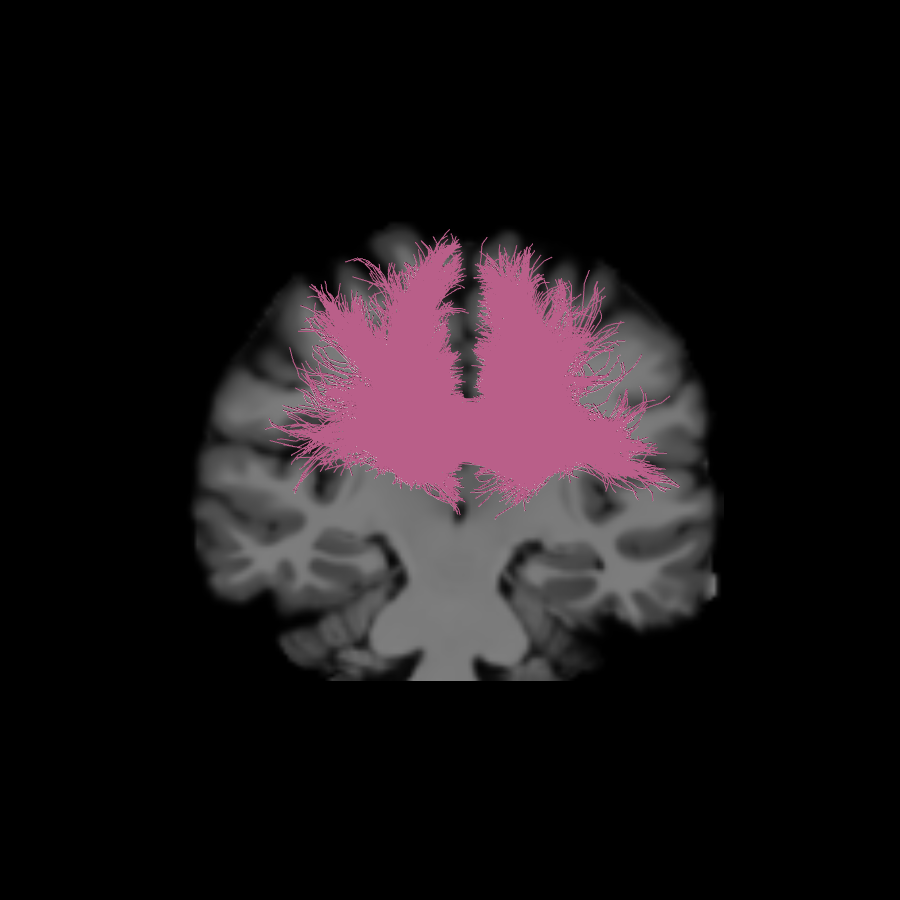} &
\includegraphics[scale=0.95, trim=2.25in 1.95in 2.25in 1.745in, clip=true, width=0.15\linewidth, keepaspectratio=true]{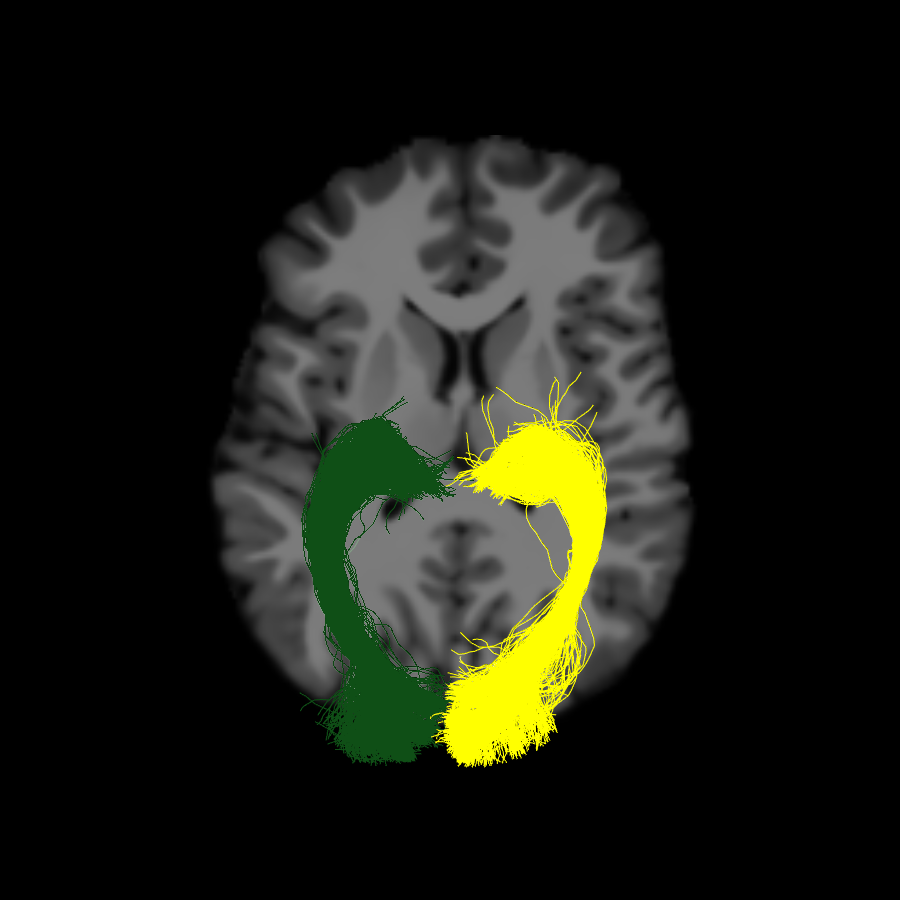} &
\includegraphics[scale=0.95, trim=1.5in 2.25in 1.5in 2.6in, clip=true, width=0.205\linewidth, keepaspectratio=true]{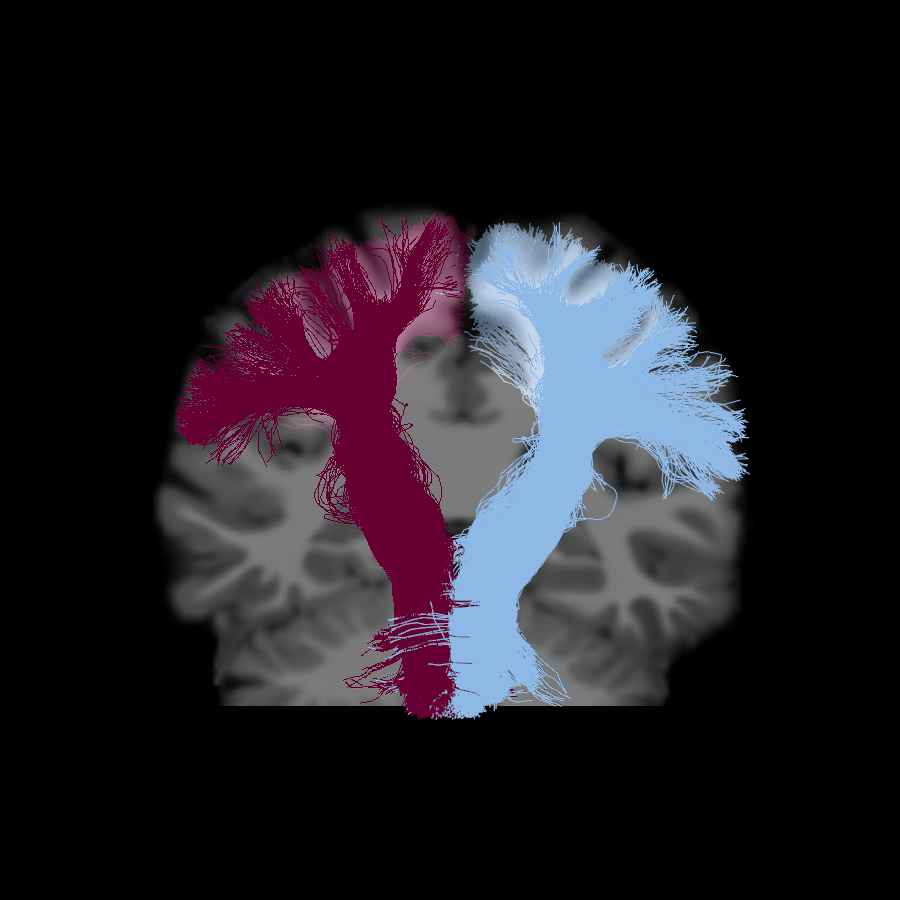} \\
\includegraphics[scale=0.95, trim=1.5in 2.25in 1.5in 2.6in, clip=true, width=0.205\linewidth, keepaspectratio=true]{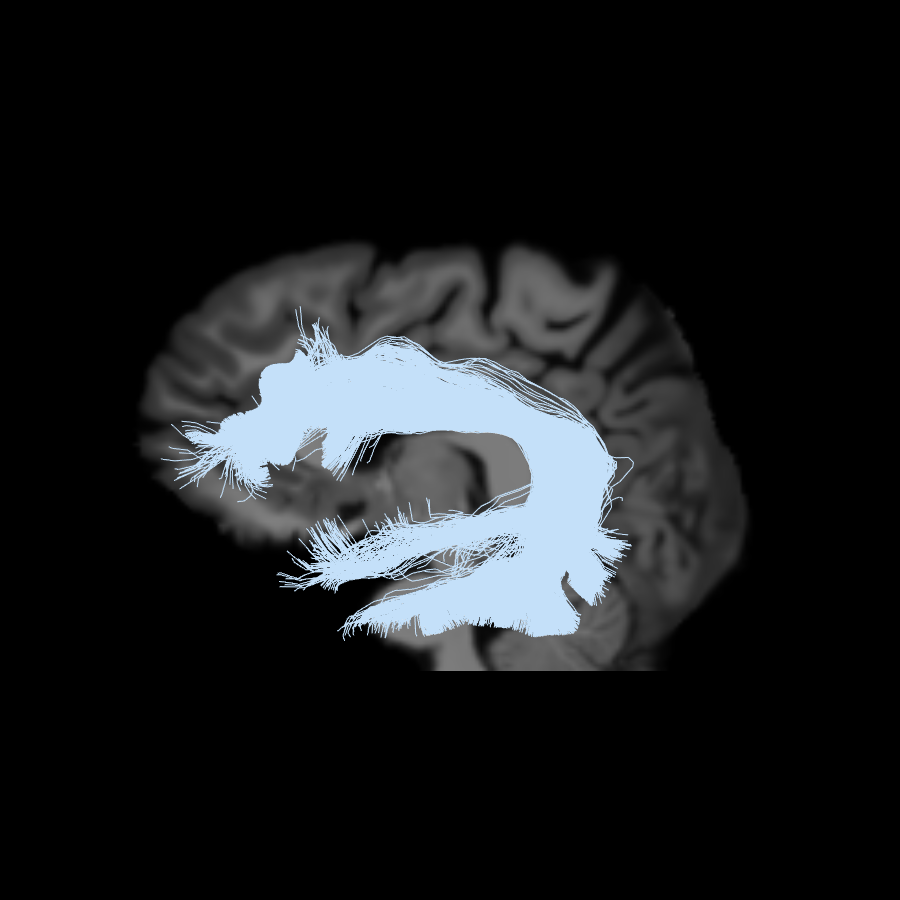} &
\includegraphics[scale=0.95, trim=1.5in 2.25in 1.5in 2.6in, clip=true, width=0.205\linewidth, keepaspectratio=true]{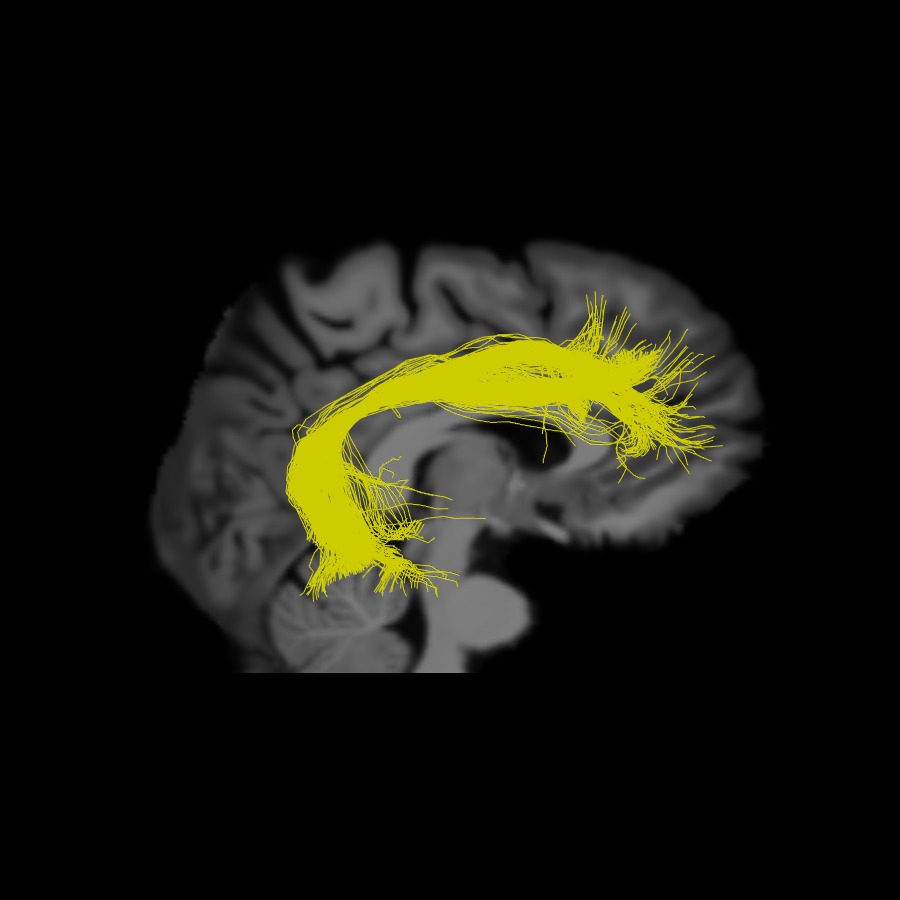} &
\includegraphics[scale=0.95, trim=1.5in 2.25in 1.5in 2.6in, clip=true, width=0.205\linewidth, keepaspectratio=true]{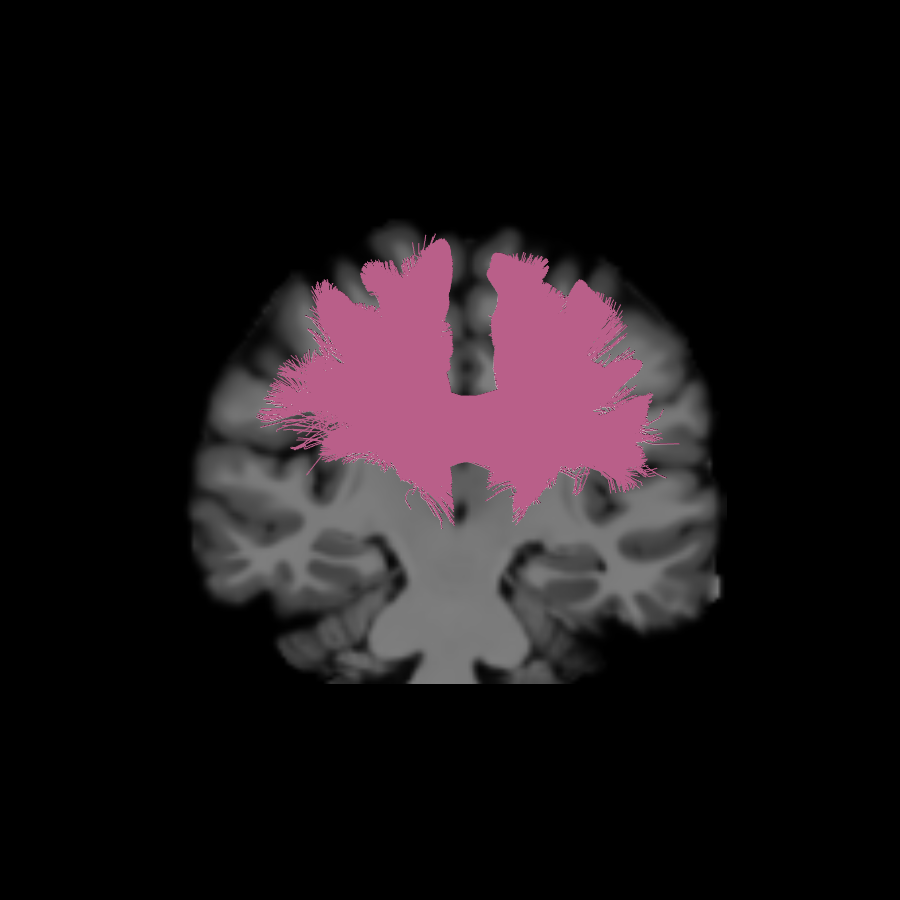} &
\includegraphics[scale=0.95, trim=2.25in 1.95in 2.25in 1.745in, clip=true, width=0.15\linewidth, keepaspectratio=true]{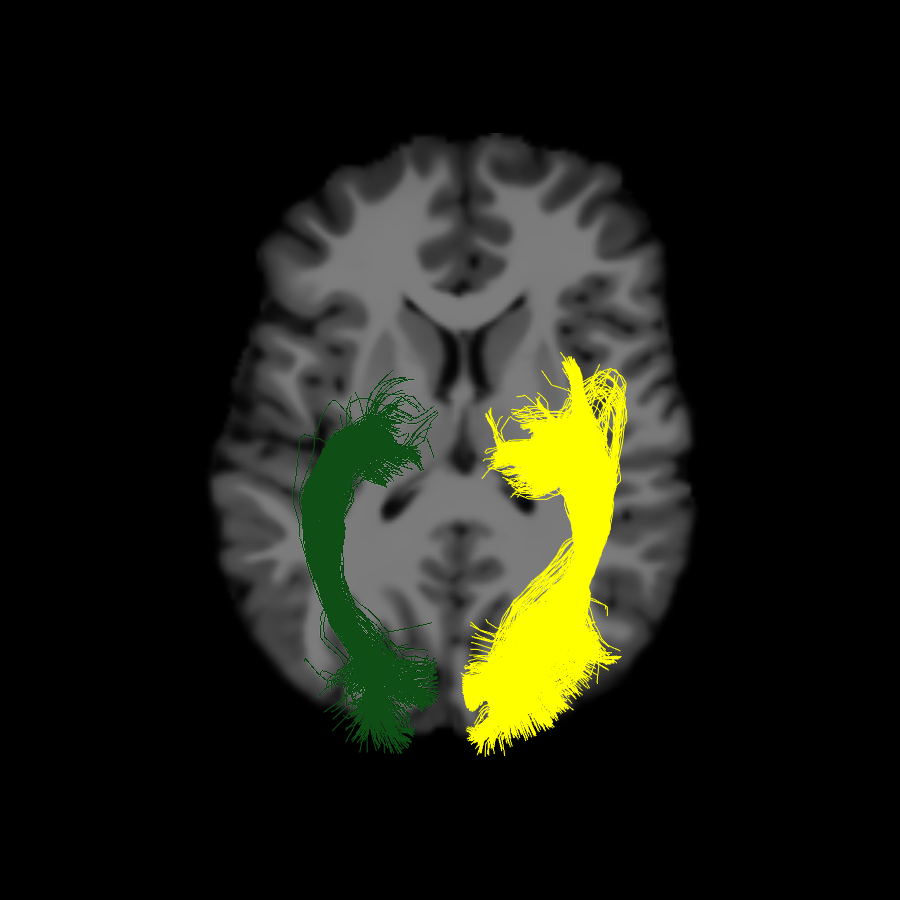} &
\includegraphics[scale=0.95, trim=1.5in 2.25in 1.5in 2.6in, clip=true, width=0.205\linewidth, keepaspectratio=true]{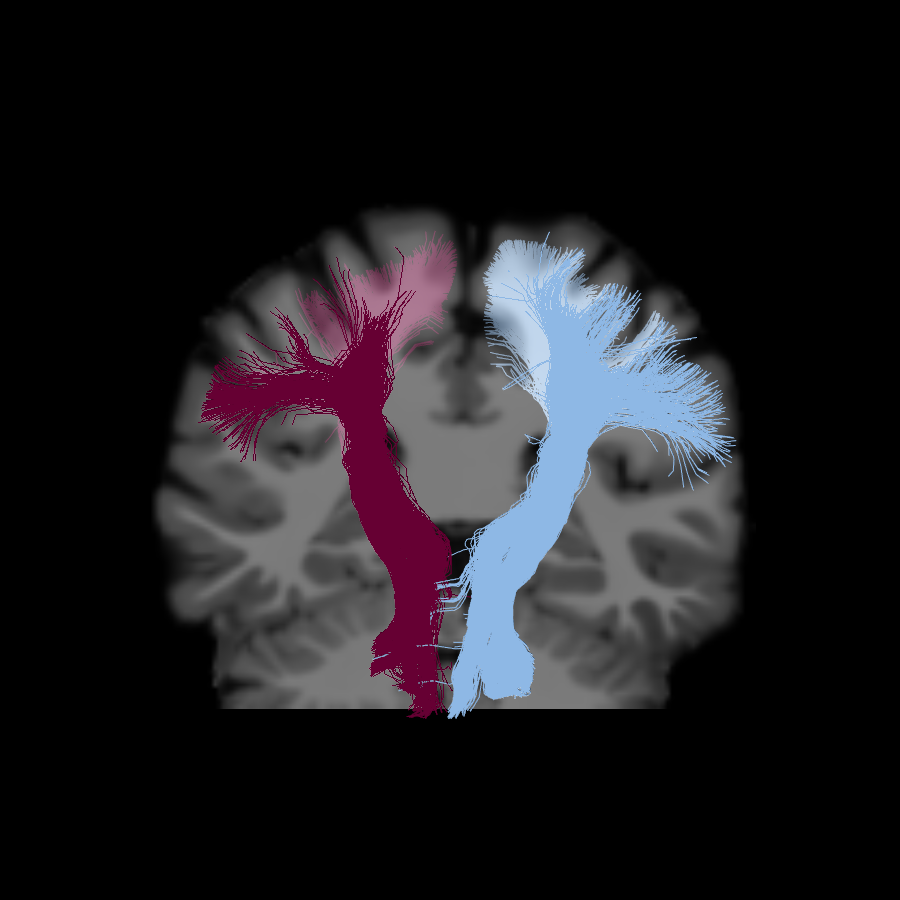} \\
\includegraphics[scale=0.95, trim=1.5in 2.25in 1.5in 2.6in, clip=true, width=0.205\linewidth, keepaspectratio=true]{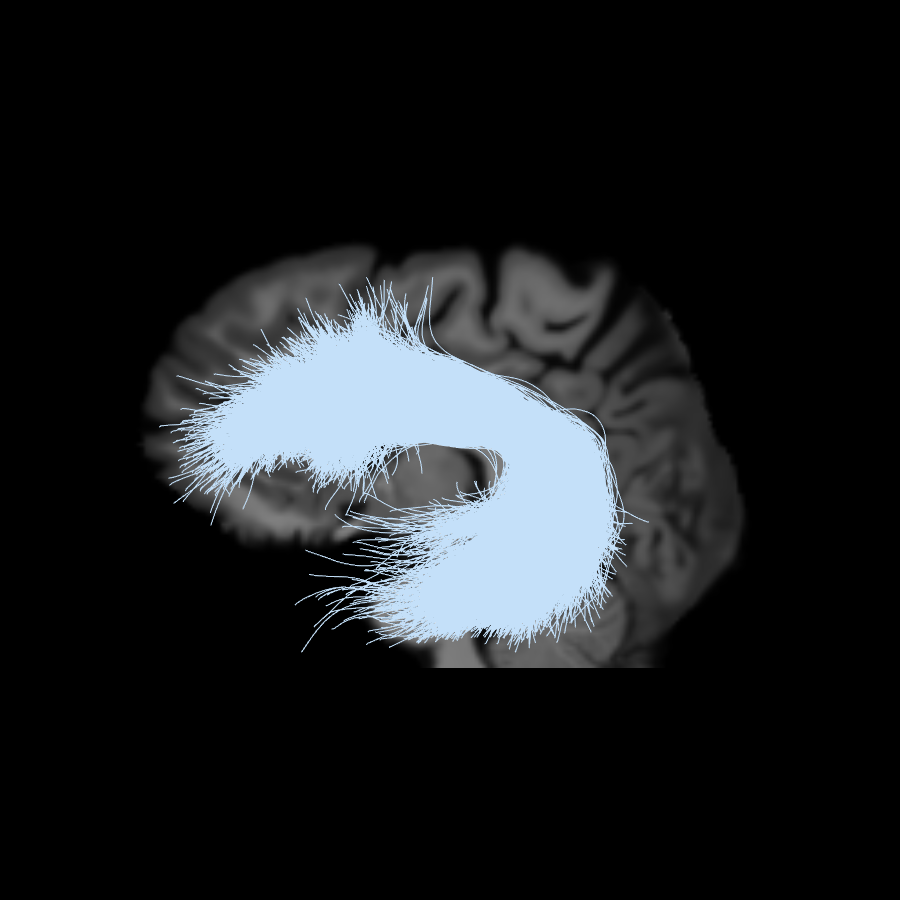} &
\includegraphics[scale=0.95, trim=1.5in 2.25in 1.5in 2.6in, clip=true, width=0.205\linewidth, keepaspectratio=true]{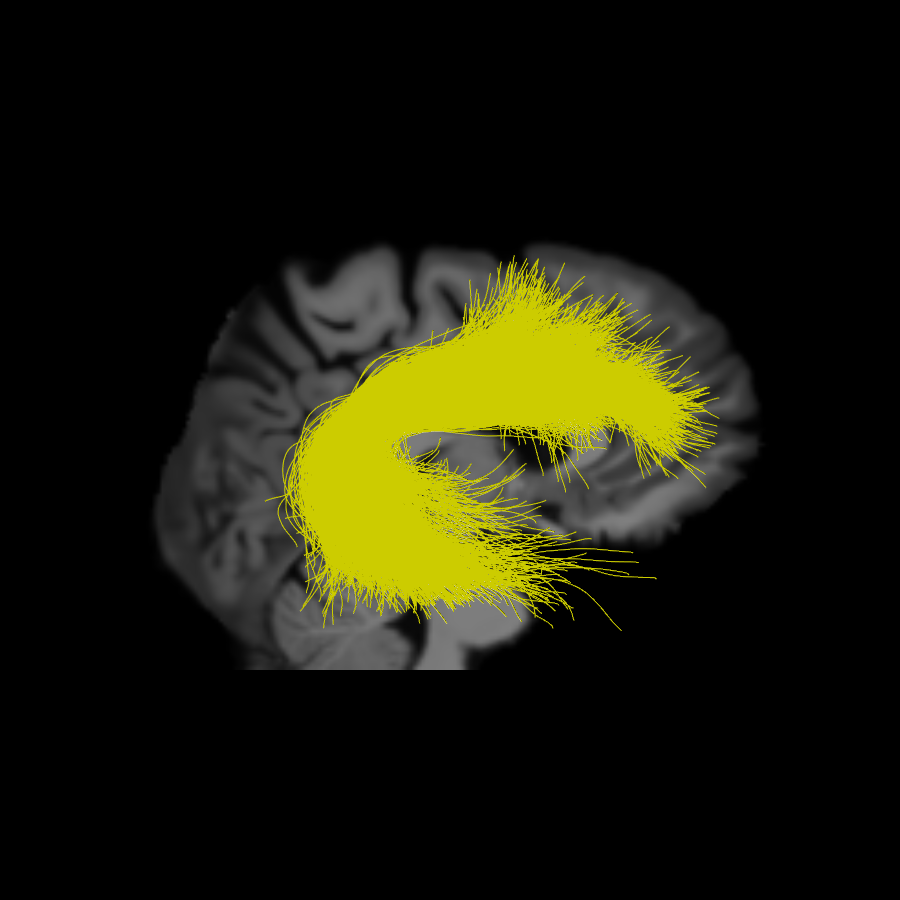} &
\includegraphics[scale=0.95, trim=1.5in 2.25in 1.5in 2.6in, clip=true, width=0.205\linewidth, keepaspectratio=true]{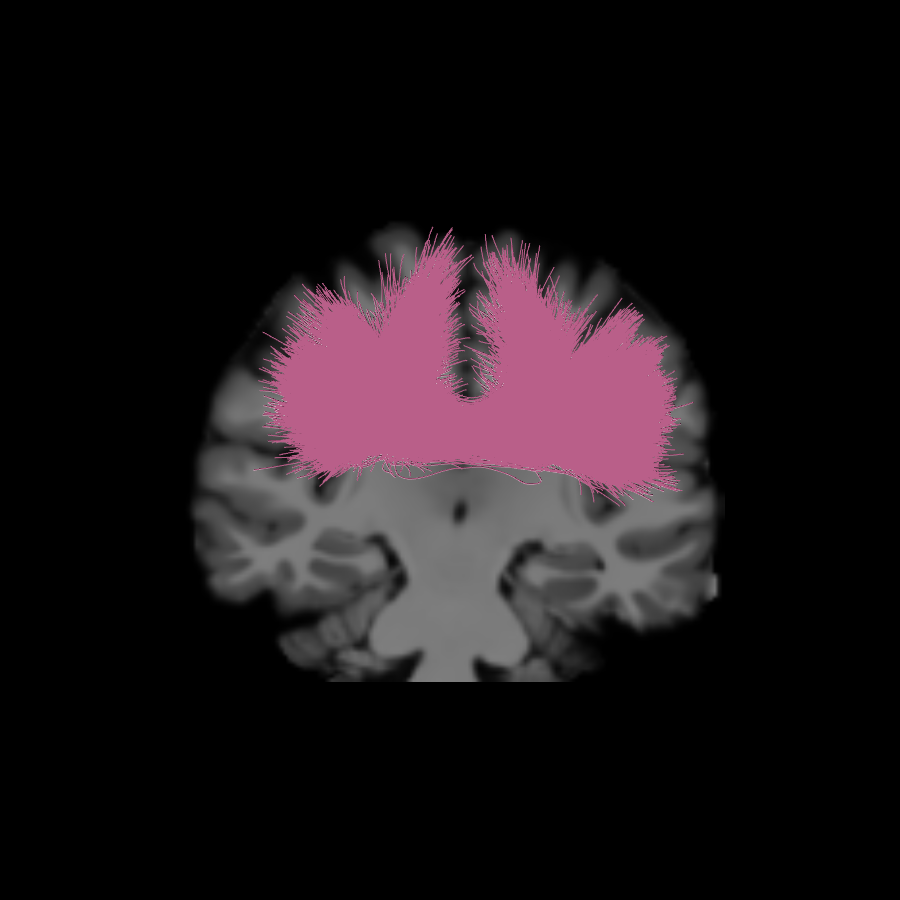} &
\includegraphics[scale=0.95, trim=2.25in 1.95in 2.25in 1.745in, clip=true, width=0.15\linewidth, keepaspectratio=true]{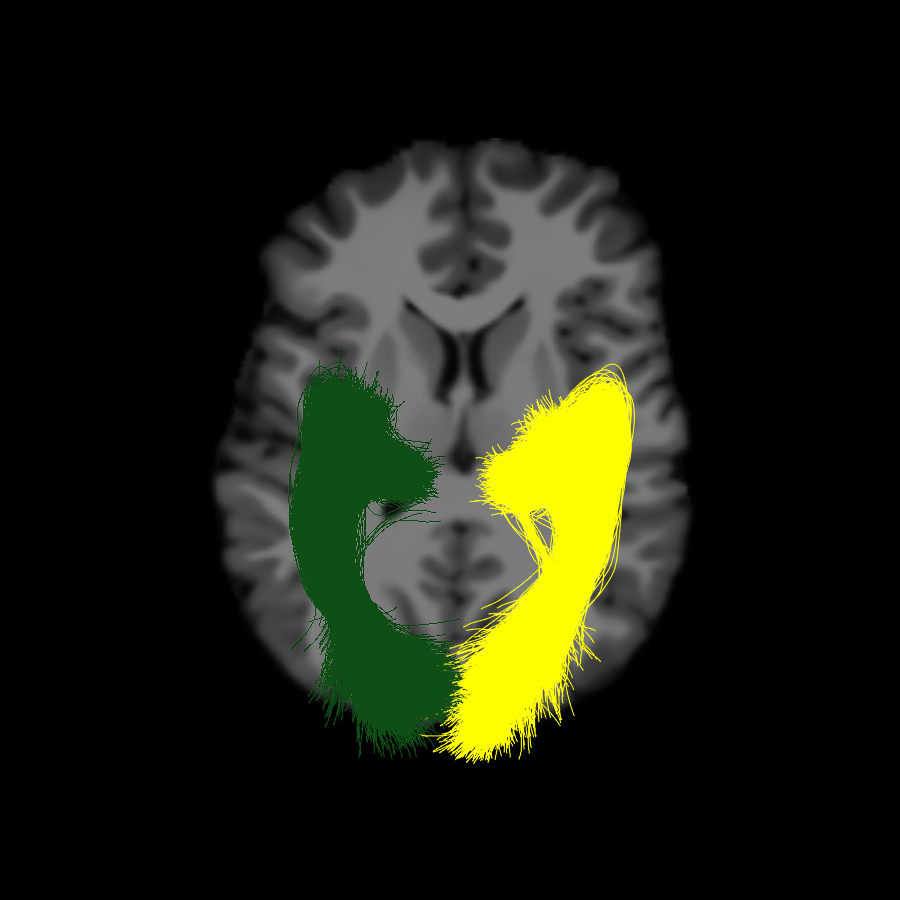} &
\includegraphics[scale=0.95, trim=1.5in 2.25in 1.5in 2.6in, clip=true, width=0.205\linewidth, keepaspectratio=true]{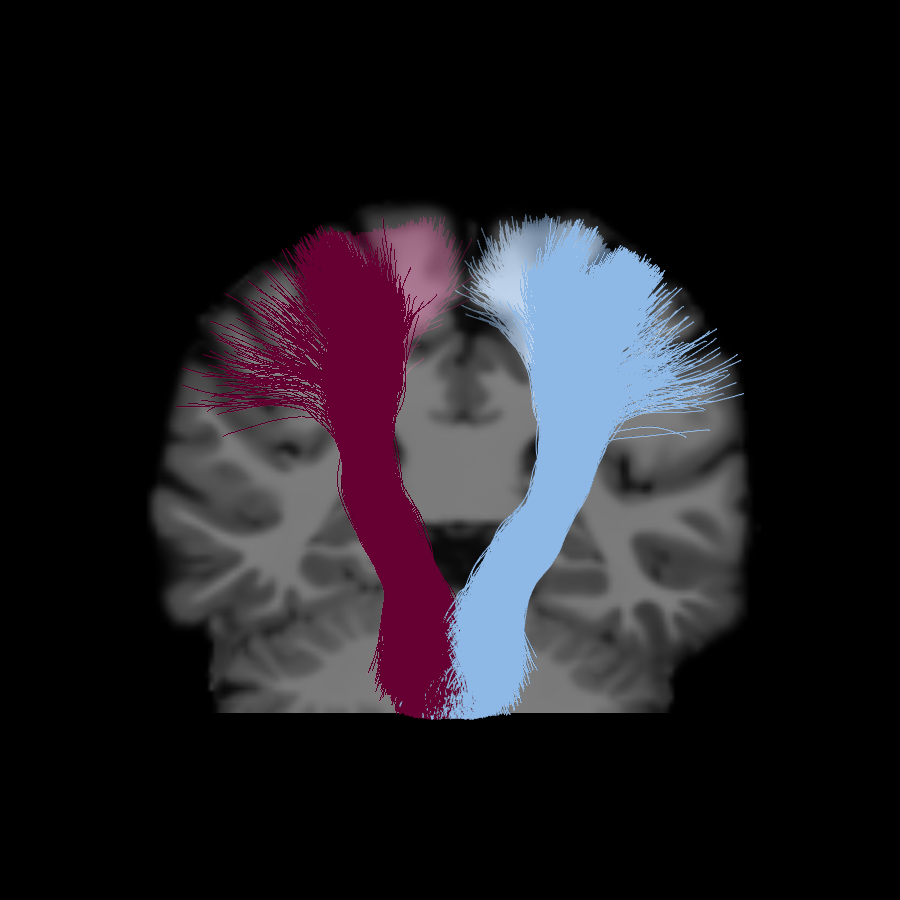} \\
\includegraphics[scale=0.95, trim=1.5in 2.25in 1.5in 2.6in, clip=true, width=0.205\linewidth, keepaspectratio=true]{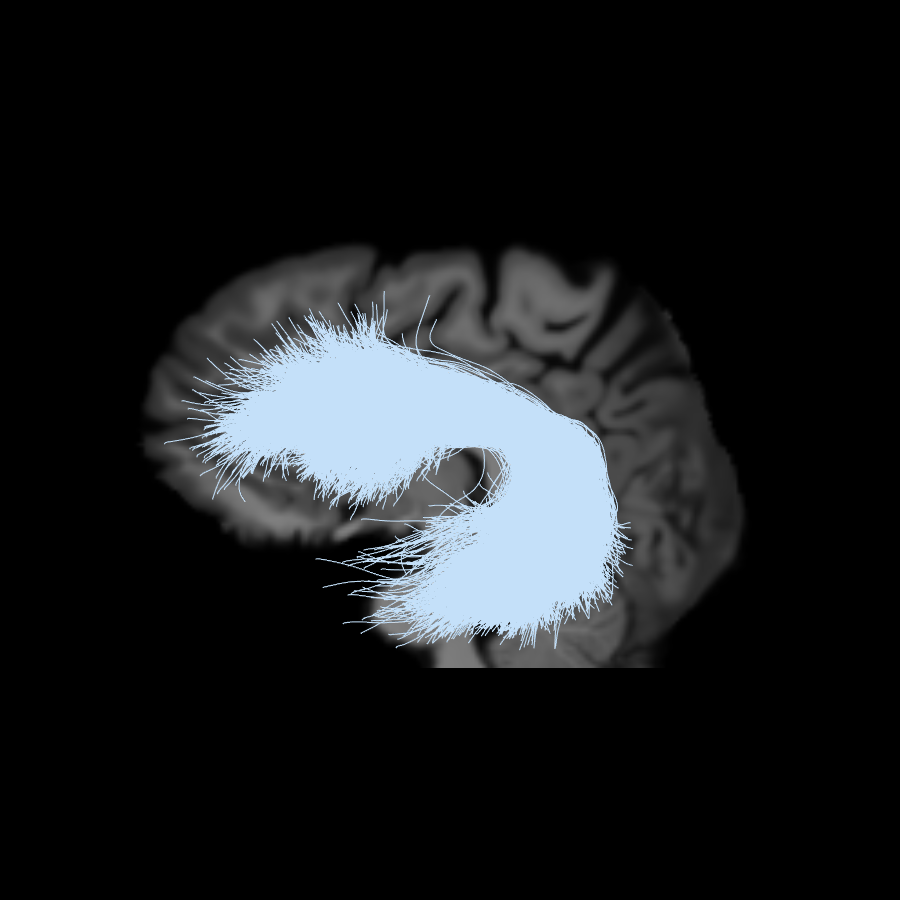} &
\includegraphics[scale=0.95, trim=1.5in 2.25in 1.5in 2.6in, clip=true, width=0.205\linewidth, keepaspectratio=true]{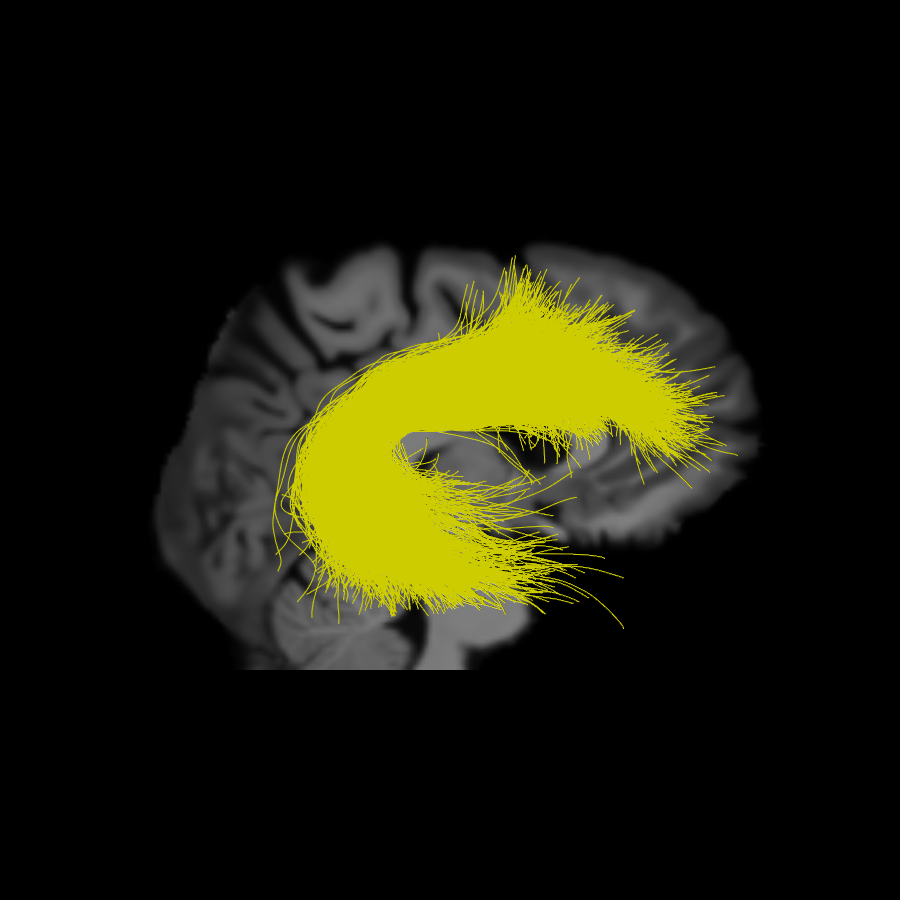} &
\includegraphics[scale=0.95, trim=1.5in 2.25in 1.5in 2.6in, clip=true, width=0.205\linewidth, keepaspectratio=true]{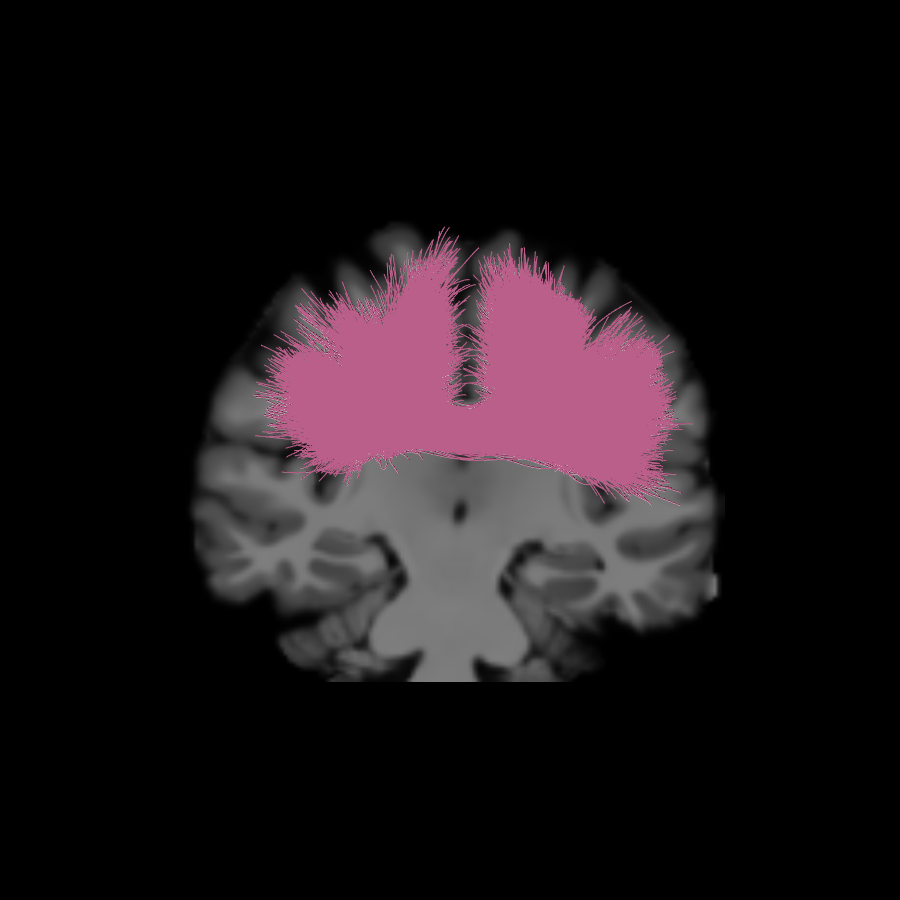} &
\includegraphics[scale=0.95, trim=2.25in 1.95in 2.25in 1.745in, clip=true, width=0.15\linewidth, keepaspectratio=true]{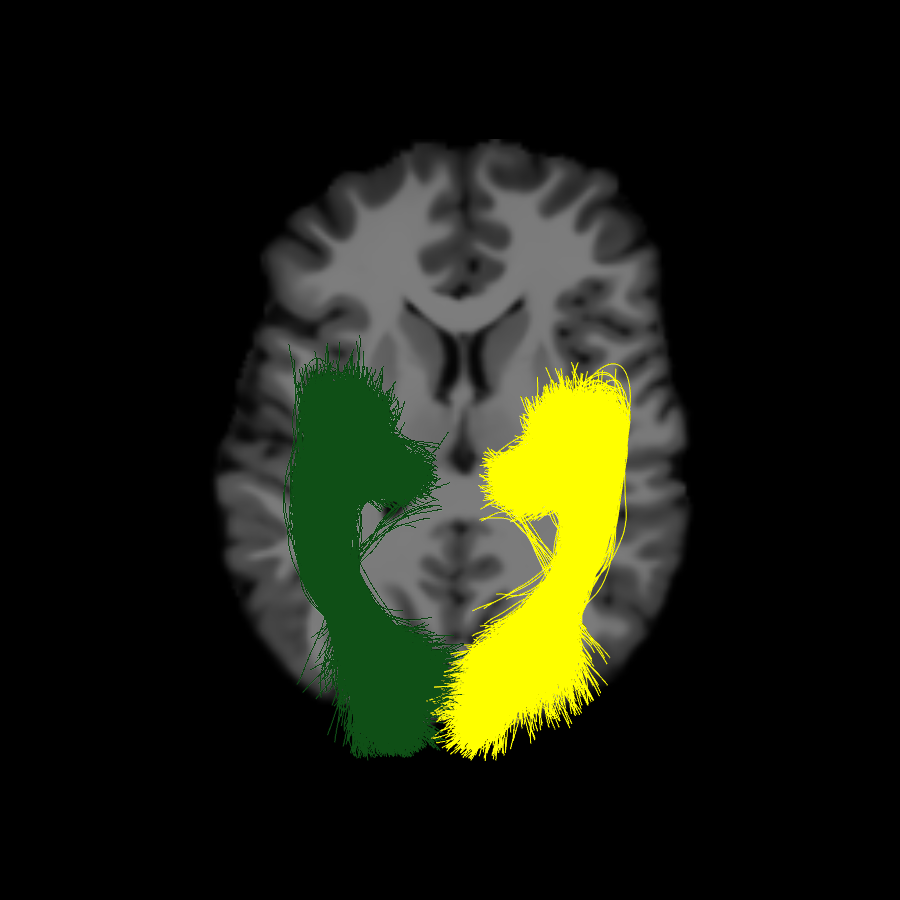} &
\includegraphics[scale=0.95, trim=1.5in 2.25in 1.5in 2.6in, clip=true, width=0.205\linewidth, keepaspectratio=true]{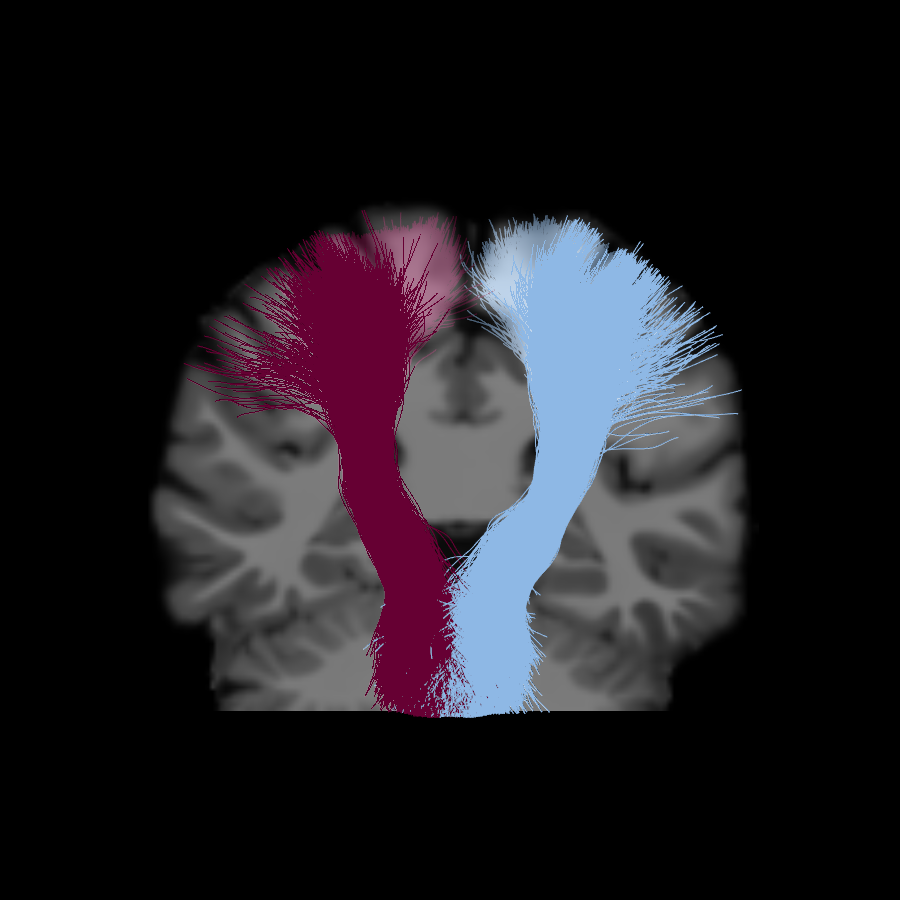} \\
\textbf{(i)} & \textbf{(ii)} & \textbf{(iii)} & \textbf{(iv)} & \textbf{(v)} \\
\end{tabular}
\end{subtable}
\caption{\label{fig:tractoinferno_subj}Baseline tractography and GESTA tractography for the TractoInferno dataset subject \textit{s1}. From top to bottom: deterministic; probabilistic; PFT; SET; GESTA-Det; GESTA-Prob tractography. Latent-generated streamlines are evaluated with the \textit{ADGC}\textsubscript{R} criterion. Bundles: (i) AF\_L; (ii) AF\_R; (iii) CC\_Fr\_1; (iv) OR\_ML; (v) PYT. Views have been chosen to best visualize the bundles.}
\end{figure*}

\begin{figure*}[!htbp]\ContinuedFloat
\begin{subtable}{\linewidth}
\caption{TractoInferno baselines \vs generative tractography. Subject \textit{s2}.}
\centering
\setlength{\tabcolsep}{0pt}
\begin{tabular}{ccccc}
\includegraphics[scale=0.95, trim=1.5in 2.25in 1.5in 2.6in, clip=true, width=0.205\linewidth, keepaspectratio=true]{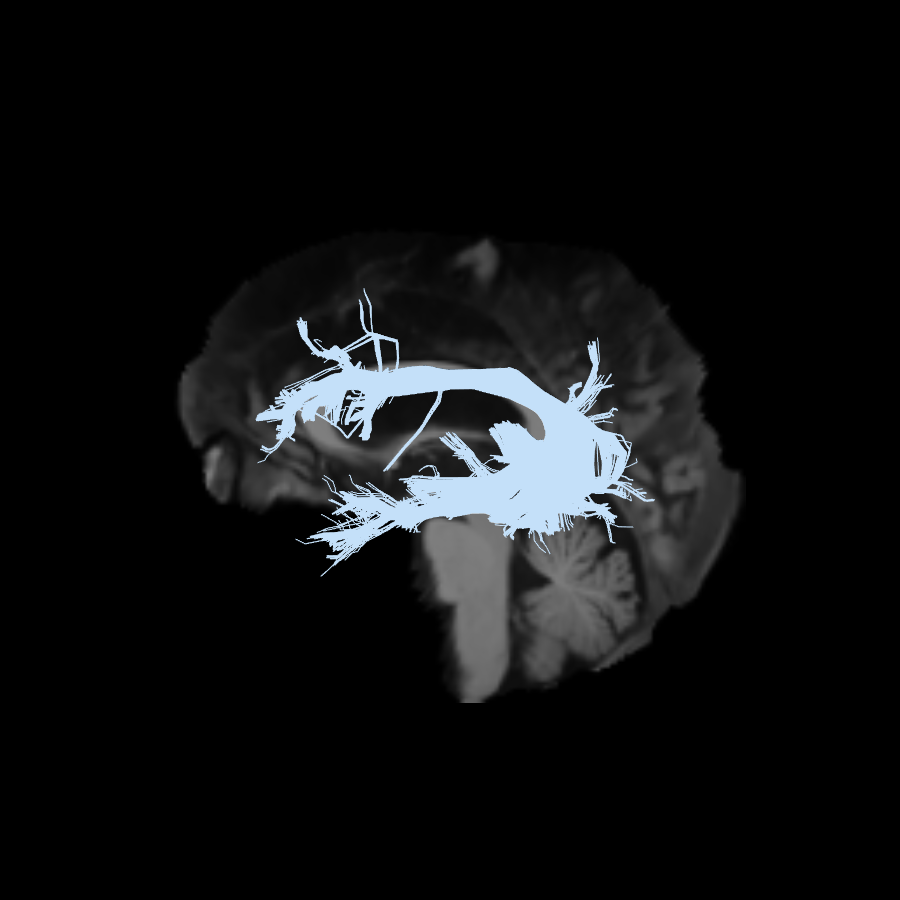} &
\includegraphics[scale=0.95, trim=1.5in 2.25in 1.5in 2.6in, clip=true, width=0.205\linewidth, keepaspectratio=true]{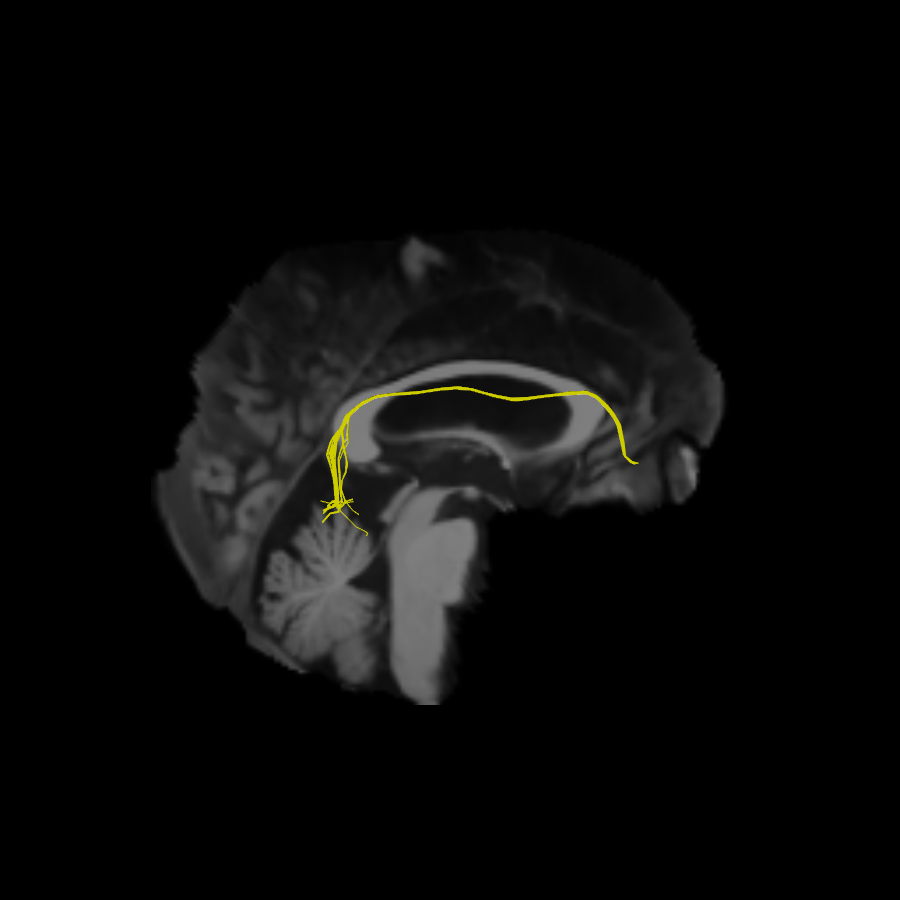} &
\includegraphics[scale=0.95, trim=1.5in 2.25in 1.5in 2.6in, clip=true, width=0.205\linewidth, keepaspectratio=true]{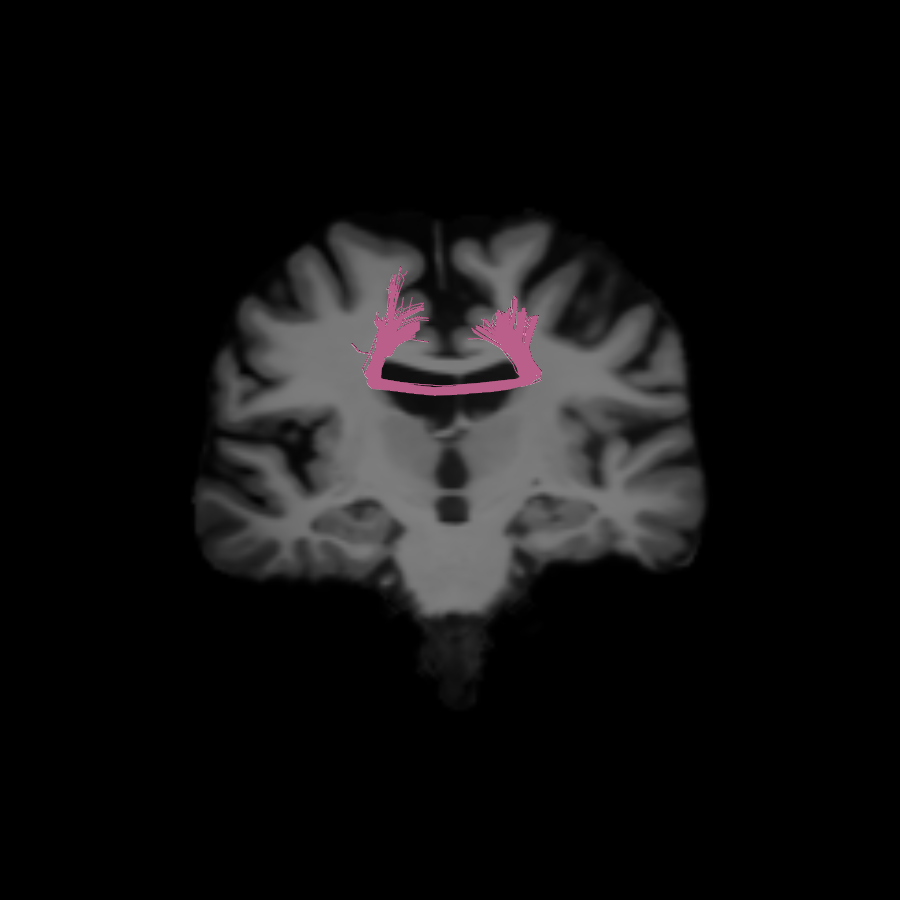} &
\includegraphics[scale=0.95, trim=2.25in 1.95in 2.25in 1.745in, clip=true, width=0.15\linewidth, keepaspectratio=true]{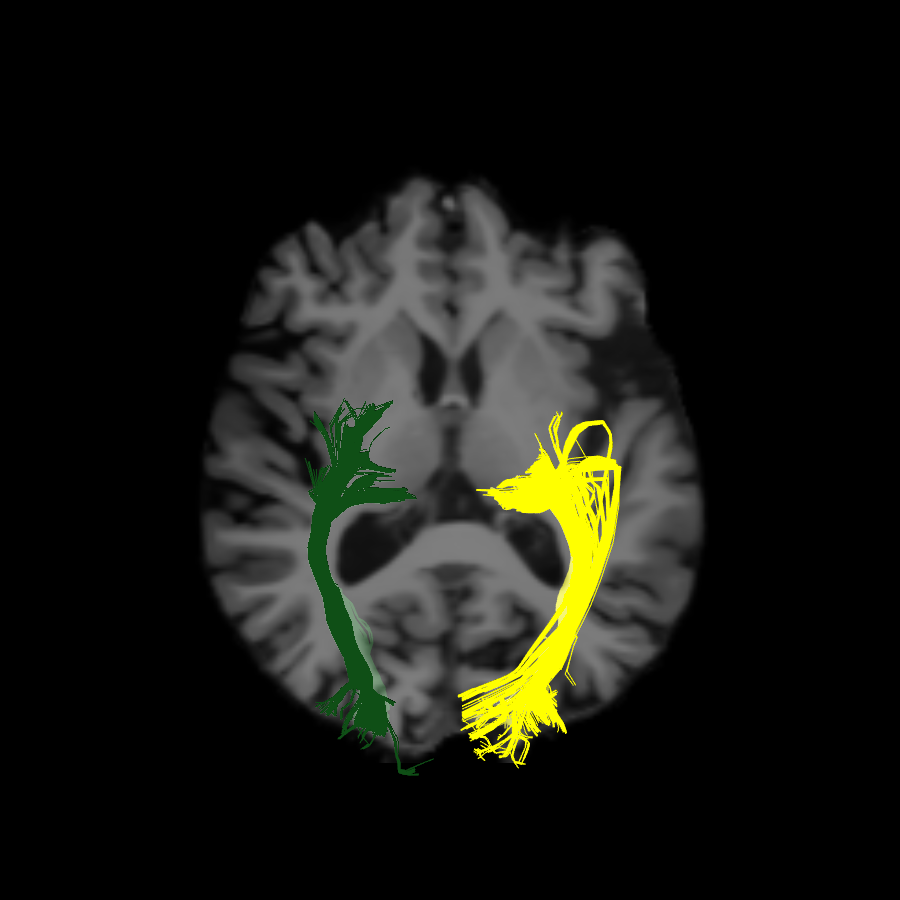} &
\includegraphics[scale=0.95, trim=1.5in 2.25in 1.5in 2.6in, clip=true, width=0.205\linewidth, keepaspectratio=true]{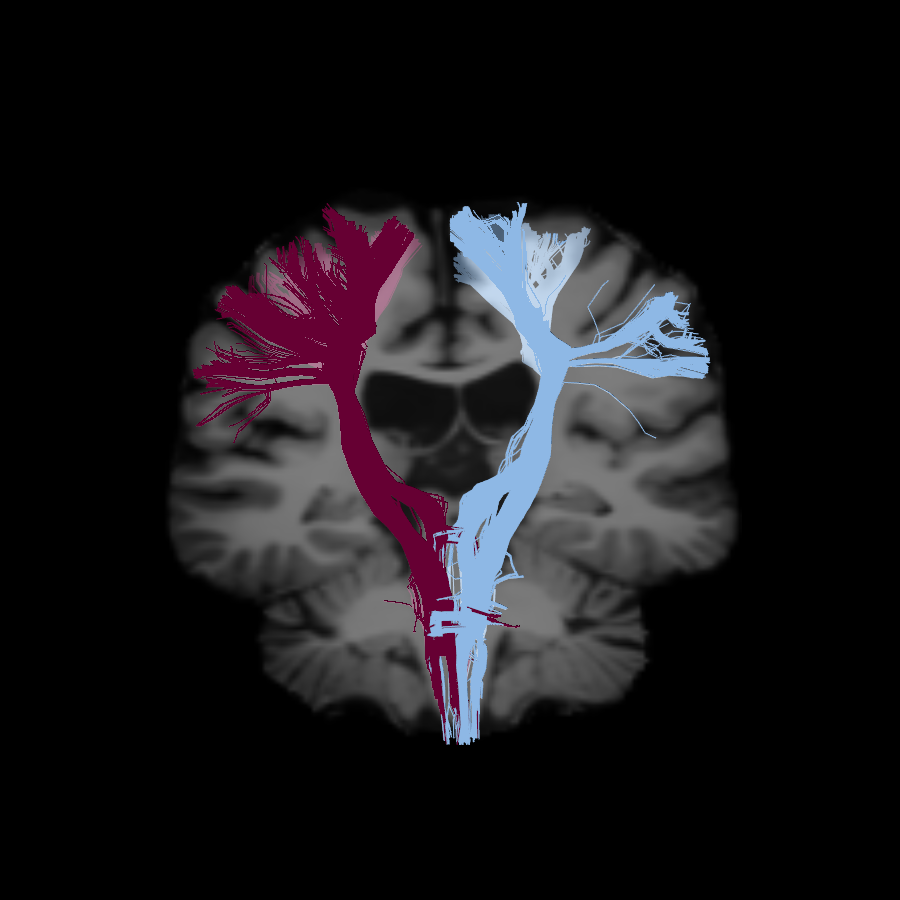} \\
\includegraphics[scale=0.95, trim=1.5in 2.25in 1.5in 2.6in, clip=true, width=0.205\linewidth, keepaspectratio=true]{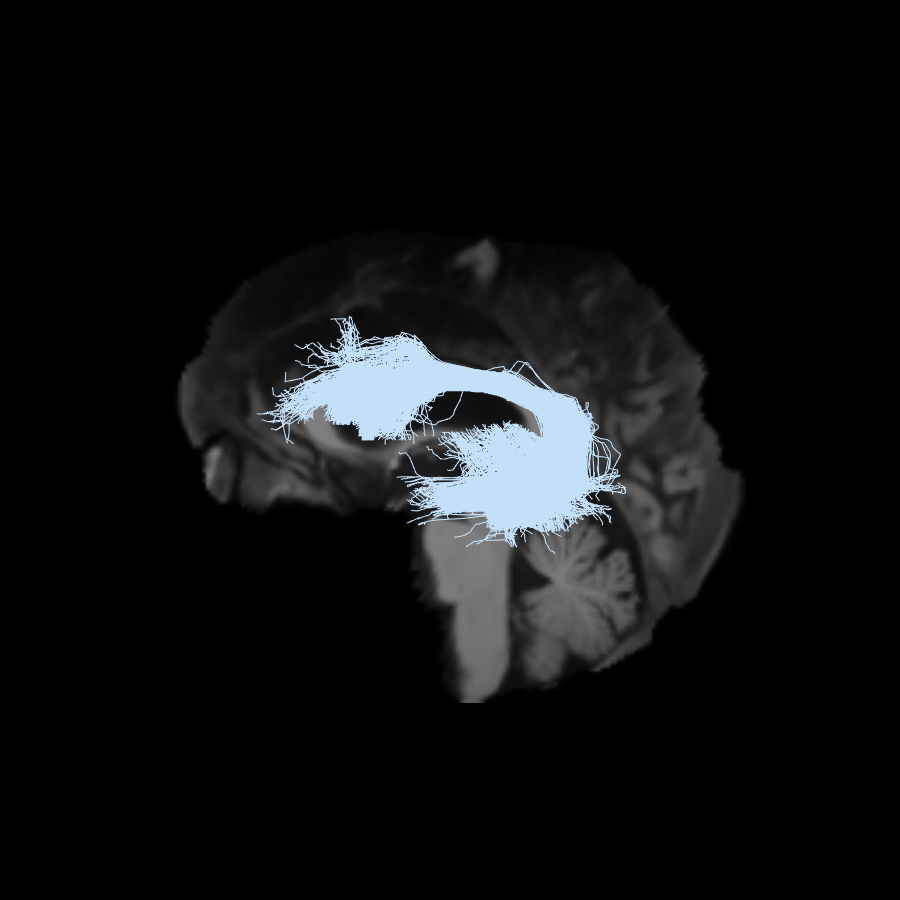} &
\includegraphics[scale=0.95, trim=1.2in 2.0in 1.15in 2.325in, clip=true, width=0.205\linewidth, keepaspectratio=true]{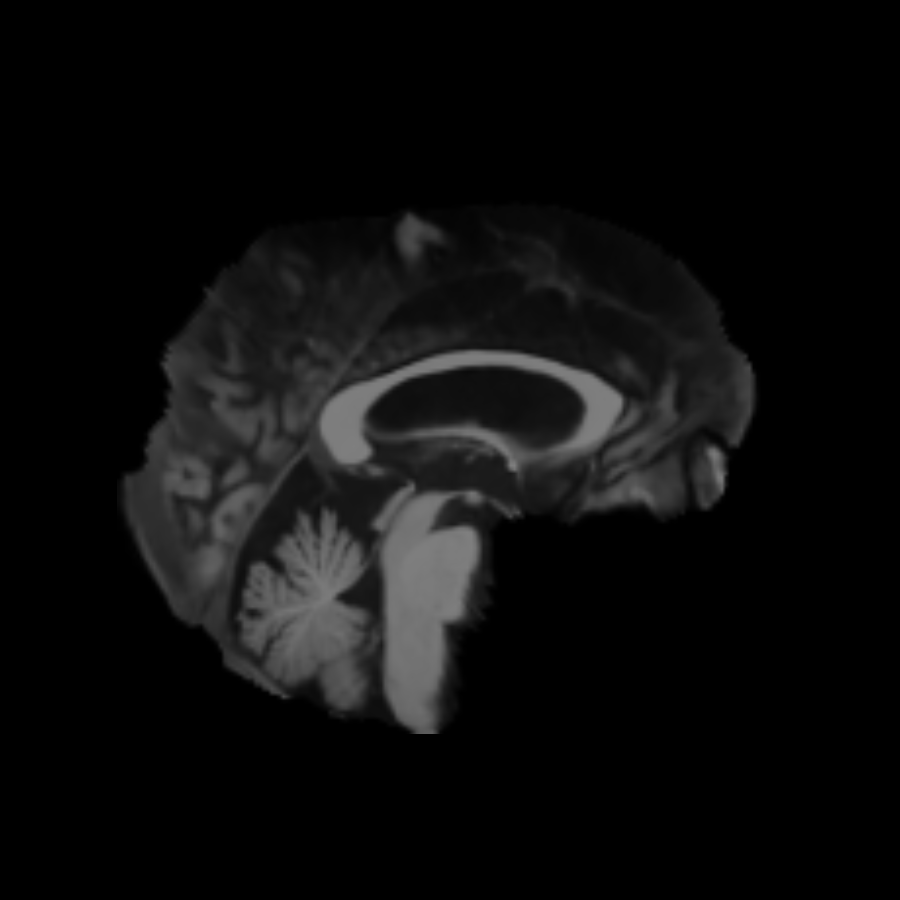} &
\includegraphics[scale=0.95, trim=1.3in 2.5in 1.3in 2.025in, clip=true, width=0.205\linewidth, keepaspectratio=true]{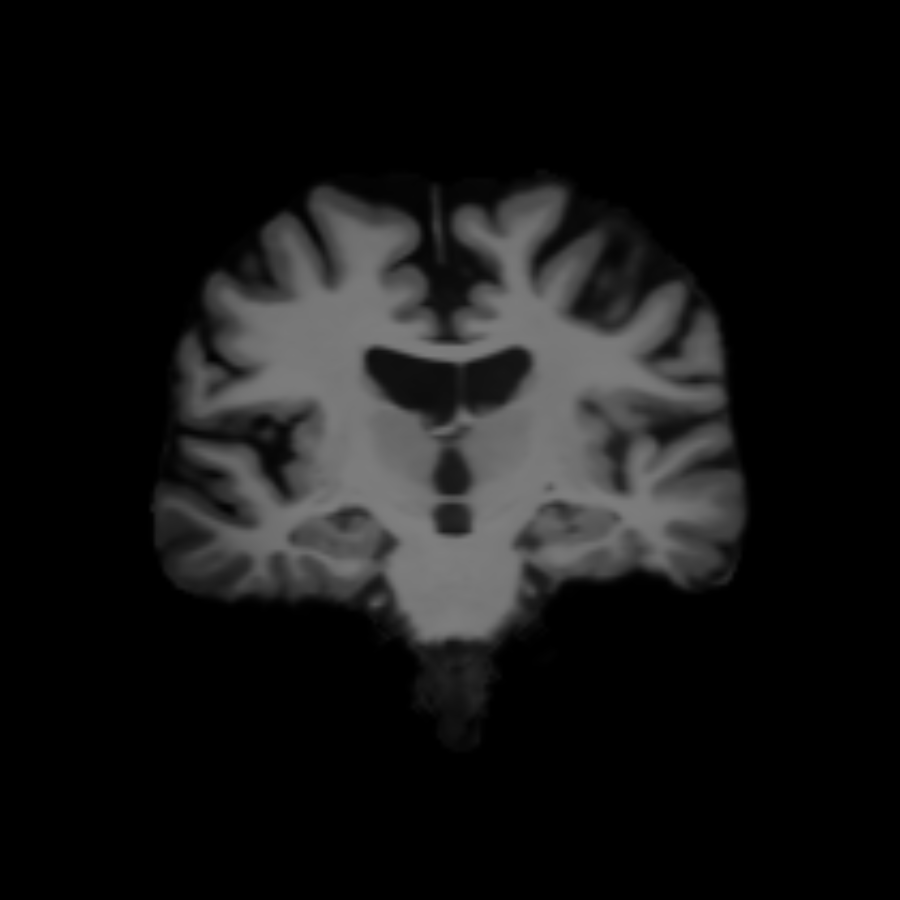} &
\includegraphics[scale=0.95, trim=2.25in 1.95in 2.25in 1.745in, clip=true, width=0.15\linewidth, keepaspectratio=true]{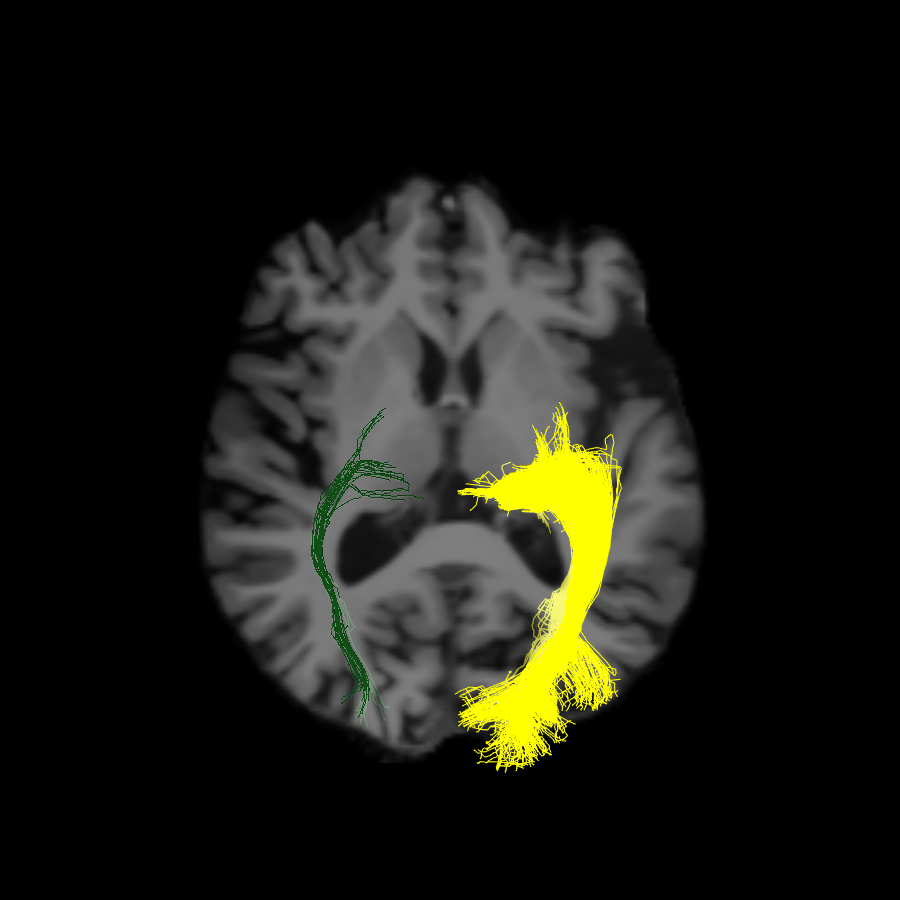} &
\includegraphics[scale=0.95, trim=1.5in 2.25in 1.5in 2.6in, clip=true, width=0.205\linewidth, keepaspectratio=true]{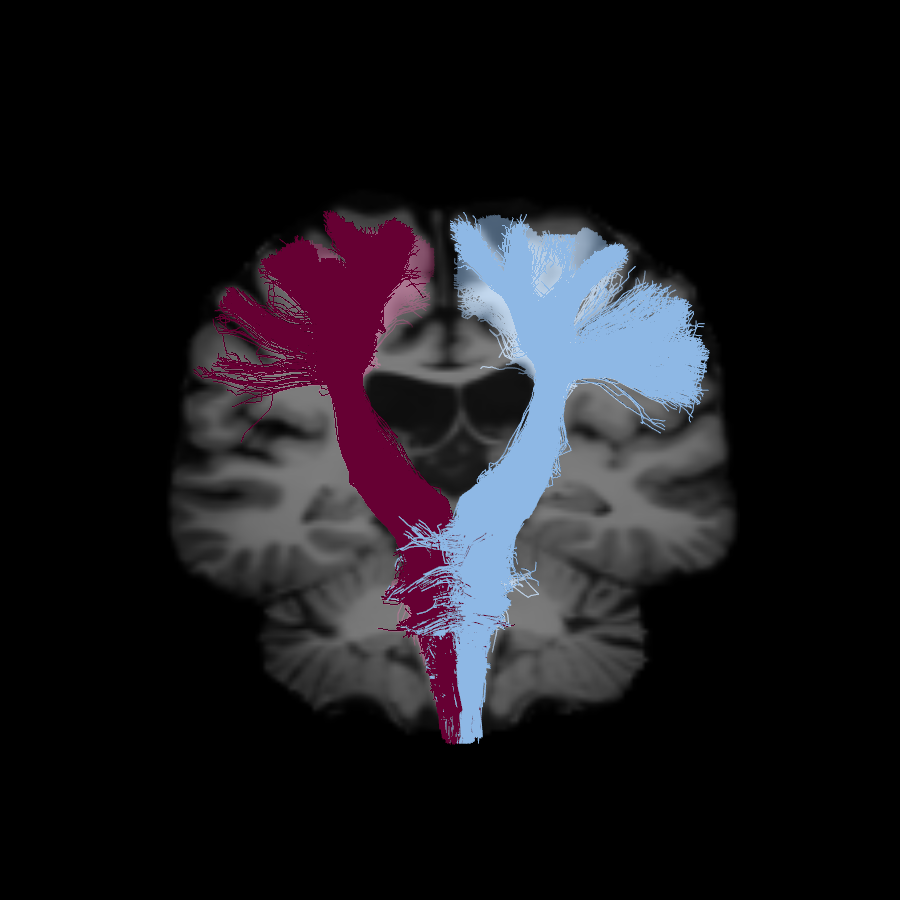} \\
\includegraphics[scale=0.95, trim=1.5in 2.25in 1.5in 2.6in, clip=true, width=0.205\linewidth, keepaspectratio=true]{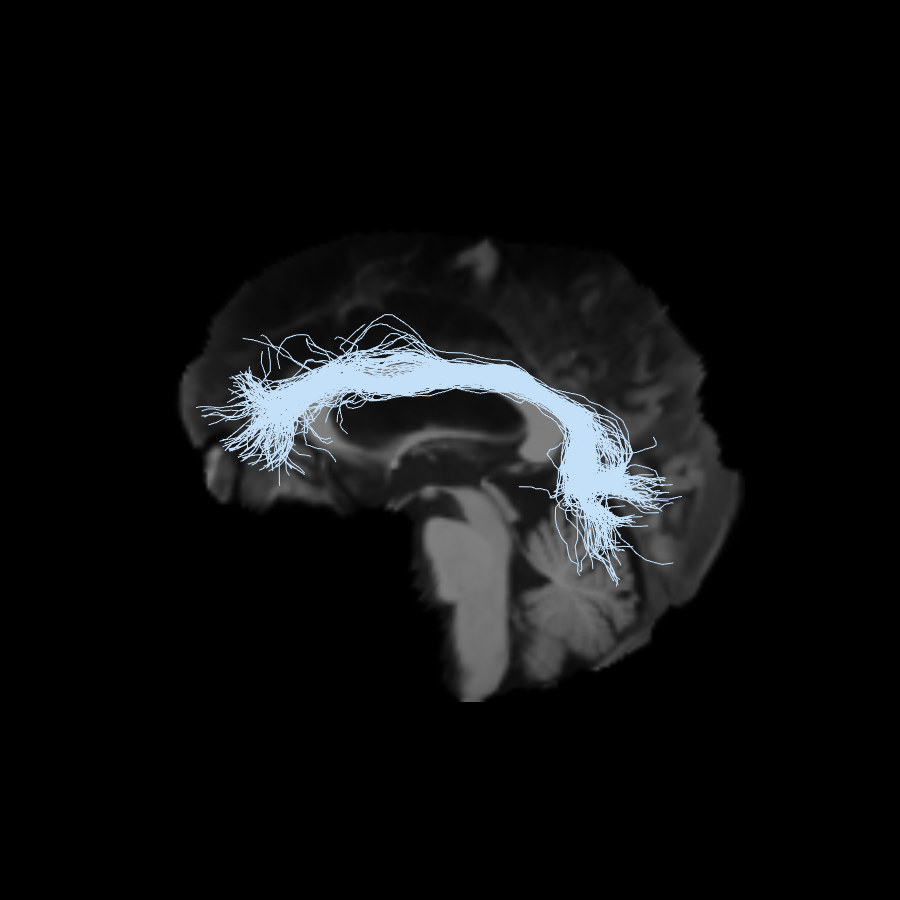} &
\includegraphics[scale=0.95, trim=1.5in 2.25in 1.5in 2.6in, clip=true, width=0.205\linewidth, keepaspectratio=true]{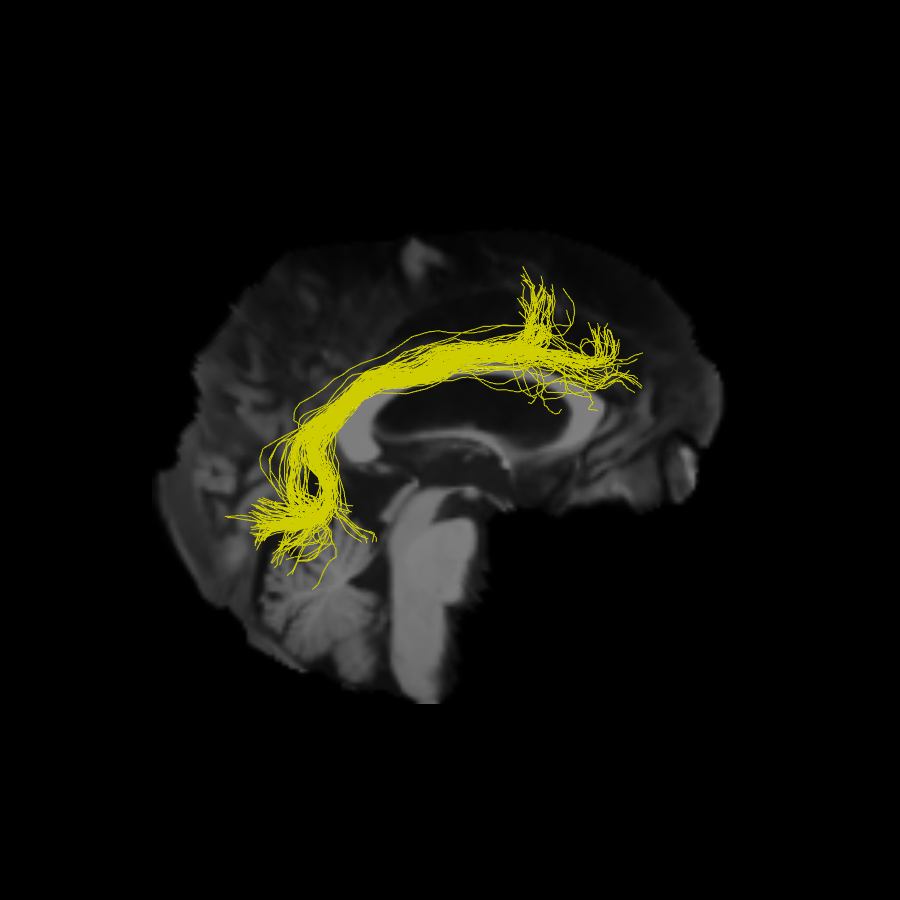} &
\includegraphics[scale=0.95, trim=1.5in 2.25in 1.5in 2.6in, clip=true, width=0.205\linewidth, keepaspectratio=true]{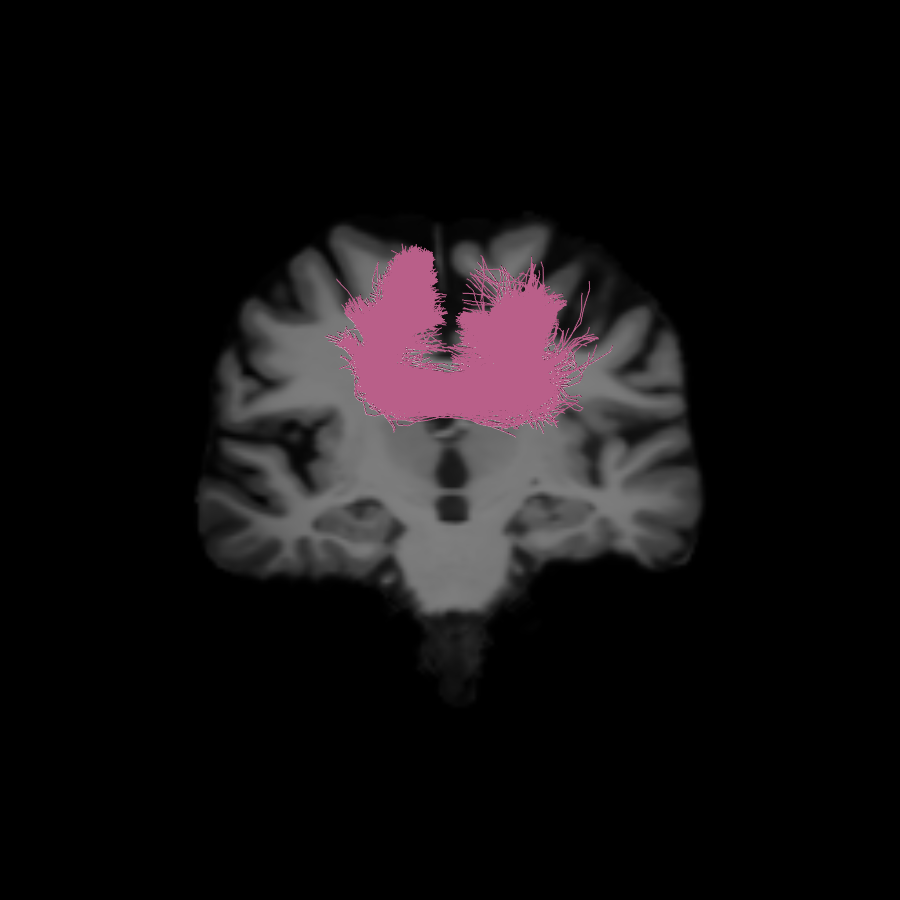} &
\includegraphics[scale=0.95, trim=2.25in 1.95in 2.25in 1.745in, clip=true, width=0.15\linewidth, keepaspectratio=true]{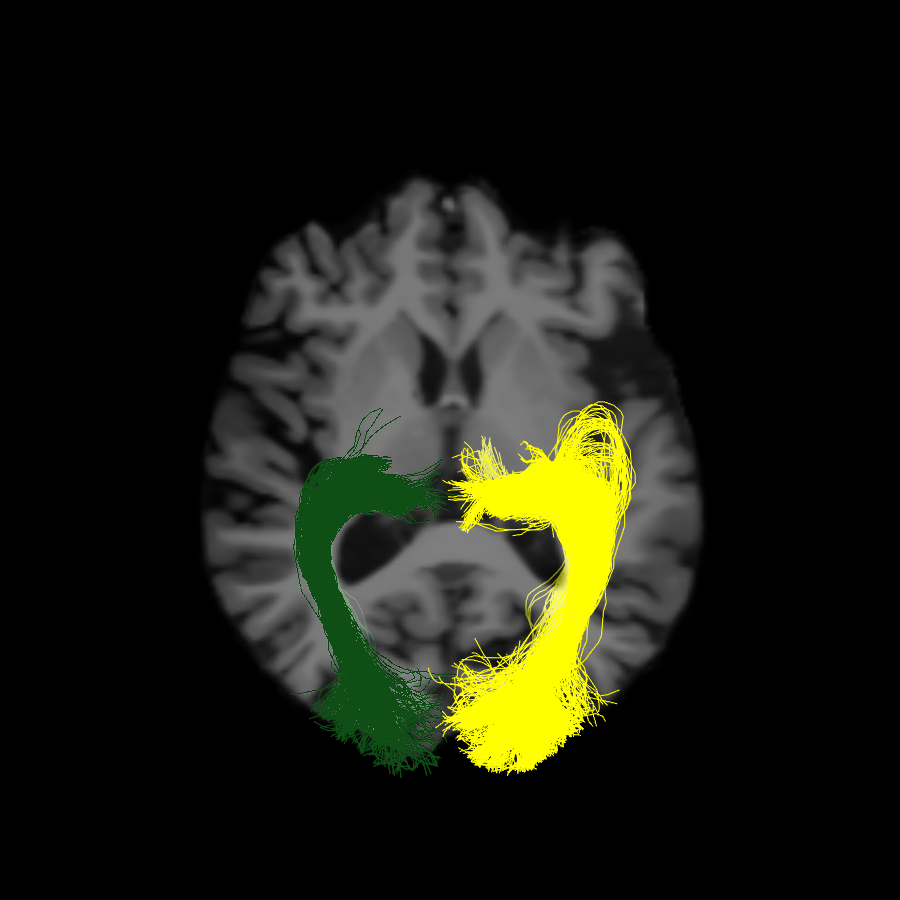} &
\includegraphics[scale=0.95, trim=1.5in 2.25in 1.5in 2.6in, clip=true, width=0.205\linewidth, keepaspectratio=true]{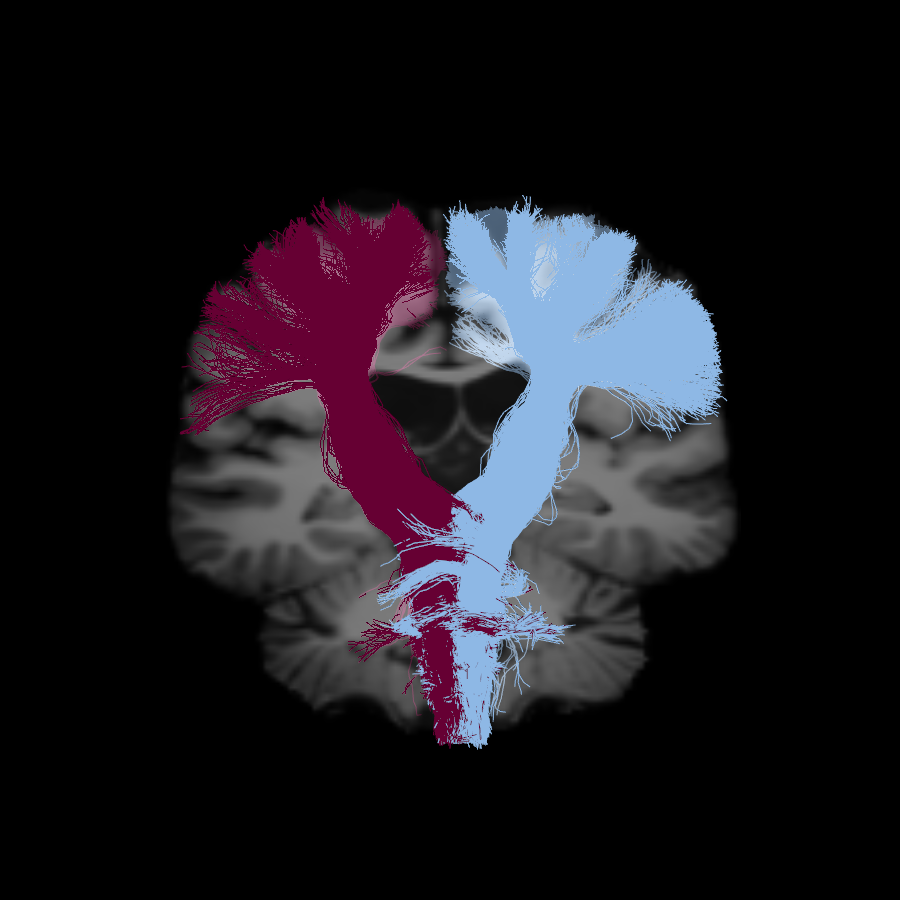} \\
\includegraphics[scale=0.95, trim=1.5in 2.25in 1.5in 2.6in, clip=true, width=0.205\linewidth, keepaspectratio=true]{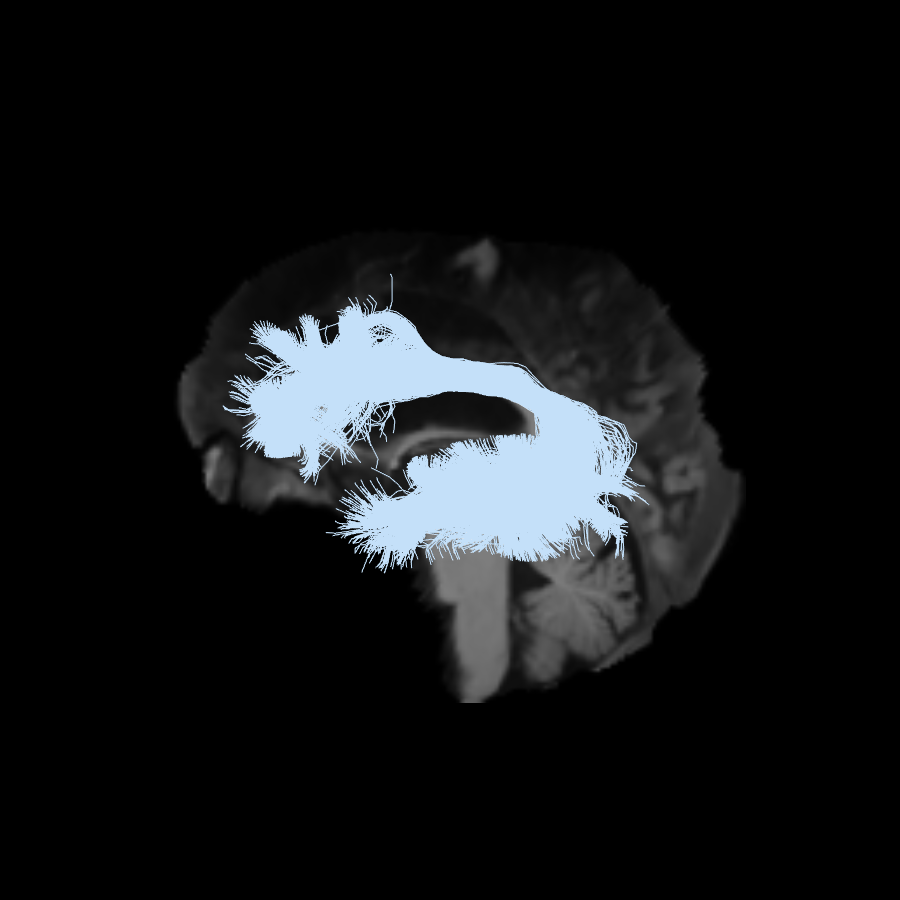} &
\includegraphics[scale=0.95, trim=1.5in 2.25in 1.5in 2.6in, clip=true, width=0.205\linewidth, keepaspectratio=true]{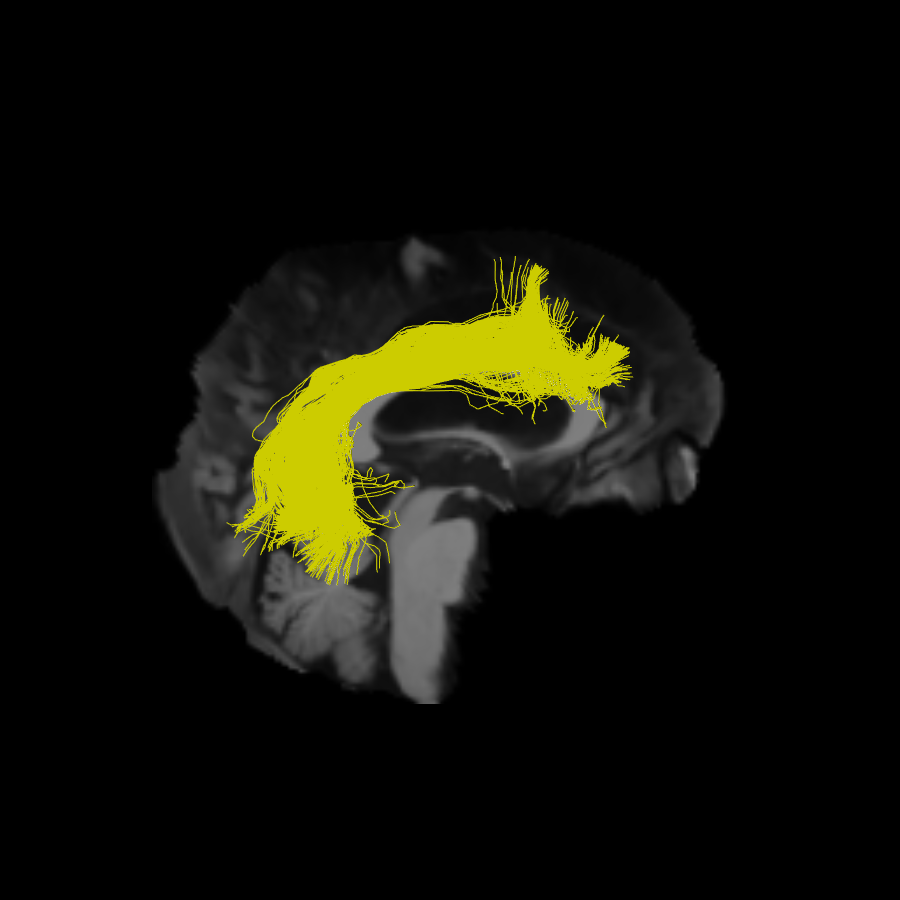} &
\includegraphics[scale=0.95,, trim=1.3in 2.5in 1.3in 2.025in, clip=true, width=0.205\linewidth, keepaspectratio=true]{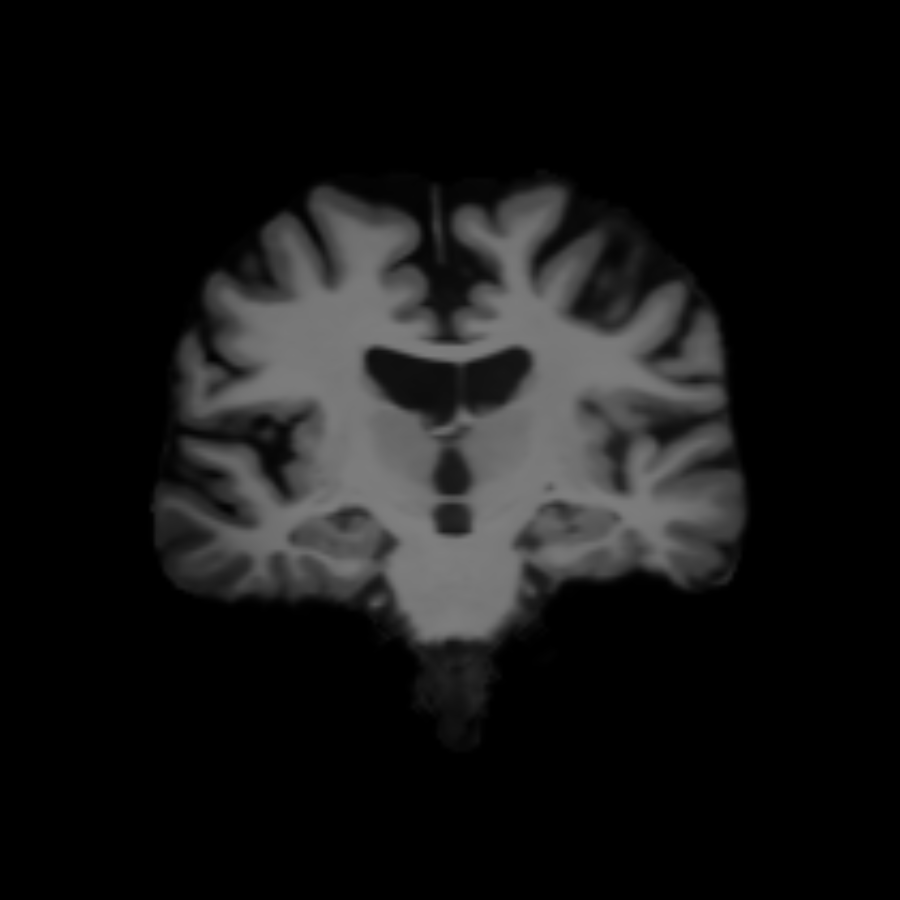} &
\includegraphics[scale=0.95, trim=2.25in 1.95in 2.25in 1.745in, clip=true, width=0.15\linewidth, keepaspectratio=true]{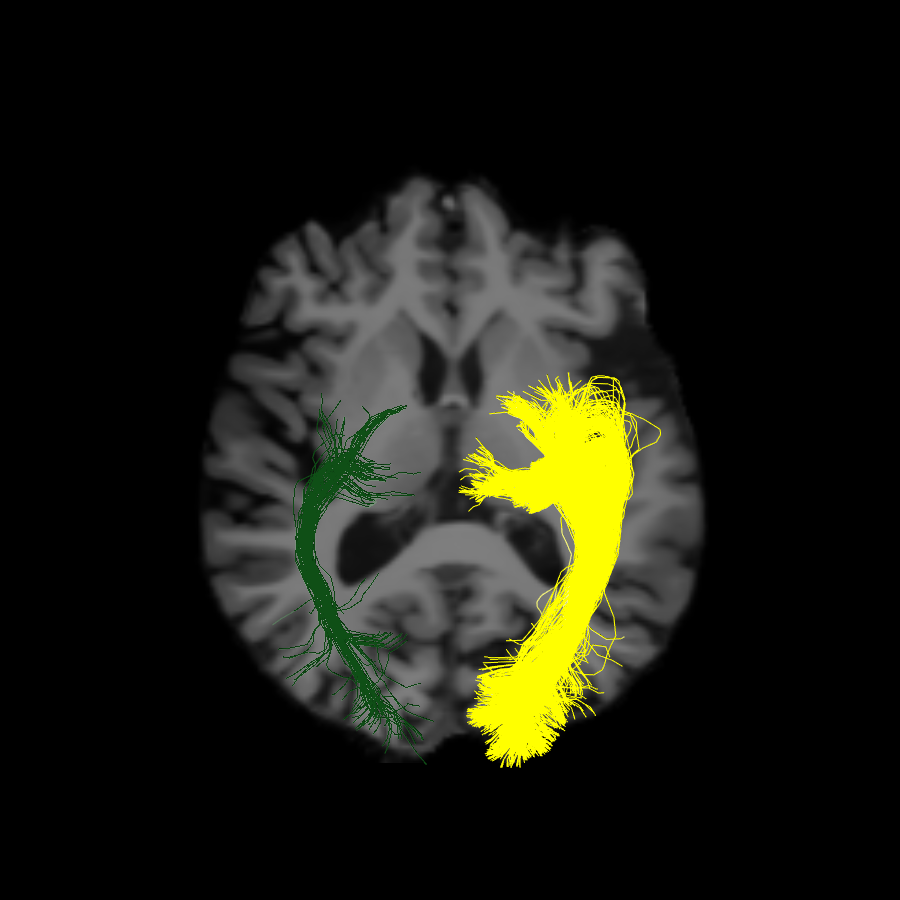} &
\includegraphics[scale=0.95, trim=1.5in 2.25in 1.5in 2.6in, clip=true, width=0.205\linewidth, keepaspectratio=true]{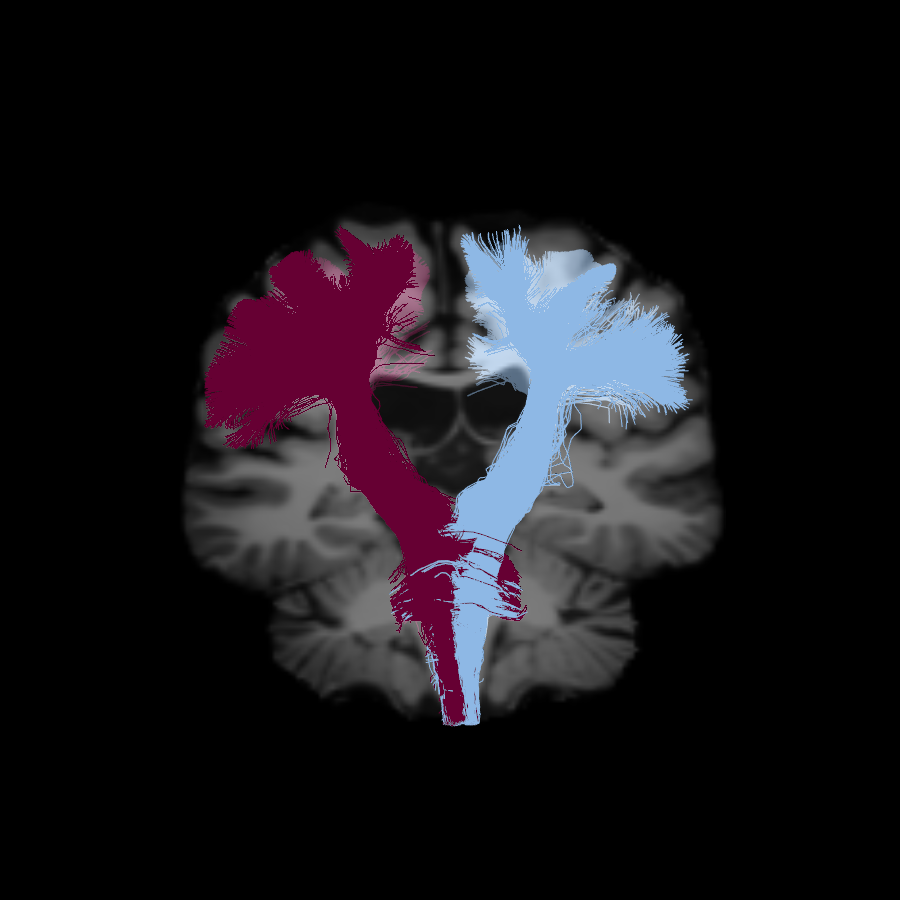} \\
\includegraphics[scale=0.95, trim=1.5in 2.25in 1.5in 2.6in, clip=true, width=0.205\linewidth, keepaspectratio=true]{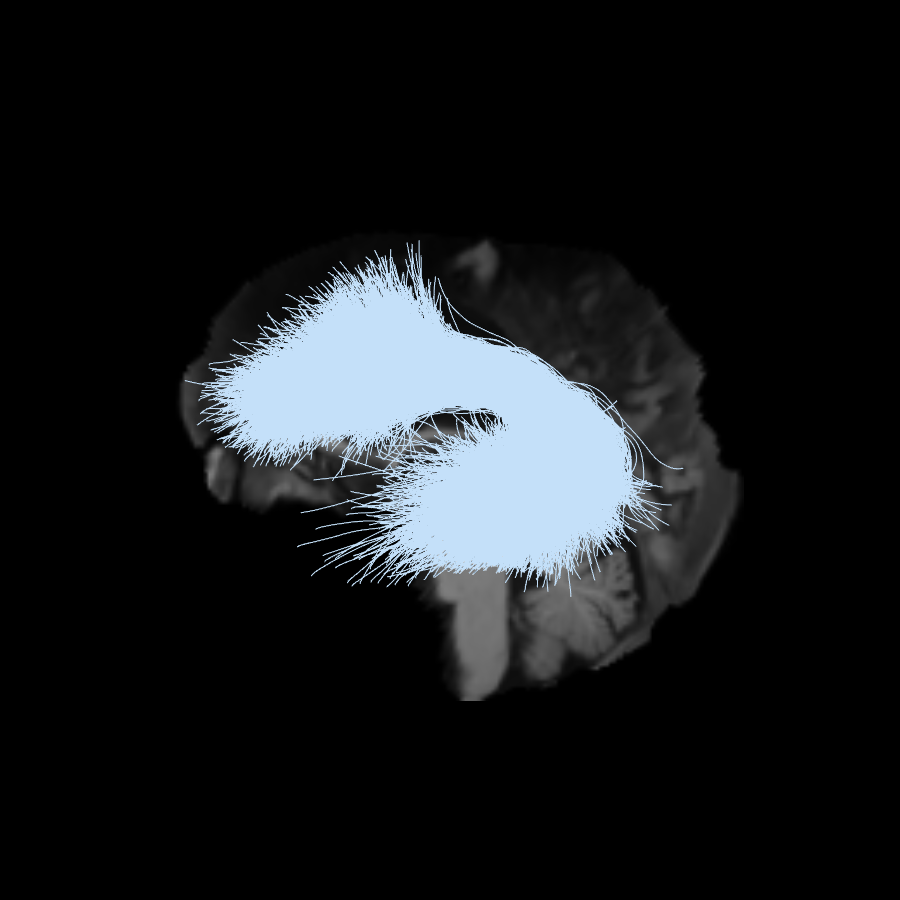} &
\includegraphics[scale=0.95, trim=1.5in 2.25in 1.5in 2.6in, clip=true, width=0.205\linewidth, keepaspectratio=true]{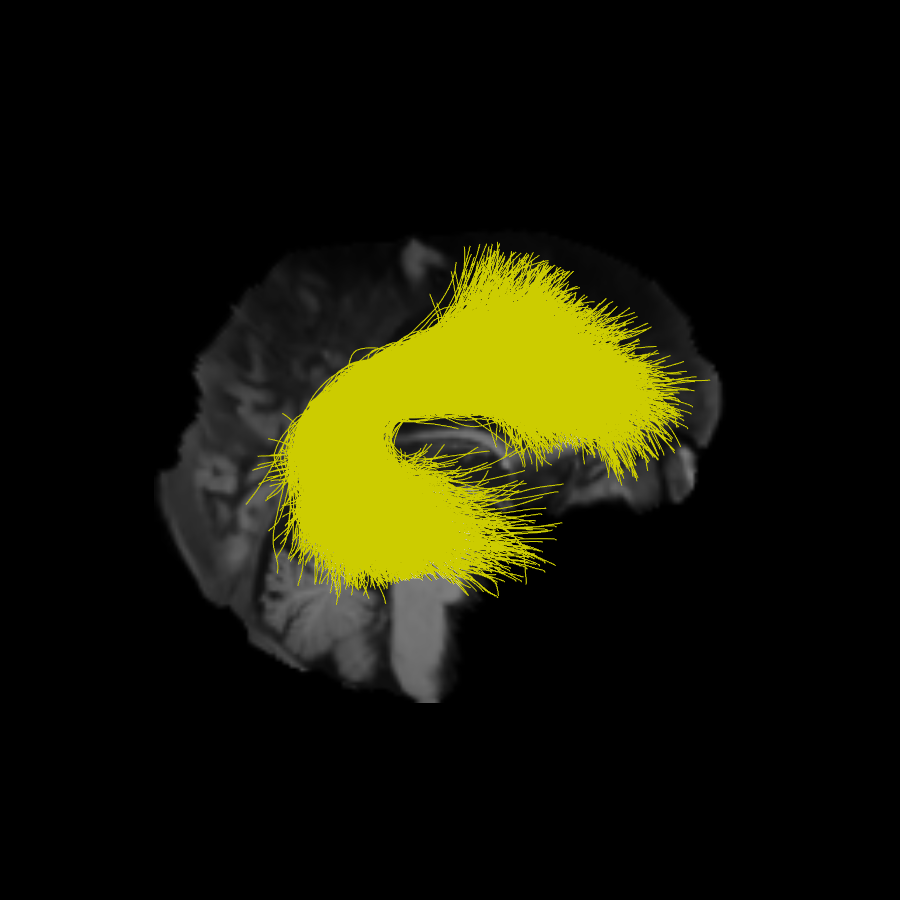} &
\includegraphics[scale=0.95, trim=1.5in 2.6in 1.5in 2.25in, clip=true, width=0.205\linewidth, keepaspectratio=true]{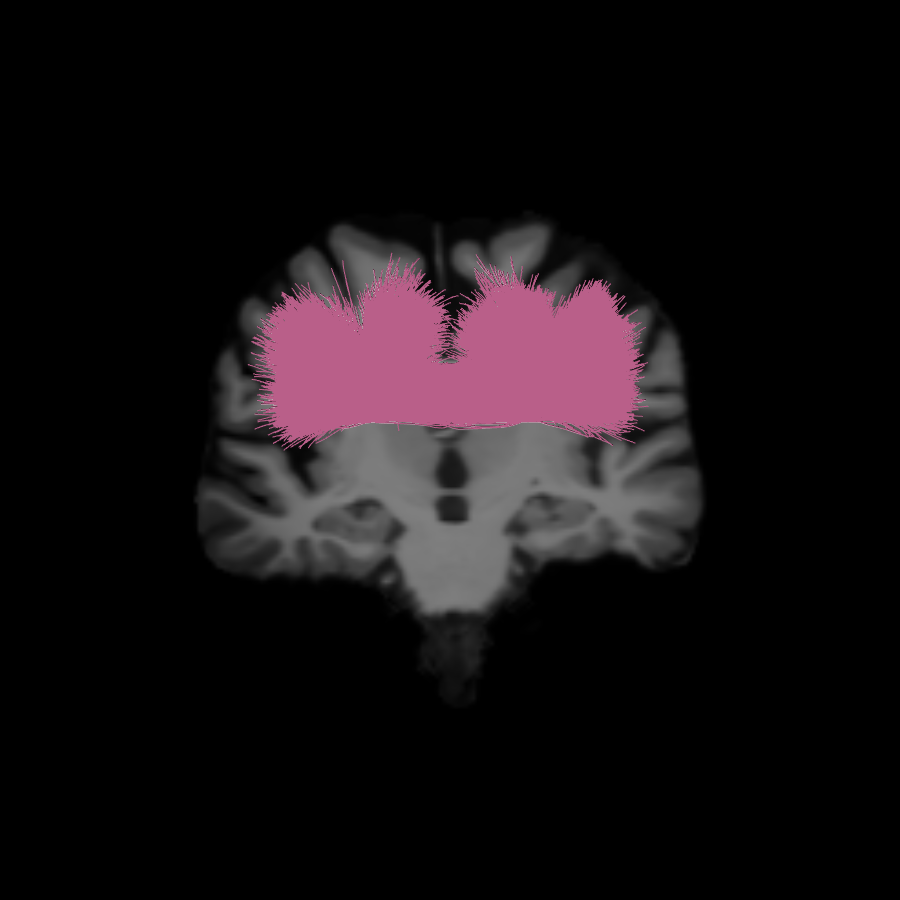} &
\includegraphics[scale=0.95, trim=2.25in 1.75in 2.25in 1.95in, clip=true, width=0.15\linewidth, keepaspectratio=true]{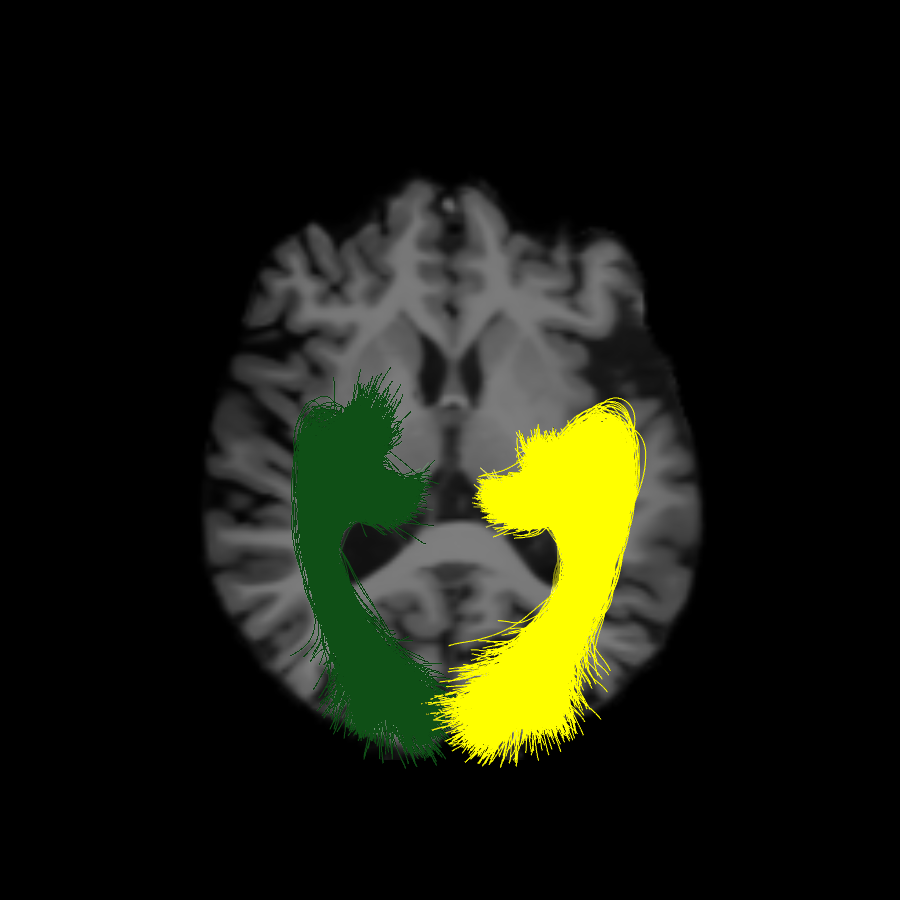} &
\includegraphics[scale=0.95, trim=1.5in 2.15in 1.5in 2.7in, clip=true, width=0.205\linewidth, keepaspectratio=true]{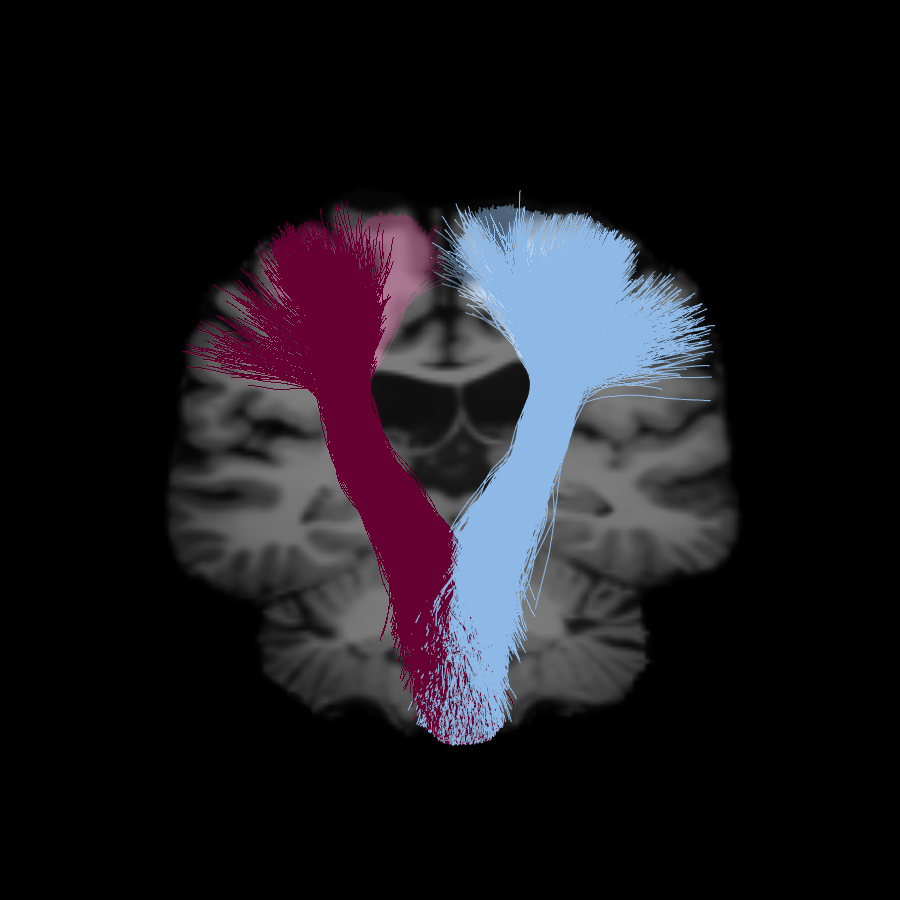} \\
\includegraphics[scale=0.95, trim=1.5in 2.25in 1.5in 2.6in, clip=true, width=0.205\linewidth, keepaspectratio=true]{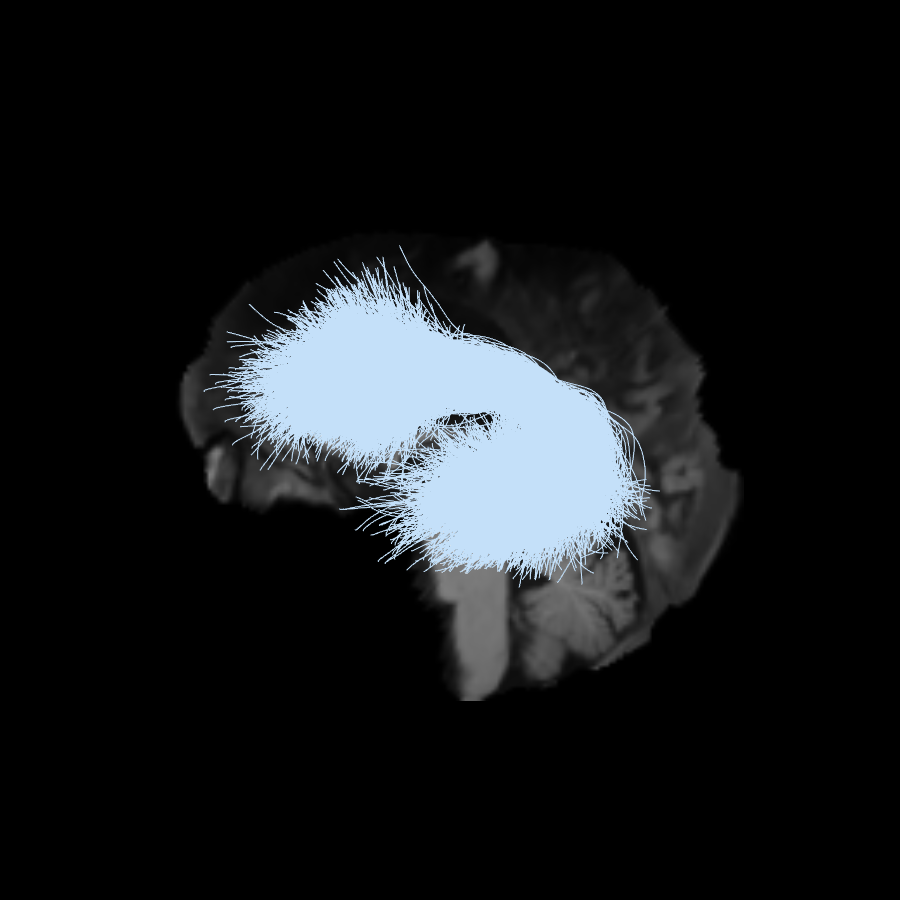} &
\includegraphics[scale=0.95, trim=1.5in 2.25in 1.5in 2.6in, clip=true, width=0.205\linewidth, keepaspectratio=true]{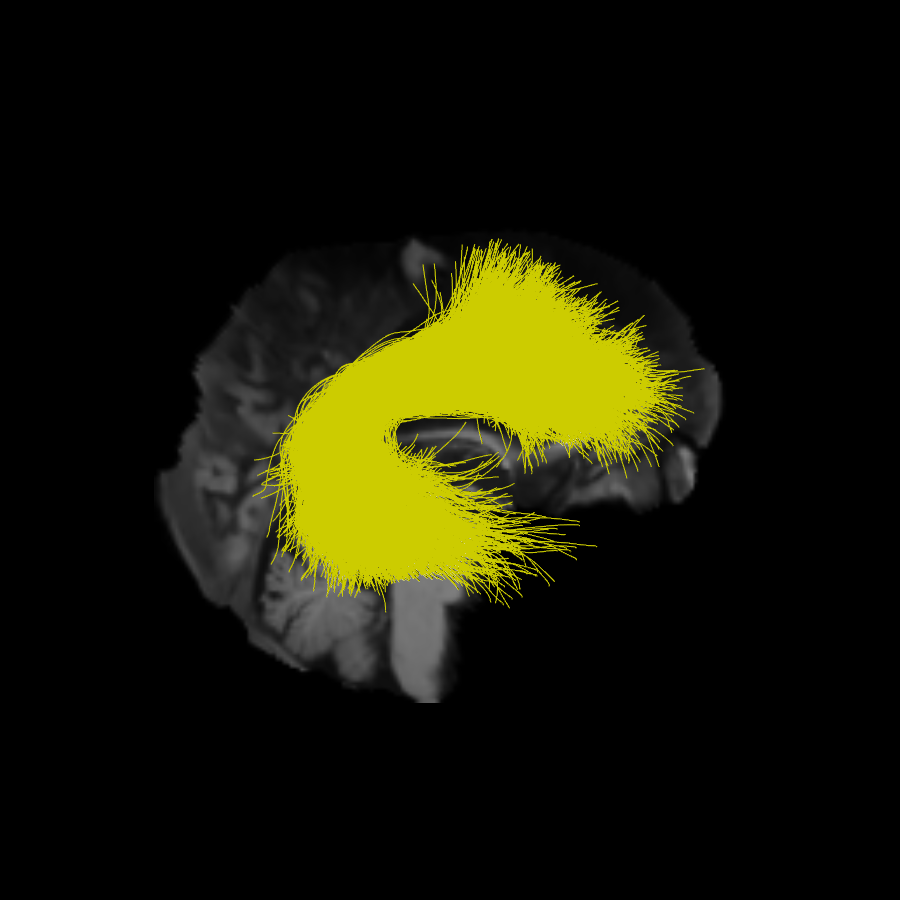} &
\includegraphics[scale=0.95, trim=1.5in 2.6in 1.5in 2.25in, clip=true, width=0.205\linewidth, keepaspectratio=true]{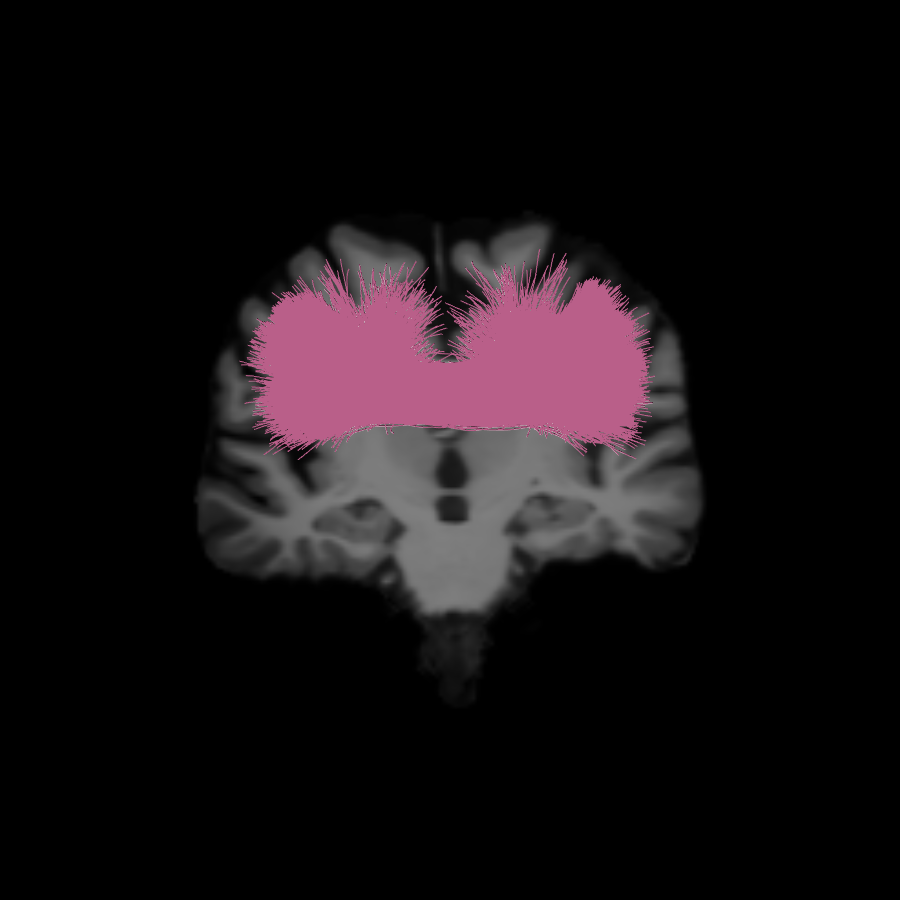} &
\includegraphics[scale=0.95, trim=2.25in 1.75in 2.25in 1.95in, clip=true, width=0.15\linewidth, keepaspectratio=true]{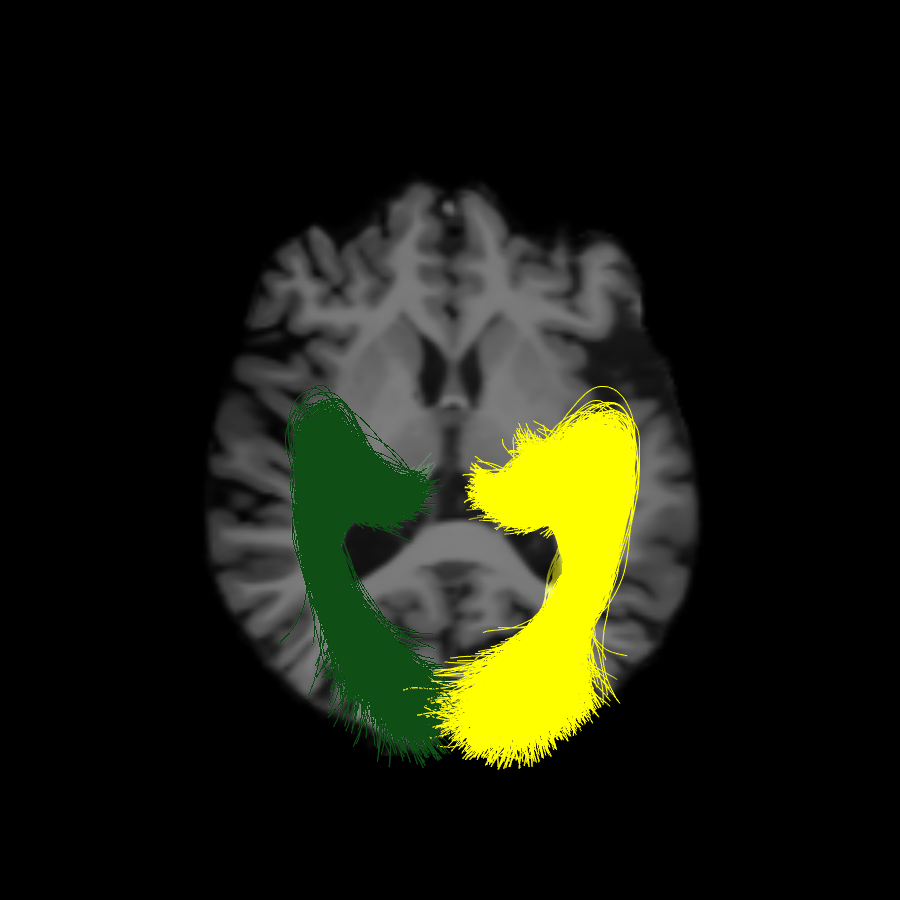} &
\includegraphics[scale=0.95, trim=1.5in 2.15in 1.5in 2.7in, clip=true, width=0.205\linewidth, keepaspectratio=true]{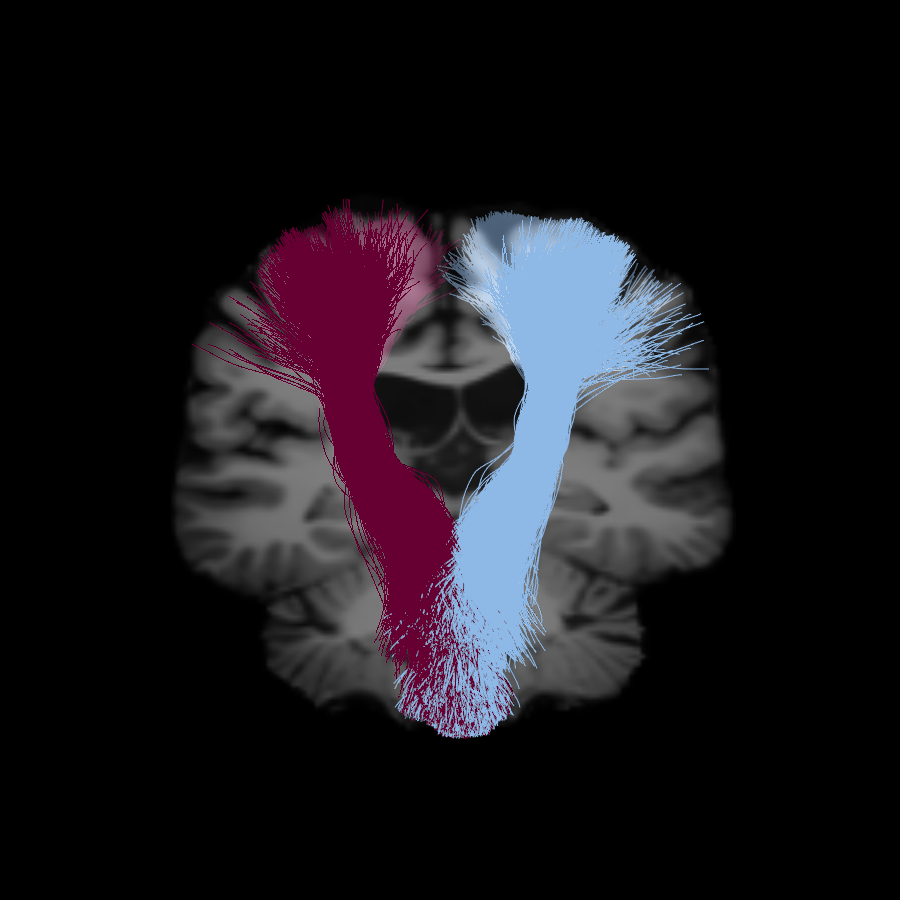} \\
\textbf{(i)} & \textbf{(ii)} & \textbf{(iii)}  & \textbf{(iv)}  & \textbf{(v)} \\
\end{tabular}
\end{subtable}
\caption{\label{fig:tractoinferno_subj}Baseline tractography and GESTA tractography for the TractoInferno dataset subject \textit{s2}. From top to bottom: deterministic; probabilistic; PFT; SET; GESTA-Det; GESTA-Prob tractography. Latent-generated streamlines are evaluated with the \textit{ADGC}\textsubscript{R} criterion. Bundles: (i) AF\_L; (ii) AF\_R; (iii) CC\_Fr\_1; (iv) OR\_ML; (v) PYT. Views have been chosen to best visualize the bundles.}
\end{figure*}

\clearpage
\newpage

\section{Discussion}
\label{sec:discussion}

We presented GESTA (\textit{Generative Sampling in Bundle Tractography using Autoencoders}), a generative method to provide new, anatomically plausible streamlines that enhance the spatial coverage of an existing tractogram. We demonstrate that the method can be particularly useful to reliably fill bundles that are poorly populated in tractograms. The method allows to circumvent the difficulties of a given tracking method on regions where a local orientation signal may prevent a streamline to be propagated due to an excessively low value, or privilege a given orientation. GESTA employs a novel global approach, based on a latent space sampling method, to generate complete streamlines in a bundle-wise fashion. As shown by our results, the produced generative tractograms provide an increase in the WM volume coverage over the baselines' across a wide set of datasets and underlying tractography methods (see the Supplementary Materials for additional evidence), even when using exclusively the available sets of seeds (\textit{unassisted seeding mode}). When using auxiliary seeds from a helper atlas (\textit{assisted seeding mode}), our generative framework allows to generate streamlines for bundles that conventional tractography methods are unable to extract. The newly generated streamlines can be used to leverage the results of an existing tractogram: GESTA can be conveniently used on the result of any tractography method to improve the provided spatial coverage. Thus, conventional local orientation integration methods, global optimization methods, microstructure-informed methods, or yet deep learning-based tracking methods (where the training data might be implicitly impacted by the mentioned effect) might equally benefit from the proposed approach.

Sampling from the latent space has the advantage of by-passing difficulties imposed by the \begin{enumerate*}[label=(\roman*)]\item diffusion image quality (streamlines could be generated from very low resolution images); and \item hard-to-track regions.\end{enumerate*} Our approach is able to generate streamlines for regions that are hard-to-track provided that reliable bundle models are available at learning and to sample from. Meanwhile, other methods such as Bundle-Specific Tractography (BST) \citep{Rheault:Neuroimage:2019} or dynamically seeded Surface Enhanced Tractography (SET) \citep{StOnge:BrainConnectivity:2021}, might find it difficult to traverse some hard-to-track regions given that they are still guided by some streamline-, diffusion- or anatomically-based signal.

GESTA uses the latent space as a ``streamline yard'', and, as such, it can work on streamlines extracted with any tractography method. As shown by the variety of features of the raw diffusion data underlying the datasets employed, or the tractography methods used to extract the seeding streamlines, the proposed generative streamline sampling method is independent of the diffusion acquisition sequence, or the choice of a particular local orientation reconstruction method. Similarly, our generative tractography process does not require computing a response function, any compartment volume fraction, or any other local construct.

As demonstrated by the results in \citet{Maffei:Neuroimage:2022, Maier-Hein:NatureComm:2017}, conventional streamline propagation methods, even when creating millions of potential pathways, fail to reconstruct appropriately the expected bundles. This is also verified in the \textit{in vivo} human tractography TractoInferno dataset subject, where conventional tractography methods are unable to extract streamlines for some bundles on healthy adult data (see the bundle detection score in table \ref{tab:tractoinferno_generative_tractography_measures} and missed bundles in figure \ref{fig:tractoinferno_subj}). A number of works (\eg \citet{Rheault:JNeuralEng:2020, Schilling:Neuroimage:2019, Schilling:HBM:2021, Zhang:Neuroimage:2022}) have analyzed the potential causes, many of which arise from the spatial constraints existing in the WM, the diffusion local orientation constructs, and the biases in streamline propagation of conventional methods. As such, the issues persist across benchmark competitions or datasets \citep{Schilling:MRI:2019}, independent of using state-of-the-art tractography methods, seeding profusely or adaptively/iteratively, or despite using ample local orientation aperture cones. The generative bundle tractography method proposed in this work is able to extract streamlines in such circumstances, leading to an improvement in the WM volume coverage. Thus, GESTA proposes to overcome these limitations by using streamlines to ``track differently''.

\subsection{Latent space sampling aspects}
\label{subsec:disc:latent_space_sampling}

Our experiments show that some latent space regions benefit from using a more localized scope in order to be appropriately sampled. These regions likely correspond to bundles with a large spatial extent or having a diversity of streamline configurations, such as the corpus callosum. We hypothesize that such bundles are encoded by a more complex distribution in the learned latent space. An unconstrained autoencoder, unconditional on any prior information, such as the one used in this work, might have a limited ability to provide a latent space that offers a uniform behavior for sampling purposes. Thus, in order to obtain a presumably more tractable, spatially localized distribution that might be easier to sample from, some bundles can be further split into coherent streamline groups. In this work, the corpus callosum (initially available as a single bundle in the ISMRM 2015 Tractography Challenge dataset) is split into the \num{6} groups defined in \citet{Rheault:Zenodo:2021}, and each group is sampled separately.

The results in figures \ref{fig:ismrm2015_generative_streamlines} and \ref{fig:tractoinferno_subj} suggest that some other bundles might have also benefited from an additional split to be able to better sample some of the seeds (\eg the most lateral CST streamlines). This would likely allow the rejection sampling method to obtain a better fit to some of the less densely packed streamline seeds, which should offer better results when sampling them.

We did not investigate whether using another set of seed streamlines, chosen according to some non-random policy, would have allowed to sample the entire corpus callosum without the need to split it. Similarly, we did not investigate whether other sampling methods are able to more easily fit and sample from such complex distributions.

In our experiments, we consider the fornix a commissural bundle (and thus composed of a single component), following the bundles used in \citet{Maier-Hein:NatureComm:2017}. However, other works (both describing the human white matter anatomy, such as \citet{Catani:Neuroimage:2002, Catani:Cortex:2008}, and presenting white matter dissection methods, such as \citet{Schilling:Neuroimage:2021}) model the fornix as a pair of left and right hemisphere component projection bundles. In our experiments, we did not take steps to ensure that the seed streamlines were equally distributed across hemispheres, and thus, when the number of seeds is low and due to the random sampling effect, components of a hemisphere might get under-represented. Such an imbalance impacts the generative sampling process, and ultimately the improvement in the white matter occupancy.

Although the bandwidth used to fit the proposed distribution influences the number of accepted samples by the rejection sampling method, this generally comes at the cost of an increased likelihood of streamlines not complying with the anatomical plausibility criteria downstream. Given enough computational resources and time, eventually a larger compliance ratio might be obtained by using different sampling parameters and by aggregating results.

Our generative framework produces virtually no ``prematurely terminated'' streamlines given a set of \textit{well-behaved} streamline seeds (see section \ref{subsec:generative_global_features} for additional insight). In this context, ``prematurely terminated'' designates those streamlines that fall notably short of the lower-end of the length distribution expected from the bundle they belong to. In conventional streamline propagation methods, a non-negligible amount of streamlines exit the white matter mask without reaching their target endpoints in the cortex due to unfavorable local orientation decisions. The extremely low rate of length-wise discarded streamlines observed experimentally, and the overlap being conserved when applying the \textit{ADGC} criterion on the ``Fiber Cup'' (see table \ref{tab:fibercup_generative_tractography_measures}), shows that GESTA is able to yield better behaved streamlines length-wise. GESTA may still provide streamlines that do not reach the cortex, as shown by the decrease in the \textit{ADGC} criterion overlap figures on the ISMRM 2015 Tractography Challenge and TractoInferno datasets (tables \ref{tab:ismrm2015_generative_tractography_measures} and \ref{tab:tractoinferno_generative_tractography_measures}); however, both the gray matter mask and the seeds used might have affected such results. As required by the application, the streamline plausibility evaluation framework admits adjustments on its criteria (\eg including other tissues and structures to the masks, etc.), parameters and thresholds.

\subsection{Generative tractography evaluation framework}
\label{subsec:generative_evaluation_framework}

The proposed streamline compliance criteria are designed such that constraints are increasingly restrictive, \textit{ADG} being the less demanding criterion, and \textit{ADGC} being the most demanding. Streamlines reaching the GM largely depend on the seed streamlines given the lack of explicit modeling in our framework. Some randomly selected seeds, which have been considered as being plausibles by the underlying scoring methods, may actually not reach the cortex, and hence the sampled streamlines might show the same behavior, ultimately being discarded by the \textit{ADGC\textsubscript{B}} (or \textit{ADGC\textsubscript{R}}) criterion. At the same time, in the proposed setting, the GM occupancy constraint is considered as a \textit{hard} constraint, as it is a binary criterion. The WM occupancy ratio constraint, where the requirement is relaxed by introducing a volume-based tissue occupancy threshold, is considered a \text{soft} criterion. Investigating whether using a spatially-localized, variable ratio or window (\eg softening the constraint only at the tissue interface or using partial volume effect (PVE) maps obtained from T1-weighted data) would offer a more suitable plausibility evaluation framework is considered to be beyond the aims of this work.

We decided to explicitly avoid generating streamline evaluation criteria that include partially fulfilling constraints of different nature. Specifically, we do not consider the case where streamlines could reach the GM at only one endpoint, but be within a given distance from the cortex at the other endpoint. Although this could be a valid criterion to assess the plausibility of streamlines, we chose to model separately the influence of criteria that differ in their nature. Also, this allows to keep the number and complexity of the plausibility evaluation criteria to a set of limited, well-established measures.

The allowed streamline to local orientation angle gap is relaxed for the BIL\&GIN callosal homotopic and TractoInferno datasets. Allowing a \ang{30} cone when propagating local orientations into streamlines is a common choice. Given that our autoencoder is trained solely using streamline data, no local information is encoded in our latent space, and hence, the constraint is relaxed in order to overcome this limitation.

Experimental evidence \citep{Maier-Hein:NatureComm:2017} has shown that some of these requirements might need to be relaxed to avoid being excessively restrictive when evaluating the results of tracking methods. In the context of this work, WM and GM masks are slightly dilated, and the streamline local orientation to peak alignment criterion uses a soft condition (rather than strictly enforcing to comply with an alignment cone at each streamline segment) (see section \ref{subsec:evaluation_framework_parameterization}). The softened criterion allows the generated streamlines to overcome the bottleneck effect at locations along their trajectory, where conventional tracking methods are likely to give rise to false positive pathways \citep{Schilling:HBM:2021}. As shown in \citet{Canales:Neuroimage:2019}, current spherical deconvolution methods' ability to distinguish crossing fiber populations declines considerably below the \numrange[range-phrase=-,range-units=single]{45}{35}{\si{\degree}} inter-fiber angle range. Thus, provided that appropriate seed streamlines exist, the generative framework presented in this work could be used in regions that are hard to traverse for conventional local reconstruction and streamline propagation methods.

\subsection{Experimental choices}
\label{subsec:experimental_choices}

In our experiments, the seed streamlines are chosen randomly. We did not investigate strategies to increase the diversity of the selected reference streamlines. The proposed procedure allows to test our method without providing it any particular advantageous context. However, as revealed in figures \ref{fig:fibercup_generative_streamlines}, \ref{fig:ismrm2015_generative_streamlines}, and \ref{fig:tractoinferno_subj}, as seeds might not be distributed uniformly along all possible pathways, the sampling method can draw samples preferentially from a given region of the latent space with a larger density. This is a consequence of the underlying (seed) data distribution being biased over some particular area or configuration, which results in an estimate of the distribution reflecting such bias.

Note that the number of available seeds varies largely across bundles. The choice is further restricted by the use of the test set in the synthetic data experiments, and the subsampling used on them to demonstrate the generative ability of GESTA. This can constrain the diversity of the latent-generated streamlines. When a low streamline count is available for seeding, auxiliary seeds from an atlas can be used to allow the sampling procedure to appropriately generate new streamline candidates, as proposed in the TractoInferno experimental setting.

\subsection{Limitations}
\label{subsec:limitations}

The quality of the generated streamlines is fundamentally tied to the richness of the seed streamlines. The tractography data used in this work, and the training and testing splits, had not been generated so as to make all possible (or known) intra-bundle configurations sufficiently available at generation time. As a consequence, the anatomical plausibility of the generative streamlines covering such particular structures gets affected.

GESTA does not fully reconcile the sensitivity \vs specificity tradeoff. Results show that the generative method incurs a higher overreach than conventional tractography methods. This is explained by the lack of local control of the generative process, the WM occupancy criterion relaxation employed, and the explicit dilation of the WM mask in the evaluation framework to avoid discarding streamlines that are slightly off-limits. Imposing constraints on the latent space might help overcoming such limitation.

The time to yield new streamline data points varies considerably across bundles (see section \ref{subsec:sampling_measures}). The rejection sampling method requires to estimate the distribution of the seeds in the latent space. We hypothesize that a complex distribution (potentially, one having a large number of seeds) is costly to approximate successfully. A more localized sampling scheme, together with latent space distribution-enforcing models, might help alleviating this.

Due to the lossy nature of the encoder and the decoder, the latent-generated streamlines describe smoother trajectories than those of the seed streamlines. This might impact the plausibility of the generative streamlines at the boundaries of the bundles.

The autoencoder requires that all tractograms used at train and test time be in the same space (for example, the standard-MNI space). Thus, our framework is limited by any potential accuracy loss derived from the spatial transformations involved in the registration process.

\section{Future work}
\label{sec:future_work}

In this work, the same generative sampling parameters (namely, the Parzen estimator bandwidth factor) are used to generate streamlines across different bundles. This allows to consistently evaluate the effect of a given set of parameters on the used success measures. However, the framework allows to provide different parameters to different bundles, which might improve the success of the method at generating streamlines for a given bundle.

Similarly, we use the same feature thresholds for all bundles to determine the plausibility of streamlines sampled from the latent space. The framework could be adapted to use bundle-specific thresholds. Using a tighter match around bundle-specific values requires an agreement on such values. Studying the restrictiveness and effects of such values, if any, is out of the scope of this work, and might be addressed in the future.

An enhanced latent space seeding method might favorably shape the underlying streamline distribution for the presented sampling method. For example, using a strategy to favor dissimilar reference picks could possibly allow the sampling method to generate candidates that would span a richer set of intra-bundle configurations. This would eventually lead to larger improvements in the white matter spatial coverage. The impact of other seeding methods (\eg adaptive, bundle-aware, etc., or based on bundle summarizing methods) in order to draw streamline samples more uniformly across all latent space regions remains to be studied. Similarly, investigating whether other sampling methods are more robust to the data distribution is left as a future work.

The proposed generative framework, being unconditional, lacks explicit spatial constraints to harness the streamline candidate generation. Self-supervised approaches for autoencoders, as the one proposed by \citet{Chen:MICCAI:2021}, could be investigated as a means to improve the learned latent space. Similarly, introducing additional constraints in the latent space might be useful to provide an autoencoder capable of generating outputs with some desired attributes. Providing some further level of guarantee on the generative streamlines' behavior, such as learning appropriate termination patterns, would require introducing some additional parameterization to train the autoencoder. Likewise, a closer streamline feature-to-latent dimension space correlation might improve the latent space-based generative capabilities offered by such a model. Leveraging the autoencoder to improve the learned latent representations, and thus increasing the sampled streamlines' success rate in terms of their plausibility according to some criterion by introducing constraints into the training process, is left for a separate piece of work.

We have demonstrated that GESTA can be used in regions or bundles where other conventional tractography methods miss to reconstruct streamlines. The \textit{in vivo} human brain datasets used in this work correspond to healthy adult data: demonstrating the effectiveness of GESTA on other clinically-relevant application data, including cognitively impaired aging population or brain tumor subject data, may be considered in a different body of work.

\section{Conclusions}
\label{sec:conclusions}

In this work, we introduced the first deep generative bundle tractography framework. The generative tractography framework uses \begin{enumerate*}[label=(\roman*)]\item the latent space of a deep autoencoder; \item a set of seed streamlines; \item a data sampling method; and \item a streamline plausibility evaluation framework\end{enumerate*} to globally reconstruct anatomically plausible streamlines bundle-wise from the trained latent space of the autoencoder. Our generative tractography method is able to reliably yield plausible streamlines for bundles that conventional streamline propagation methods fail to extract. We demonstrate that the latent space-generated streamlines can be used to enhance the spatial coverage of an existing tractogram and its white matter bundles. The GESTA (\textit{Generative Sampling in Bundle Tractography using Autoencoders}) framework is shown to successfully generate anatomically plausible streamlines across a variety of both synthetic and \textit{in vivo} human brain datasets by exploiting the latent space of an already trained autoencoder. GESTA unlocks the potential of deep autoencoders for generative tractography, and the proposed framework provides the basis to build increasingly proficient deep learning generative tractography methods. Ultimately, it may help tractometry and connectomics derivatives by providing a more anatomically reliable white matter mapping.

\section*{Acknowledgments}
\label{sec:acknowledgments}

This work has been partially supported by the Centre d'Imagerie M\'{e}dicale de l’Universit\'{e} de Sherbrooke (CIMUS); the Axe d'Imagerie M\'{e}dicale (AIM) of the Centre de Recherche du CHUS (CRCHUS); and the R\'{e}seau de Bio-Imagerie du Qu\'{e}bec (RBIQ)/Quebec Bio-imaging Network (QBIN) (FRSQ - R\'{e}seaux de recherche th\'{e}matiques File: 35450). This research was enabled in part by support provided by Calcul Qu\'{e}bec (\href{https://www.calculquebec.ca/en/}{www.calculquebec.ca}) and the Digital Research Alliance of Canada Advanced Research Computing service (\href{https://alliancecan.ca/en/services/advanced-research-computing}{www.alliancecan.ca}). We also thank the research chair in Neuroinformatics of the Universit\'{e} de Sherbrooke. Finally, thanks to F\'{e}lix Dumais, Carl Lemaire, Emmanuelle Renauld, \'{E}tienne St-Onge, and Antoine Th\'{e}berge for their insightful comments and discussions. Data were provided in part by the Human Connectome Project, WU-Minn Consortium (Principal Investigators: David Van Essen and Kamil Ugurbil; 1U54MH091657) funded by the 16 NIH Institutes and Centers that support the NIH Blueprint for Neuroscience Research; and by the McDonnell Center for Systems Neuroscience at Washington University.

\bibliographystyle{abbrvnat}
\setcitestyle{authoryear,open={((},close={))}}
\bibliography{./bibliography/bibliography.bib}

\appendix
\section{}
\label{sec:appendix}

\subsection{Streamline evaluation framework parameterization}
\label{subsec:evaluation_framework_parameterization}

In order to determine the anatomical plausibility of the streamlines generated from the latent space, a number of choices are made concerning each of the evaluated aspects. An analysis on the thresholding values presented in table \ref{tab:plausibility_criteria_values} is provided in the following:

\begin{itemize}
\item Streamline geometry: the streamline length is set to be within the \SIrange{20}{220}{\si{\milli\metre}} range. The streamline winding is set to be below \ang{330} for the ``Fibercup'' and ISMRM 2015 Tractography Challenge datasets; and \ang{360} for the BIL\&GIN callosal homotopic data, and the TractoInferno data. The value is kept below \ang{360} for the synthetic and single-subject experiments in order to demonstrate that GESTA is still able to generate streamlines that comply with stricter winding requirements. The value is set to \ang{360} for the BIL\&GIN callosal homotopic and TractoInferno datasets following \citet{Legarreta:MIA:2021, Petit:OHBM:2019}, where it is experimentally observed that a non-negligible proportion of callosal streamlines describe closed loops.
\item Local orientation alignment geometry: \num{75}{\percent} of the streamline local orientation values are required to be within a \ang{30} cone with respect to the closest local fODF peak for the ``Fibercup'' and ISMRM 2015 Tractography Challenge datasets, which had been tracked using the same local tractography method, and \ang{40} for the BIL\&GIN callosal homotopic and TractoInferno data. Such a degree of tolerance allows to account for differences in the tracking methods and parameters, which possibly result in tractograms of varying intrinsic attributes. At the same time, this might allow to traverse regions that are hard-to-track for conventional streamline propagation methods. \ang{90} values are masked in the check to accommodate for the peak support absent segments (such as those overreaching the WM mask at the streamline endpoints). The fODF peak maps used in this work contain the \num{5} largest peaks at each voxel.
\item White matter tissue occupancy: streamlines are required to lie within the WM mask boundaries. We chose to skip the last $N =$ \num{10} consecutive vertices at each streamline point when checking the WM occupancy. The generative framework provides streamline samples whose endpoints may exceed the boundaries of the WM. If the occupancy requirement is not relaxed, such streamlines would be discarded. Our streamlines have a fixed number of vertices, and thus the step size is variable across streamlines: shortest streamlines (in physical units) have a smaller step size when compared to longest streamlines. After investigation, skipping $N =$ \num{10} consecutive vertices at each endpoint was chosen as a conservative upper bound to guarantee that the WM occupancy criterion would not penalize excessively the shortest streamlines. The WM tissue occupation is set as a binary criterion for the ``Fiber Cup'' dataset, but is softened for the rest of the datasets to require at least \num{95}{\percent} of the streamline vertices to be contained in the WM tissue. Such a relaxation comes as a necessary compromise for the partial volume effects and streamline trajectories slightly veering off locally.
\item Gray matter tissue connectivity: streamline endpoints are required to be located in the GM mask.
\end{itemize}

The GESTA framework does not have an accurate control over the streamlines' trajectory, or their starting and termination vertices. Thus, the WM and GM tissue masks are dilated using \num{2} iterations and a structuring element with a connectivity equal to \num{1}. The generative streamline candidates are trimmed to the corresponding brain mask. The brain masks are also eroded using \num{2} iterations and a structuring element with a connectivity equal to \num{1}, except for the ``Fiber Cup'' dataset mask, which is used ``as is'' given the particular binary nature of the available structural data.

Due to the aforementioned lack of a precise control over the streamline termination, latent-generated raw streamlines are found to overreach the GM mask frequently. The generated raw streamlines are thus trimmed to lie within the computed brain masks for each dataset or subject. Using an eroded version of the brain masks allows to ensure that streamlines would not enter the CSF. Thus, we favor streamlines whose endpoints lay in the dilated GM mask for the \textit{ADGC\textsubscript{B}} (or \textit{ADGC\textsubscript{R}}) criterion. In this work, we use binary GM tissue maps. We do not apply additional post-processing to the generative streamlines to only report on the generative ability of the autoencoder-based framework.

\subsection{Streamline evaluation measure definition}
\label{subsec:evaluation_framework_definition}

In order to assess whether a latent space-generated streamline is plausible, the following definitions are considered:
\begin{itemize}
\item The streamline length is defined as its arc length, that is, the sum of the Euclidean distances between each pair of consecutive vertices in $\mathbb{R}^{3}$ over the entire streamline trajectory.
\item The streamline winding is defined as the total turning angle projected on the best fitting plane. The best fitting plane's basis vectors are found through a singular value decomposition (SVD) of the streamline vertices matrix. The total angle is then computed as the cumulative signed angle between each pair of consecutive vertices.
\item The streamline local orientation alignment is defined as the minimum value between the orientation vector defined by each consecutive streamline vertices and the vectors interpolated from the fODF peaks at each streamline vertex being considered.
\item The WM occupancy is defined as the ratio of streamline vertices contained in the WM mask over the total number of streamline vertices.
\item The GM occupancy is defined as an indicator function that yields a value of \num{1} when both endpoints of a streamline are contained within the GM mask, and \num{0} otherwise.
\end{itemize}

\subsection{Ensemble tracking baselines}
\label{subsec:ensemble_tracking_baselines}

The ensemble tracking baseline \citep{Joanisse:ISMRM:2021} used for the ``Fiber Cup'' dataset was generated out of \num{36} different tractograms obtained with different parameters: local orientations were computed using Constrained Spherical Deconvolution (CSD) \citep{Tournier:Neuroimage:2004, Tournier:Neuroimage:2007, Descoteaux:TMI:2009}, and probabilistic local tractography was used in all cases; \num{10} seeds were employed per voxel; the fODF sampling was done using three (\num{3}) different sphere discretizations containing \{\numlist[list-separator={,},list-final-separator={,}]{100;200;724}\} vertices; the tracking step size was varied in the \{\numlist[list-separator={,},list-final-separator={,}]{0.1;0.5;1.0}\} \si{\milli\meter} set; and the maximum aperture angle between tracking steps was varied in the \{\numlist[list-separator={,},list-final-separator={,}]{10;20;30;40}\}\si{\degree} set. All parameter permutations resulted in a total of \num{36} tractograms that were concatenated to provide the final result.

The ensemble tracking results used as baselines \citep{Maier-Hein:NatureComm:2017} for the ISMRM 2015 Tractography Challenge dataset were obtained as the combination of three (\num{3}) different tractograms computed using global tractography \citep{Reisert:Neuroimage:2011}, the Fiber Assignment by Continuous Tracking (FACT) \citep{Mori:AnnalsNeurol:1999}, and the 2nd order integration over Fibre Orientation Distributions (iFOD2) probabilistic tractography \citep{Tournier:ISMRM:2010} methods. Tractography reconstruction and processing parameters were varied for each of the submissions. Readers are referred to the supplementary materials in \citet{Maier-Hein:NatureComm:2017} for further details.

\subsection{Sampling measures}
\label{subsec:sampling_measures}

Tables \ref{tab:fibercup_sampling_measures}, and \ref{tab:ismrm2015_sampling_measures} show the latent space rejection sampling time performance as a function of the seed streamline count for the ``Fiber Cup'' and ISMRM 2015 Tractography Challenge datasets.

As shown in table \ref{tab:ismrm2015_sampling_measures}, the generative tractography takes approximately
\SI{4.94}{\hour} and \SI{58.35}{\hour} for the $P$ = \num{3}\percent and $P = $\num{100}\percent seed streamline ratio cases (\num{878} and \num{28891} total seed streamlines, respectively) to generate \num{405000} streamlines for the ISMRM 2015 Tractography Challenge dataset.

It detaches from the analysis of the figures that the time required to complete the generative tractography is not linear in terms of the seed streamline count across all bundles. However, generally, it is verified that a larger amount of seeds translates into a longer sampling time. Note that for each seed ratio all seed streamlines were re-computed (\ie independent experiments were run for each ratio), and the seed placement was not optimized (seeds were chosen randomly following a uniform probability distribution). Thus, the underlying seed distribution presumably influences the generative sampling efficiency. Analyzing the optimal seed placement in terms of the sampling time efficiency and tractography measure quality is part of the future work. Measures were made on a conventional desktop machine (Intel(R) Xeon(R) W-2133 CPU @ \SI{3.60}{\giga\hertz} \num{6} core processor; \SI{16} GB RAM; NVIDIA GeForce GTX 1080 Ti \SI{12} GB graphics card). Note that the sampling takes place on the CPU and that the method was not optimized.

\begin{table*}[!htbp]
\caption{\label{tab:fibercup_sampling_measures}``Fiber Cup'' bundle-wise sampling measures: seed streamline count and sampling time. Time values are in \si{\second}.}
\centering
\begin{tabular}{ccccccccc}
\toprule
& \multicolumn{8}{c}{\textbf{Seed ratio} ($P${\percent})} \\
\cmidrule(lr){2-9}
& \multicolumn{2}{c}{3} & \multicolumn{2}{c}{5} & \multicolumn{2}{c}{10} & \multicolumn{2}{c}{100} \\
\cmidrule(lr){2-3}\cmidrule(lr){4-5}\cmidrule(lr){6-7}\cmidrule(lr){8-9}
\textbf{Bundle} & \textbf{Seed count} & \textbf{Time} & \textbf{Seed count} & \textbf{Time} & \textbf{Seed count} & \textbf{Time} & \textbf{Seed count} & \textbf{Time}\\
\midrule
1 & 5 & 1.79 & 9 & 1.69 & 17 & 2.75 & 165 & 1.4\\
2 & 7 & 3.4 & 12 & 7.57 & 23 & 2.53 & 226 & 1.49\\
3 & 5 & 3.8 & 8 & 3.93 & 16 & 2.5 & 153 & 1.63\\
4 & 6 & 2.65 & 10 & 14.21 & 19 & 2.18 & 181 & 1.3\\
5 & 4 & 2.04 & 7 & 1.83 & 13 & 2.54 & 223 & 1.7\\
6 & 4 & 1.19 & 6 & 1.85 & 12 & 2.22 & 112 & 2.02\\
7 & 3 & 1.71 & 4 & 1.5 & 8 & 3.09 & 77 & 1.53\\
\midrule
Total & 34 & 16.58 & 56 & 32.58 & 108 & 17.8 & 1037 & 11.07\\
\bottomrule
\end{tabular}
\end{table*}

\begin{table*}[!htbp]
\caption{\label{tab:ismrm2015_sampling_measures}ISMRM 2015 Tractography Challenge dataset bundle-wise sampling measures: seed streamline count and sampling time. Time values are in \si{\second}.}
\centering
\begin{tabular}{ccccccccc}
\toprule
& \multicolumn{8}{c}{\textbf{Seed ratio} ($P${\percent})} \\
\cmidrule(lr){2-9}
& \multicolumn{2}{c}{3} & \multicolumn{2}{c}{5} & \multicolumn{2}{c}{10} & \multicolumn{2}{c}{100} \\
\cmidrule(lr){2-3}\cmidrule(lr){4-5}\cmidrule(lr){6-7}\cmidrule(lr){8-9}
\textbf{Bundle} & \textbf{Seed count} & \textbf{Time} & \textbf{Seed count} & \textbf{Time} & \textbf{Seed count} & \textbf{Time} & \textbf{Seed count} & \textbf{Time}\\
\midrule
CC\_Fr\_1 & 15 & 25.63 & 24 & 10.67 & 48 & 47.43 & 472 & 205.76\\
CC\_Fr\_2 & 96 & 1923.96 & 160 & 1419.94 & 320 & 1176.5 & 3199 & 29419.91\\
CC\_Oc & 13 & 11.42 & 22 & 23.82 & 44 & 41.62 & 432 & 307.54\\
CC\_Pa & 30 & 35.78 & 50 & 136.77 & 99 & 59.24 & 990 & 412.86\\
CC\_Pr\_Po & 61 & 4583.58 & 101 & 1990.92 & 201 & 2043.69 & 2003 & 19003.06\\
Cing-L & 36 & 2870.65 & 60 & 4262.46 & 120 & 4465.06 & 119 & 5946.96\\
Cing-R & 44 & 3821.84 & 73 & 3957.7 & 146 & 2220.1 & 1456 & 7686.81\\
CST-L & 16 & 4.68 & 27 & 7.97 & 53 & 15.49 & 524 & 55.37\\
CST-R & 18 & 57.84 & 30 & 14.93 & 60 & 48.37 & 600 & 335.61\\
Fx & 8 & 172.62 & 14 & 909.68 & 27 & 223.74 & 262 & 3334\\
FPT-L & 23 & 489.88 & 39 & 361.88 & 77 & 1396.93 & 761 & 16348.2\\
FPT-R & 60 & 2003.71 & 99 & 1958.92 & 198 & 2094.75 & 1971 & 115878.85\\
ICP-L & 20 & 28.66 & 33 & 14.31 & 65 & 66.71 & 646 & 2163.14\\
ICP-R & 13 & 12.63 & 21 & 6.95 & 41 & 30.58 & 405 & 1608.58\\
ILF-L & 71 & 50.59 & 119 & 59.14 & 237 & 35.83 & 2364 & 615.87\\
ILF-R & 62 & 122.5 & 103 & 150.24 & 206 & 54.53 & 2054 & 1027.60\\
MCP & 53 & 515.64 & 88 & 462.19 & 176 & 467.77 & 1752 & 123.82\\
OR-L & 5 & 20.86 & 8 & 32.85 & 16 & 11.41 & 154 & 200.59\\
OR-R & 11 & 14.59 & 17 & 21.88 & 34 & 16.07 & 340 & 79.18\\
POPT-L & 24 & 4.47 & 40 & 8.12 & 79 & 12.73 & 785 & 35.85\\
POPT-R & 11 & 13.79 & 24 & 5.04 & 47 & 17.78 & 466 & 51.22\\
SCP-L & 4 & 3.1 & 7 & 3.58 & 13 & 3.22 & 126 & 10.57\\
SCP-R & 5 & 5.21 & 7 & 4.3 & 14 & 3.55 & 137 & 17.08\\
SLF-L & 67 & 217.32 & 111 & 248.44 & 221 & 332.77 & 2205 & 916.83\\
SLF-R & 63 & 638.91 & 105 & 286.41 & 209 & 100.14 & 2085 & 1150.56\\
UF-L & 25 & 14.06 & 41 & 39.83 & 82 & 71.03 & 815 & 228.87\\
UF-R & 21 & 135.87 & 35 & 104.91 & 69 & 408.38 & 690 & 2878.04\\
\midrule
Total & 878 & 17799.79 & 1458 & 16503.88 & 2902 & 15465.4 & 28891 & 210042.72\\
\bottomrule
\end{tabular}
\end{table*}

\newpage
\clearpage

\subsection{Generative global features}
\label{subsec:generative_global_features}

GESTA provides streamlines that are consistent with the global geometrical features of streamlines belonging to a given bundle. In this work, such features are measured in terms of the streamline length and winding. Provided that the appropriate seeds are available, the rejection sampling procedure yields virtually no ``prematurely terminated'' streamlines, nor do the described pathways contain sharp bends or turns. Both these effects might occur frequently in local orientation propagation methods.

\begin{figure*}[!htbp]
\centering
\setlength{\tabcolsep}{0pt}
\begin{tabular}{cc}
\includegraphics[scale=0.95, trim=0.15in 0.15in 0.15in 0.15in, clip=true, width=0.45\linewidth, keepaspectratio=true]{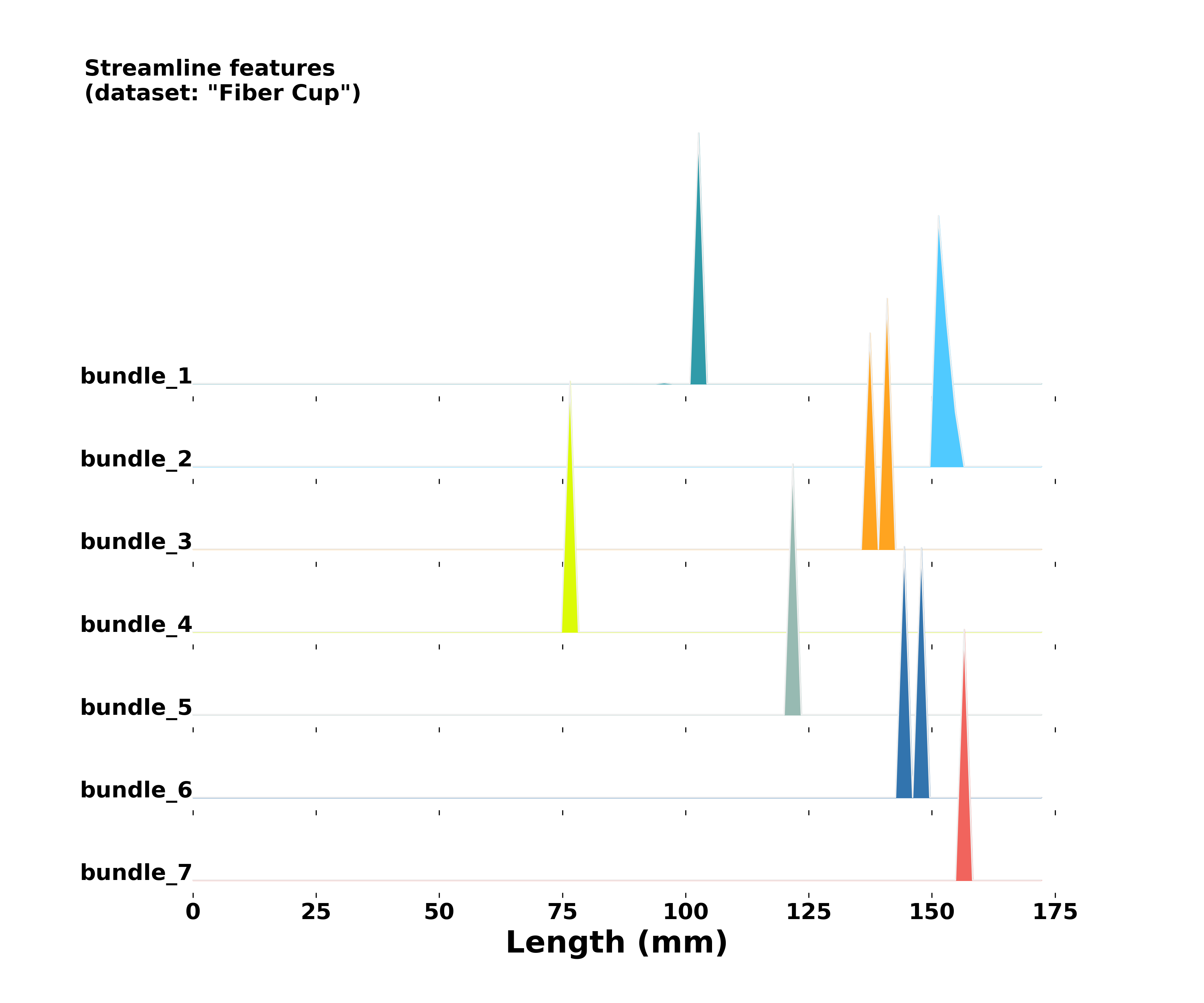} &
\includegraphics[scale=0.95, trim=0.15in 0.15in 0.15in 0.15in, clip=true, width=0.45\linewidth, keepaspectratio=true]{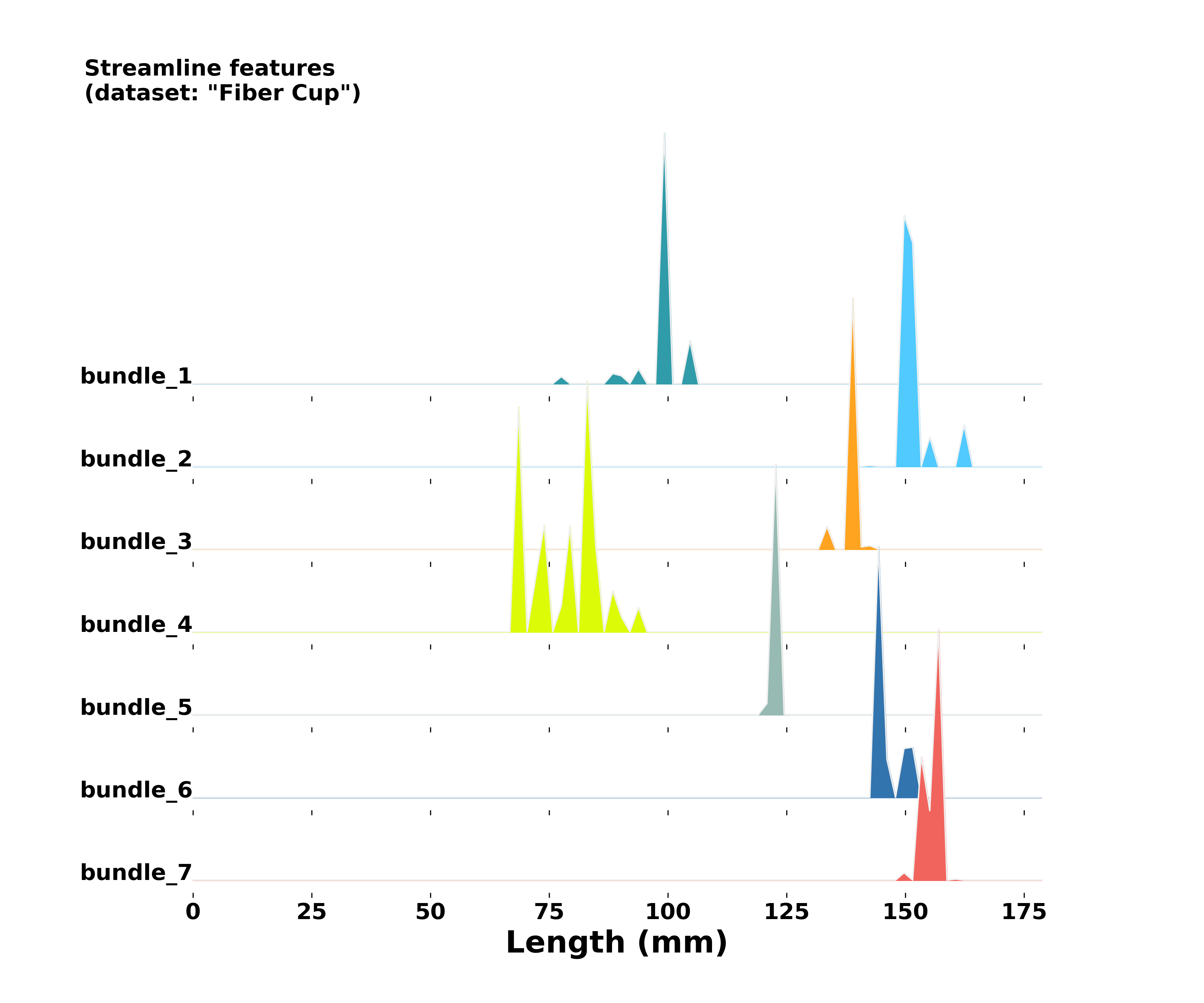} \\
\textbf{(a)} & \textbf{(b)} \\
\includegraphics[scale=0.95, trim=0.15in 0.15in 0.15in 0.15in, clip=true, width=0.45\linewidth, keepaspectratio=true]{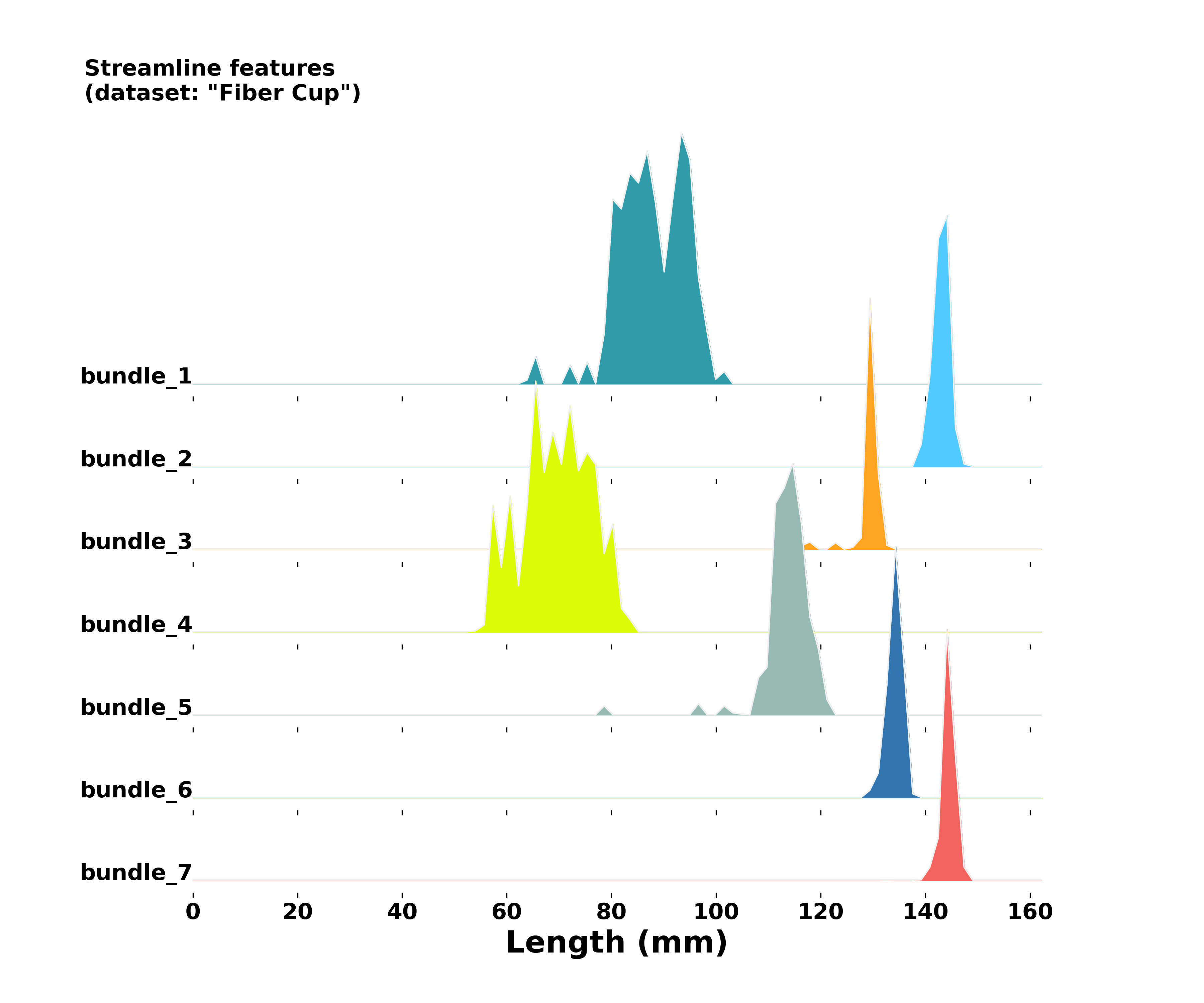} &
\includegraphics[scale=0.95, trim=0.15in 0.15in 0.15in 0.15in, clip=true, width=0.45\linewidth, keepaspectratio=true]{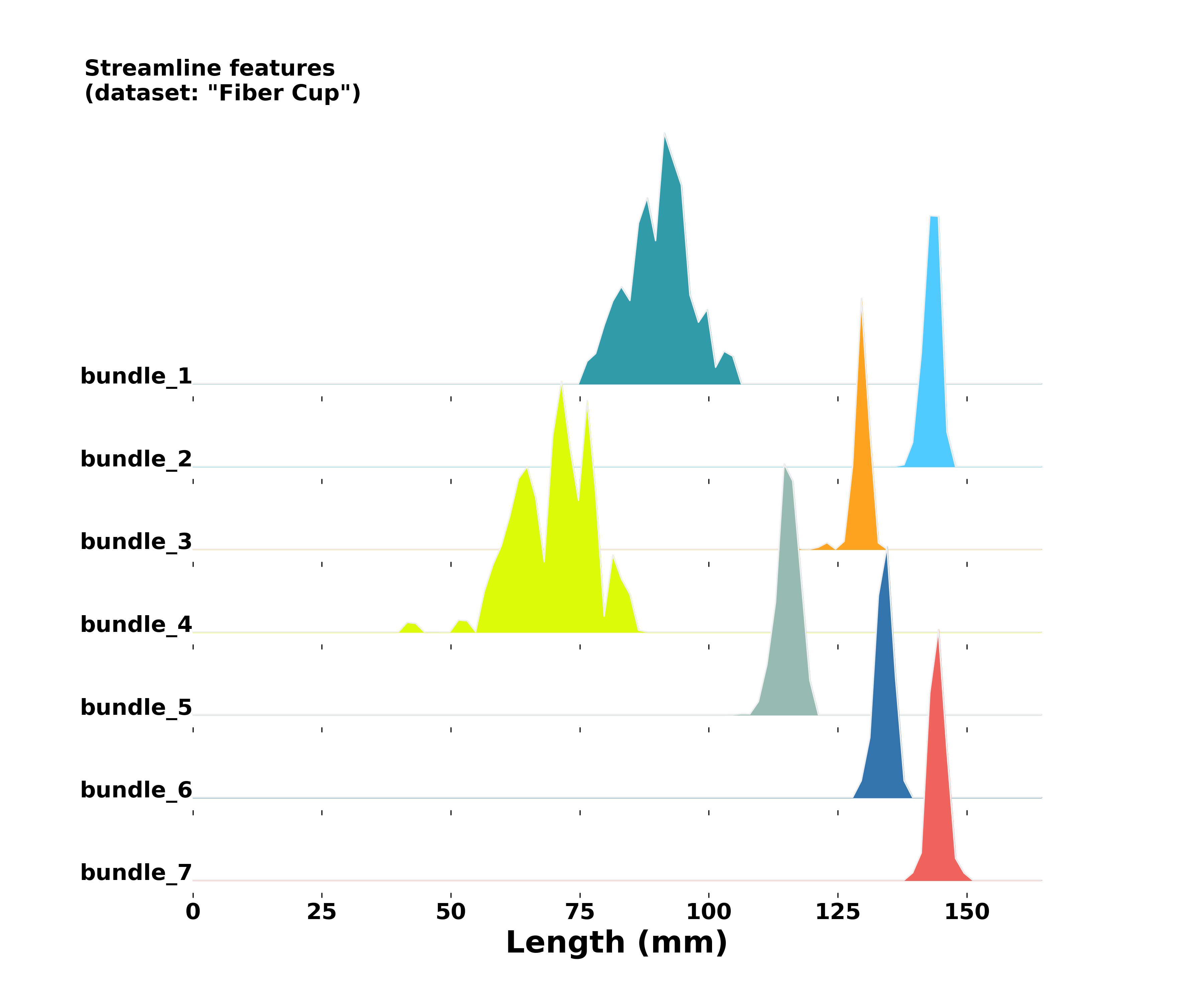} \\
\textbf{(c)} & \textbf{(d)} \\
\end{tabular}
\caption{\label{fig:fibercup_length_distribution}Streamline length distribution values for the ``Fiber Cup'' dataset: (a, b) seed streamlines; (c, d) latent space-generated streamlines; left: $P = 10${\percent}; right: $P = 100${\percent}. Note that the ranges are different.}
\end{figure*}

\begin{figure*}[!htbp]
\centering
\setlength{\tabcolsep}{0pt}
\begin{tabular}{cc}
\includegraphics[scale=0.95, trim=0.15in 0.15in 0.15in 0.15in, clip=true, width=0.45\linewidth, keepaspectratio=true]{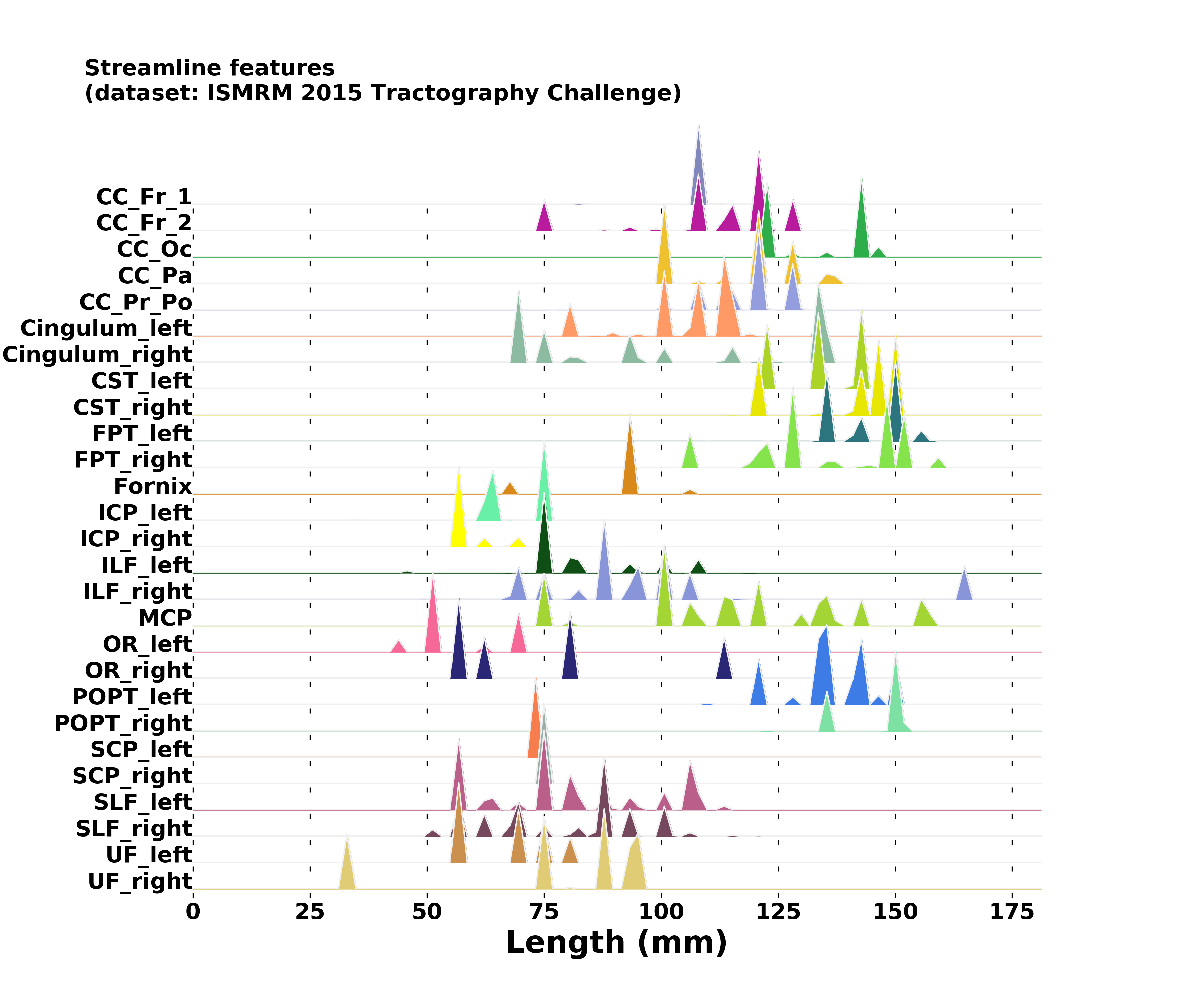} &
\includegraphics[scale=0.95, trim=0.15in 0.15in 0.15in 0.15in, clip=true, width=0.45\linewidth, keepaspectratio=true]{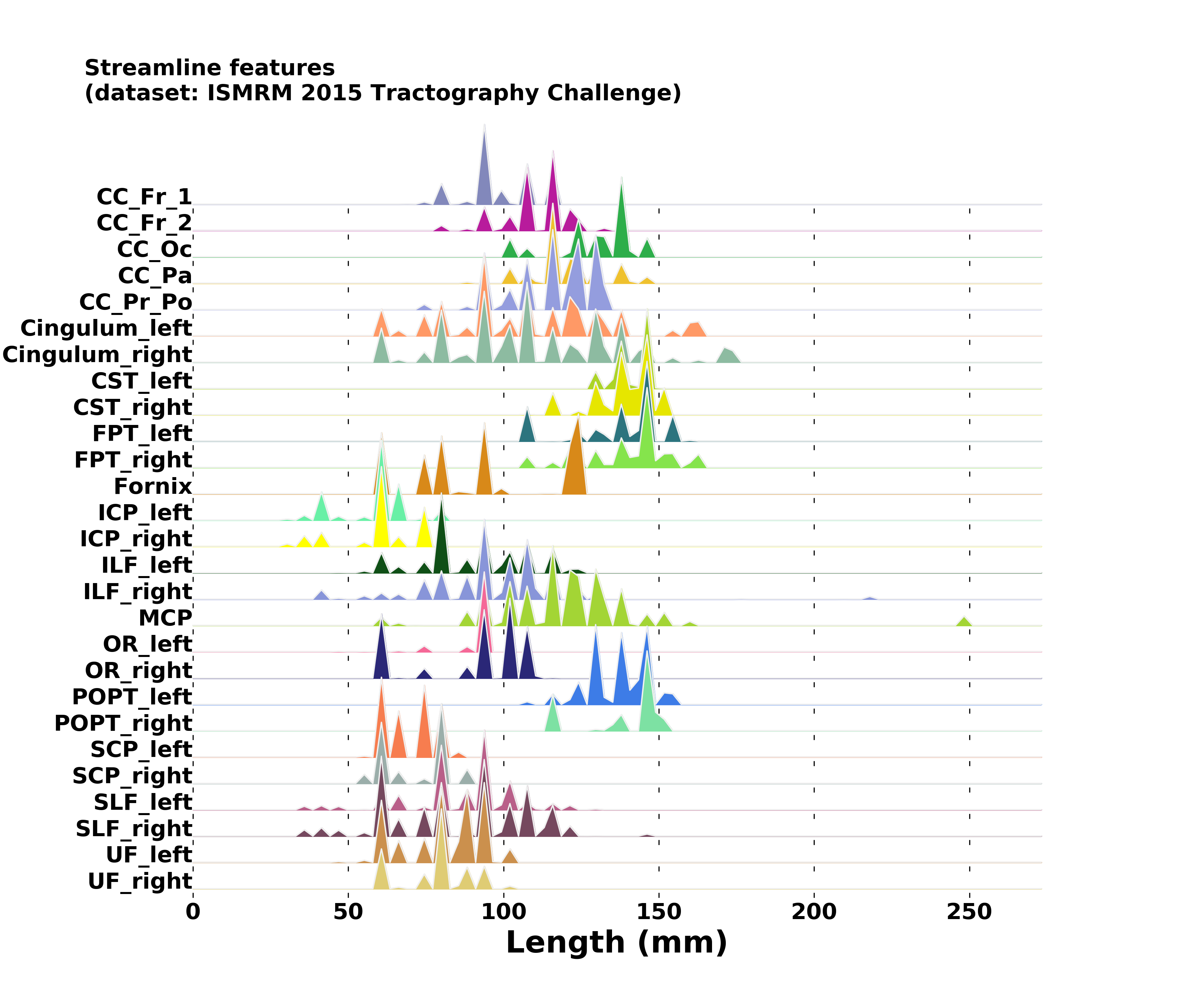} \\
\textbf{(a)} & \textbf{(b)} \\
\includegraphics[scale=0.95, trim=0.15in 0.15in 0.15in 0.15in, clip=true, width=0.45\linewidth, keepaspectratio=true]{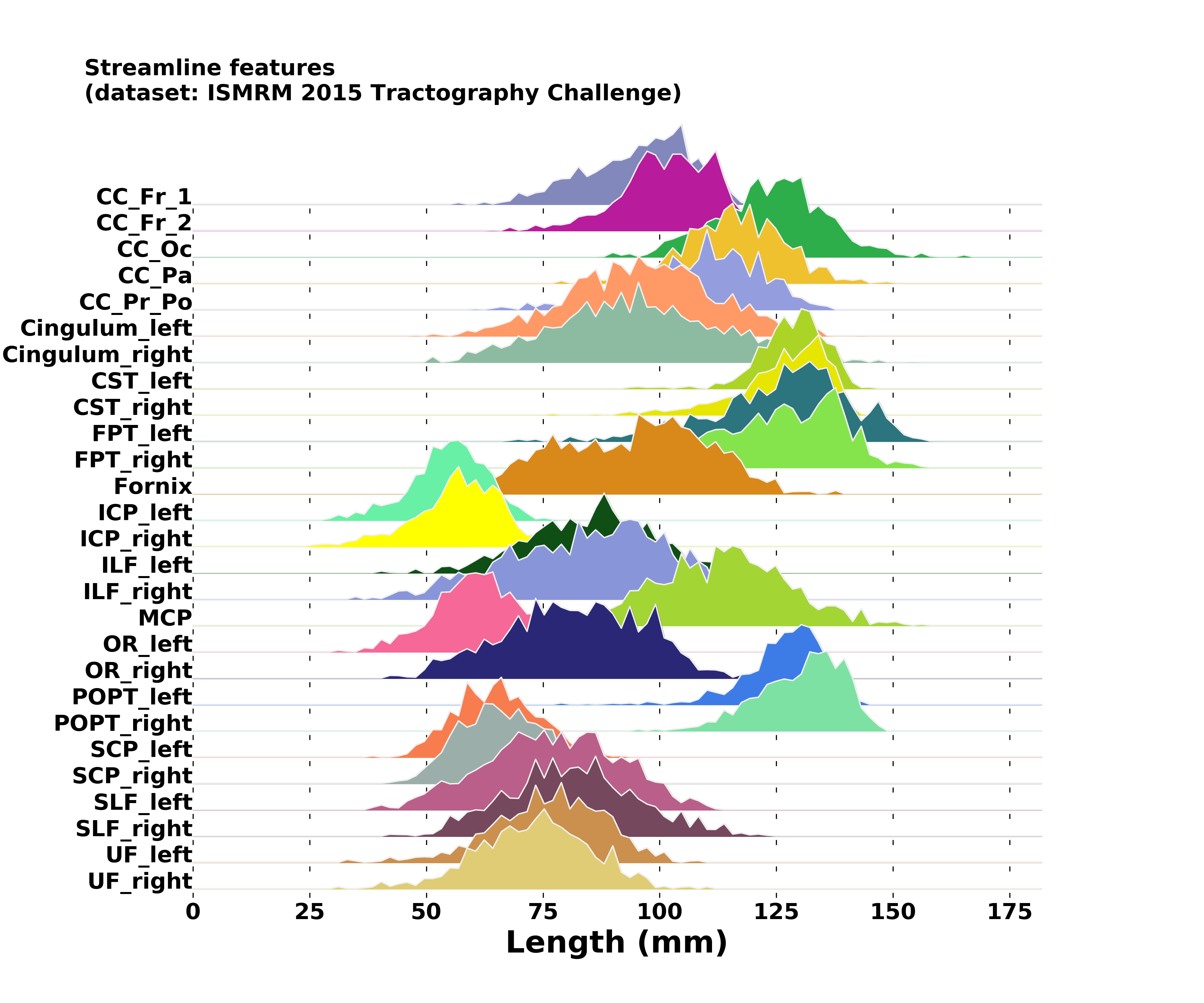} &
\includegraphics[scale=0.95, trim=0.15in 0.15in 0.15in 0.15in, clip=true, width=0.45\linewidth, keepaspectratio=true]{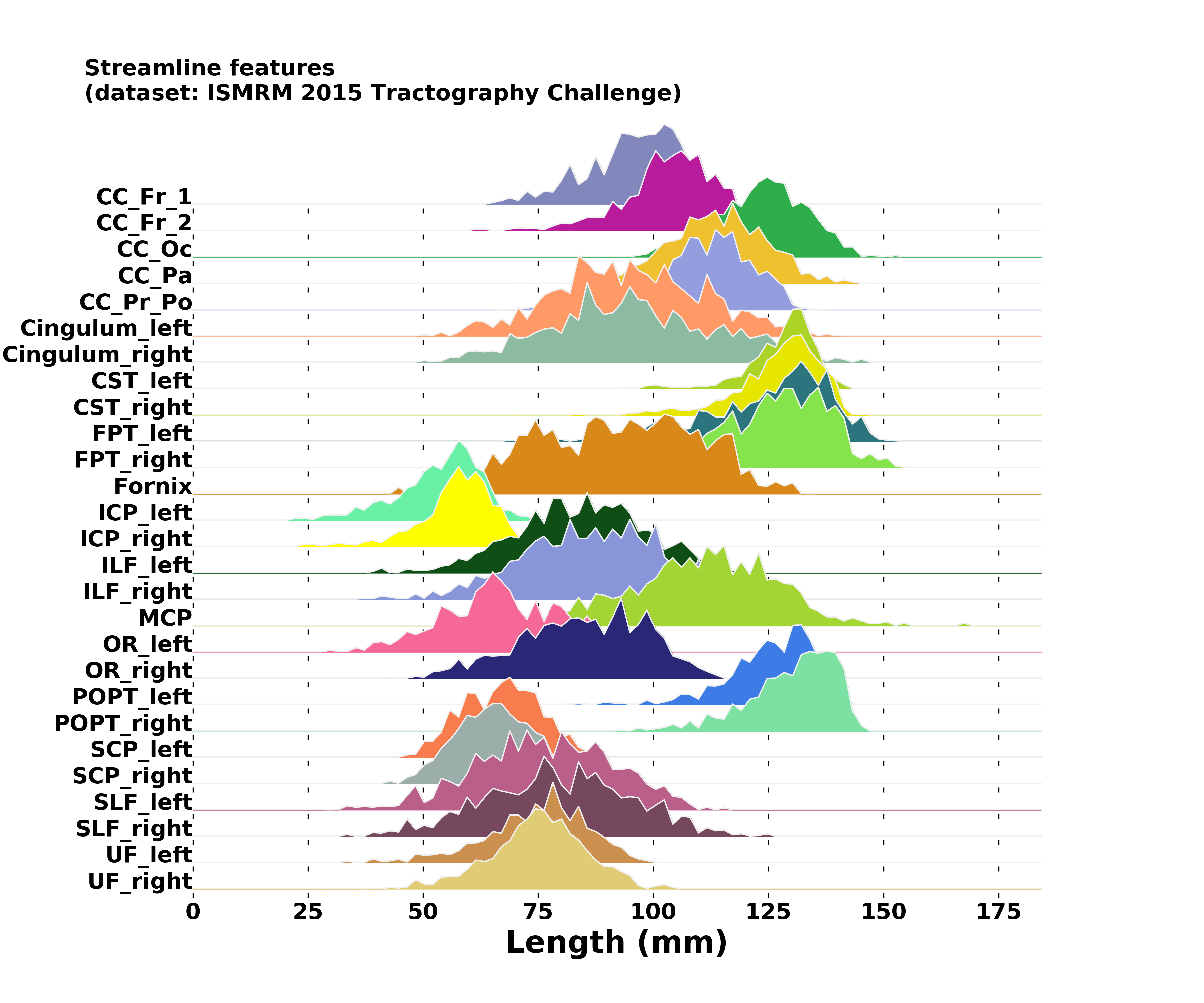} \\
\textbf{(c)} & \textbf{(d)} \\
\end{tabular}
\caption{\label{fig:ismrm2015_length_distribution}Streamline length distribution values for the ISMRM 2015 Tractography Challenge  dataset: (a, b) seed streamlines; (c, d) latent space-generated streamlines; left: $P = 10${\percent}; right: $P = 100${\percent}. Note that the ranges are different.}
\end{figure*}

Table \ref{tab:geometry_plausibility} shows the acceptance rate for the latent space-generated streamlines on each dataset. Additionally, figures \ref{fig:fibercup_length_distribution} and \ref{fig:ismrm2015_length_distribution} show the streamline length density plots for the ``Fiber Cup'' and ISMRM 2015 Tractography Challenge dataset seed streamlines and the corresponding generative tractography streamlines (prior to being evaluated for their plausibility) for the $P = 10${\percent} and $P = 100${\percent} cases.



\begin{table*}[!htbp]
\caption{\label{tab:geometry_plausibility}Acceptance rate for the latent space-generated streamlines. Mean and standard deviation values across bundles. Note that the acceptance rate for these features remains invariable across the ADG and the ADGC criteria. ``c/h'' denotes ``callosal homotopic'' for the BIL\&GIN data.}
\centering
\begin{tabular}{ccccccc}
\toprule
& & \multicolumn{5}{c}{\textbf{Dataset}}\\
\cmidrule(lr){3-7}
\textbf{Seed ratio ($P${\percent})} & \textbf{Measure} & ``Fiber Cup'' & ISMRM 2015 & BIL\&GIN c/h & \multicolumn{2}{c}{TractoInferno} \\
\cmidrule(lr){6-7}
& & & & & Det & Prob \\
\midrule
\multirow{2}{*}{3} & Length & 100 & 100 (0.01) & - & - & - \\
& Winding & 100 & 100 & - & - & - \\
\midrule
\multirow{2}{*}{5} & Length & 100 & 100 (0.01) & - & - & - \\
& Winding & 100 & 100 & - & - & - \\
\midrule
\multirow{2}{*}{10} & Length & 100 & 100 & - & - & - \\
& Winding & 100 & 100 & - & - & - \\
\midrule
\multirow{2}{*}{100} & Length & 100 & 100 (0.01) & 99.67 (1.37) & - & - \\
& Winding & 100 & 100 & 99.76 (0.44) & - & - \\
\midrule
\multirow{2}{*}{-} & Length & - & - & - & 100 & 100 \\
& Winding & - & - & - & 99.97 (0.07) & 99.87 (0.39) \\
\end{tabular}
\end{table*}

\newpage
\clearpage

\subsection{Acronyms}
\label{subsec:acronyms}

The bundles considered for the ISMRM 2015 Tractography Challenge dataset include: corpus callosum (CC); left/right cingulum (Cing); left/right cortico-spinal tract (CST); fornix (Fx); left/right fronto-pontine tract (FPT); left/right inferior cerebellar peduncle (ICP); left/right inferior longitudinal fasciculus (ILF); middle cerebellar peduncle (MCP); left/right optic radiation (OR); left/right parieto-occipital pontine tract (POPT); left/right superior cerebellar peduncle (SCP); left/right superior longitudinal fasciculus (SLF); and left/right uncinate fasciculus (UF).

The corpus callosum splits used for the ISMRM 2015 Tractography Challenge dataset are: frontal lobe (most anterior part) (CC\_Fr\_1); frontal lobe (most posterior part) (CC\_Fr\_2); occipital lobe (CC\_Oc); parietal lobe (CC\_Pa); pre-/post-central gyri (CC\_Pr\_Po); and temporal lobe (CC\_Te).

The BIL\&GIN callosal homotopic data had been defined using the following \num{26} gyral-based regions of interest: angular gyrus (AG); cingulum (Cing); cuneus (Cu); fusiform gyrus (FuG); hippocampus (Hippo); inferior frontal gyrus (IFG); inferior occipital gyrus (IOG); inferior temporal gyrus (ITG); insula (Ins); lateral fronto-orbital gyrus (LFOG); lingual gyrus (LG); middle frontal gyrus (MFG); middle fronto-orbital gyrus (MFOG); middle occipital gyrus (MOG); middle temporal gyrus (MTG); parahippocampal gyrus (PHG); post-central gyrus (PoCG); pre-central gyrus (PrCG); pre-cuneus (PrCu); rectus gyrus (RG); superior frontal gyrus (SFG); supramarginal gyrus (SMG); superior occipital gyrus (SOG); superior parietal gyrus (SPG); superior temporal gyrus (STG); and temporal pole (TPole).

The TractoInferno dataset bundles used in this work include: left/right arcuate fasciculus (AF); corpus callosum, frontal lobe (most anterior part) (CC\_Fr\_1); left/right optic radiation and Meyer's loop (OR\_ML); and left/right pyramidal tract (PYT).

\section*{Supplementary material}
\label{sec:suppl_material}

\subsection*{Human Connectome Project (HCP) data experiment}
\label{sec:hcp_dataset}

A subject from the Human Connectome Project (HCP) dataset \citep{Glasser:NatureNeurosc:2016} is used to perform whole-brain generative tractography on an \textit{in vivo} dataset. The data were acquired using an HCP scanner equipped with high-end hardware enabling diffusion encoding gradient strengths of \SI{100}{\milli\tesla\per\metre} at \SI{1.25}{\milli\metre} isotropic spatial resolution along \num{270} gradient directions. The available preprocessed data is used here.

The tractography data was reconstructed using global tracking \citep{Reisert:Neuroimage:2011}. The HCP data were registered \citep{Avants:Neuroimage:2011} to the ISMRM 2015 Tractography Challenge dataset space following \citep{Legarreta:MIA:2021}. The Tractometer tool \citep{Cote:MIA:2013, Maier-Hein:NatureComm:2017} was employed to obtain the bundles of interest, as the data had been registered to the ISMRM 2015 Tractography Challenge data space. The corpus callosum, initially available as a single bundle, is split into the \num{6} groups defined in \citet{Rheault:Zenodo:2021}, and each group is sampled separately. $P = $100\percent seed streamlines are used to generate $N =$ \num{15000} streamlines for each bundle using the \textit{unassisted seeding mode} of GESTA. Table \ref{tab:hcp_plausibility_criteria_values} shows the values used for the streamline plausibility assessment.

\begin{table*}[!htbp]
\caption{\label{tab:hcp_plausibility_criteria_values}Plausibility criteria values for the HCP dataset. LOA: local orientation angle; WM: white matter occupancy; T/F: binary requirement (true/false). Note that \textit{ADGC} includes the \textit{ADG} criteria.}
\centering
\begin{tabular}{cc|c}
\hline
& & \textbf{HCP}\\
\cmidrule{2-3}
\multirow{4}{*}{\textit{ADG}} & Length (\si{\milli\metre}) & \numrange[range-phrase=--]{20}{220} \\
& Winding (\si{\degree}) & \num{< 340} \\
& LOA-to-fODF peak (\si{\degree}) & \num{< 40} \\
& WM ratio ({\percent}) & \num{> 95} \\
\midrule
\textit{ADGC} & GM & T/F \\
\end{tabular}
\end{table*}

The overlap and overreach are used for evaluation purposes. As the data had been registered to the ISMRM 2015 Tractography Challenge dataset space in \citet{Legarreta:MIA:2021}, and as no re-training is done in this work, it naturally follows measuring the performance in terms of the overlap and overreach with the ISMRM 2015 Tractography Challenge dataset models.

Table \ref{tab:hcp_generative_tractography_measures} shows the bundle overlap and overreach of the latent-generated streamlines for both streamline plausibility criteria compared to those of the seed streamlines on the HCP dataset. On average, GESTA increases the overlap significantly by proposing new, anatomically plausible streamlines using the available set of seed streamlines. An increase is observed in the overreach as a result of using a soft requirement for the white matter volume occupancy criterion and a slightly dilated WM mask.

\begin{table*}[!htp]
\caption{\label{tab:hcp_generative_tractography_measures}HCP dataset reconstructed seed streamlines' and generative streamlines' overlap and overreach. $N =$ \num{15000} streamlines are generated for each bundle with a bandwidth factor of value \num{1.0}, and the plausibility is evaluated using the \textit{ADG\textsubscript{R}} and \textit{ADGC\textsubscript{R}} criteria. Mean and standard deviation values across bundles.}
\centering
\begin{tabular}{ccccc|cc}
\toprule
& & & \multicolumn{2}{c}{\textit{ADG\textsubscript{R}}} & \multicolumn{2}{c}{\textit{ADGC\textsubscript{R}}} \\
\cmidrule(lr){4-5}\cmidrule(lr){6-7}
\textbf{Seed ratio} ($P${\percent}) & \textbf{OL} & \textbf{OR} & \textbf{OL} ($\uparrow$) & \textbf{OR} ($\downarrow$) & \textbf{OL} ($\uparrow$) & \textbf{OR} ($\downarrow$) \\
\midrule
100 & 0.29 (0.17) & 0.12 (0.12) & 0.74 (0.15) & 1.36 (0.71) & 0.62 (0.15) & 0.88 (0.48) \\
\bottomrule
\end{tabular}
\end{table*}

\begin{figure*}[!htbp]
\centering
\setlength{\tabcolsep}{0pt}
\begin{tabular}{cccc}
\includegraphics[scale=0.95, trim=2.55in 2.5in 2.55in 3.0in, clip=true, width=0.225\linewidth, keepaspectratio=true]{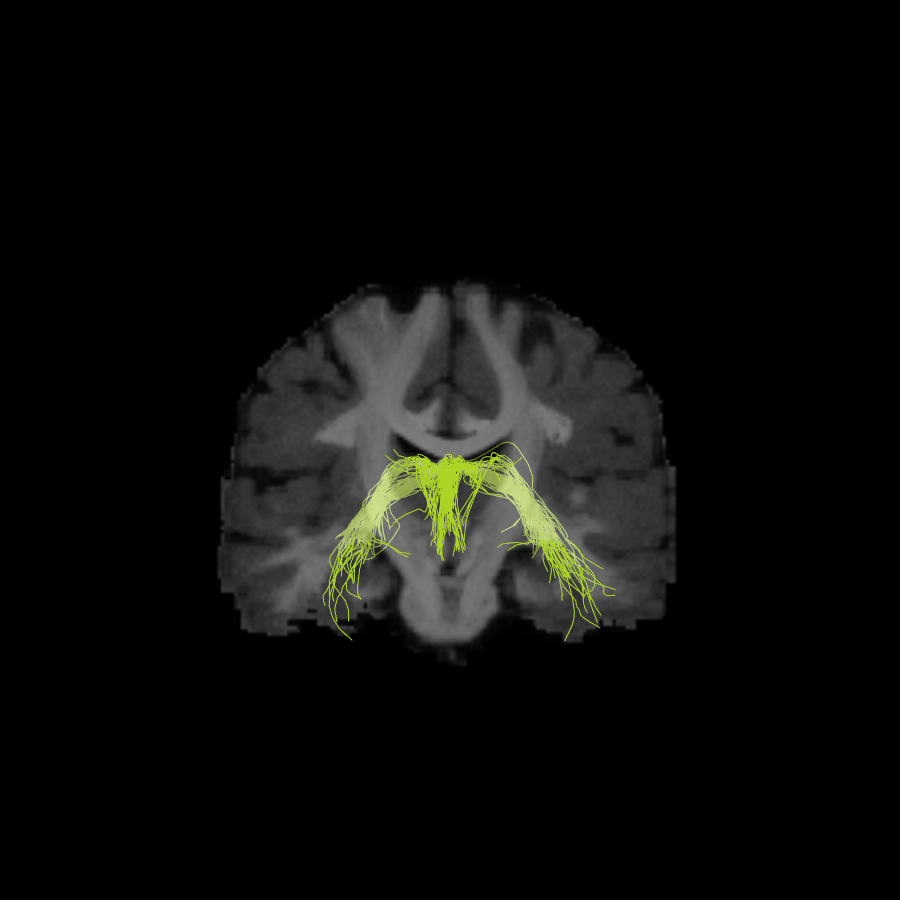} &
\includegraphics[scale=0.95, trim=1.75in 2.4in 1.75in 1.2in, clip=true, width=0.215\linewidth, keepaspectratio=true]{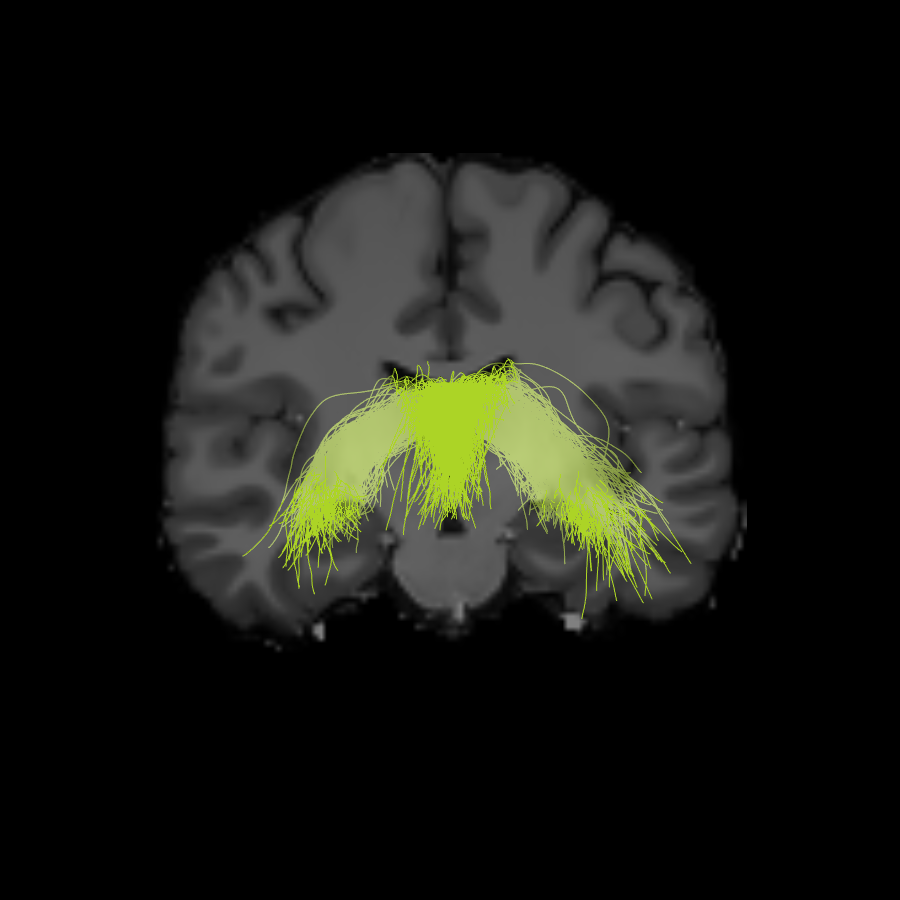} &
\hspace{0.01in}
\includegraphics[scale=0.95, trim=2.75in 3.15in 2.75in 2.75in, clip=true, width=0.225\linewidth, keepaspectratio=true]{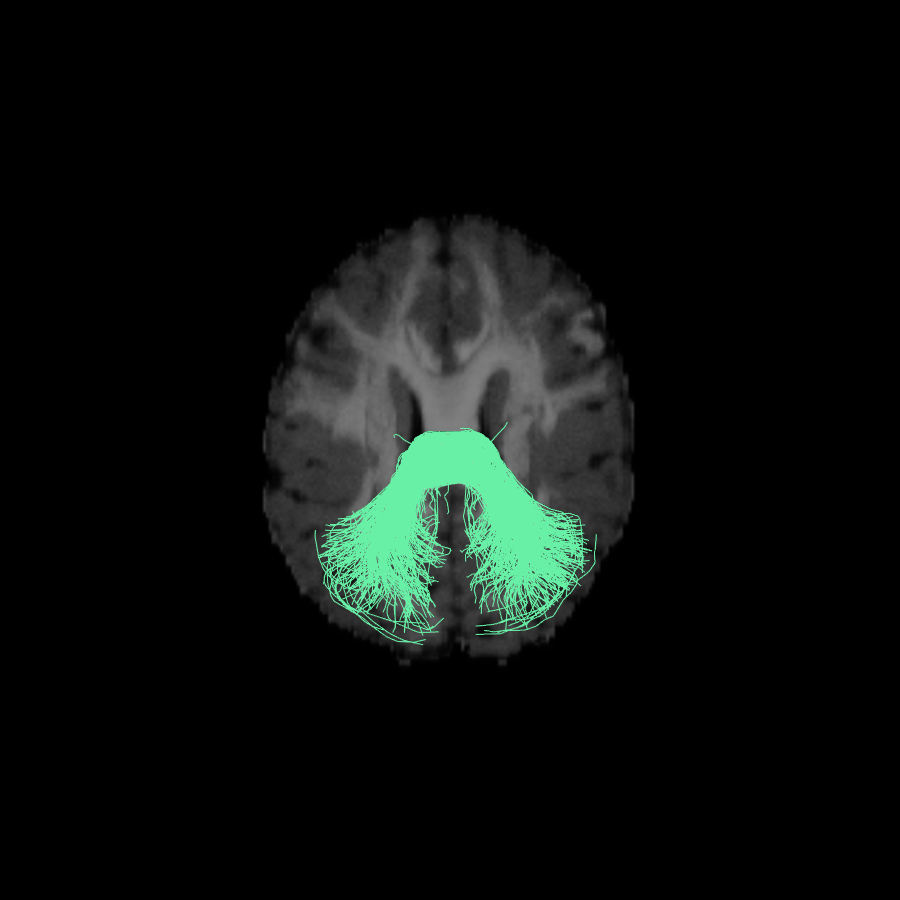} &
\includegraphics[scale=0.95, trim=1.95in 2.0in 1.95in 2.0in, clip=true, width=0.215\linewidth, keepaspectratio=true]{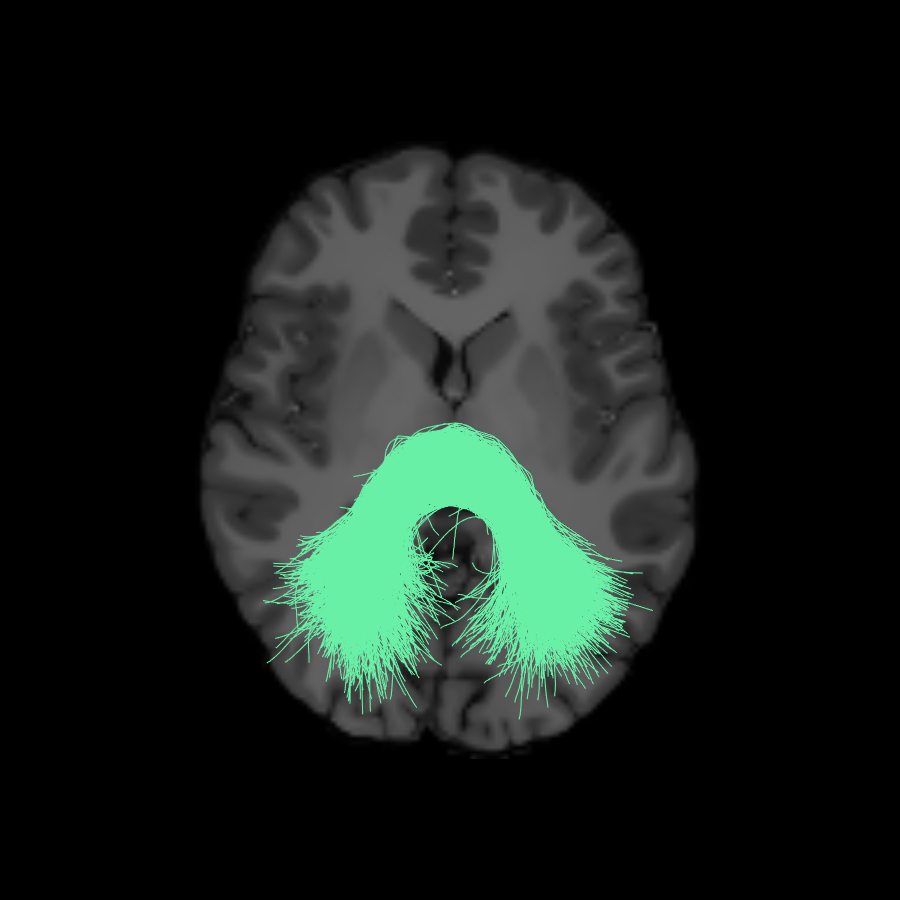} \\
\multicolumn{2}{c}{\textbf{(a)}} & \multicolumn{2}{c}{\textbf{(b)}} \\
\includegraphics[scale=0.95, trim=2.5in 2.25in 2.5in 3.25in, clip=true, width=0.225\linewidth, keepaspectratio=true]{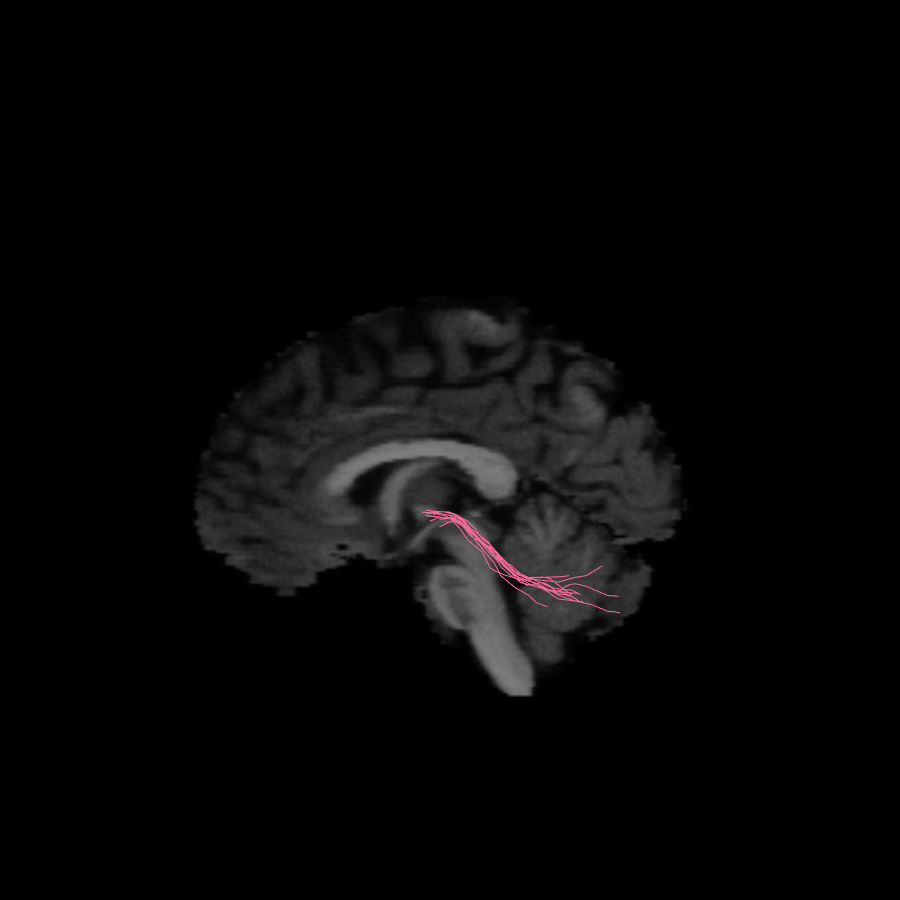} &
\includegraphics[scale=0.95, trim=1.95in 2.05in 1.95in 2.05in, clip=true, width=0.215\linewidth, keepaspectratio=true]{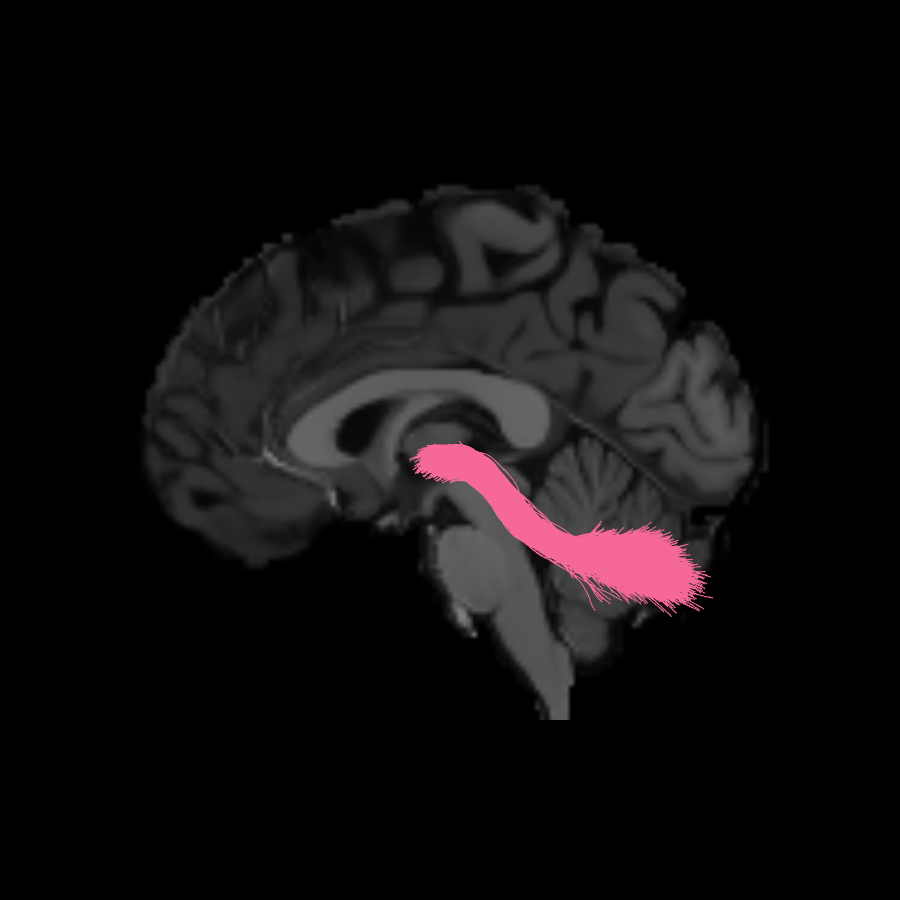} &
\hspace{0.01in}
\includegraphics[scale=0.95, trim=2.5in 2.25in 2.5in 3.25in, clip=true, width=0.225\linewidth, keepaspectratio=true]{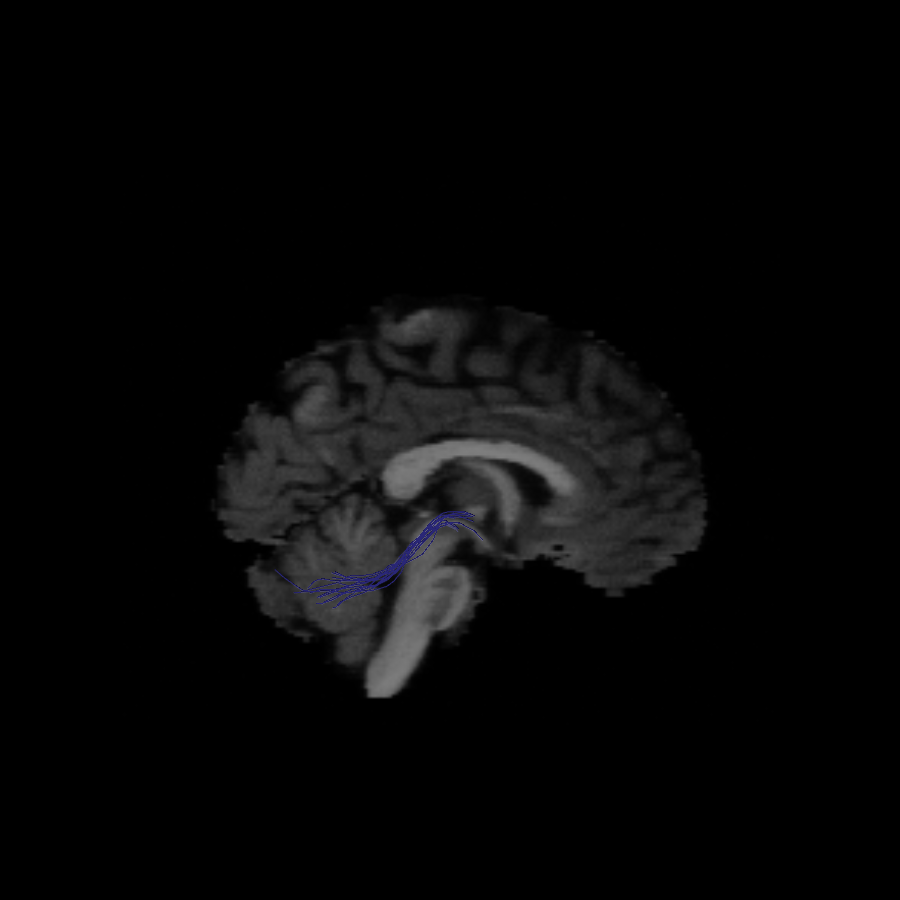} &
\includegraphics[scale=0.05, trim=1.95in 2.05in 1.95in 2.05in, clip=true, width=0.215\linewidth, keepaspectratio=true]{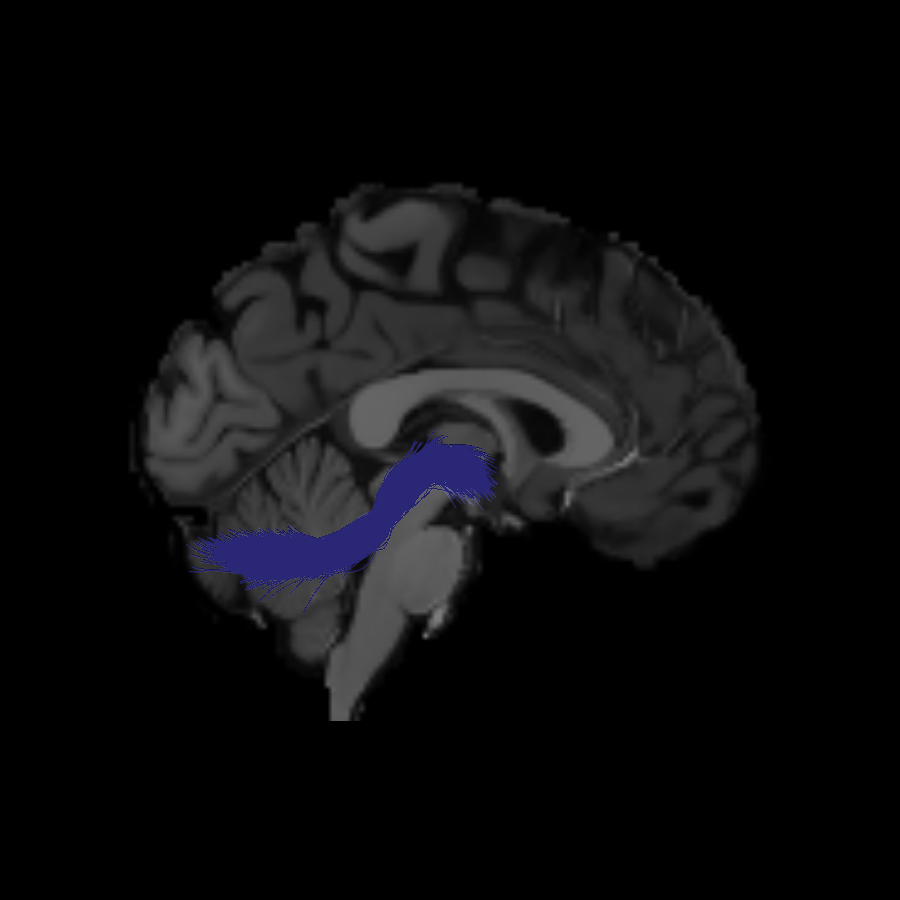} \\
\multicolumn{2}{c}{\textbf{(c)}} & \multicolumn{2}{c}{\textbf{(d)}} \\
\includegraphics[scale=0.95, trim=2.5in 2.25in 2.5in 3.25in, clip=true, width=0.225\linewidth, keepaspectratio=true]{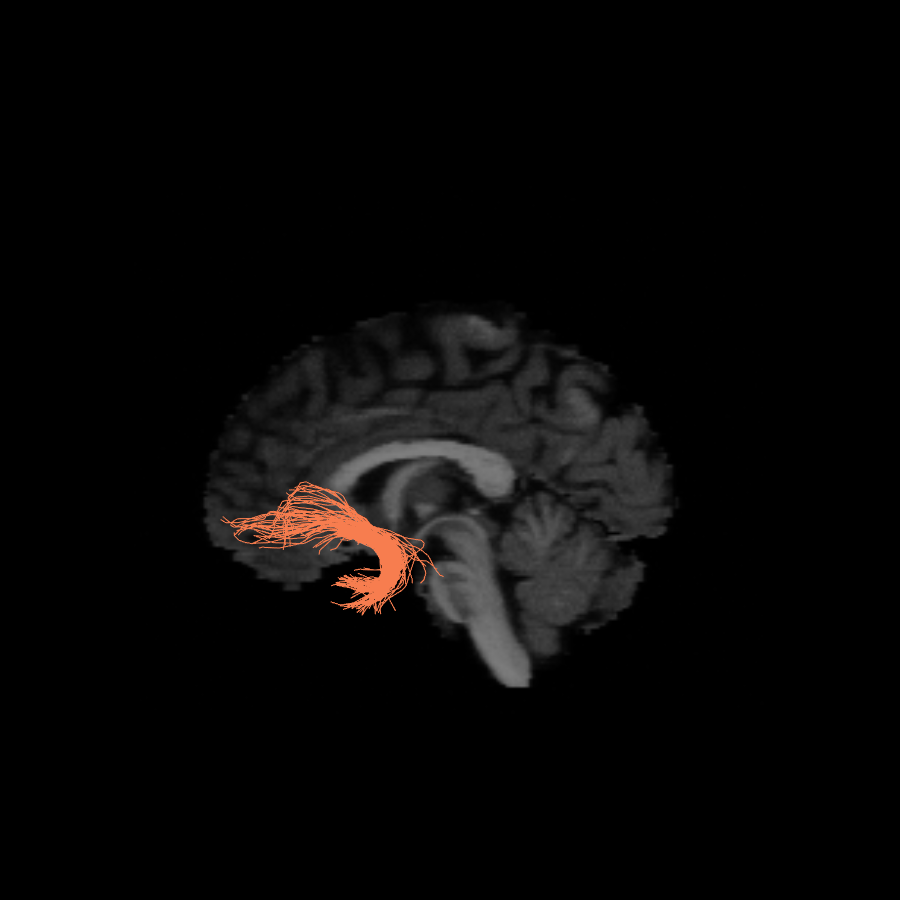} &
\includegraphics[scale=0.95, trim=1.95in 2.05in 1.95in 2.05in, clip=true, width=0.215\linewidth, keepaspectratio=true]{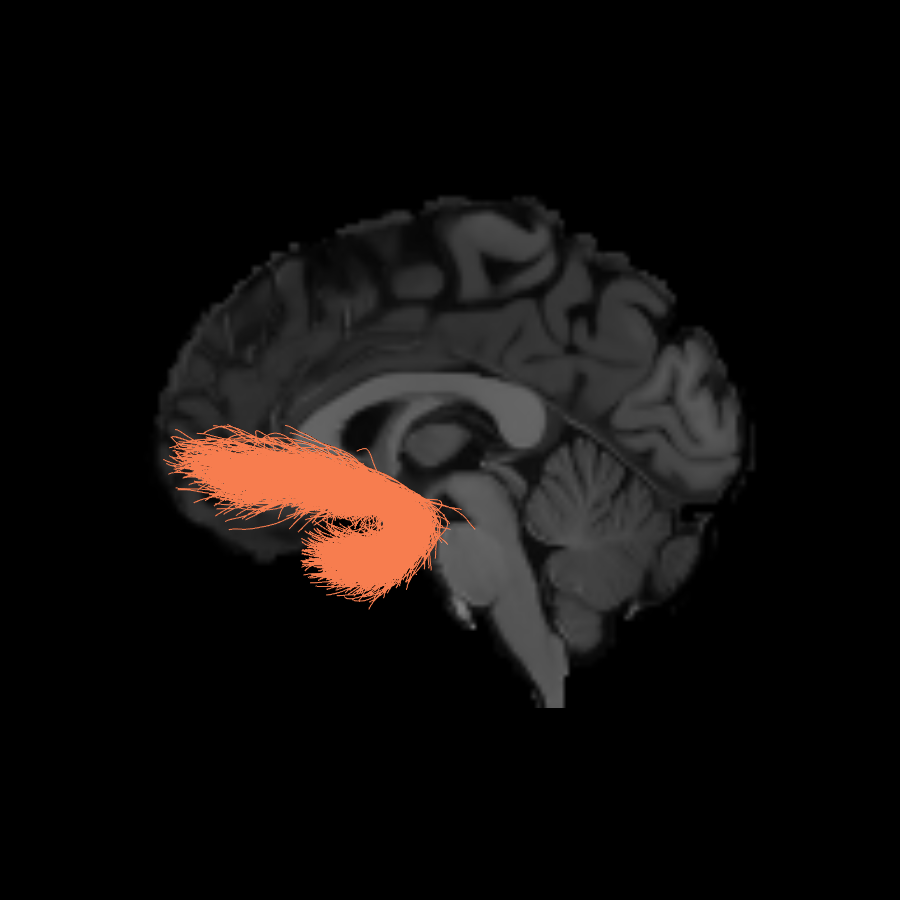} &
\hspace{0.01in}
\includegraphics[scale=0.95, trim=2.5in 2.25in 2.5in 3.25in, clip=true, width=0.225\linewidth, keepaspectratio=true]{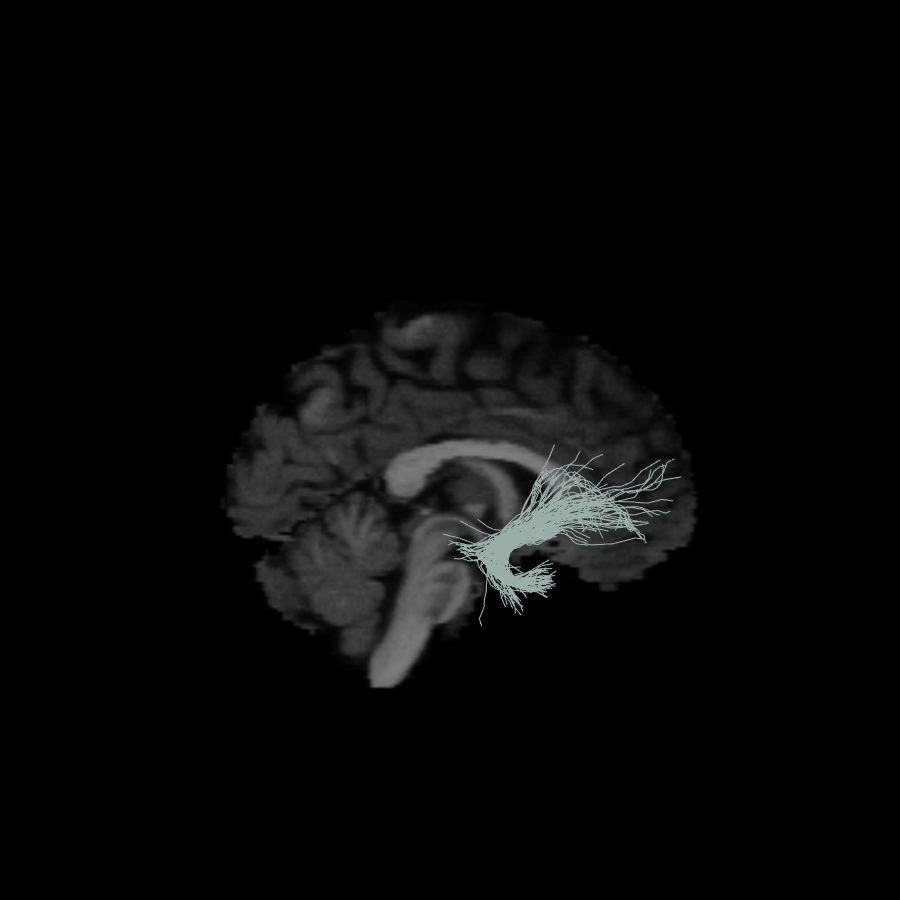} &
\includegraphics[scale=0.95, trim=1.95in 2.05in 1.95in 2.05in, clip=true, width=0.215\linewidth, keepaspectratio=true]{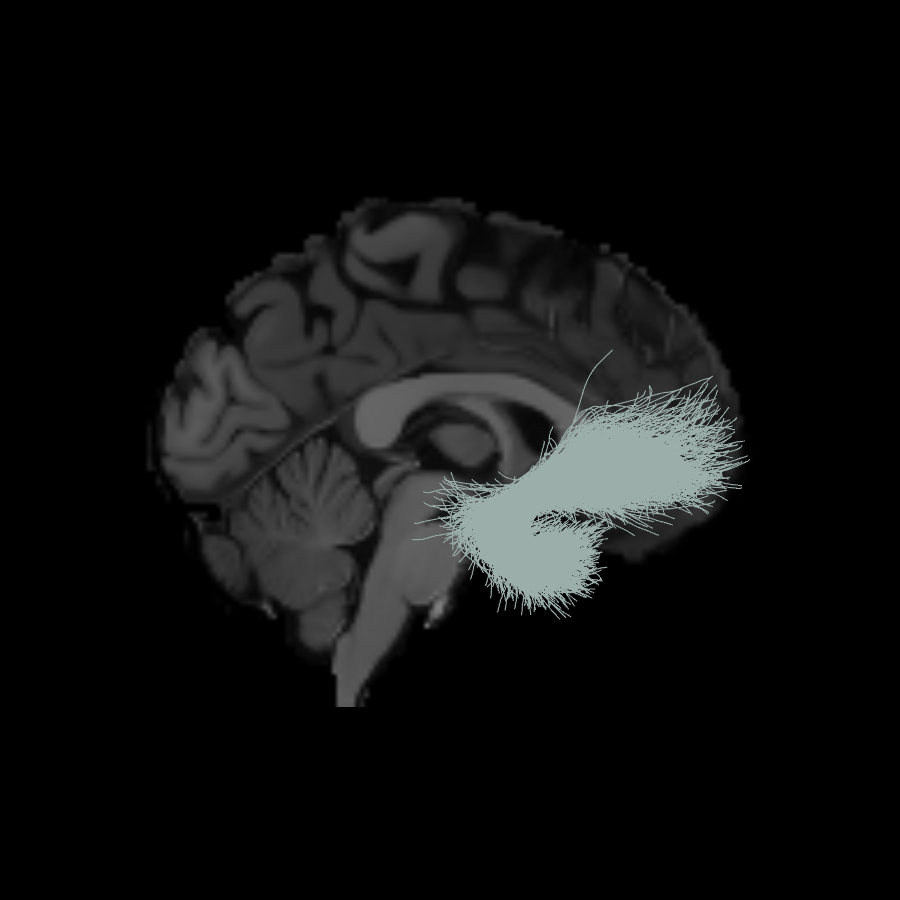} \\
\multicolumn{2}{c}{\textbf{(e)}} & \multicolumn{2}{c}{\textbf{(f)}} \\
\includegraphics[scale=0.95, trim=2.5in 2.25in 2.5in 3.25in, clip=true, width=0.225\linewidth, keepaspectratio=true]{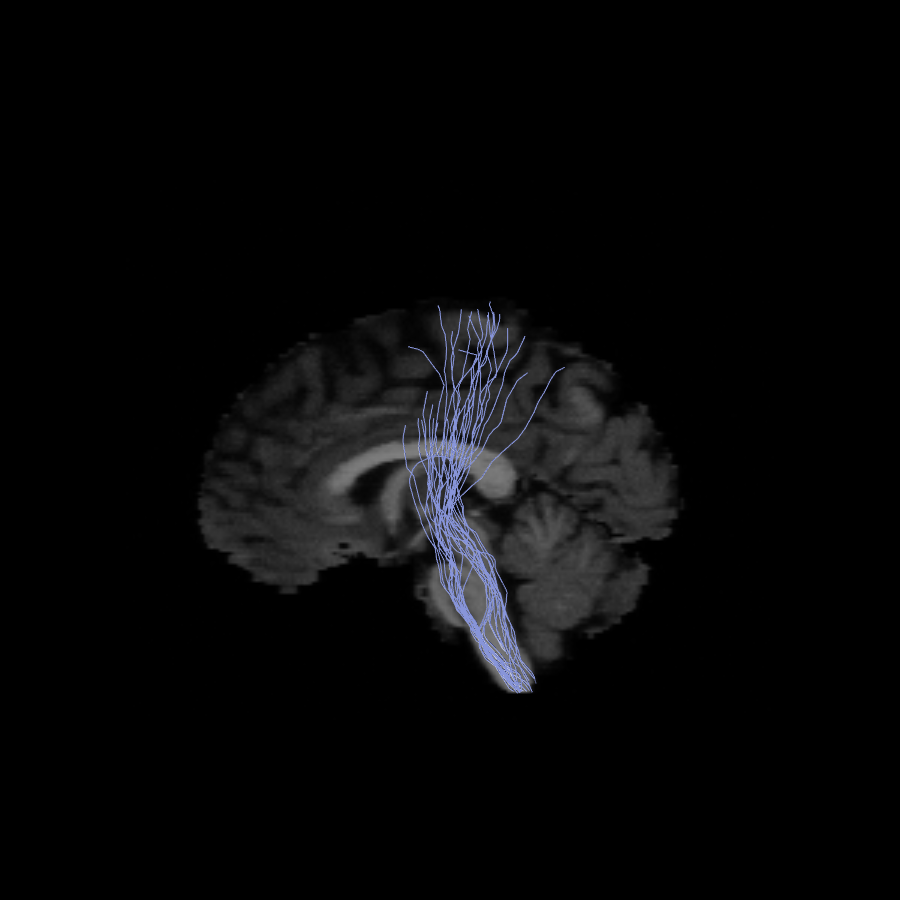} &
\includegraphics[scale=0.95, trim=1.95in 2.05in 1.95in 2.05in, clip=true, width=0.215\linewidth, keepaspectratio=true]{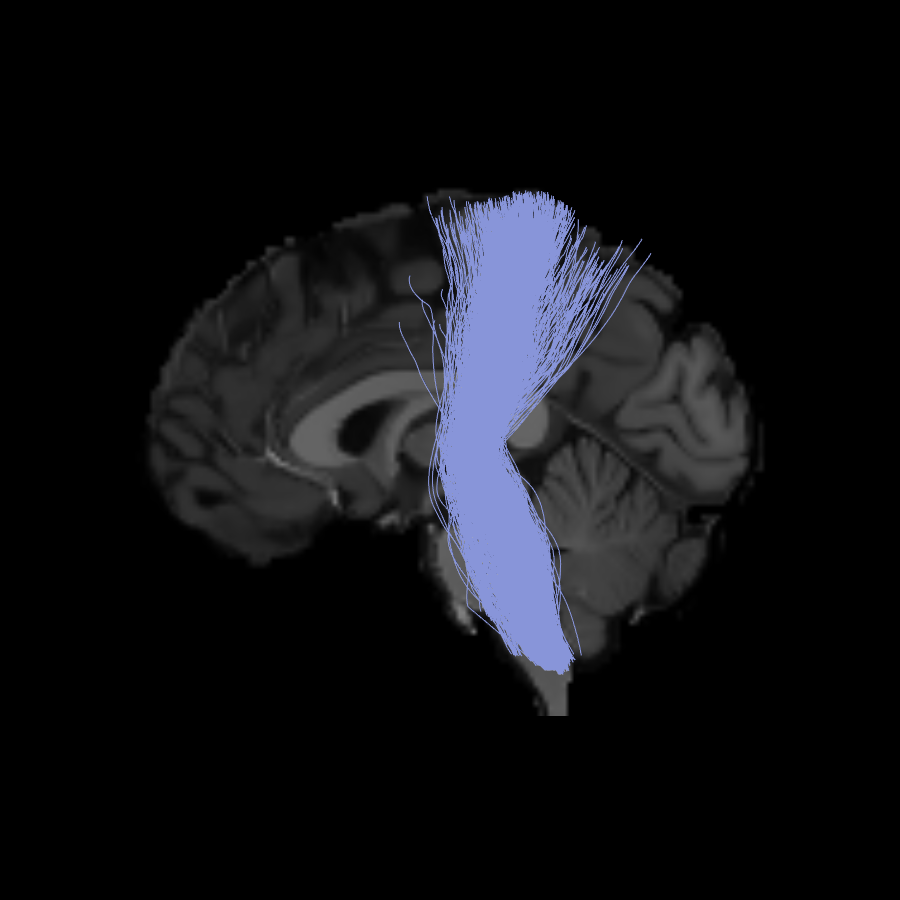} &
\hspace{0.01in}
\includegraphics[scale=0.95, trim=2.5in 2.25in 2.5in 3.25in, clip=true, width=0.225\linewidth, keepaspectratio=true]{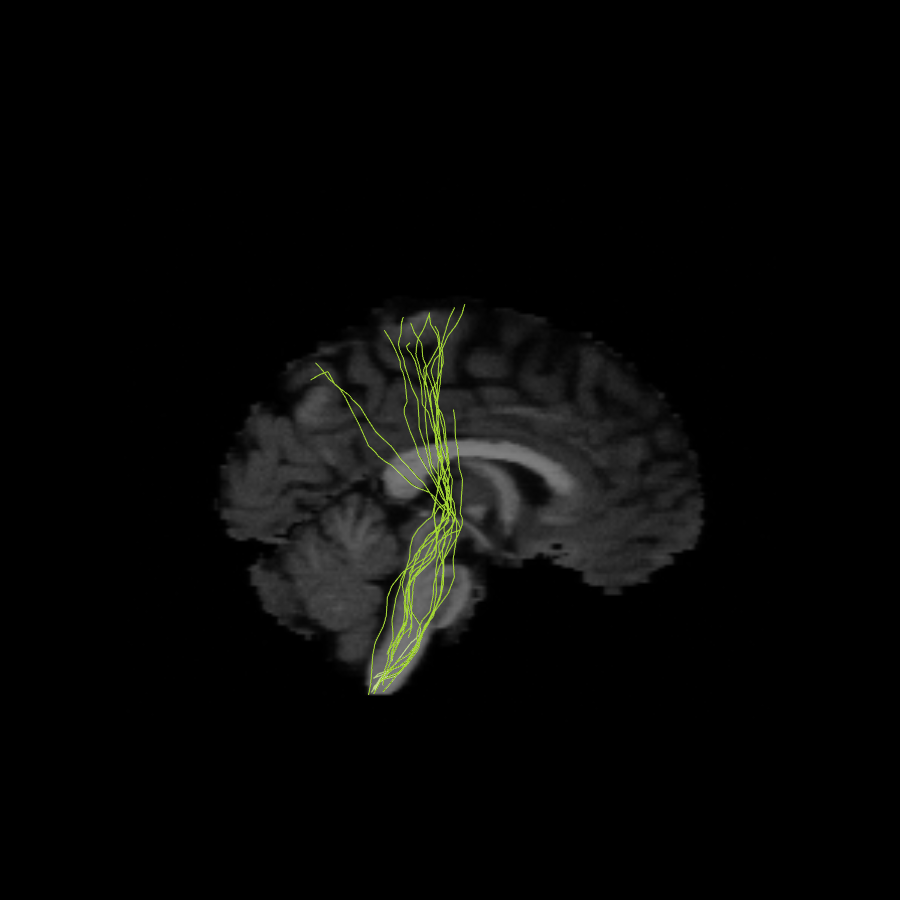} &
\includegraphics[scale=0.95, trim=1.95in 2.05in 1.95in 2.05in, clip=true, width=0.215\linewidth, keepaspectratio=true]{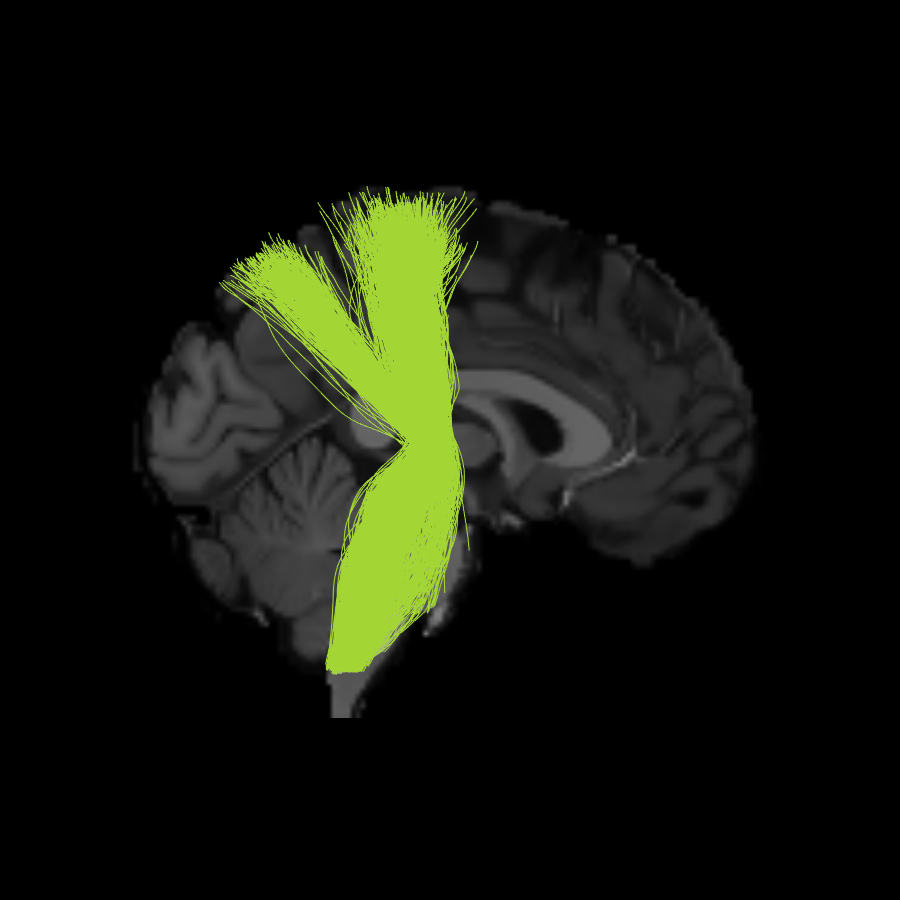} \\
\multicolumn{2}{c}{\textbf{(g)}} & \multicolumn{2}{c}{\textbf{(h)}} \\
\includegraphics[scale=0.95, trim=2.75in 3.15in 2.75in 2.75in, clip=true, width=0.225\linewidth, keepaspectratio=true]{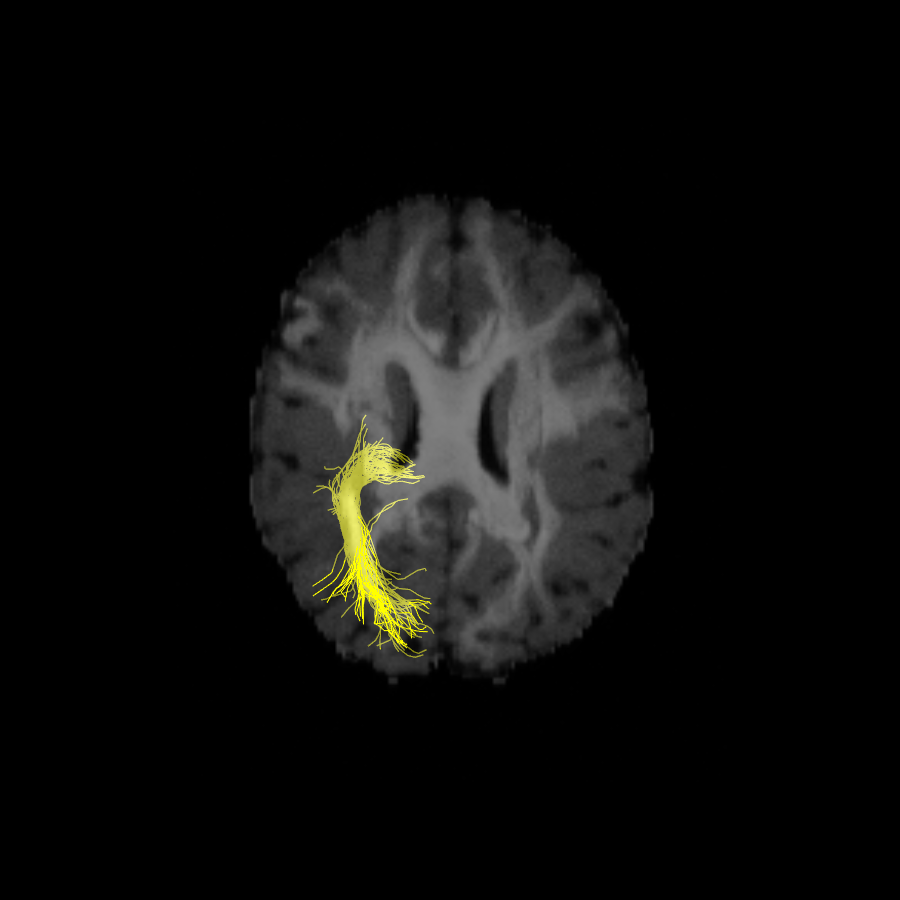} &
\includegraphics[scale=0.95, trim=1.95in 2.0in 1.95in 2.0in, clip=true, width=0.215\linewidth, keepaspectratio=true]{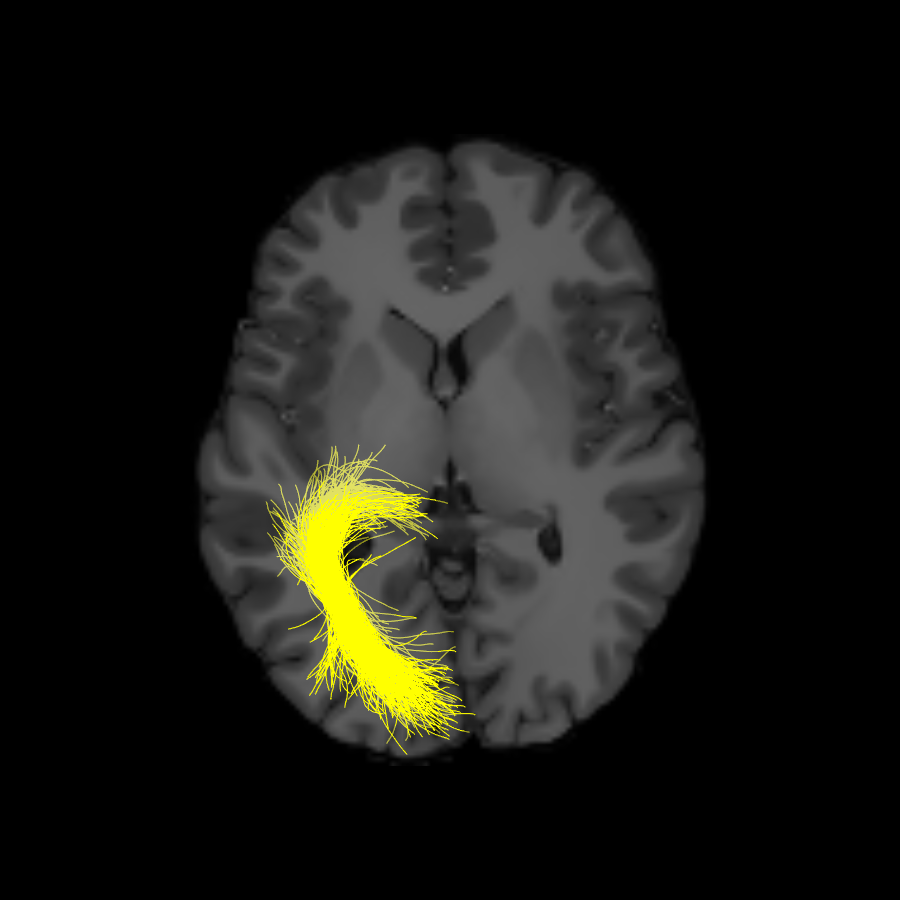} &
\hspace{0.01in}
\includegraphics[scale=0.95, trim=2.75in 3.15in 2.75in 2.75in, clip=true, width=0.225\linewidth, keepaspectratio=true]{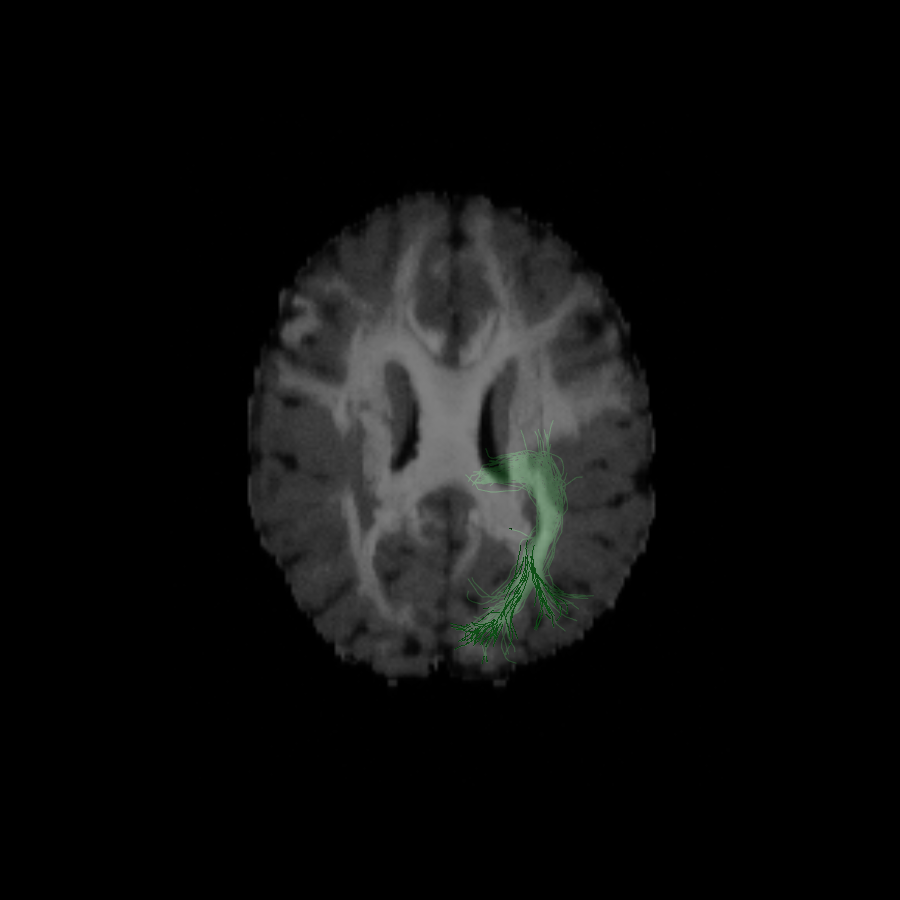} &
\includegraphics[scale=0.95, trim=1.95in 2.0in 1.95in 2.0in, clip=true, width=0.215\linewidth, keepaspectratio=true]{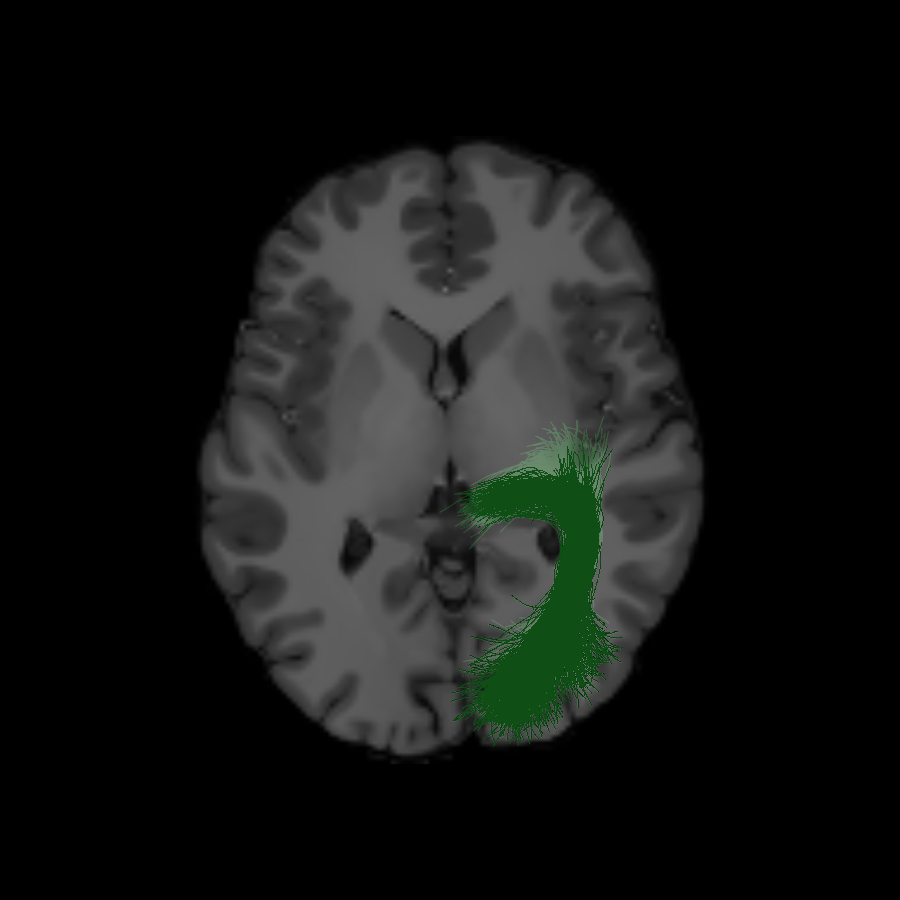} \\
\multicolumn{2}{c}{\textbf{(i)}} & \multicolumn{2}{c}{\textbf{(j)}} \\
\end{tabular}
\caption{\label{fig:hcp_generative_bundles}GESTA applied to the (a) fornix; (b) MCP; (c) left SCP; (d) right SCP; (e) left UF; (f) right UF; (g) left POPT; (h) right POPT; (i) left OR; and (j) right OR bundles of the HCP dataset. For each bundle: left: seed streamlines; right: GESTA-generated  plausible streamlines. All available streamlines are used for seeding in the latent space, and streamlines are evaluated using the \textit{ADGC\textsubscript{R}} criterion. Seed streamlines are in the ISMRM 2015 Tractography Challenge dataset space; latent-generated streamlines are in subject space.}
\end{figure*}

Ten (\num{10}) bundles obtained using the proposed generative tractography method for the HCP dataset are shown in figure \ref{fig:hcp_generative_bundles}. As it is the case with the ``Fiber Cup'' and ISMRM 2015 Tractography Challenge datasets, GESTA populates each of the bundles with anatomically plausible streamlines using a limited set of seed streamlines: it increases by \num{2.6} times the average overlap across bundles for the \textit{ADG\textsubscript{R}} criterion. Although the number of seeds varies notably across bundles (see the difference between the MCP and both POPT bundles), the rejection sampling procedure is still able to produce new samples using an extremely modest set of seed streamlines.

\end{document}